%% file: main.tex
\documentclass[a4paper,12pt,notitlepage]{report}

\input{preamble/thesis}

\title{Embodied Spatial Intelligence: From 3D Perception to 3D Reasoning}
\author{Jiading Fang \\ Toyota Technological Institute at Chicago (TTIC)}

\begin{document}

\input{contents/title_page}
\clearpage

\onehalfspacing

\pagenumbering{roman}

\input{contents/abstract}

\input{contents/acknowledgement}
\clearpage

\input{contents/toc}
\newpage

\pagenumbering{arabic}

\input{contents/introduction}
\clearpage

\include{contents/chap_scene}

\section{Implicit camera estimation for robust scene understanding: Self-supervised camera self-calibration}\label{sec:self-calib}
\input{papers/selfcalib/sections/abstract}
\subsection{Introduction}
\input{papers/selfcalib/sections/introduction}

\subsection{Related Work}
\input{papers/selfcalib/sections/related}
\subsection{Methodology}
\input{papers/selfcalib/sections/methodology}
\subsection{Experiments}
\input{papers/selfcalib/sections/experiments}
\subsection{Conclusion}
\input{papers/selfcalib/sections/conclusion}
\clearpage

\section{Implicit generalizable depth field modeling by 3D data augmentation: DeFiNe}\label{sec:define}
\input{papers/define/sections/00abstract}

\subsection{Introduction}

\input{papers/define/sections/01introduction}

\subsection{Related Work}
\input{papers/define/sections/02related}

\subsection{Methodology}

\input{papers/define/sections/03methodology}

\subsection{Experiments}

\input{papers/define/sections/04experiments}

\subsection{Conclusion}
\input{papers/define/sections/05conclusion}

\clearpage

\section{Implicit representation as modular maps for scalable scene modelling: NeRFuser}\label{sec:nerfuser}
\input{papers/nerfuser/sections/abstract}
\subsection{Introduction}

\input{papers/nerfuser/sections/introduction}

\subsection{Related Work}
\input{papers/nerfuser/sections/related_work}
\subsection{Methodology}
\input{papers/nerfuser/sections/methodology}

\subsection{Experiments}
\input{papers/nerfuser/sections/experiments}

\subsection{Conclusion}
\input{papers/nerfuser/sections/conclusion}
\clearpage

\include{contents/chap_reason}

\section{Benchmarking Mapping and Navigation Capabilities of LLMs: MANGO}\label{sec:mango}
\input{papers/mango/sections/abstract}
\subsection{Introduction}
\input{papers/mango/sections/introduction}
\subsection{Related Work}
\input{papers/mango/sections/related_work}

\subsection{A Benchmark for Text-Based Mapping and Navigation}
\input{papers/mango/sections/benchmark}
\subsection{Experiments}
\input{papers/mango/sections/experiments}

\subsection{Conclusion}
\input{papers/mango/sections/conclusion}
\clearpage

\section{3D Referring Expression Resolution through LLMs with iterative reasoning: Transcrib3D}\label{sec:transcrib3d}
\input{papers/transcrib3d/sections/abstract}
\subsection{Introduction}
\input{papers/transcrib3d/sections/intro}

\subsection{Related Work}

\input{papers/transcrib3d/sections/related_works}

\subsection{Methodology}

\input{papers/transcrib3d/sections/methodology}

\subsection{Experiments}
\input{papers/transcrib3d/sections/experiments}
\subsection{Discussion}
\input{papers/transcrib3d/sections/discussion}

\subsection{Conclusion}

\input{papers/transcrib3d/sections/conclusion}
\clearpage

\section{State-maintaining language model for long-horizon embodied reasoning: Statler}\label{sec:statler}
\input{papers/statler/sections/abstract}

\subsection{Introduction}
\input{papers/statler/sections/introduction}
\subsection{Related Work}
\input{papers/statler/sections/related_work}
\subsection{Motivating Example}
\input{papers/statler/sections/motivating_example}
\subsection{Methodology}
\input{papers/statler/sections/methodology}

\subsection{Experiments}
\input{papers/statler/sections/experiments}

\subsection{Conclusion}
\input{papers/statler/sections/conclusion}
\clearpage

\input{contents/conclusion}

{\small
\bibliographystyle{IEEEtranN}
\bibliography{references}
}

\end{document}

%% file: preamble/thesis.tex
\usepackage{graphicx} %
\usepackage[square, numbers, sort]{natbib}
\usepackage{mathtools}
\usepackage{amsmath}
\usepackage{amssymb}
\usepackage{svg}
\usepackage{bm}
\usepackage{booktabs}
\usepackage{multirow}
\usepackage{tabularx}
\usepackage{wrapfig}
\usepackage{makecell}
\usepackage{titlesec}
\usepackage{xspace}
\usepackage{microtype}      %
\usepackage{url}            %
\usepackage[hidelinks]{hyperref}
\usepackage{csquotes}
\expandafter\def\expandafter\quote\expandafter{\quote\small}

\titleformat{\paragraph}
{\normalfont\normalsize\bfseries}{\theparagraph}{1em}{}
\titlespacing*{\paragraph}
{0pt}{3.25ex plus 1ex minus .2ex}{1.5ex plus .2ex}

\makeatletter
\renewcommand{\theparagraph}{\thesubsection.\arabic{paragraph}}
\@addtoreset{paragraph}{subsection}
\makeatother

\usepackage[inline]{enumitem}
\usepackage{subcaption}

\newcolumntype{Y}{>{\centering\arraybackslash}X}

\usepackage[precision=2, unit=mm]{lengthconvert}
\usepackage{tikz}
\usetikzlibrary{patterns,calc,fit,math, positioning}
\tikzset{
    response/.style 2 args = {
        draw, line width=1pt, inner sep=0, outer sep=0,
        fit=(#1) (#2)}
}

\newcommand{\greencheck}{}%
\DeclareRobustCommand{\greencheck}{%
  \tikz\fill[scale=0.3, color=green!50!black]
  (0,.35) -- (.25,0) -- (1,.7) -- (.25,.15) -- cycle;%
}

\newcommand{\redx}{}%
\DeclareRobustCommand{\redx}{%
    \tikz[scale=0.23, color=red!50!black] {
    \draw[line width=0.7,line cap=round] (0,0) to [bend left=6] (1,1);
    \draw[line width=0.7,line cap=round] (0.2,0.95) to [bend right=3] (0.8,0.05);}
}

\newcommand{\nfrac}[2]{$#1$~\raisebox{1.2pt}{\scalebox{0.75}{$#2$}}}
\newcommand{\nfracb}[2]{$\bm{#1}$~\raisebox{1.2pt}{\scalebox{0.75}{$\bm{#2}$}}}

\usepackage{newfloat}
\DeclareFloatingEnvironment{codefloat}
\usepackage{listings}

\newcommand{\codefont}{\fontfamily{lmtt}\selectfont}
\usepackage{listings}

\usepackage{parcolumns}
\lstdefinestyle{datalogstyle}{
	basicstyle={\codefont\scriptsize},  %
	xleftmargin={6pt},
    xrightmargin={6pt},
	numbers=left,
    columns=flexible,
    breakindent=0pt,
    breaklines=true, 
	frame=tb,
	stepnumber=1,
	firstnumber=1,
	numberfirstline=true,
	tabsize=2,
	extendedchars=true,
	breaklines=true,
	columns=fullflexible,
	keepspaces=true,
	escapeinside={@}{@},
	firstnumber=last,
	captionpos=b, 
	commentstyle=\color{black!65},
	numberstyle=\tiny\color{black!65},
	stringstyle=\color{codepurple},
	breakatwhitespace=false, 
	keepspaces=true,              
    mathescape=true, 
	numbersep=5pt,                  
	showspaces=false,                
	showstringspaces=false,
	showtabs=false,
	aboveskip={0.8\baselineskip},
	belowskip={0.2\baselineskip},
}
\usepackage{soul}

\definecolor{aigold}{RGB}{244,210, 1} 
\definecolor{aigreen}{RGB}{213, 245, 227}

\definecolor{humanpurple}{RGB}{235, 222, 240}

\definecolor{commentgray}{RGB}{86, 101, 115}

\definecolor{aired}{RGB}{255,180,181}

\usepackage[noabbrev,capitalize]{cleveref} %
\crefname{equation}{Equation}{Equations}   %
\crefname{section}{Section}{Sections}      %
\crefname{footnote}{Footnote}{Footnotes}   
\crefname{listing}{Prompt}{Prompts}
\crefname{assumption}{Assumption}{Assumptions}
\crefname{line}{Line}{Lines}   %
\crefname{paragraph}{Paragraph}{Paragraphs}

\usepackage{xpatch}
\usepackage{setspace}

\newcommand{\data}[1]{\texttt{\codefont#1}}
\newcommand{\code}[1]{\data{#1}}  %
\newcommand{\mvar}[1]{\code{\textit{#1}\hspace{-1pt}}} %

\usepackage{pifont}
\newcommand{\circone}{\ding{172}\xspace}
\newcommand{\circtwo}{\ding{173}\xspace}
\newcommand{\circthree}{\ding{174}\xspace}

\newcommand{\defeq}{\mathrel{\stackrel{\textnormal{\tiny def}}{=}}}

\definecolor{lightgreen}{rgb}{0.56, 0.93, 0.56}
\definecolor{mygreen}{rgb}{0.2, 0.47, 0.15}
\definecolor{moonstoneblue}{rgb}{0.45, 0.66, 0.76}
\definecolor{mypink}{rgb}{1.0, 0.75, 0.80}
\definecolor{myorange}{rgb}{1.0, 0.65, 0.0}
\definecolor{mykhaki}{rgb}{0.76, 0.69, 0.57}
\definecolor{light_cornflower_blue_2}{rgb}{0.64,0.76,0.96}
\definecolor{light_yellow_2}{rgb}{1.0, 0.9, 0.6}
\definecolor{light_green_2}{rgb}{0.71, 0.84, 0.66}
\definecolor{light_green_1}{rgb}{0.57, 0.76, 0.49}
\definecolor{dark_green_1}{rgb}{0.42, 0.66, 0.31}

\usepackage{float}
\floatstyle{plain}
\newfloat{prompt}{thp}{lop}
\floatname{prompt}{Prompt}

\usepackage[frozencache,cachedir=_minted-main]{minted}
\usepackage[most]{tcolorbox}

\usepackage{etoolbox}
\AtBeginEnvironment{tcolorbox}{\footnotesize}

\usepackage[verbose=true,letterpaper]{geometry}
\AtBeginDocument{
  \newgeometry{
    textheight=9in,
    textwidth=5.5in,
    top=1in,
    headheight=12pt,
    headsep=25pt,
    footskip=30pt
  }
}

\usepackage[nottoc,notlot,notlof]{tocbibind}

%% file: contents/title_page.tex
\definecolor{ttic_blue}{RGB}{55,93,137}
\begin{tikzpicture}[overlay,remember picture]
    \draw [line width=1mm] [ttic_blue]
        ($ (current page.north west) + (0.9in,-0.9in) $)
        rectangle
        ($ (current page.south east) + (-0.9in,0.9in) $);
\end{tikzpicture}
\begin{center}
\Large \MakeUppercase{{\bf Embodied Spatial Intelligence: From implicit scene modeling to spatial reasoning}}\\
\normalsize
\vspace{0.25in}
by\\Jiading Fang\\
\vspace{1in}
A thesis submitted\\
in partial fulfillment of the requirements for\\
the degree of\\
\vspace{0.3in}
Doctor of Philosophy in Computer Science\\
\vspace{0.3in}
at the\\
\vspace{0.3in}
\MakeUppercase{Toyota Technological Institute at Chicago}\\
Chicago, Illinois\\
\vspace{0.3in}
August, 2024\\
\vspace{1.0in}
Thesis Committee:\\
\vspace{0.1in}
Matthew R.\ Walter (Thesis Advisor)\\
Gregory Shakhnarovich\\
Adrien Gaidon\\
Mac Schwager\\
\end{center}
\thispagestyle{empty}
\newpage

%% file: contents/abstract.tex
\chapter*{Abstract}
\addcontentsline{toc}{chapter}{Abstract}

The central goal of robotics is to build autonomous agents that can serve alongside humans in the real, three-dimensional world. To be effective, these robots must observe their environment and act to change it according to human instructions, ideally delivered in natural language. This thesis defines the synthesis of these abilities—perceiving and acting within human-inhabited spaces through natural interaction—as "Embodied Spatial Intelligence." While recent advances in Large Language Models (LLMs) have brought the prospect of conversational robots closer to reality, significant challenges remain. Specifically, this work addresses two fundamental questions: how to build the right scene representation for environmental understanding, and how to create the right task representation for planning and action.

To develop more efficient and scalable representations for robot perception, Chapter \ref{chap:scene} presents contributions that leverage implicit neural modeling. This work enhances the robustness, generalization, and scalability of scene representations through three primary contributions. Section \ref{sec:self-calib} introduces a self-supervised learning method for camera self-calibration, providing robustness against disturbances to camera parameters. Section \ref{sec:define} details a novel approach for constructing continuous implicit depth fields that achieve state-of-the-art accuracy and extrapolation, enabled by new 3D data augmentation techniques. Finally, Section \ref{sec:nerfuser} demonstrates how to scale typically small-scale implicit representations to the level of buildings and city blocks using novel registration and blending methods.

To create better task representations that enable robots to plan and act upon observed spatial concepts, Chapter \ref{chap:reason} investigates and facilitates the use of LLMs for spatial reasoning tasks. This chapter explores three key areas. Section \ref{sec:mango} proposes a novel benchmark, generated from text-based games, to evaluate the mapping and navigation capabilities of LLMs. Section \ref{sec:transcrib3d} presents a method that leverages LLMs to understand 3D spatial relationships from natural language referring expressions, using transcribed 3D scene descriptions. Lastly, Section \ref{sec:statler} introduces a technique to improve long-horizon, LLM-based decision-making by incorporating a "state" summary of action history, generated by a secondary LLM. Collectively, these contributions advance our understanding of how robots can comprehend spatial relationships, while also identifying key challenges and proposing effective solutions.

%% file: contents/acknowledgement.tex
\chapter*{Acknowledgments}
\addcontentsline{toc}{chapter}{Acknowledgments}

The journey toward completing this PhD has been the most intellectually challenging yet rewarding period of my life. It would not have been possible without the support, guidance, and encouragement of many individuals, to whom I am deeply grateful.

First and foremost, I would like to express my heartfelt thanks to my advisor, Professor Matthew Walter. You provided me with the invaluable opportunity to explore my academic interests and nurtured my growth as a researcher. Your unwavering support and trust have been the foundation of my development, allowing me the freedom to pursue my curiosity and inspiring my creativity. Your positive outlook on life and dedication to fostering a collaborative and friendly lab environment have made my PhD journey not only fruitful but also enjoyable. Your commitment to robotics education and its societal impact has deeply influenced my perspective on our responsibilities as researchers. I consider myself incredibly fortunate to have had you as my advisor.

I am also profoundly grateful to Professor Greg Shakhnarovich for his insightful discussions and mentorship in the field of computer vision. Your critical feedback has continually pushed my work to new heights, and your support throughout this journey has been invaluable.

A special thank you goes to my lab mates, who have made the past six years truly memorable. I am particularly thankful to Igor Vasiljevic, whose passion for 3D vision ignited my own interests in the field, and who has been a constant collaborator and source of inspiration. My gratitude also extends to Shengjie Lin, whose partnership in numerous projects around 3D perception and reasoning has deepened my understanding of the field. I would also like to thank Falcon Dai, Chip Schaff, Andrea Daniele, and Takuma Yoneda for their collaborative efforts and support in making our shared projects possible.

I would also like to extend my sincere thanks to my collaborators at Toyota Research Institute (TRI), Vitor Guizilini, Rares Ambrus, and Adrien Gaidon. Your insights and the invaluable resources you provided were instrumental in the success of many of my projects. I deeply appreciate your contributions and the opportunity to work alongside you.

I am deeply appreciative of all the TTIC faculty and staff who have provided mentorship and support over the years. You have created a unique, safe, and collaborative environment at TTIC, one that feels like a second home, allowing us to do our best intellectual work.

Last but certainly not least, I want to express my deepest gratitude to my parents, Yunhong Huang and Fengjie Fang, for their unconditional love and unwavering support throughout my life. You are the backbone that has enabled me to pursue my dreams, and for that, I am forever grateful.

%% file: contents/toc.tex
\tableofcontents

\listoffigures

\listoftables

%% file: contents/introduction.tex
\chapter{Introduction}

Building intelligent robotic agents that are both mechanically capable and can understand their environments to operate reliably alongside humans is a long-standing goal of artificial intelligence. A system that achieves this capability may be described as possessing \emph{Embodied Spatial Intelligence}. With the recent advent of large language models (LLMs) and their vision-language derivatives (VLMs), human–robot interaction has advanced markedly. However, significant challenges remain in robust spatial understanding, a critical component for safe and effective robot operation~\cite{kamath2023whatsup, chen2024spatialvlm, cheng2024spatialrgpt}. Addressing this gap is a central debate in modern robotics, with proposed solutions ranging from further scaling of 2D vision systems to the deep integration of 3D sensory signals. This thesis contributes to this debate by examining the fundamental role of 3D in the era of modern machine learning.

Robotics is the discipline of creating embodied agents that can intelligently interact with and modify the physical world—a three-dimensional space. At its core, a robotic system learns a mapping from observations to actions that alter the state of that world. Let the observation space be $\mathcal{O}$ (e.g., images, proprioception, language) and let $\mathrm{T}_x\mathbb{R}^3$ denote the tangent space of $\mathbb{R}^3$ at a point $x$ (representing, for instance, instantaneous twists in $\mathrm{SE}(3)$ or velocity fields). The fundamental problem is to learn a mapping:
\begin{equation}
\mathcal{M}:\ \mathcal{O}\to \mathrm{T}_x\mathbb{R}^3.
\end{equation}
Because actions ultimately operate in 3D, effective robotic behavior necessitates 3D understanding. The key research question is therefore \emph{where} and \emph{how} such understanding should be embedded within the system.

This thesis decomposes the mapping $\mathcal{M}$ into two consecutive functions, $\mathcal{V}$ and $\mathcal{F}$:
\begin{equation}
\mathcal{M} = \mathcal{F} \circ \mathcal{V},
\end{equation}
where the first, the perception map $\mathcal{V}$, constructs a 3D representation in a target space $\mathcal{T}\subseteq \mathbb{R}^3$ from observations,
\begin{equation}
\mathcal{V}:\ \mathcal{O}\to \mathcal{T},
\end{equation}
and the second, the action map $\mathcal{F}$, leverages this representation to produce actions in an action space $\mathcal{A}$,
\begin{equation}
\mathcal{F}:\ \mathcal{T}\to \mathcal{A}\subseteq \mathrm{T}_x\mathbb{R}^3.
\end{equation}
This decomposition structures our investigation into two parts: first, building the 3D representation, and second, using it to act.

\section{From 2D to 3D: Building Implicit Scene Representations}
Historically, the robotics community has pursued 3D understanding through both hardware and software. On the hardware side, advances in 3D sensing (e.g., structured light, LiDAR, event cameras) have made direct 3D acquisition increasingly accessible. On the software side, strong geometric biases are often encoded into models, from point-based networks like PointNet~\cite{Qi2017PointNetDH} to volumetric and sensor-fusion methods~\cite{KinectFusion, Hornung2013OctoMap, Newcombe2011DTAM, Sun2018PWCNet}. In contrast, biological perception relies primarily on two 2D sensors (our eyes) and motion, from which depth is recovered via binocular and monocular cues.

Drawing inspiration from this biological model, this section investigates the perception map $\mathcal{V}$ for the case where observations are 2D images ($\mathcal{O}\subseteq\mathbb{R}^2$) and the target representation is 3D ($\mathcal{T}\subseteq\mathbb{R}^3$). We examine monocular depth estimation, the learnability of general 3D inductive biases, and the challenge of creating scalable scene representations.

\subsection{Monocular Depth Estimation: A Canonical 2D–3D Mapping}
A canonical instance of $\mathcal{V}$ is monocular depth estimation, where a single RGB image is mapped to a depth image. From the principles of projective geometry, this problem is ill-posed: a single 2D projection can correspond to an infinite family of 3D scenes. In practice, however, the problem is not intractable.

Evidence comes from individuals with monocular vision. Following the loss of stereopsis, the human visual system often adapts through neuroplasticity. While near-field depth perception is initially impaired, it is often recovered over time by learning to exploit monocular cues such as perspective, occlusion, and motion parallax. This natural existence proof motivates machine learning approaches to monocular depth.

The classical objection is that learning-based methods solve an ill-posed problem for which multi-view geometry provides a geometrically guaranteed solution. However, the objective is not to recover all geometrically possible scenes, but rather to learn the statistical regularities of scenes that typically occur in the physical world. This "typical" set occupies a lower-dimensional manifold within the space of all possible scenes—an observation that underlies many modern learning problems. This principle was powerfully demonstrated by self-supervised methods; for example, the foundational work of Monodepth2~\cite{monodepth2} showed that depth and camera pose could be learned jointly from monocular video sequences alone. Inspired by this self-supervised paradigm, we extended this approach to \emph{jointly} estimate camera parameters alongside monocular depth—an even more under-constrained problem that nonetheless proves effective in practice (Section~\ref{sec:self-calib}).

\subsection{Are General 3D Inductive Biases Learnable?}
While self-supervised monocular depth methods perform well, they are limited by photometric assumptions and the instability of co-training pose networks. Consequently, many 3D vision models have instead relied on strong, hand-engineered inductive biases (e.g., cost volumes). Although geometrically precise and data-efficient, these biases can limit robustness to real-world noise and scenarios where classical geometry is itself ill-posed (e.g., non-overlapping frames).

The recent availability of large, high-quality 3D datasets has made direct supervised training a viable alternative. This raises a fundamental question: are \emph{general} 3D inductive biases learnable from data? In other words, can a single architecture learn to perform diverse geometric tasks like depth estimation, pose estimation, and novel view synthesis from an arbitrary number of input views?

Our work on the Depth Field Network (DeFiNe)~\cite{guizilini2022depth} explores this question. Inspired by architectures that replace hard-coded geometric priors with learnable transformers~\cite{yifan2021input}, DeFiNe ingests image sequences and learns a latent scene embedding that can be decoded into multiple outputs, including depth, color, and novel views. A follow-up, DeLiRa~\cite{guizilini2023delira}, extended this paradigm to show that co-training on light-field data improves the quality of all decoded fields. Compared to models with strong inductive biases, DeFiNe exhibits superior zero-shot generalization across domains (Section~\ref{sec:define}).

This trend has recently culminated in models like VGGT~\cite{wang2025vggt}, which demonstrates that a largely vanilla transformer trained on massive 3D datasets can serve as a single backbone for recovering multiple geometric signals. This provides a concrete instance of the “bitter lesson” in the 3D domain: with sufficient scale, much of the useful inductive bias previously thought to require specialized architectures is, in fact, learnable.

\subsection{Scalable Scene Representations}
Robots must operate in environments that range from connected indoor spaces to unbounded outdoor areas. While explicit 3D representations like voxel grids can readily aggregate information into global maps, implicit neural representations are typically capacity-limited and lack established aggregation mechanisms. This limitation affects both generalizable methods like DeFiNe and optimization-based approaches like NeRF~\cite{Mildenhall2020NeRFRS}.

We address this scalability challenge with NeRFuser~\cite{fang2023nerfuser}, a framework that aligns partially overlapping, independently trained NeRFs and renders the joint space by blending their contributions. This method offers a practical path toward representing large-scale scenes without requiring a single monolithic model. As detailed in Section~\ref{sec:nerfuser}, the distributed nature of NeRFuser also enables applications like asynchronous, privacy-preserving map-building across a fleet of robots~\cite{asadi2024di}.

\section{Acting in 3D: Task-Specific Requirements}
The perception map $\mathcal{V}$ yields a 3D representation of the world. The action map $\mathcal{F}$ must translate this representation into physical action. The specific requirements for this map depend on the task. We categorize embodied tasks along two axes—response time and spatial range—which highlight different demands on the underlying 3D system:

\begin{itemize}
    \item \textbf{Short Response, Short Range} (e.g., manipulation). Requires precise, local 3D cues for high-frequency control.
    \item \textbf{Short Response, Long Range} (e.g., autonomous driving). Demands fast reaction to distant objects, often relying on monocular cues.
    \item \textbf{Long Response, Long Range} (e.g., indoor navigation). Favors persistent, global world models for long-horizon planning.
    \item \textbf{Long Response, Short Range} (e.g., multi-step task planning). Places high demands on memory, state tracking, and reliable spatial grounding.
\end{itemize}
These four regimes motivate the technical contributions of this thesis.

\paragraph*{Short Response, Short Range (Manipulation).} Here, “short range” refers to the workspace reachable by a manipulator, and “short response” denotes control frequencies above $\sim 10$\,Hz. Tasks include tabletop manipulation, where objects are often fragile and grippers may lack tactile sensing, necessitating precise, high-rate control. Such systems commonly employ stereo cameras, often augmented with depth sensors. We find that explicit 3D grounding improves reliability, particularly for object selection and disambiguation in clutter. Chapter~\ref{sec:transcrib3d} (Transcrib3D; \cite{fang2024transcrib3d}) shows that transcribing a local 3D scene into structured text enables tool-augmented reasoning for referring-expression resolution, a key precursor to manipulation.

\paragraph*{Short Response, Long Range (Autonomous Driving).} High-speed motion requires rapid reactions. At typical highway following distances, perception is dominated by monocular cues, even for binocular systems. This aligns with the observation that humans with monocular vision can judge long distances well~\cite{cutting1995perceiving}. Consequently, modern autonomous driving systems often prioritize 360$^\circ$ camera coverage over large stereo overlaps. Nonetheless, precise 3D remains crucial for centimeter-level localization, long-range obstacle avoidance, and robustness in adverse conditions.

\paragraph*{Long Response, Long Range (Navigation and Exploration).} Many applications involve long-horizon planning without stringent real-time constraints. In indoor navigation, low-speed motion and reliable local obstacle avoidance shift the primary challenge from control to planning. World representations can range from simple topological graphs to dense 3D maps, depending on task requirements. To probe whether language models can perform such tasks using textual history alone, we developed MANGO~\cite{ding2024mango}, a benchmark for navigation in text-based environments. We find that even capable LLMs perform poorly on tasks trivial for humans, suggesting a need for explicit mapping and reasoning capabilities (Section~\ref{sec:mango}).

\paragraph*{Long Response, Short Range (Multi-step Task Planning).} High-level, multi-stage tasks have traditionally been tackled with discrete optimization, often assuming fully observable state. In practice, specifications are often given in natural language and state estimation is uncertain. While LLM-based planners like SayCan~\cite{pmlr-v205-ichter23a} have demonstrated impressive capabilities, they are prone to hallucination over long horizons. To address this, we propose Statler~\cite{yoneda2023statler}, a dual-LLM architecture in which one agent maintains an explicit world state while another proposes short-term plans based on that state (Chapter~\ref{sec:statler}).

\section{Thesis Scope and Roadmap}
This thesis develops methods and systems along two primary axes:
\begin{enumerate}
    \item \textbf{Implicit, robust, and scalable 3D scene representations} that improve generalization across domains and robustness to calibration error.
    \item \textbf{Hybrid systems for persistent spatial reasoning} that reveal the limitations of current LLMs and introduce new architectures for maintaining state over long horizons.
\end{enumerate}

Chapter~\ref{sec:self-calib} presents implicit camera self-calibration via self-supervised learning~\cite{fang2022selfcalib}. Chapter~\ref{sec:define} introduces DeFiNe, a continuous depth field designed for generalization~\cite{guizilini2022depth}. Chapter~\ref{sec:nerfuser} addresses the scaling of implicit representations via NeRF registration and blending~\cite{fang2023nerfuser}.

On the reasoning axis, Chapter~\ref{sec:mango} benchmarks the mapping and navigation abilities of LLMs, highlighting significant gaps relative to human performance~\cite{ding2024mango}. Chapter~\ref{sec:transcrib3d} proposes Transcrib3D, which translates 3D scenes into text to enable tool-augmented, iterative reasoning~\cite{fang2024transcrib3d}. Chapter~\ref{sec:statler} introduces Statler, a dual-LLM system that maintains explicit world state for robust, long-horizon task execution~\cite{yoneda2023statler}.

Together, these contributions support three central claims: (a) robust 3D understanding is essential for embodied agents; (b) many of the inductive biases required for this understanding are learnable at scale; and (c) a practical path toward embodied spatial intelligence lies in strategically integrating high-quality 3D structure with large-scale 2D foundation models. (A broader discussion of this synthesis appears in the Conclusion.)

%% file: contents/chap_scene.tex
\chapter{Robotic Scene Representations from Implicit Modeling}\label{chap:scene}

Ever since the introduction from Euclid, people has been using geometry to describe the world we see. However, real world scenes are usually too complex to be regular, which means analytical solutions are almost impossible to find. To be able to describe the real world, people started to use sampling methods to create discrete representations, or \textit{explicit} scene representations. Typical examples include pointcloud and meshes as discrete sets of points and (points and edges), which are also widely used in robotic applications \cite{schoenberger2016sfm, schoenberger2016mvs, mur2015orb}.

More recently, the maturity of neural network training has made it possible to directly model the physical scene as a continuous function over the space, or \textit{implicit} neural representations. Typical examples include NeRF \cite{mildenhall2020nerf} or Neural-SDF \cite{sitzmann2020implicit}. The continuous nature of implicit neural representations forgoes the limitation of resolution, which not only promises better accuracy, but more importantly, enables data-driven training approaches. A good example is the application of self-supervised learning to monocular depth estimation creating orders of magnitude more data for training to achieve SOTA results \cite{monodepth2}. 

This advancement also greatly benefit the area of robotics. Scene representations are products of perception systems, and are prerequisites to spatially intelligent robotic agents. Their jobs are to form structured information from sequences or sets of 2D camera captures to represent the 3D scene, geometrically and semantically. Unlike more traditional methods that work in settings with ideal conditions, scene representations that built for robotic systems have extra sets of requirements to work for complicated conditions that could happen in real-world applications. Such complexity is hard to model with explicit geometric constraints, but easier with implicit modeling with data-driven approaches.

For example, the camera parameters could drift from its calibrated values during driving, thus we need a system that is \textit{robust} to such perturbations; When executing tasks, robots can be asked to travel long distance or over large scale scenes, thus it requires the scene representation to be \textit{scalable}; Robots are expected to work in a variety of environments, thus the learning system for building the scene representation should be \textit{generalizable}, meaning that the performance remain high even when the test environments are different to the training data. In the following section, I will present my paper to addressed the "Robustness" \cite{fang2022selfcalib}, "Generalizability" \cite{guizilini2022depth}, "Scalability" \cite{fang2023nerfuser} problems respectively.

%% file: papers/selfcalib/sections/abstract.tex
\input{papers/selfcalib/figures/teaser}

Camera calibration is integral to robotics and computer vision algorithms that seek to infer geometric properties of the scene from visual input streams. 
In practice, calibration is a laborious procedure
requiring specialized data collection and careful tuning.
This process must be repeated whenever the parameters of the camera change, which can be a frequent occurrence for mobile robots and autonomous vehicles. 
In contrast, self-supervised depth and ego-motion estimation approaches can bypass explicit calibration by inferring per-frame projection models that optimize a view-synthesis objective.
In this paper, we extend this approach to explicitly calibrate a wide range of cameras from raw videos in the wild. We propose a learning algorithm to regress  per-sequence calibration parameters using an efficient family of general camera models.
Our procedure achieves self-calibration results with sub-pixel reprojection error, outperforming other learning-based methods.  We validate our approach on a wide variety of camera geometries, including perspective, fisheye, and catadioptric.  Finally, we show that our approach leads to improvements in the downstream task of depth estimation, achieving state-of-the-art results on the EuRoC dataset with greater computational efficiency than contemporary methods.

%% file: papers/selfcalib/figures/teaser.tex
\begin{figure}[H]
\centering
\subfloat[Input]{
\includegraphics[width=0.32\textwidth,height=1.8cm]{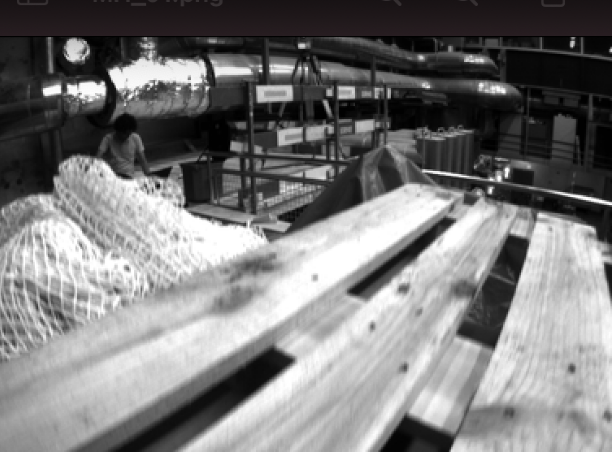}}%
\subfloat[Predicted depth]{
\includegraphics[width=0.32\textwidth,height=1.8cm]{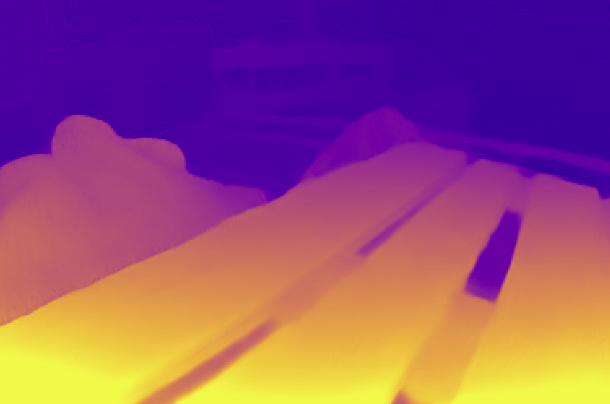}}%
\subfloat[Rectified image]{
\includegraphics[width=0.32\textwidth,height=1.8cm]{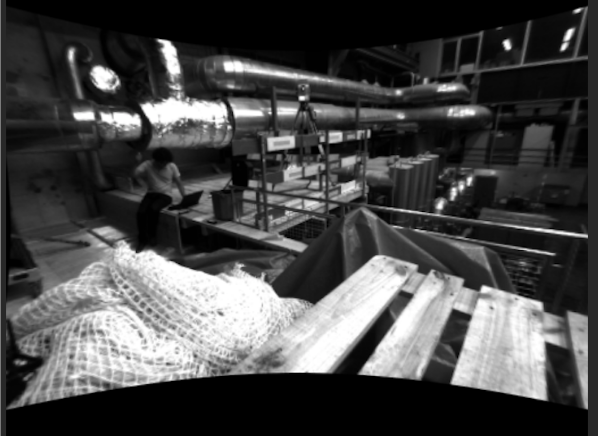}}%
\\
\subfloat[Camera parameter re-calibration]{\includegraphics[width=0.8\textwidth]{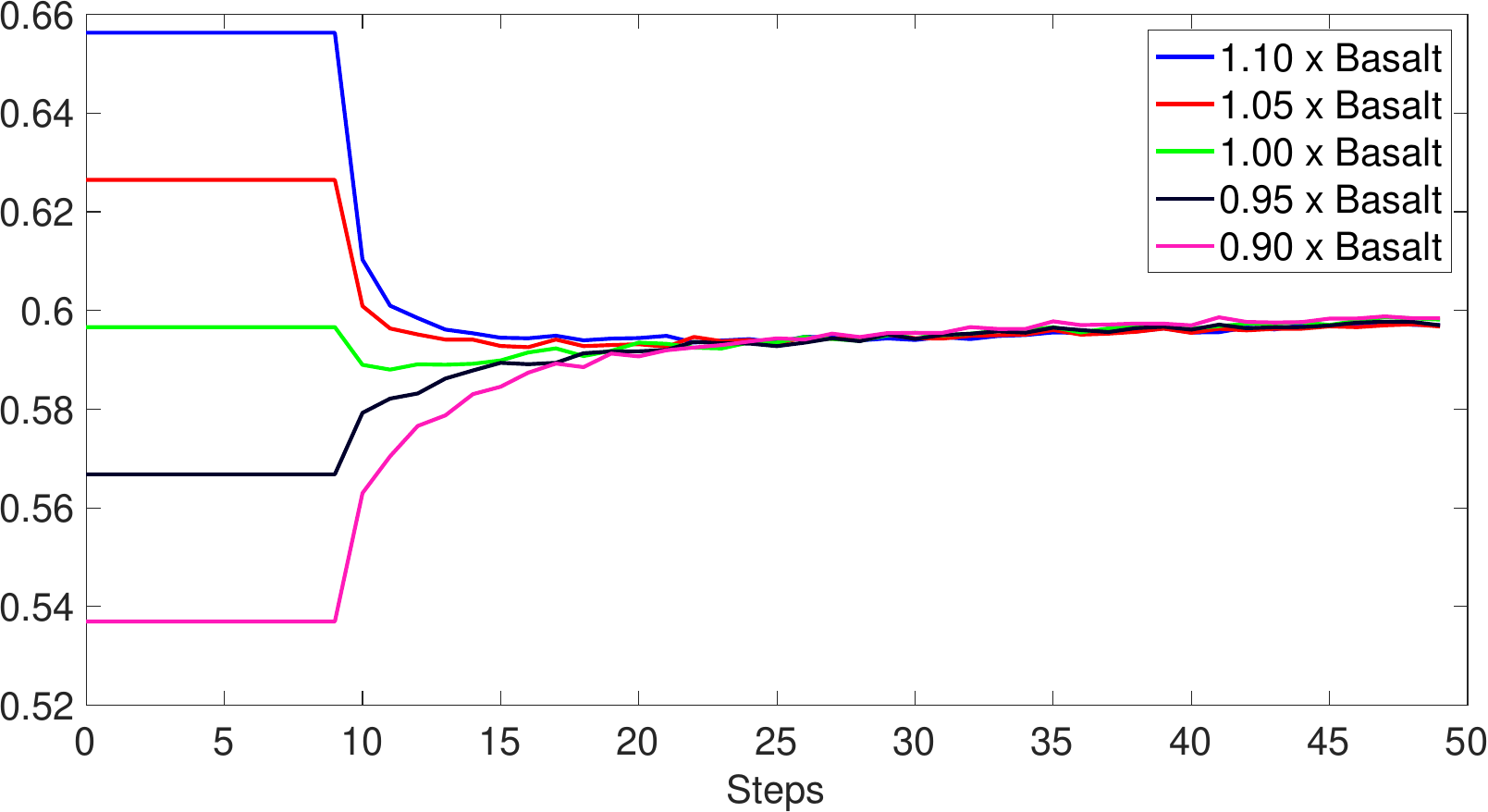}}

\caption{\textbf{Our self-supervised self-calibration procedure} can recover accurate parameters for a wide range of cameras using a structure-from-motion objective on raw videos (EuRoC dataset, top), enabling on-the-fly re-calibration and robustness to intrinsics perturbation (bottom).}
\label{fig:teaser}
\end{figure}

%% file: papers/selfcalib/sections/introduction.tex
Cameras provide rich information about the scene, while being small, lightweight, inexpensive, and power efficient. Despite their wide availability, camera calibration largely remains a manual, time-consuming process that typically requires collecting images of known targets (e.g., checkerboards) as they are deliberately moved in the scene~\cite{zhang2000flexible}. While applicable to a wide range of camera models~\cite{scaramuzza2006flexible,kannala2006generic,grossberg2001general}, this process is tedious and has to be repeated whenever the camera parameters change. A number of methods perform calibration ``in the wild''~\cite{caprile1990using, pollefeys1997stratified, cipolla1999camera}. However, they rely on strong assumptions about the scene structure, which cannot be met during deployment in unstructured environments. Learning-based methods relax these assumptions, and regress camera parameters directly from images, either by using labelled data for supervision~\cite{bogdan2018deepcalib} or by extending the framework of self-supervised depth and ego-motion estimation~\cite{garg2016unsupervised, zhou2017unsupervised} to also learn per-frame camera parameters~\cite{gordon2019depth, vasiljevic2020neural}.

While these methods enable learning accurate depth and ego-motion without calibration, they are either over-parameterized~\cite{vasiljevic2020neural} or limited to near-pinhole cameras~\cite{gordon2019depth}. In contrast, we propose a self-supervised camera calibration algorithm capable of learning expressive models of different camera geometries in a computationally efficient manner. In particular, our approach adopts a family of general camera models~\cite{usenko2018double} that scales to higher resolutions than previously possible, while still being able to model highly complex geometries such as catadioptric lenses. Furthermore, our framework learns camera parameters per-sequence rather than per-frame, resulting in self-calibrations that are more accurate and more stable than those achieved using contemporary learning methods. We evaluate the reprojection error of our approach compared to conventional target-based calibration routines, showing comparable sub-pixel performance despite only using raw videos at training time.

Our contributions can be summarized as follows:
\begin{itemize}
\item We propose to self-calibrate a variety of generic camera models from raw video using self-supervised depth and pose learning as a proxy objective, providing for the first time a calibration evaluation of camera model parameters \textbf{learned purely from self-supervision}.
\item We demonstrate the utility of our framework on challenging and radically different datasets, learning depth and pose on perspective, fisheye, and catadioptric images without architectural changes.
\item We achieve \textbf{state-of-the-art depth evaluation results on the challenging EuRoC MAV dataset} by a large margin, using our proposed self-calibration framework. 
\end{itemize}

%% file: papers/selfcalib/sections/related.tex
\textbf{Camera Calibration.}
Traditional calibration for a variety of camera models uses targets such as checkerboards or AprilTags to generate 2D-3D correspondences, which are then used in a bundle adjustment framework to recover relative poses as well as intrinsics~\cite{zhang2000flexible, hartley2003multiple}. Targetless methods typically make strong assumptions about the scene, such as the existence of vanishing points and known (Manhattan world) scene structure~\cite{caprile1990using, pollefeys1997stratified, cipolla1999camera}.  
While highly accurate, these techniques require a controlled setting and manual target image capture to re-calibrate. Several models are implemented in OpenCV~\cite{bradski2000opencv}, kalibr~\cite{rehder2016extending}.
These methods require specialized settings to work, limiting their generalizability. %

\textbf{Camera Models.}
The pinhole camera model is ubiquitous in robotics and computer vision~\cite{leonard08,urmson2008autonomous} and is especially common in recent deep learning architectures for depth estimation~\cite{zhou2017unsupervised}.  There are two main families of models for high-distortion cameras. The first is the ``high-order polynomial'' distortion family that includes pinhole radial distortion~\cite{fryer1986lens}, omnidirectional~\cite{scaramuzza2006flexible}, and Kannala-Brandt~\cite{kannala2006generic}. The second is the ``unified camera model'' family that includes the Unified Camera Model (UCM)~\cite{geyer2000unifying}, Extended Unified Camera Model (EUCM)~\cite{khomutenko2015enhanced},
and Double Sphere Camera Model (DS)~\cite{usenko2018double}. Both families are able to achieve low reprojection errors for a variety of different camera geometries~\cite{usenko2018double}, however the unprojection operation of the ``high-order polynomial'' models requires solving for the root of a high-order polynomial, typically using iterative optimization, which is a computationally expensive operation. Further, the process of calculating gradients for these models is non-trivial.
In contrast, the ``unified camera model'' family has an easily computed, closed-form unprojection function. While our framework is applicable to high-order polynomial models, we choose to focus on the unified camera model family in this paper.

\textbf{Learning Camera Calibration.}
Work in learning-based camera calibration can be divided into two types: \emph{supervised} approaches that leverage ground-truth calibration parameters or synthetic data to train single-image calibration regressors; and \emph{self-supervised} methods that utilize only image sequences. Our proposed method falls in the latter category, and aims to self-calibrate a camera system using only image sequences.
Early work on applying CNNs to camera calibration focused on regressing the focal length~\cite{workman2015deepfocal} or horizon lines~\cite{workman2016horizon}; synthetic data was used for distortion calibration~\cite{rong2016radial}  and fisheye rectification~\cite{yin2018fisheyerecnet}.  Using panorama data to generate images with a wide variety of intrinsics, \citet{lopez2019deep} are able to estimate both extrinsics (tilt and roll) and intrinsics (focal length and radial distortion).  DeepCalib~\cite{bogdan2018deepcalib} takes a similar approach:  given a panoramic dataset, generate projections with different focal lengths. Then, they train a CNN to regress from a set of synthetic images $I$ to their (known) focal lengths $f$. Typically, training images are generated by taking crops of the desired focal lengths from $360$ degree panoramas~\cite{hold2018perceptual, zhu2020single}. While this can be done for any kind of image, and does not require image sequences, it does require access to panoramic images. Furthermore, the warped ``synthetic'' images are not the true 3D-2D projections. This approach has been extended to pan-tilt-zoom~\cite{zhang2020deepptz} and fisheye~\cite{yin2018fisheyerecnet} cameras.
Methods also exist for specialized problems like undistorting portraits~\cite{zhao2019learning}, monocular 3D reconstruction~\cite{yin2021learning}, and rectification~\cite{yang2021progressively, liao2021deep}.

After the publication of this paper, ~\cite{hagemann2023deep} extends our method, and improves the performance by adding a differentiable bundle-adjustment layer. On the other hand, leveraging the same principle, differentiable camera models have also been extended to use in extrinsic calibration~\cite{kanai2023robust}.

\textbf{Self-Supervised depth and ego-motion}. Self-supervised learning has also been used to learn camera parameters from geometric priors.  Gordon et al.~\cite{gordon2019depth} learn a pinhole and radial distortion model, while Vasiljevic et al.~\cite{vasiljevic2020neural} learn a generalized central camera model applicable to a wider range of camera types, including catadioptric. These methods both learn calibration on a per-frame basis, and do not offer a calibration evaluation of their learned camera model.  Furthermore, while \citet{vasiljevic2020neural} is much more general than \citet{gordon2019depth}, it is limited to fairly low resolutions by the complex and approximate generalized projection operation. In our work, we trade some degree of generality (i.e., a global, central vs.\ per-pixel model) for a closed-form and efficient projection operation and ease of calibration evaluation.

%% file: papers/selfcalib/sections/methodology.tex
First, we describe the self-supervised monocular depth learning framework that we use as proxy for self-calibration. Then we describe the family of unified camera models we consider and how we learn their parameters end-to-end. 

\begin{figure}[!t]
    \centering
    \includegraphics[width=\linewidth]{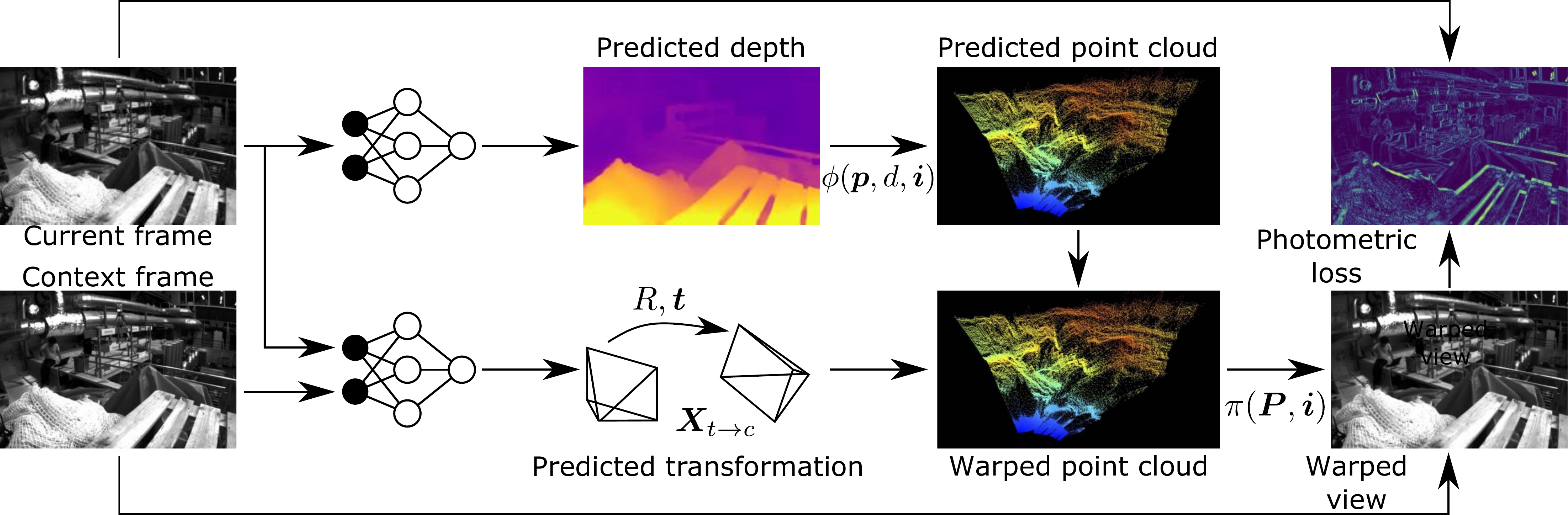}
    \caption{\textbf{Our self-supervised self-calibration architecture.}  We use gradients from the photometric loss to update the parameters of a unified camera model (Fig.\ \ref{fig:ucm_figure}).
    }
    \label{fig:ssl}
\vspace*{-2mm}
\end{figure}

\paragraph{Self-Supervised Monocular Depth Estimation}
\label{subsec:monodepth}
Self-supervised depth and ego-motion architectures consist of a depth network that produces depth maps $\hat{D}_{t}$ for a target image $I_t$, as well as a pose network that predicts the relative rigid-body transformation between target $t$ and context $c$ frames, $\hat{\bm{X}}^{t \to c} = \begin{psmallmatrix}{\hat{\bm{R}}^{t\to c}} & {\hat{\bm{t}}^{t\to c}}\\ \bm{0} & \bm{1}\end{psmallmatrix} \in \text{SE(3)}$.
We train the networks jointly by minimizing the photometric reprojection error between the actual target image $I_t$ and a synthesized image $\hat{I}_t$ generated by projecting pixels from the context image $I_c$ (usually preceding or following $I_t$ in a sequence) onto the target image $I_t$ using the predicted depth map $\hat{D}_{t}$ and ego-motion $\hat{\bm{X}}^{t \to c}$~\cite{zhou2017unsupervised}. See Figure~\ref{fig:ssl} for an overview.
The general pixel-warping operation is defined as:
\begin{equation}%
    \hat{\bm{p}}^t = \pi \left({\hat{\bm{R}}}^{t \rightarrow c} \phi (\bm{p}^t, \hat{d}^t, \bm{i}) + \bm{\hat{t}}^{t \rightarrow c}, \bm{i}\right),
\end{equation}\label{eq:warp_mono}%
where $\bm{i}$ are camera intrinsic parameters modeling the geometry of the camera, which is required for both projection of 3D points $\bm{P}$ onto image pixels $\bm{p}$ via $\pi(\bm{P},\bm{i}) = \bm{p}$ and unprojection via $\phi(\bm{p}, \hat{d},\bm{i}) = \bm{P}$ assuming an estimated pixel depth of $\hat{d}$.
The camera parameters $\bm{i}$ are generally the standard pinhole model~\cite{hartley2003multiple} defined by the $3 \times 3$ intrinsic matrix $\bm{K}$, but can include any differentiable model such as the Unified Camera Model family~\cite{usenko2018double} as described next.

\begin{figure}[!th]
    \centering
    \includegraphics[width=0.6\linewidth]{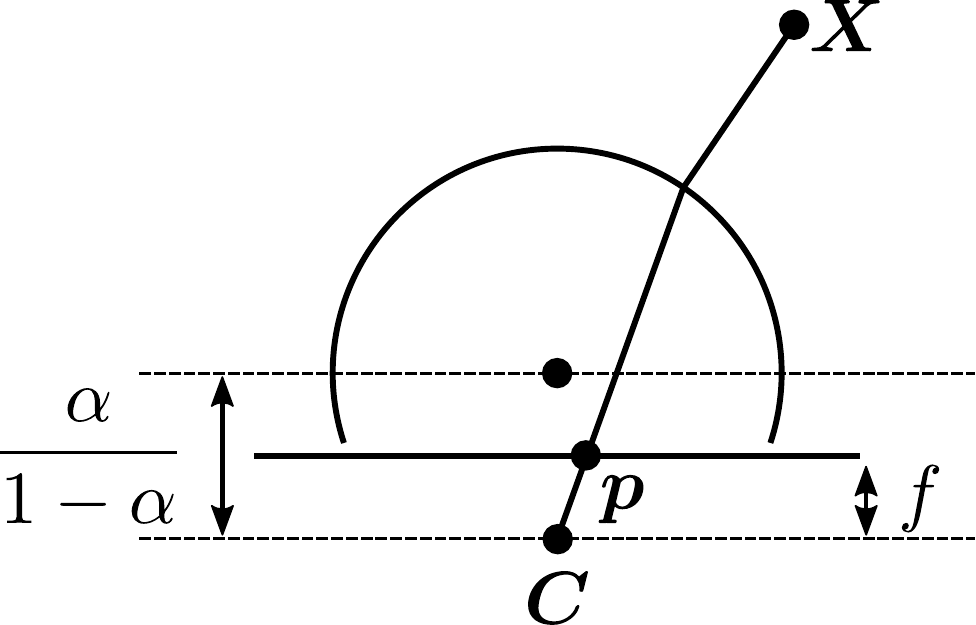}
    \caption{\textbf{The Unified Camera Model~\cite{usenko2018double} used in our self-calibration pipeline}. Points are projected onto a unit sphere before being projected onto an image plane of a standard pinhole camera offset by $\frac{\alpha}{1-\alpha}$ from the sphere center.}
    \label{fig:ucm_figure}
\vspace*{-2mm}
\end{figure}

\paragraph{End-to-End Self-Calibration}
\label{subsec:ucm}
UCM~\cite{geyer2000unifying} is a parametric global central camera model that uses only five parameters to represent a diverse set of camera geometries, including perspective, fisheye, and catadioptric. A 3D point is projected onto a unit sphere and then projected onto the image plane of a pinhole camera, shifted by $\frac{\alpha}{1-\alpha}$ from the center of the sphere (Fig.~\ref{fig:ucm_figure}). EUCM and DS are two extensions of the UCM model. EUCM replaces the unit sphere with an ellipse as the first projection surface, and DS replaces the one unit sphere with two unit spheres in the projection process.  We self-calibrate all three models (in addition to a pinhole baseline) in our experiments. For brevity, we only describe the original UCM and refer the reader to \citet{usenko2018double} for details on the EUCM and DS models.

There are multiple parameterizations for UCM \cite{geyer2000unifying}, and we use the one from~\citet{usenko2018double} since it has better numerical properties.  UCM extends the pinhole camera model $(f_x, f_y, c_x, c_y)$ with only one additional parameter $\alpha$. The 3D-to-2D projection of $\bm{P}=(x,y,z)$  is defined as

\begin{equation} \label{eq:ucm_proj}
    \bm{\pi}(\bm{P}, \bm{i}) = \begin{bmatrix}
f_x \frac{x}{\alpha d + (1-\alpha)z} 
\\ 
f_y \frac{y}{\alpha d + (1-\alpha)z} 
\end{bmatrix} +
\begin{bmatrix}
c_x \\
c_y
\end{bmatrix}
\end{equation}
where the camera parameters are $\bm{i} = (f_x, f_y, c_x, c_y, \alpha)$ and $d=\sqrt{x^2+y^2+z^2}$

The unprojection operation of pixel $\bm{p} = (u,v,1)$ at estimated depth $\hat{d}$ is:
\begin{equation}\label{eq:ucm_unproj}
    \phi(\bm{p}, \hat{d},\bm{i}) = \hat{d} \frac{\xi + \sqrt{1 + (1-\xi^2)r^2}}{1 + r^2}\begin{bmatrix} m_x \\ m_y \\ 1\end{bmatrix} - \begin{bmatrix} 0 \\ 0 \\ \hat{d} \zeta \end{bmatrix}
\end{equation}

where
\begin{subequations}
\begin{align}
    m_x &= \frac{u - c_x}{f_x}(1- \alpha)&
    m_y &= \frac{v - c_y}{f_y}(1- \alpha)&\\
    r^2 &= m_{x}^2 + m_{x}^2&
    \zeta &= \frac{\alpha}{1-\alpha}&
\end{align}
\end{subequations}

As shown in Equations~\ref{eq:ucm_proj} and \ref{eq:ucm_unproj}, the UCM camera model provides  closed-form projection and unprojection functions that are both differentiable.
Therefore, the overall architecture is end-to-end differentiable with respect to both neural network parameters (for pose and depth estimation) and camera parameters. This enables learning self-calibration end-to-end from the aforementioned view synthesis objective alone.
At the start of self-supervised depth and pose training, rather than pre-calibrating the camera parameters, we initialize the camera with ``default'' values based on image shape only (for a detailed discussion of the initialization procedure, please see Section~\ref{sec:perturbation_test}). 
Although the projection \eqref{eq:ucm_proj} and unprojection \eqref{eq:ucm_unproj} are initially inaccurate, they quickly converge to highly accurate camera parameters with sub-pixel re-projection error (see Table~\ref{table:reproj_error}). 

As we show in our experiments, our method combines flexibility with computational efficiency. Indeed, our approach enables learning from heterogeneous datasets with potentially vastly differing sensors for which separate parameters $\bm{i}$ are learned. As most of the parameters (in the depth and pose networks) are shared thanks to the decoupling of the projection model, this enables scaling up in-the-wild training of depth and pose networks. Furthermore, our method is efficient, with only one extra parameter relative to the pinhole model. This enables learning depth for highly-distorted catadioptric cameras at a much higher resolution than previous over-parametrized models ($1024 \times 1024$ vs. $384 \times 384$ for~\citet{vasiljevic2020neural}). 
Note that, in contrast to prior works~\cite{gordon2019depth, vasiljevic2020neural}, we learn intrinsics per-sequence rather than per-frame.
This increases stability compared to per-frame methods that exhibit frame-to-frame variability~\cite{vasiljevic2020neural}, and can be used over sequences of varying sizes.

%% file: papers/selfcalib/sections/experiments.tex
In this section we describe two sets of experimental validations for our architecture: (i) calibration, where we find that the re-projection error of our learned camera parameters compares favorably to target-based traditional calibration toolboxes; and (ii) depth evaluation, where we achieve state-of-the-art results on the challenging EuRoC MAV dataset.

\paragraph{Datasets}

Self-supervised depth and ego-motion learning uses monocular sequences~\cite{zhou2017unsupervised, godard2019digging, gordon2019depth, packnet} or rectified stereo pairs~\cite{godard2019digging, superdepth} from forward-facing cameras~\cite{geiger2012we,packnet,caesar2020nuscenes}. Given that our goal is to learn camera calibration from raw videos in challenging settings, we use the standard KITTI dataset as a baseline, and focus on the more challenging and distorted EuRoC~\cite{burri2016euroc} fisheye sequences.

\noindent\textbf{{KITTI~\cite{geiger2012we}}}
We use this dataset to show that our self-calibration procedure is able to accurately recover pinhole intrinsics alongside depth and ego-motion. Following related work~\cite{zhou2017unsupervised, godard2019digging, gordon2019depth, packnet} we use the training protocol of~\cite{eigen2014depth}, including filtering static images as described by~\citet{zhou2017unsupervised}. The resulting training set contains of $39810$ images, with $697$ images left for evaluation. 

\noindent\textbf{{EuRoC~\cite{burri2016euroc}}} The dataset consists of a set of indoor MAV sequences with general six-DoF motion. Consistent with recent work~\cite{gordon2019depth}, we train using center-cropping and down-sample the images to a $384 \times 256$ resolution, while training and evaluating on the same split. For calibration evaluation, we follow~\citet{usenko2018double} and use the calibration sequences from the dataset. We evaluate the UCM, EUCM and DS camera models in terms of re-projection error.

\noindent\textbf{OmniCam~\cite{schonbein2014calibrating}} A challenging outdoor catadioptric sequence, containing 12000 frames captured by an autonomous car rig. As this dataset does not provide ground-truth depth information, we only provide qualitative results.

\paragraph{Training Protocal}
We implement the group of unified camera models described in ~\cite{usenko2018double} as differentiable PyTorch~\cite{paszke2017automatic} operations, modifying the self-supervised depth and pose architecture of~\citet{godard2019digging} to jointly learn depth, pose, and the unified camera model intrinsics. We use a learning rate of $2$e-$4$ for the depth and pose network and $1$e-$3$ for the camera parameters.  We use a StepLR scheduler with $\gamma=0.5$ and a step size of $30$. All of the experiments are run for $50$ epochs. The images are augmented with random vertical and horizontal flip, as well as color jittering. We train our models on a Titan X GPU with 12\,GB of memory, with a batch size of $16$ when training on images with a resolution of $384 \times 256$. We note that our method requires significantly less memory than that of~\citet{vasiljevic2020neural} which learns a generalized camera model parameterized through a per-pixel ray surface. 

\input{papers/selfcalib/tables/calib_table}

\paragraph{Camera Self-Calibration}

We evaluate the results of the proposed self-calibration method on the EuRoC dataset; detailed depth estimation evaluations are provided in Sec.~\ref{subsec:results_depth}. To our knowledge, ours is the first direct calibration evaluation of self-supervised intrinsics learning; although~\citet{gordon2019depth} compare \textit{ground-truth} calibration to their per-frame model, they do not evaluate the re-projection error for their learned parameters. 

Following~\citet{usenko19nfr}, we evaluate our self-supervised calibration method on the family of unified camera models: UCM, EUCM, and DS, as well as the perspective (pinhole) model. As a lower bound, we use the Basalt~\cite{usenko19nfr} toolbox and compute camera calibration parameters for each unified camera model using the calibration sequences of the EuRoC dataset. We note that unlike Basalt, our method regresses the intrinsic calibration parameters directly from raw videos, without using any of the calibration sequences. 

\input{papers/selfcalib/tables/euroc_calib}

\begin{figure}[!t]
  \centering
  \includegraphics[width=0.49\linewidth]{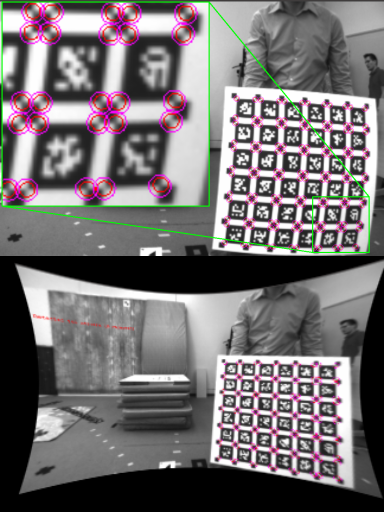}\hfil
  \includegraphics[width=0.49\linewidth]{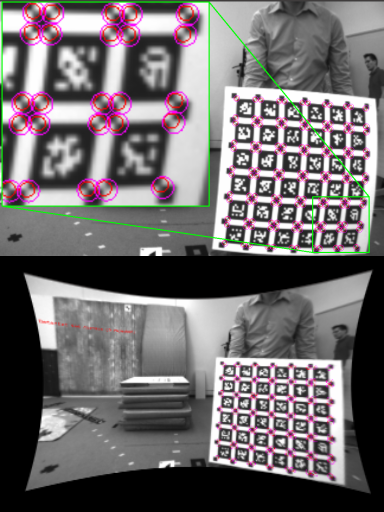}
\caption{\textbf{EuRoC rectification results} using images from the calibration sequences. Each column visualizes the results rendered using (left) the Basalt calibrated intrinsics and (right) our learned intrinsics. The top row shows that detected (small circles) and reprojected (big circles) corners are close using both calibration methods. The bottom row shows the same images after rectification.
}\label{fig:rectification}
\end{figure}
Table~\ref{table:reproj_error} summarizes our re-projection error results. We use the EuRoC AprilTag~\cite{olson2011apriltag} calibration sequences with Basalt to measure re-projection error using the full estimation procedure (Table~\ref{table:reproj_error} --- \textit{Target-based}) and learned intrinsics (Table~\ref{table:reproj_error} --- \textit{Learned}).  For consistency, we optimize for both intrinsics and camera poses for the baselines and only for the camera poses for the learned intrinsics evaluation. Note that with learned intrinsics, UCM, EUCM and DS models all achieve sub-pixel mean projection error despite the camera parameters having been learned from raw video data.

\input{papers/selfcalib/tables/perturbation_test}

\begin{figure}[!t]
  \centering
    \subfloat[$f_x$]{\includegraphics[width=0.475\textwidth]{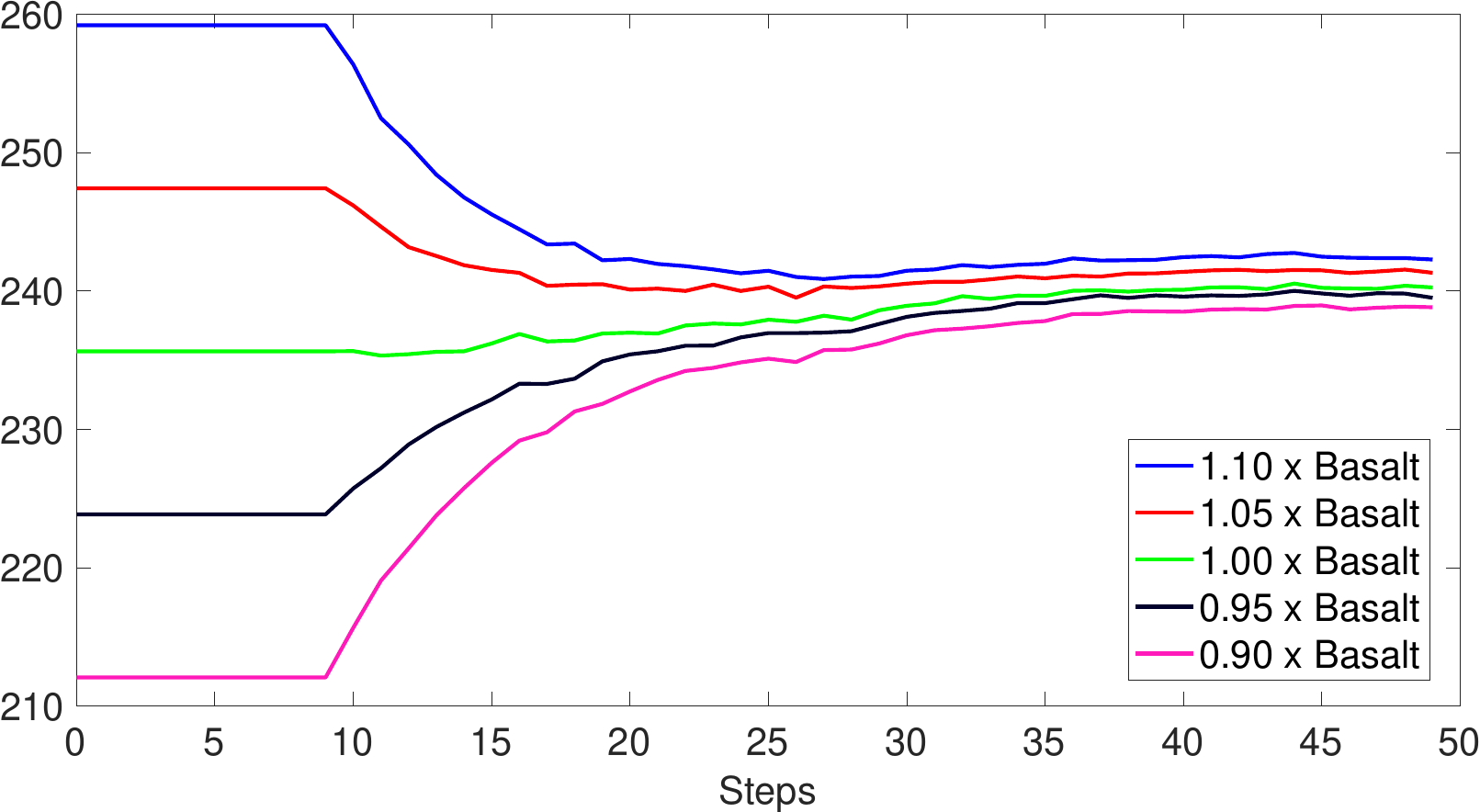}}\hfill
    \subfloat[$f_y$]{\includegraphics[width=0.475\textwidth]{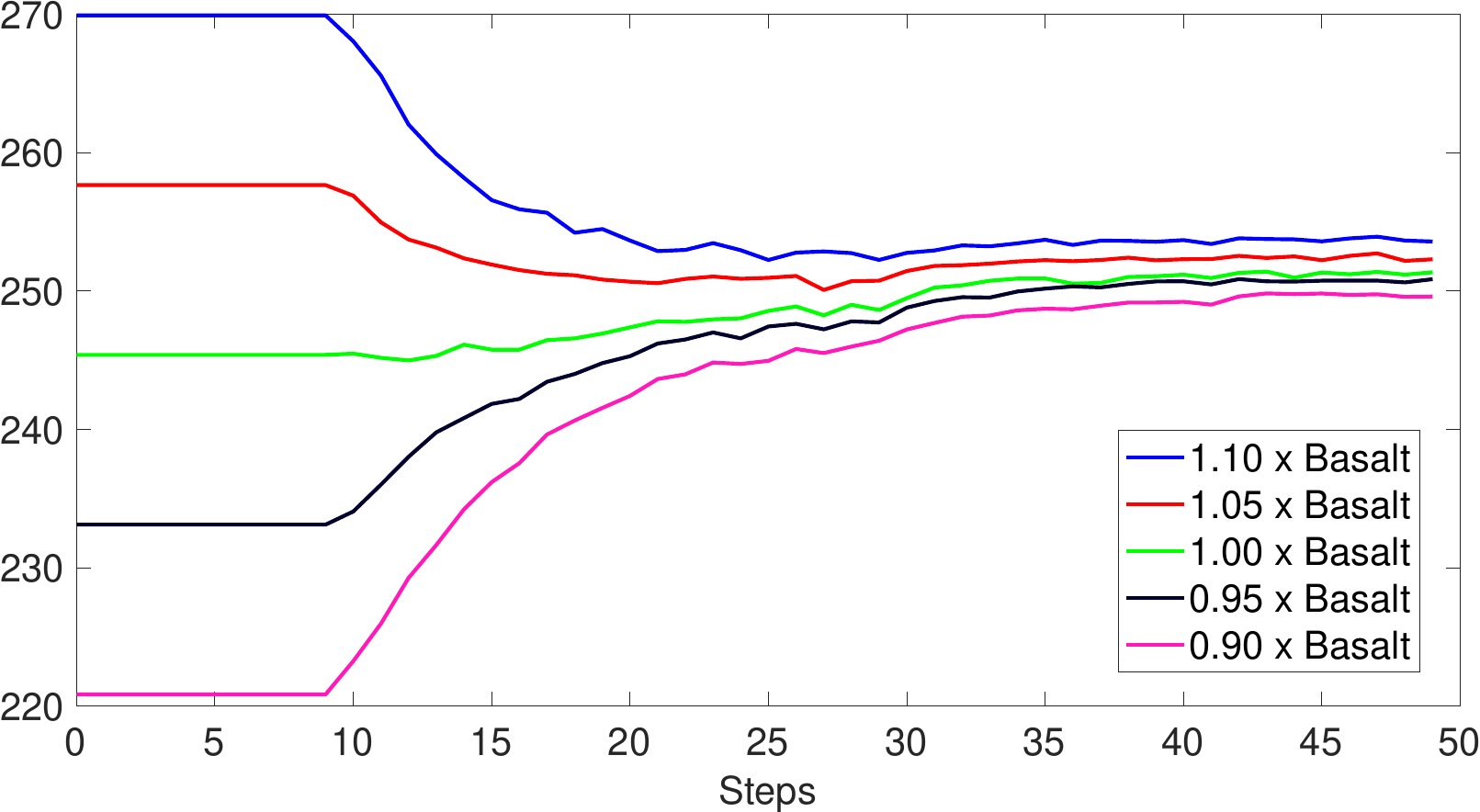}}\\
    \subfloat[$c_x$]{\includegraphics[width=0.475\textwidth]{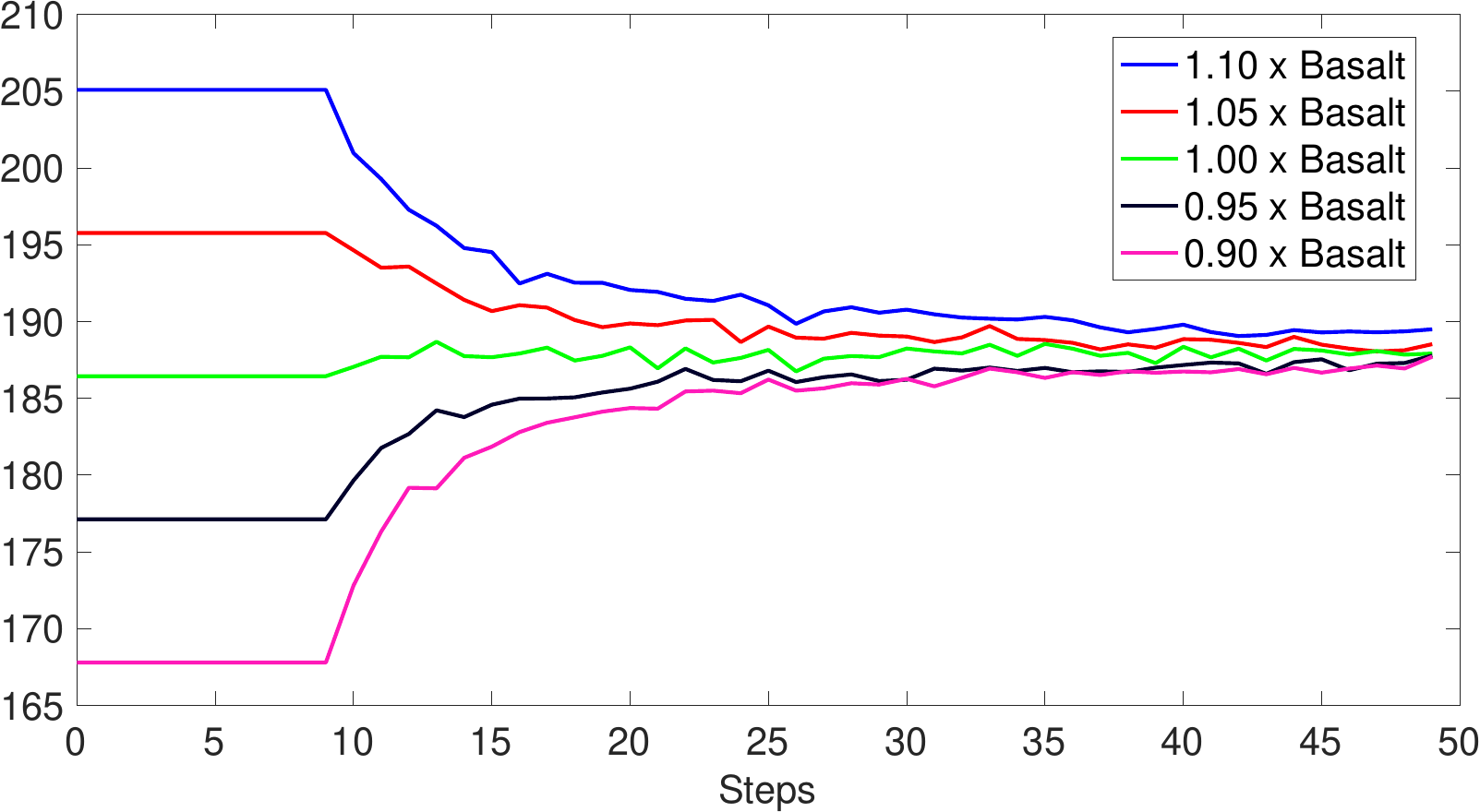}}\hfill
    \subfloat[$c_y$]{\includegraphics[width=0.475\textwidth]{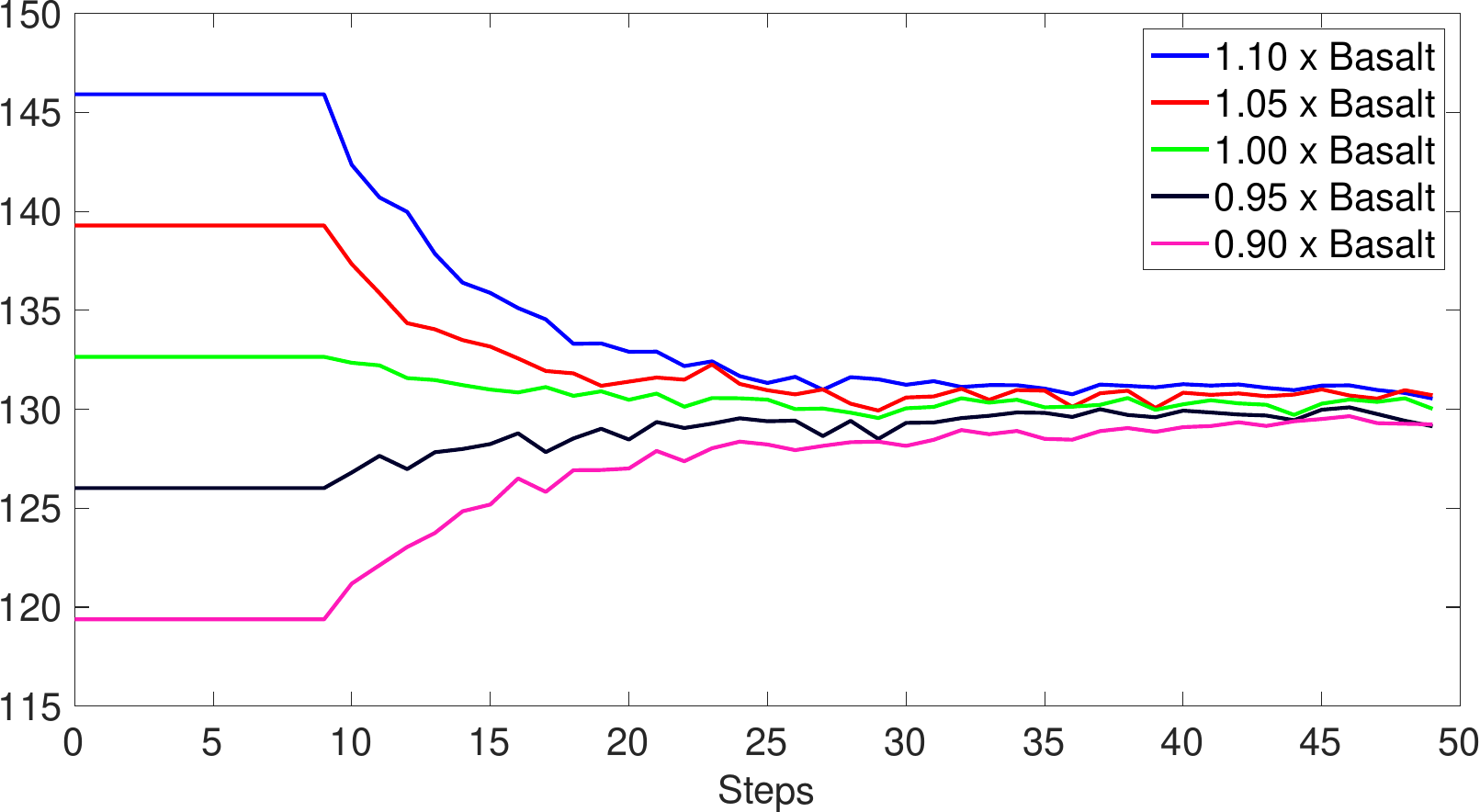}}\\
    \subfloat[$\alpha$]{\includegraphics[width=0.475\textwidth]{papers/selfcalib/pics/convergence/alpha}}\hfill
    \subfloat[$\beta$]{\includegraphics[width=0.475\textwidth]{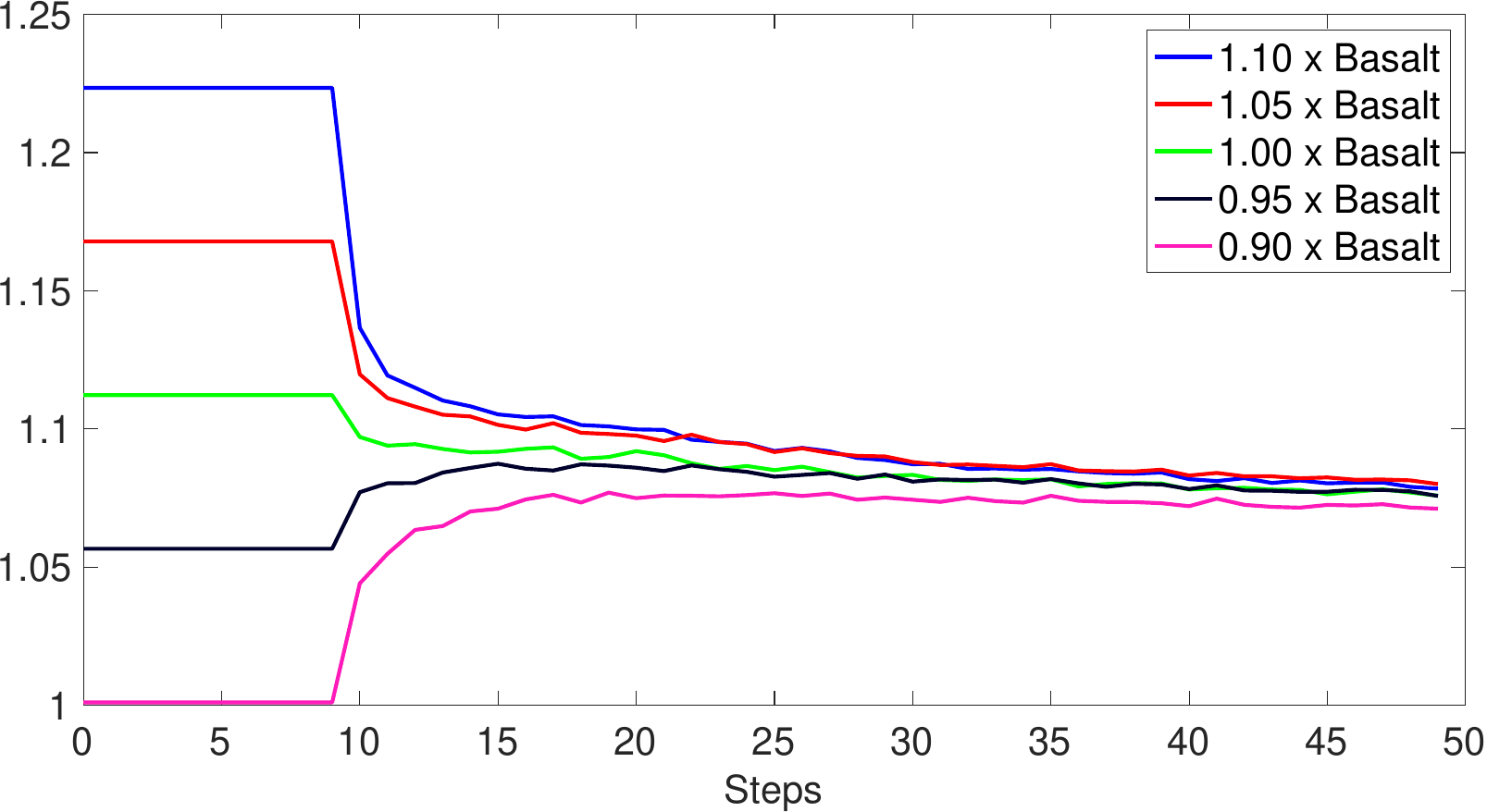}}
  \caption{\textbf{EuRoC perturbation test}, showing how our proposed learning-based method is able to recover from changes in camera parameters for online self-calibration.}

  \label{fig:perturbation}
  \vspace{-3mm}
\end{figure}

Table~\ref{table:intrinsic_numbers_compare} compares the target-based calibrated parameters to our learned parameters for different camera models trained on the \textit{cam0} sequences of the EuRoC dataset. Though the parameter vectors were initialized with no prior knowledge of the camera model and updated purely based on gradients from the reprojection error, they converge to values very close to the output of a procedure that uses bundle adjustment on calibrated image sequences.

\paragraph{Camera Rectification}

Using our learned camera parameters, we rectify calibration sequences on the EuRoC dataset to demonstrate the quality of the calibration.  EuRoC was captured with a fisheye camera and exhibits a high degree of radial distortion that causes the straight edges of the checkerboard grid to be curved.  In Figure~\ref{fig:rectification}, we see that our learned parameters allow for the rectified grid to track closely to the true underlying checkerboard.

\paragraph{Camera Re-calibration: Perturbation Experiments}
\label{sec:perturbation_test}
Thus far, we have assumed to have no prior knowledge of the camera calibration. In many real-world robotics settings, however, one may want to re-calibrate a camera based on a potentially incorrect prior calibration.  Generally, this requires the capture of new calibration data.  Instead, we can initialize our parameter vectors with this initial calibration (in this setting, a perturbation of Basalt calibration of the EUCM model) and see the extent to which self-supervision can nudge the parameters back to their ``true value''.   

Given Basalt parameters $I_c=[f_x, f_y, c_x, c_y, \alpha, \beta]$, we perturb them as $I_{1.1}=1.1\times I_c$, $I_{1.05}=1.05\times I_c$, $I_{0.95}=0.95\times I_c$, $I_{0.9}=0.9\times I_c$ and initialize the camera parameters at the beginning of training with these values. All runs have warm start, i.e., freezing the gradients for the intrinsics for the first $10$ epochs while we train the depth and pose networks.  As Figure~\ref{fig:perturbation} shows, our method converges to within $3\%$ of the Basalt estimate for each parameter.  Table~\ref{table:perturbation} provides the values of the converged parameters along with the mean projection error (MRE) for each experiment.

\input{papers/selfcalib/tables/kitti_depth}
\input{papers/selfcalib/tables/euroc_depth}

\begin{figure}[H]
  \centering
 \subfloat[EuRoC]{
  \includegraphics[width=0.32\linewidth]{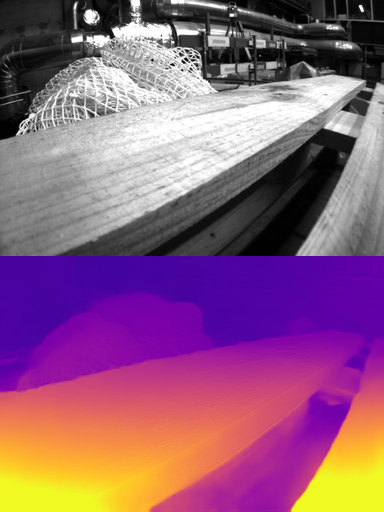}
  \includegraphics[width=0.32\linewidth]{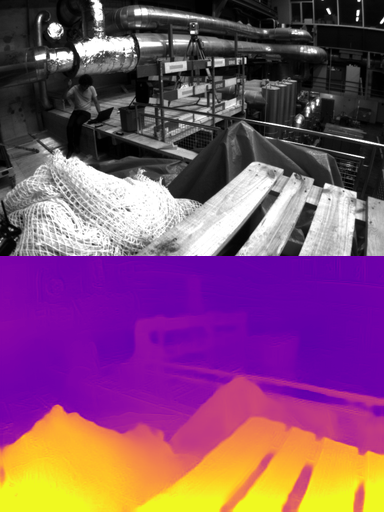}
  \includegraphics[width=0.32\linewidth]{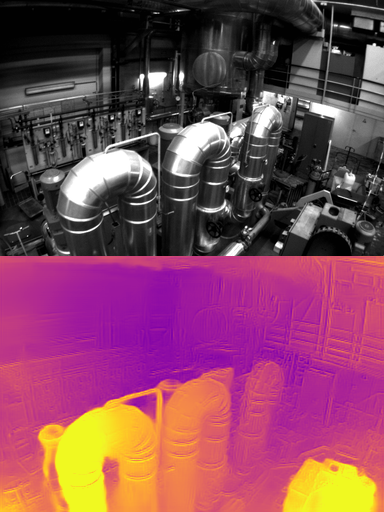}
}\label{fig:depth_euroc}
\subfloat[OmniCam]{
  \includegraphics[width=0.32\linewidth]{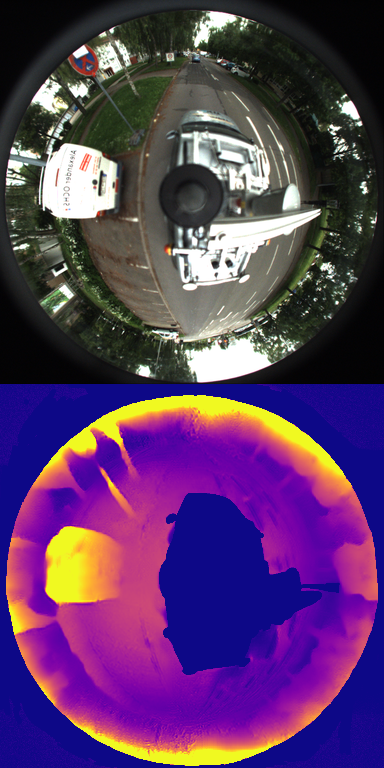}
  \includegraphics[width=0.32\linewidth]{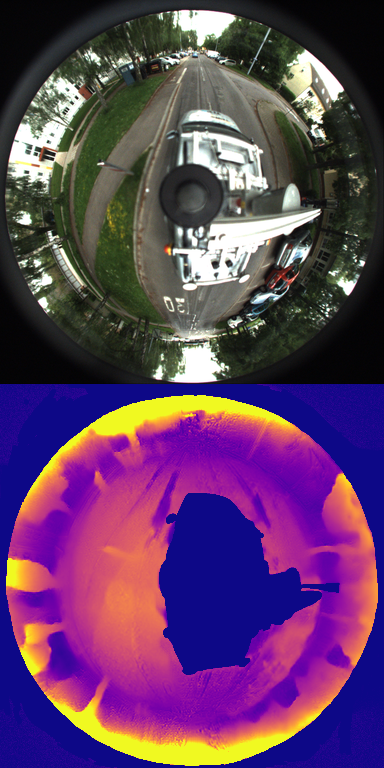}
  \includegraphics[width=0.32\linewidth]{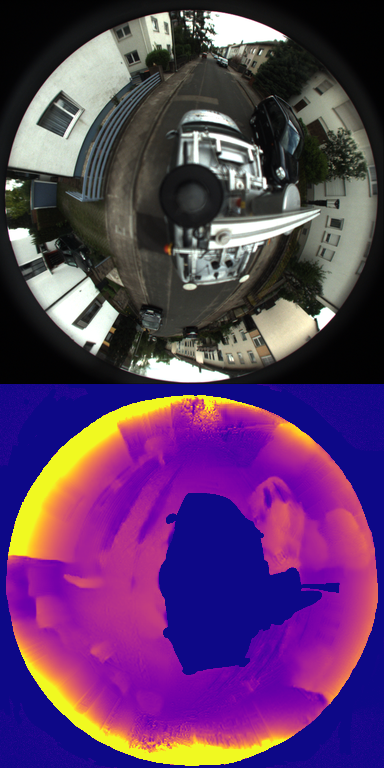}
 }
  \caption{\textbf{Qualitative depth estimation results} on non-pinhole datasets with (a) fisheye and (b) catadioptric images.}
  \label{fig:depth_omnicam}
\end{figure}

\paragraph{Depth Estimation}
\label{subsec:results_depth}
While we use depth and pose estimation as proxy tasks for camera self-calibration, the unified camera model framework allows us to achieve meaningful results compared to prior camera-learning-based approaches (see Figures~\ref{fig:pointcloud_euroc} and \ref{fig:depth_omnicam}). 

\noindent\textbf{KITTI results.} Table~\ref{table:kitti_depth} presents the results of our method on the KITTI dataset. We note that our approach is able to model the simple pinhole setting, achieving results that are on par with approaches that are  tailored specifically to this camera geometry. Interestingly, we see an increase in performance using the UCM model, which we attribute to the ability to further account for and correct calibration errors.

\noindent\textbf{EuRoC results.}
Compared to KITTI, EuRoC is a significantly more challenging dataset that involves cluttered indoor sequences with six-DoF motion. Compared to the per-frame distorted camera models of \citet{gordon2019depth} and \citet{vasiljevic2020neural}, we achieve significantly better absolute relative error, especially with EUCM, where the error is reduced by $16\%$ (see Table~\ref{table:euroc_depth_MH}). We also train NRS~\cite{vasiljevic2020neural} on this dataset for further comparison, using the official repository.  

\input{papers/selfcalib/tables/multi_dataset_depth}

\input{papers/selfcalib/figures/pointclouds}

\noindent\textbf{Combining heterogeneous datasets.}
One of the strengths of the unified camera model is that it can represent a wide variety of cameras without prior knowledge of their specific geometry. As long as we know which sequences come from which camera, we can learn separate calibration vectors that share the same depth and pose networks. This is particularly useful as a way to improve performance on smaller datasets, since it enables one to take advantage of unlabeled data from other sources. To evaluate this property, we experimented with mixing KITTI and EuRoC.  In this experiment, we reshaped the KITTI images to match those in the EuRoC dataset (i.e.,  $384 \times 256$). As Table~\ref{table:multi_depth} shows, our algorithm is able to take advantage of the KITTI images to improve performance on the EuRoC depth evaluation.

\paragraph{Computational Cost}

Our work is closely related to the learned general camera model (NRS) of~\citet{vasiljevic2020neural} given that in both works the parameters of a central general camera model are learned in a self-supervised way.  Being a per-pixel model, NRS is more general than ours and can handle settings where there is local distortion, which a global camera necessarily cannot model.  However, the computational requirements of the per-pixel NRS are significantly higher. For example, we train on EuRoC images with a resolution of $384 \times 256$ with a batch size of $16$, which consumes about $6$\,GB of GPU memory.  Each epoch takes about $15$ minutes.

On the same GPU, NRS uses $16$\,GB of GPU memory with a batch size of $1$ to train on the same sequences, running one epoch in about $120$ minutes. This is due to the high-dimensional (yet approximate) projection operation required for a generalized camera. Thus, we trade some degree of generality for significantly higher efficiency than prior work, with higher accuracy on the EuRoC dataset (see Table~\ref{table:euroc_depth_MH}).

%% file: papers/selfcalib/tables/calib_table.tex
\captionsetup[table]{skip=6pt}

\begin{table}[t!]
\renewcommand{\arraystretch}{1.1}
\centering
{
\setlength{\tabcolsep}{0.3em}
\begin{tabular}{lcc}
\toprule
\multirow{2}{*}{\bf{Method}} & \multicolumn{2}{c}{\emph{Mean Reprojection Error}} 
\\
\cmidrule{2-3}
  & 
\emph{Target-based} & 
\emph{Learned}
\\
\midrule
Pinhole & 1.950 & 2.230\\
UCM~\cite{geyer2000unifying}  &
0.145 & 0.249 \\
EUCM~\cite{khomutenko2015enhanced}  &
0.144 & 0.245 \\
DS~\cite{usenko2018double}  &
0.144  & 0.344 \\
\bottomrule
\end{tabular}
}
\caption{
\textbf{Mean reprojection error on EuRoC} at $256 \times 384$ resolution for UCM, EUCM and DS models using (left) AprilTag-based toolbox calibration Basalt ~\cite{usenko19nfr} and (right) our self-supervised learned (L) calibration.
Note that despite using no ground-truth calibration targets, our self-supervised procedure produces sub-pixel reprojection error.
}
\label{table:reproj_error}
\end{table}

%% file: papers/selfcalib/tables/euroc_calib.tex
\captionsetup[table]{skip=6pt}

\begin{table}[t!]
\renewcommand{\arraystretch}{1.1}
\centering
{
\setlength{\tabcolsep}{0.3em}
\begin{tabular}{lcccccccc}
\toprule
\textbf{Method}  & 
$f_x$ &
$f_y$ &
$c_x$ &
$c_y$ & 
$\alpha$ &
$\beta$ &
$\xi$ &
$w$\\
\midrule
UCM (L) &
237.6 & 247.9 & 187.9 & 130.3 & 0.631 & \multirow{2}{*}{---} & \multirow{2}{*}{---}& \multirow{2}{*}{---}\\ %
UCM (B) & 235.4 & 245.1 & 186.5 & 132.6 & 0.650 \\ %
\midrule
EUCM (L) & 237.4 & 247.7 & 186.7 & 129.1 & 0.598 & 1.075 & \multirow{2}{*}{---} & \multirow{2}{*}{---}\\
EUCM (B) & 235.6 & 245.4 & 186.4 & 132.7 & 0.597 & 1.112\\
\midrule
DS (L) & 184.8 & 193.3 & 187.8 & 130.2 & 0.561 & \multirow{2}{*}{---} & -0.232 & \multirow{2}{*}{---}\\
DS (B) & 181.4 & 188.9 & 186.4 & 132.6 &  0.571 & & -0.230\\
\bottomrule
\end{tabular}
}
\caption{\textbf{Intrinsic calibration evaluation of different methods} on the EuRoC dataset, where B denotes intrinsics obtained from Basalt, and L denotes learned intrinsics.}
\label{table:intrinsic_numbers_compare}
\end{table}

%% file: papers/selfcalib/tables/perturbation_test.tex
\captionsetup[table]{skip=6pt}

\begin{table}[t!]
\renewcommand{\arraystretch}{1.1}
\centering
{
\setlength{\tabcolsep}{0.3em}
\begin{tabular}{lccccccc}
\toprule
\textbf{Perturbation}  & 
$f_x$ &
$f_y$ &
$c_x$ &
$c_y$  & 
$\alpha$ &
$\beta$ &
\textbf{MRE}

\\
\midrule

$I_{1.10}$ init & 242.3 & 253.6 & 189.5 & 130.7 & 0.5984 & 1.080 & 0.409\\
$I_{1.05}$ init & 241.3 & 252.3 & 188.5 & 130.5 & 0.5981 & 1.078 & 0.367\\
$I_c$ init & 240.2 & 251.4 & 187.9 & 130.0  & 0.5971 & 1.076 & 0.348\\
$I_{0.95}$ init & 239.5 & 250.9 & 187.8 & 129.2  & 0.5970 & 1.076 & 0.332\\
$I_{0.90}$ init & 238.8 & 249.6 & 187.7 & 129.1 & 0.5968 & 1.071 & 0.298\\

\midrule

$I_c$ & 235.6 & 245.4 & 186.4 & 132.7 & 0.597 & 1.112 & 0.144\\
\bottomrule
\end{tabular}
}
\caption{
\textbf{EUCM perturbation test results.} With perturbed initialization, all intrinsic parameters achieve sub-pixel convergence for mean reprojection error (\textbf{MRE}), with only a small offset to the Basalt calibration numbers.
}
\label{table:perturbation}
\end{table}

%% file: papers/selfcalib/tables/kitti_depth.tex
\captionsetup[table]{skip=6pt}

\begin{table}[t!]
\renewcommand{\arraystretch}{1.1}
\centering
{
\setlength{\tabcolsep}{0.2em}
\begin{tabular}{llcccc}
\toprule
\textbf{Method}  &  Camera &
Abs Rel$\downarrow$ &
Sq Rel$\downarrow$ &
RMSE$\downarrow$ &
$\delta_{1.25}$ $\uparrow$ %
\\
\toprule
\citet{gordon2019depth} &
K &
0.129 & 0.982 & 5.23 & 0.840\\ %
\citet{gordon2019depth} & 
L(P) &
0.128 & 0.959 & 5.23 & 0.845 \\ %
\citet{vasiljevic2020neural} &
K(NRS) &
0.137 & 0.987 & 5.33 & 0.830 \\ %
\citet{vasiljevic2020neural} &
L(NRS) &
0.134 & 0.952 & 5.26 & 0.832 \\ %
\midrule
Ours & 
L(P) &
0.129 & \textbf{0.893} & 4.96 & 0.846 \\ %
Ours &
L(UCM) &
\textbf{0.126} & 0.951 & \textbf{4.89} &  \textbf{0.858} \\ %
\bottomrule
\end{tabular}
}
\caption{
\textbf{Quantitative depth evaluation on the KITTI \cite{burri2016euroc} dataset}, using the standard \emph{Eigen} split and the \emph{Garg} crop, for distances up to 80m (with median scaling). K and L($\cdot$) denote known and learned intrinsics, respectively.
 P means pinhole model.}
\label{table:kitti_depth}
\end{table}

%% file: papers/selfcalib/tables/euroc_depth.tex
\captionsetup[table]{skip=6pt}

\begin{table}[t!]
\renewcommand{\arraystretch}{1.1}
\centering
{
\setlength{\tabcolsep}{0.3em}
\begin{tabular}{lccccc}
\toprule
\textbf{Method}  & 
Camera  & 
Abs Rel$\downarrow$ &
Sq Rel$\downarrow$ &
RMSE$\downarrow$ &
$\alpha_{1}$ $\uparrow$ %
\\
\midrule
\citet{gordon2019depth} & 
PB %
&
0.332 & 0.389 & 0.971 & 0.420 \\ %
\citet{vasiljevic2020neural}
&
NRS %
&
0.303 & 0.056 & 0.154 & 0.556 \\ %
\midrule

Ours & UCM & 0.282 & 0.048 & 0.141 & 0.591 \\ %
Ours &EUCM & \textbf{0.278} & \textbf{0.047} & \textbf{0.135} & \textbf{0.598} \\ %
Ours &DS & \textbf{0.278} & 0.049 & 0.141 & 0.584 \\ %

\bottomrule
\end{tabular}
}
\caption{
\textbf{Quantitative depth evaluation of different methods on the EuROC \cite{burri2016euroc} dataset}, using the evaluation procedure in~\cite{gordon2019depth} with center cropping.  The training data consists of ``Machine Room'' sequences and the evaluation is on the ''Vicon Room 201'' sequence (with median scaling). PN means plum-bob model.}
\vspace{-3mm}
\label{table:euroc_depth_MH}
\end{table}

%% file: papers/selfcalib/tables/multi_dataset_depth.tex
\captionsetup[table]{skip=6pt}

\begin{table}[t!]
\renewcommand{\arraystretch}{1.1}
\centering
{
\setlength{\tabcolsep}{0.3em}
\begin{tabular}{lcccccc}
\toprule
\textbf{Dataset}  & 
Abs Rel$\downarrow$ &
Sq Rel$\downarrow$ &
RMSE$\downarrow$ &
$\alpha_{1}$ $\uparrow$ & 
$\alpha_{2}$ $\uparrow$ &
$\alpha_{3}$ $\uparrow$
\\
\midrule
EuRoC~\cite{gordon2019depth}  &
0.265 & \textbf{0.042} & 0.130 & 0.600 & 0.882 & \textbf{0.966} \\
EuRoC+KITTI &
\textbf{0.244} & 0.044 & \textbf{0.117} & \textbf{0.742} & \textbf{0.907} & 0.961 \\

\bottomrule
\end{tabular}
}
\caption{
\textbf{Quantitative multi-dataset depth evaluation} on EuRoC (without cropping and with median scaling). 
}
\label{table:multi_depth}
\end{table}

%% file: papers/selfcalib/figures/pointclouds.tex
\begin{figure}[!t]
 \centering
 \includegraphics[width=1.0\linewidth]{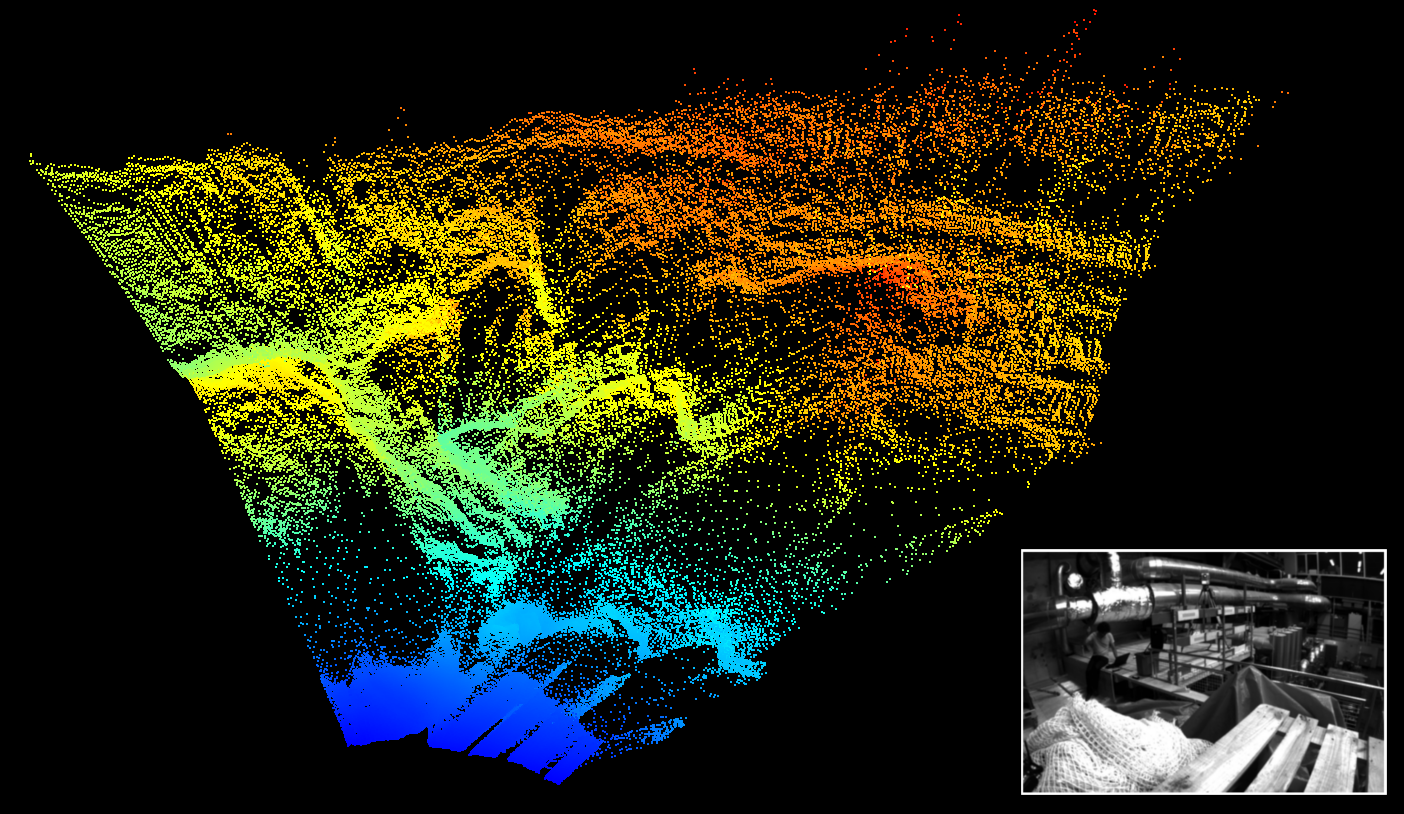}
\caption{\textbf{Self-supervised monocular pointcloud} for EuRoC, obtained by unprojecting predicted depth with our learned camera parameters (input image on the bottom right).} \label{fig:pointcloud_euroc}
\vspace{-3mm}
\end{figure}

%% file: papers/selfcalib/sections/conclusion.tex
We proposed a procedure to self-calibrate a family of general camera models using self-supervised depth and pose estimation as a proxy task.  We rigorously evaluated the quality of the resulting camera models, demonstrating sub-pixel calibration accuracy comparable to manual target-based toolbox calibration approaches. Our approach generates per-sequence camera parameters, and can be integrated into any learning procedure where calibration is needed and the projection and un-projection operations are interpretable and differentiable. As shown in our experiments, our approach is particularly amenable to online re-calibration, and can be used to combine datasets of different sources, learning independent calibration parameters while sharing the same depth and pose network. 

%% file: papers/define/sections/00abstract.tex
\begin{figure}[H]
\centering
\subfloat{
\includegraphics[width=1.0\textwidth]{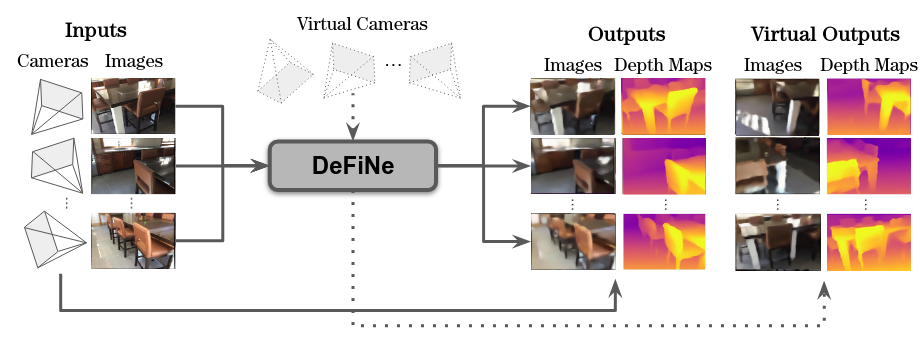}
}
\vspace{-3mm}
\caption{Our \textbf{Depth Field Networks (DeFiNe)} achieve state of the art in multi-view depth estimation, while also enabling predictions from arbitrary viewpoints.%
}
\label{fig:teaser}
\end{figure}

Modern 3D computer vision leverages learning to boost geometric reasoning, mapping image data to classical structures such as cost volumes or epipolar constraints to improve matching. 
These architectures are specialized according to the particular problem, and thus require significant task-specific tuning, often leading to poor domain generalization performance.
Recently, generalist Transformer architectures have achieved impressive results in tasks such as optical flow and depth estimation by encoding geometric priors as inputs rather than as enforced constraints.
In this paper, we extend this idea and propose to learn an implicit, multi-view consistent scene representation, introducing a series of 3D data augmentation techniques as a geometric inductive prior to increase view diversity.  We also show that introducing view synthesis as an auxiliary task further improves depth estimation.
Our Depth Field Network\xspace (DeFiNe\xspace) achieve state-of-the-art results in stereo and video depth estimation without explicit geometric constraints, and improve on zero-shot domain generalization by a wide margin.

%% file: papers/define/sections/01introduction.tex
Estimating 3D structure from a pair of images is a cornerstone problem of computer vision. Traditionally, this is treated as a correspondence problem, whereby one applies a homography to stereo-rectify the images and then matches pixels (or patches) along epipolar lines to obtain disparity estimates. %
Contemporary approaches to stereo are specialized variants of classical methods, relying on correspondences to compute cost volumes, epipolar losses, bundle adjustment objectives, or projective multi-view constraints, among others. These are either baked directly into the model architecture or enforced as part of the loss function.

Applying the principles of classical vision in this way has given rise to architectures that achieve state-of-the-art results on tasks such as stereo depth estimation~\cite{kendall2017end,lipson2021raft}, optical flow~\cite{raft}, and multi-view depth~\cite{deepv2d}. However, this success comes at a cost: each architecture is specialized and purpose-built for a single task. %
Great strides have been made to alleviate the dependence on strong geometric assumptions~\cite{gordon2019depth,vasiljevic2020neural}, and two recent trends allow us to \textit{decouple} the task from the architecture: (i) implicit representations and (ii) generalist networks.  Our work draws upon both of these directions. 

Implicit representations of geometry and coordinate-based networks have recently achieved incredible popularity in the vision community. This direction is pioneered by work on neural radiance fields (NeRF)~\cite{mildenhall2020nerf,xie2021neural}, where a point- and ray-based parameterization along with a volume rendering objective allow simple MLP-based networks to achieve state-of-the-art view synthesis results.  Follow-up works extend this coordinate-based representation to the pixel domain~\cite{pixelnerf}, allowing predicted views to be conditioned on image features. 
The second trend in computer vision has been the use of generalist architectures, pioneered by Vision Transformers~\cite{dosovitskiy2020image}. Emerging as an attention-based architecture for NLP, Transformers have been used for a diverse set of tasks, including depth estimation~\cite{li2021revisiting,ranftl2021vision}, optical flow~\cite{jaegle2021perceiverio}, and image generation~\cite{esser2021taming}. Transformers have also been applied to geometry-free view synthesis~\cite{rombach2021geometry}, demonstrating that attention can learn long-range correspondence between views for 2D-3D tasks. 
Representation Transformers (SRT)~\cite{sajjadi2021scene} use the Transformer encoder-decoder model to learn scene representations for view synthesis from sparse, high-baseline data with no geometric constraints.  However, owing to the quadratic scaling of the self-attention module, experiments are limited to low-resolution images and require very long training periods. %

To alleviate the quadratic complexity of self-attention, the Perceiver architecture~\cite{jaegle2021perceiver} disentangles the dimensionality of the latent representation from that of the inputs by fixing the size of the latent representation. Perceiver~IO~\cite{jaegle2021perceiverio} extends this architecture to allow for arbitrary outputs, with results on optical flow estimation that outperform traditional cost-volume based methods. Similarly, the recent Input-level Inductive Bias (IIB) architecture %
\cite{yifan2021input} uses image features and camera information as input to Perceiver~IO to directly regress stereo depth, outperforming baselines that use explicit geometric constraints. %
Building upon these works, we propose to learn a \textit{geometric scene representation} for depth synthesis from novel viewpoints, including estimation, interpolation, and extrapolation. We expand the IIB framework to the scene representation setting, taking sequences of images and predicting a consistent multi-view latent representation suitable for different downstream tasks.  
Taking advantage of the query-based nature of the Perceiver~IO architecture, we propose a series of 3D augmentations that increase viewpoint density and diversity during training, thus encouraging (rather than enforcing) multi-view consistency. Furthermore, we show that the introduction of view synthesis as an auxiliary task, decoded from the same latent representation, improves depth estimation without additional ground truth.

We test our model on the popular ScanNet benchmark~\cite{dai2017scannet}, achieving state-of-the-art real-time results for stereo depth estimation and competitive results for video depth estimation, without relying on memory- or compute-intensive operations such as cost volume aggregation and test-time optimization.  We show that our 3D augmentations lead to significant improvements over baselines that are limited to the viewpoint diversity of training data.
Furthermore, our zero-shot transfer results from ScanNet to 7-Scenes~\cite{shotton2013scene} improve the state-of-the-art by a large margin, demonstrating that our method generalizes better than specialized architectures, which suffer from poor performance on out-of-domain data.
Our contributions are summarized as follows:
\begin{itemize}
\item We use a generalist Transformer-based architecture to learn a depth estimator from an arbitrary number of posed images. In this setting, we (i) \textbf{propose a series of 3D augmentations} that improve the geometric consistency of our learned latent representation; and (ii) show that \textbf{jointly learning view synthesis as an auxiliary task improves depth estimation}.
\item Our Depth Field Networks\xspace (DeFiNe\xspace) not only achieve \textbf{state-of-the-art stereo depth estimation results} on the widely used ScanNet dataset, but also exhibit superior generalization properties with \textbf{state-of-the-art results on zero-shot transfer to 7-Scenes.}
\item DeFiNe\xspace also \textbf{enables depth estimation from arbitrary viewpoints}.  We evaluate this novel generalization capability in the context of \emph{interpolation} (between timesteps), and \emph{extrapolation} (future timesteps).
\end{itemize}

%% file: papers/define/sections/02related.tex
\paragraph{Monocular Depth Estimation.}
Supervised depth estimation---the task of estimating per-pixel depth given an RGB image and a corresponding ground-truth depth map---dates back to the pioneering work of Saxena et al.~\cite{saxena2005learning}. %
Since then, deep learning-based architectures designed for supervised monocular depth estimation have become increasingly sophisticated~\cite{eigen2015predicting,eigen2014depth,fu2018deep,laina2016deeper,lee2019big}, generally offering improvements over the standard encoder-decoder convolutional architecture. %
Self-supervised methods provide an alternative to those that rely on ground-truth depth maps at training time, and are able to take advantage of the new availability of large-scale video datasets. Early self-supervised methods relied on stereo data~\cite{godard2017unsupervised}, and then progressed to fully monocular video sequences~\cite{zhou2017unsupervised}, with increasingly sophisticated losses~\cite{shu2020feature} and architectures~\cite{monodepth2,packnet,watson2021temporal}. 

In a follow-up study, ZeroDepth~\cite{guizilini2023towards} extends our input-level geometry bias framework to metric-aware monocular depth estimation with SOTA results.

\paragraph{Multi-view Stereo.}
Traditional multi-view stereo approaches have dominated even in the deep learning era. COLMAP~\cite{schonberger2016structure} remains the standard framework for structure-from-motion, incorporating sophisticated bundle adjustment and keypoint refinement procedures, at the cost of speed. With the goal of producing closer to real-time estimates, multi-view stereo learning approaches adapt traditional cost volume-based approaches to stereo~\cite{kendall2017end,chang2018pyramid} and multi-view~\cite{yao2018mvsnet,im2019dpsnet} depth estimation, often relying on known extrinsics to warp views into the frame of the reference camera. 
Recently, iterative refinement approaches that employ recurrent neural networks have made impressive strides in optical flow estimation~\cite{raft}. Follow-on work applies this general recurrent correspondence architecture to stereo depth~\cite{lipson2021raft}, scene-flow~\cite{teed2021raft}, and even SLAM~\cite{teed2021droid}.  While their results are impressive, recurrent neural networks can be difficult to train, and test-time optimization increases inference time over a single forward pass.

Recently, Transformer-based architectures~\cite{attention_all} have replaced CNNs in many geometric estimation tasks.  The Stereo Transformer~\cite{li2021revisiting} architecture replaces cost volumes with a correspondence approach inspired by sequence-to-sequence modeling.
The Perceiver~IO~\cite{jaegle2021perceiverio} architecture constitutes a large departure from cost volumes and geometric losses. For the task of optical flow, Perceiver~IO feeds positionally encoded images through a Transformer~\cite{jaegle2021perceiver}, rather than using a cost volume for processing.  %
IIB~\cite{yifan2021input} adapts the Perceiver~IO architecture to generalized stereo estimation, proposing a novel epipolar parameterization as an additional input-level inductive bias.  Building upon this baseline, 
we propose a series of geometry-preserving 3D data augmentation techniques designed to promote the learning of a \emph{geometrically-consistent latent scene representation}. We also introduce novel view synthesis as an auxiliary task to depth estimation, decoded from the same latent space. %
Our video-based representation (aided by our 3D augmentations) allows us to generalize to novel viewpoints, rather than be restricted to the stereo setting. %

In a subsequent work, ~\cite{xu2023unifying} builds a unified flow, stereo and depth estimation, with a similar principle to learn multi-modal feature using cross-attention.

\paragraph{Video Depth Estimation.}
Video and stereo depth estimation methods generally produce monocular depth estimates at test time. %
ManyDepth~\cite{watson2021temporal} combines a monocular depth framework with multi-view stereo, aggregating predictions in a cost volume and thus enabling multi-frame inference at test-time.
Recent methods accumulate predictions at train and test time, either with generalized stereo~\cite{ummenhofer2017demon} or with sequence data~\cite{zhou2018deeptam}.  DeepV2D~\cite{deepv2d} incorporates a cost-volume based multi-view stereo approach with an incremental pose estimator to iteratively improve depth and pose estimates at train and test time. 

Another line of work draws on the availability of monocular depth networks that perform accurate but \textit{multi-view inconsistent} estimates at test time~\cite{luo2020consistent}. In this setting, additional geometric constraints are enforced to finetune the network and improve multi-view consistency through epipolar constraints. %
Consistent Video Depth Estimation~\cite{luo2020consistent} refines COLMAP~\cite{schonberger2016structure} results with a monocular depth network constrained to be multi-view consistent. Subsequent work jointly optimizes depth and pose for added robustness to challenging scenarios with poor calibration~\cite{kopf2021robust}. A recent  framework incorporates many architectural elements of prior work into a Transformer-based architecture that takes video data as input for multi-view depth~\cite{long2021multi}. %
NeuralRecon~\cite{Sun_2021_CVPR} moves beyond depth-based architectures to learn Truncated Signed Distance Field (TSDF) volumes as a way to improve surface consistency. %

\paragraph{Novel View Synthesis.}
Since the emergence of neural radiance fields~\cite{mildenhall2020nerf}, implicit representations and volume rendering have emerged as the \textit{de facto} standard for view synthesis.  They parameterize viewing rays and points with positional encoding,
and need to be re-trained on a scene-by-scene basis.  Many recent improvements leverage depth supervision to improve view synthesis in a volume rendering framework~\cite{azinovic2021neural,deng2021depth,rematas2021urban,nerfingmvs,zhu2021nice}.
An alternative approach replaces volume rendering with a directly learned light field network~\cite{sitzmann2021light}, predicting color values directly from viewing rays. This is the approach we take when estimating the auxiliary view synthesis loss, due to its computational simplicity. 
Other works attempt to extend the NeRF approach beyond single scene models by incorporating learned~\textit{features}, enabling few-shot volume rendering~\cite{pixelnerf}.  Feature-based methods have also treated view synthesis as a sequence-learning task, such as the Scene Representation Transformer (SRT) architecture~\cite{sajjadi2021scene}.  %

Building upon the same principle, DeLiRa~\cite{guizilini2023delira} extends the output space to radiance field and light field, showing an improvement of performance when self-supervised across the output fields.

%% file: papers/define/sections/03methodology.tex
\input{papers/define/figures/diagram}

\vspace{-2mm}
Our proposed DeFiNe\xspace architecture (Figure~\ref{fig:diagram}) is designed with flexibility in mind, so data from different sources can be used as input and different output tasks can be estimated from the same latent space. Similar to Yifan et al.~\cite{yifan2021input}, we use Perceiver~IO~\cite{jaegle2021perceiver} as our general-purpose Transformer backbone. During the encoding stage, our model takes RGB images from calibrated cameras, with known intrinsics and relative poses. The architecture processes this information according to the modality into different pixel-wise embeddings that serve as input to our Perceiver~IO backbone.  This encoded information can be queried using only camera embeddings, producing estimates from arbitrary viewpoints.  %

\vspace{-2mm}
\paragraph{Perceiver~IO}
Perceiver~IO~\cite{jaegle2021perceiverio} is a recent extension of the Perceiver~\cite{jaegle2021perceiver} architecture. %
The Perceiver architecture alleviates one of the main weaknesses of Transformer-based methods, namely the quadratic scaling of self-attention with input size.  This is achieved by using a fixed-size $N_l \times C_l$ latent representation $\mathcal{R}$, and learning to project high-dimensional $N_e \times C_e$ encoding embeddings onto this latent space using cross-attention layers. Self-attention is performed in the lower-dimensional latent space, producing a \emph{conditioned latent representation} $\mathcal{R}_{c}$ that can be queried using $N_d \times C_d$ decoding embeddings to generate estimates, again using cross-attention layers.

\vspace{-2mm}
\paragraph{Input-Level Embeddings}

\noindent\textbf{Image Embeddings (Figure \ref{fig:rgb_embeddings}).}
Input $3 \times H \times W$ images are processed using a ResNet18~\cite{he2016deep} encoder, producing a list of features maps at increasingly lower resolutions and higher dimensionality. Feature maps at $1/4$ the original resolution are concatenated with lower-resolution feature maps, after upsampling using bilinear interpolation. The resulting image embeddings are of shape $H/4 \times W/4 \times 960$, and are used in combination with the camera embeddings from each corresponding pixel (see below) to encode visual information. \\

\noindent\textbf{Camera Embeddings (Figure \ref{fig:cam_embeddings}).}
\label{sec:camera_embeddings}
These embeddings capture multi-view scene geometry (e.g., camera intrinsics and extrinsics) as additional inputs during the learning process. Let $\textbf{x}_{ij} = (u,v)$ be an image coordinate corresponding to pixel $i$ in camera $j$, with assumed known pinhole $3 \times 3$ intrinsics $\mathbf{K}_j$ and $4 \times 4$ transformation matrix $T_j= \left[
\begin{smallmatrix}
\mathbf{R}_j & \textbf{t}_j \\
\textbf{0} & 1
\end{smallmatrix}\right]$
relative to a canonical camera $T_0$. Its origin $\mathbf{o}_j$ and direction $\textbf{r}_{ij}$ are given by:
\begin{equation}
\textbf{o}_j = - \mathbf{R}_j \mathbf{t}_j 
\quad , \quad 
\textbf{r}_{ij} = \big(\mathbf{K}_j \mathbf{R}_j \big)^{-1}  
\left[u_{ij},v_{ij},1\right]^T 
+ \textbf{t}_j
\end{equation}
Note that this formulation differs from the standard convention~\cite{mildenhall2020nerf}, which does not consider the camera translation $\textbf{t}_j$ when generating viewing rays $\textbf{r}_{ij}$. We ablate this variation in Table \ref{tab:ablation}, showing that it leads to better performance for the task of depth estimation.
These two vectors are then Fourier-encoded dimension-wise to produce higher-dimensional vectors, with a mapping of:
\begin{equation}
    x \mapsto 
    \begin{bmatrix}
        x, & \sin(f_1\pi x), & \cos(f_1\pi x), & \dots, & \sin(f_K\pi x), & \cos(f_K\pi x)
    \end{bmatrix}^\top
\end{equation}
where $K$ is the number of Fourier frequencies used ($K_o$ for the origin and $K_r$ for the ray directions), equally spaced in the interval $[1,\frac{\mu}{2}]$. The resulting camera embedding is of dimensionality 
$
2 \big( 3(K_o + 1) + 3(K_r + 1) \big) = 
6 \left( K_o + K_r + 2\right)
$. During the encoding stage, camera embeddings are produced per-pixel assuming a camera with $1/4$ the original input resolution, resulting in a total of $\frac{HW}{16}$ vectors. During the decoding stage, embeddings from cameras with arbitrary calibration (i.e., intrinsics and extrinsics) can be queried to produce virtual estimates.

\vspace{-2mm}
\paragraph{Geometric 3D Augmentations}
\label{sec:augmentations}
Data augmentation is a core component of deep learning pipelines~\cite{shorten2019survey} that improves model robustness by applying transformations to the training data consistent with the data distribution in order to introduce desired equivariant properties. 
In computer vision and depth estimation in particular, standard data augmentation techniques are usually constrained to the 2D space and include color jittering, flipping, rotation, cropping, and resizing~\cite{monodepth2,yifan2021input}. 
Recent works have started looking into 3D augmentations~\cite{sajjadi2021scene} to improve robustness to errors in scene geometry in terms of camera localization (i.e., extrinsics) and parameters (i.e., intrinsics).
Conversely, we are interested in \emph{encoding} scene geometry at the input-level, so our architecture can learn a multi-view-consistent geometric latent scene representation. Therefore, in this section we propose a series of 3D augmentations to increase the number and diversity of training views while maintaining the spatial relationship between cameras, thus enforcing desired equivariant properties within this setting.\\

\input{papers/define/figures/augmentations}
\noindent\textbf{Virtual Camera Projection.}
One of the key properties of our architecture is that it enables querying from arbitrary viewpoints, since only camera information (viewing rays) is required at the decoding stage. When generating predictions from these novel viewpoints, the network creates \emph{virtual} information consistent with the implicit structure of the learned latent scene representation, conditioned on information from the encoded views. We evaluate this capability in Section~\ref{sec:depth_synthesis}, showing superior performance relative to the explicit projection of information from encoded views. 
Here, we propose to leverage this property at training time as well, generating additional supervision in the form of \emph{virtual cameras} with corresponding ground-truth RGB images and depth maps obtained by projecting available information onto these new viewpoints (Figure~\ref{fig:virtual_projection}). This novel augmentation technique forces the learned latent scene representation to be viewpoint-independent. Experiments show that this approach provides benefits in both the (a) stereo setting, with only two viewpoints; and (b) video setting, with similar viewpoints and a dominant camera direction.

From a practical perspective, virtual cameras are generated by adding translation noise $\bm{\epsilon}_v = [\epsilon_x,\epsilon_y,\epsilon_z]_v \sim \mathcal{N}(0,\sigma_v)$ to the pose of a camera $i$. The viewing angle is set to point towards the center $\textbf{c}_i$ of the pointcloud $P_i$ generated by unprojecting information from the selected camera, which is also perturbed by $\bm{\epsilon}_c = [\epsilon_x,\epsilon_y,\epsilon_z]_c \sim \mathcal{N}(0,\sigma_v)$. When generating ground-truth information, we project the combined pointcloud from all available cameras onto these new viewpoints as a way to preserve full scene geometry. Furthermore, because the resulting RGB image and depth map will be sparse, we can improve efficiency by only querying at these specific locations. \\

\noindent\textbf{Canonical Jittering.}
When operating in a multi-camera setting, it is standard practice to select one camera to be the \emph{reference} camera, and position all other cameras relative to it~\cite{im2019dpsnet}.  One drawback of this convention is that one camera will always be at the same location (the origin of its own coordinate system) and will therefore produce the same camera embeddings, leading to overfitting. Intuitively, scene geometry should be invariant to the translation and rotation of the entire sensor suite. To enforce this property on our learned latent scene representation, we propose to inject some amount of noise to the canonical pose itself, so it is not located at the origin of the coordinate system.  
Note that this is different from methods that inject per-camera noise~\cite{novotny2017learning} with the goal of increasing robustness to localization errors. We only inject noise \emph{once}, on the canonical camera, and propagate it to other cameras, so relative scene geometry is preserved within a translation and rotation offset (Figure~\ref{fig:canonical_jittering}). However, this offset is reflected on the input-level embeddings produced by each camera, and thus forces the latent representation to be invariant to these transformations. 

In order to perform canonical jittering, we randomly sample translation $\bm{\epsilon}_t = [\epsilon_x, \epsilon_y, \epsilon_z]^\top \sim \mathcal{N}(0,\sigma_t)$ and rotation \mbox{$\bm{\epsilon}_r  = [\epsilon_\phi, \epsilon_\theta, \epsilon_\psi]^\top \sim \mathcal{N}(0,\sigma_r)$} 
errors from zero-mean normal distributions with pre-determined standard deviations.  Represented as Euler angles, we convert each set of rotation errors to a $3 \times 3$ rotation matrix $\mathbf{R}_r$. We then use the rotation matrix and translation error to create a jittered canonical transformation matrix $T_0' = \left[
\begin{smallmatrix}
\mathbf{R}_r & \bm{\epsilon}_t \\
\textbf{0} & 1
\end{smallmatrix} \right]$ that is then propagated to all other $N$ cameras, such that $T_i' = T_0' \cdot T_i$,  $\forall i \in \{1, \dots, N-1\}$.\\

\noindent\textbf{Canonical Randomization.}
As an extension to canonical jittering, we also introduce canonical randomization to encourage generalization to different relative camera configurations, while still preserving scene geometry. Assuming $N$ cameras, we randomly select $o \in \{0,\dots,N-1\}$ as the canonical index. We then compute the relative transformation matrix $T_i'$ given world-frame transformation matrix $T_i$ as $T_i' = T_i \cdot T_o^{-1}$ $\forall i \in \{0,\dots,N-1\}$. %

\vspace{-2mm}
\paragraph{Decoders}

We use task-specific decoders, each consisting of one cross-attention layer between the $N_d \times C_d$ queries and the $N_l \times C_l$ conditioned latent representation $\mathcal{R}_c$, followed by a linear layer that creates an output of size $N_d \times C_o$, and a sigmoid activation function $\sigma(x)=\frac{1}{1 + e^{-x}}$ to produce values in the interval $[0,1]$. 
We set $C_o^d = 1$ for depth estimation and $C_o^s = 3$ for view synthesis. Depth estimates are scaled to lie within the range $[d_\text{min}, d_\text{max}]$. Note that other decoders can be incorporated with DeFiNe\xspace without any modification to the underlying architecture, enabling the generation of multi-task estimates from arbitrary viewpoints.

\paragraph{Losses}

We use an L1-log loss $\mathcal{L}_{d} = \lVert \log(d_{ij}) - \log(\hat{d}_{ij})\rVert_1$ to supervise depth estimation, where $\hat{d}_{ij}$ and $d_{ij}$ are depth estimates and ground truth, respectively, for pixel $j$ at camera $i$. For view synthesis, we use an L2 loss $\mathcal{L}_{s} = \lVert \textbf{p}_{ij} - \hat{\textbf{p}}_{ij} \rVert^2$, where $\hat{\textbf{p}}_{ij}$ and $\textbf{p}_{ij}$ are RGB estimates and ground truth, respectively, for pixel $j$ at camera $i$. We use a weight coefficient $\lambda_{s}$ to balance these two losses, and another $\lambda_{v}$ to balance losses from available and virtual cameras. The final loss is of the form:
\begin{equation}
\label{eq:loss}
\mathcal{L} = \mathcal{L}_{d} + \lambda_{s} \mathcal{L}_{s} + \lambda_{v}\big(\mathcal{L}_{d,v} + \lambda_{s}(\mathcal{L}_{s,v})\big)
\end{equation}
Note that because our architecture enables querying at specific image coordinates, we can improve efficiency at training time by not computing estimates for pixels without ground truth (e.g., sparse depth maps or virtual cameras).

%% file: papers/define/figures/diagram.tex
\begin{figure}[t!]
\centering
\begin{subfigure}[t]{1.0\textwidth}
    \centering
    \includegraphics[width=\textwidth]{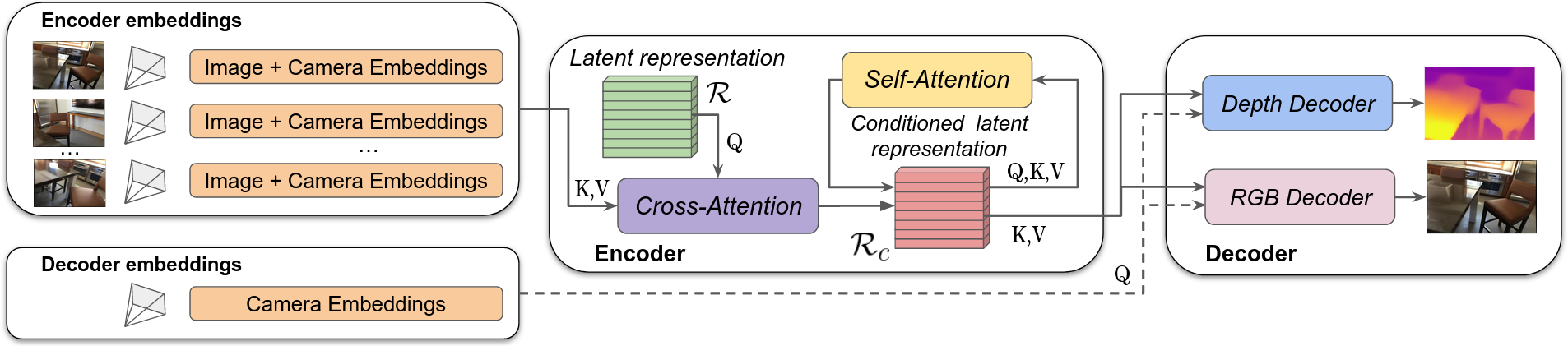}
    \caption{Architecture overview.}
    \label{fig:diagram}
\end{subfigure}
\\
\begin{subfigure}[t]{0.45\textwidth}
    \centering
    \includegraphics[height=1.5in]{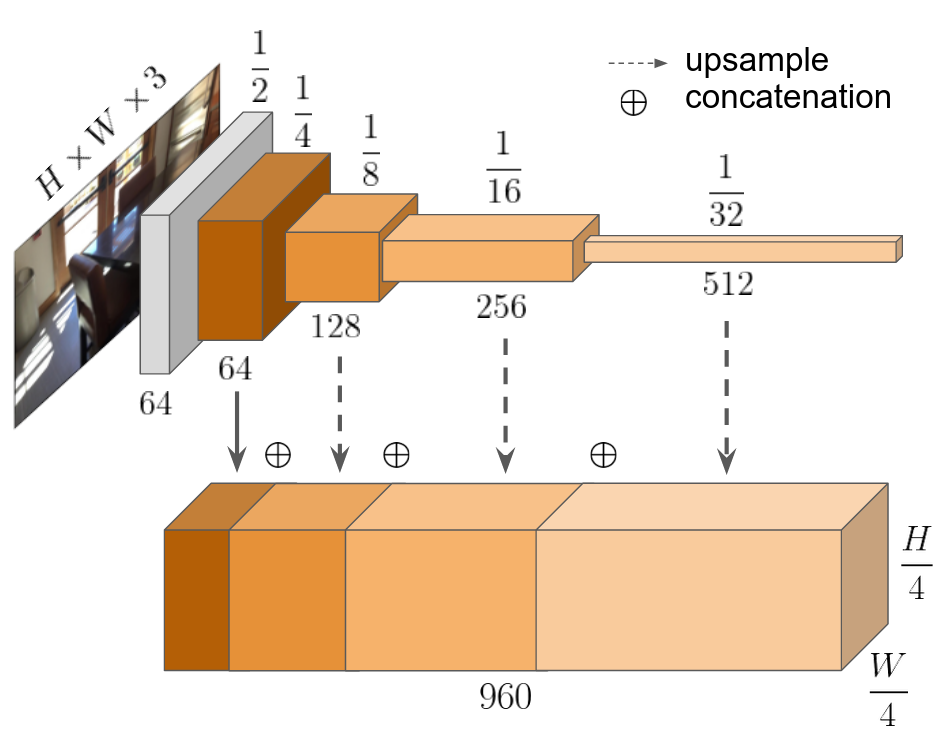}
    \caption{Image embeddings.}
    \label{fig:rgb_embeddings}
\end{subfigure}\hfill
\begin{subfigure}[t]{0.45\textwidth}
    \centering
    \includegraphics[height=1.5in]{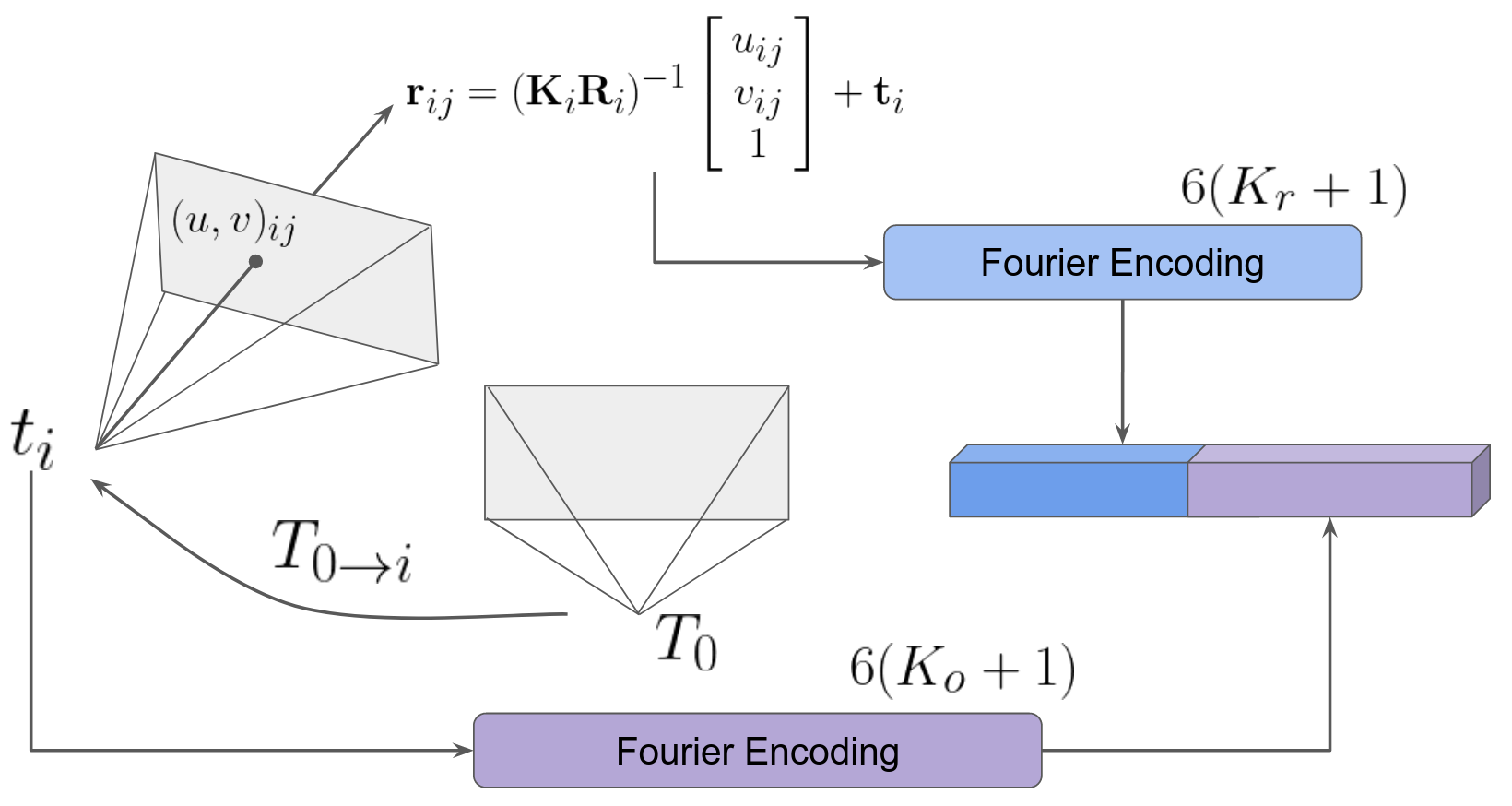}
    \caption{Camera embeddings.}
    \label{fig:cam_embeddings}
\end{subfigure}
\caption{
\textbf{Overview of our proposed DeFiNe architecture}, and the embeddings used to encode and decode information for depth and view synthesis. 
}
\label{fig:encodings}
\end{figure}

%% file: papers/define/figures/augmentations.tex
\begin{figure}[t!]
\centering
\begin{subfigure}[t]{0.48\textwidth}
    \centering
    \includegraphics[width=\textwidth]{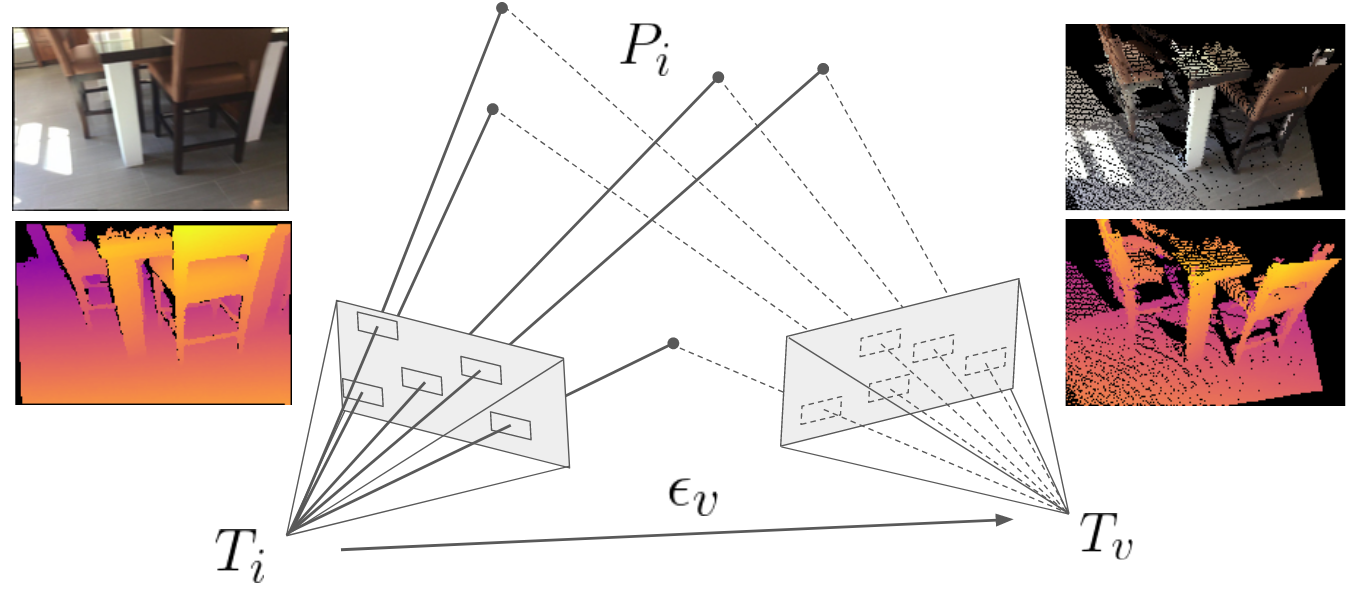}
    \caption{Virtual Camera Projection.}
    \label{fig:virtual_projection}
\end{subfigure}
\hspace{0.8mm}
\begin{subfigure}[t]{0.45\textwidth}
    \centering
    \includegraphics[width=\textwidth]{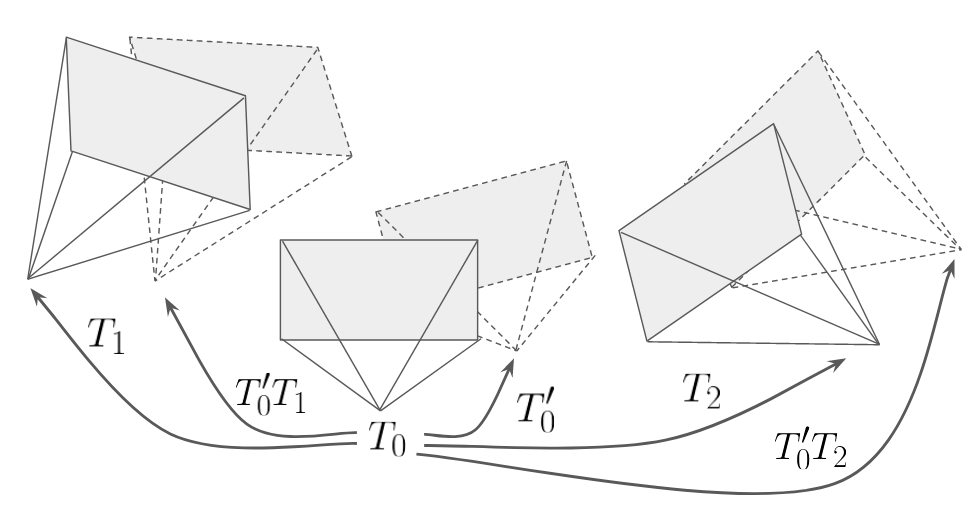}
    \caption{Canonical Jittering.}
    \label{fig:canonical_jittering}
\end{subfigure}
\caption{
\textbf{Geometric augmentations}. 
(a) Information from camera $i$ is projected onto a virtual camera at $T_v$, creating additional supervision from other viewpoints. 
(b) Noise $T_0'$ is introduced to the canonical camera at $T_0$, and then propagated to other cameras to preserve relative scene geometry. 
}
\label{fig:augmentations}
\vspace{-3mm}
\end{figure}

%% file: papers/define/sections/04experiments.tex
\paragraph{Datasets}

\noindent\textbf{ScanNet~\cite{dai2017scannet}}
We evaluate our DeFiNe\xspace for both \emph{stereo} and \emph{video} depth estimation using ScanNet,  an RGB-D video dataset that contains $2.5$ million views from around $1500$ scenes.
For the stereo experiments, we follow the same setting as Kusupati et al.~\cite{kusupati2020normal}: we downsample scenes by a factor of $20$ and use a custom split to create stereo pairs, resulting in $94212$ training and $7517$ test samples. 
For the video experiments, we follow the evaluation protocol of Teed et al.~\cite{deepv2d}, with a total of $1405$ scenes for training.
For the test set, we use a custom split to select $2000$ samples from $90$ scenes not covered in the training set. Each training sample includes a target frame and a context of $[-3,3]$ frames with stride $3$. Each test sample includes a pair of frames, with a context of $[-3,3]$ relative to the first frame of the pair with stride $3$.\\

\noindent\textbf{7-Scenes~\cite{shotton2013scene}}
We also evaluate on the test split of 7-Scenes to measure zero-shot cross-dataset performance. Collected using KinectFusion~\cite{KinectFusion}, the dataset consists of $640 \times 480$ images in $7$ settings, with a variable number of scenes in each setting.  There are $500$--$1000$ images in each scene.  We follow the evaluation protocol of Sun et al.~\cite{Sun_2021_CVPR}, median-scaling predictions using ground-truth information before evaluation.

\paragraph{Stereo Depth Estimation}
\label{sec:stereo_depth_estimation}
To test the benefits our proposed geometric 3D augmentation procedures over the IIB~\cite{yifan2021input} baseline, we first evaluate DeFiNe\xspace on the task of stereo depth estimation. Here, because each sample provides minimal information about the scene (i.e., only two frames), the introduction of additional virtual supervision should have the largest effect. We report our results in Figure~\ref{fig:depth_scannet_stereo}a and visualize examples of reconstructed pointclouds in Figure~\ref{fig:pointclouds}. DeFiNe\xspace significantly outperforms other methods on this dataset, including IIB.  Our virtual view augmentations lead to a large ($20\%$) relative improvement, showing that DeFiNe\xspace benefits from a scene representation that encourages multi-view consistency.

\input{papers/define/figures/scannet_stereo}

\input{papers/define/tables/ablation}

\paragraph{Ablation Study}
\label{sec:ablation}

We perform a detailed ablation study to evaluate the effectiveness of each component in our proposed architecture, with results shown in Table~\ref{tab:ablation}. Firstly, we evaluate performance when (Table~\ref{tab:ablation}:1) learning only depth estimation, and see that the joint learning of view synthesis as an auxiliary task leads to significant improvements. The claim that depth estimation improves view synthesis has been noted before~\cite{deng2021depth,nerfingmvs}, and is attributed to the well-known fact that multi-view consistency facilitates the generation of images from novel viewpoints.  However, our experiments also show the inverse: view synthesis improves depth estimation. Our hypothesis is that appearance is required to learn multi-view consistency since it enables visual correlation across frames. By introducing view synthesis as an additional task, we are also encoding appearance-based information into our latent representation. This leads to improvements in depth estimation even though no explicit feature matching is performed at an architecture or loss level.

\input{papers/define/figures/pointclouds}

\input{papers/define/figures/temporal_qualitative}

We also ablate different variations of our RGB encoder for the generation of image embeddings and show that (Table~\ref{tab:ablation}:2) our proposed multi-level feature map concatenation (Figure~\ref{fig:rgb_embeddings}) leads to the best results relative to the standard single convolutional layer, or (Table~\ref{tab:ablation}:3) using $1/4$-resolution ResNet18 64-dimensional feature maps. Similarly, we also ablate some of our design choices, namely (Table~\ref{tab:ablation}:4) the use of camera embeddings instead of positional encodings; (Table~\ref{tab:ablation}:5) global viewing rays (Section~\ref{sec:camera_embeddings}) instead of traditional relative viewing rays; (Table~\ref{tab:ablation}:6) the use of $\lambda_s=1$ in the loss calculation (Equation~\ref{eq:loss}) such that both depth and view synthesis tasks have equal weights; and (Table~\ref{tab:ablation}:7) the use of epipolar cues as additional geometric embeddings, as proposed by IIB~\cite{yifan2021input}. As expected, camera embeddings are crucial for multi-view consistency and global viewing rays improve over the standard relative formulation. Interestingly, using a smaller $\lambda_s$ degrades depth estimation performance, providing further evidence that the joint learning of view synthesis is beneficial for multi-view consistency. We did not observe meaningful improvements when incorporating the epipolar cues from IIB~\cite{yifan2021input}, indicating that DeFiNe\xspace is capable of directly capturing these constraints at an input-level due to the increase in viewpoint diversity. Lastly, we ablate the impact of our various proposed geometric augmentations (Section~\ref{sec:augmentations}), showing that they are key to our reported state-of-the-art performance. 

Lastly, we evaluate depth estimation from virtual cameras, using different noise levels $\sigma_v$ at test time. We also train models using different noise levels and report the results in Figure~\ref{fig:virtual_results}. From these results, we can see that the optimal virtual noise level, when evaluating at the target location, is $\sigma_v=0.25$\,m (yellow line), relative to the baseline without virtual noise (blue line). However, models trained with higher virtual noise (e.g., the orange line, with $\sigma_v=1$\,m) are more robust to larger deviations from the target location.

\paragraph{Video Depth Estimation}
\label{sec:video_depth_estimation}

To highlight the flexibility of our proposed architecture, we also experiment using video data from ScanNet following the training protocol of Tang et al.~\cite{tang2018ba}. We evaluate performance on both ScanNet itself, using their evaluation protocol~\cite{tang2018ba}, as well as zero-shot transfer (without fine-tuning) to the 7-Scenes dataset. 
Table~\ref{tab:depth_scannet_temporal} reports quantitative results, while Figure~\ref{fig:temporal_qualitative} provides qualitative examples.
On ScanNet, DeFiNe\xspace outperforms most published approaches, significantly improving over DeMoN~\cite{ummenhofer2017demon}, BA-Net~\cite{tang2018ba}, and CVD~\cite{luo2020consistent} both in terms of performance and speed. 
We are competitive with DeepV2D~\cite{deepv2d} in terms of performance, and roughly $14\times$ faster, owing to the fact that DeFiNe\xspace does not require bundle adjustment or any sort of test-time optimization. 
In fact, our inference time of $49$\,ms can be split into $44$\,ms for encoding and only $5$\,ms for decoding, enabling very efficient generation of depth maps after information has been encoded once. 
The only method that outperforms DeFiNe\xspace in terms of speed is NeuralRecon~\cite{Sun_2021_CVPR}, which uses a sophisticated TSDF integration strategy. Performance-wise, we are also competitive with NeuralRecon, improving over their reported results in one of the three  metrics (Sq.\ Rel). 

Next, we evaluate zero-shot transfer from ScanNet to 7-Scenes, which is a popular test of generalization for video depth estimation. In this setting, DeFiNe\xspace significantly improves over all other methods, including DeepV2D (which fails to generalize) and NeuralRecon ($\sim$$40\%$ improvement). We attribute this large gain to the highly intricate and specialized nature of these other architectures. In contrast, our method has no specialized module and instead learns a geometrically-consistent multi-view latent representation. 

In summary, we achieve competitive results on ScanNet and significantly improve the state-of-the-art for video depth generalization, as evidenced by the large gap between DeFiNe\xspace and the best-performing methods on 7-Scenes.

\input{papers/define/figures/scannet_temporal_deepv2d}

\paragraph{Depth from Novel Viewpoints}
\label{sec:depth_synthesis}

We previously discussed the strong performance that DeFiNe\xspace achieves on traditional depth estimation benchmarks, and showed how it improves out-of-domain generalization by a wide margin. Here, we explore another aspect of generalization that our architecture enables: viewpoint generalization. This is possible because, in addition to traditional depth estimation from RGB images, DeFiNe\xspace can also generate depth maps from arbitrary viewpoints since it only requires camera embeddings to decode estimates. We explore this capability in two different ways: \emph{interpolation} and \emph{extrapolation}. When interpolating, we encode frames at $\{t-5,t+5\}$, and decode virtual depth maps at locations $\{t-4,\dots,t+4\}$. When extrapolating, we encode frames at $\{t-5,\dots,t-1\}$, and decode virtual depth maps at locations $\{t,\dots,t+8\}$. We use the same training and test splits as in our stereo experiments, with a downsampling factor of $20$ to encourage smaller overlaps between frames. As baselines for comparison, we consider the explicit projection of 3D information from encoded frames onto these new viewpoints. We evaluate both standard depth estimation networks~\cite{monodepth2,packnet,lee2019big} as well as DeFiNe\xspace itself, that can be used to either explicitly project information from encoded frames onto new viewpoints (projection), or query from the latent representation at that same location (query).

\input{papers/define/figures/synthesis}

Figure~\ref{fig:depth_synthesis} reports results in terms of root mean squared error (RMSE) considering only valid projected pixels.  Of particular note, our multi-frame depth estimation architecture significantly outperforms other single-frame baselines. However, and most importantly, results obtained by implicit querying consistently outperform those obtained via explicit projection. This indicates that our model is able to improve upon available encoded information via the learned latent representation. Furthermore, we also generate geometrically consistent estimates for areas without valid explicit projections (Figure~\ref{fig:synthesis_qualitative}).  As the camera tilts to the right, the floor is smoothly recreated in unobserved areas, as well as the partially observed bench. Interestingly, the chair at the extreme right was not recreated, which could be seen as a failure case. However, because the chair was never observed in the first place, it is reasonable for the model to assume the area is empty, and recreate it as a continuation of the floor.

%% file: papers/define/figures/scannet_stereo.tex
\begin{figure}[t!]
\centering
\begin{subfigure}[t]{0.5\textwidth}
    \centering
    \raisebox{13mm}{
    \footnotesize %
    \begin{tabular}{l|ccc}
        \toprule
        \textbf{Method} &
        Abs.Rel$\downarrow$ & %
        RMSE$\downarrow$ &
        $\delta_{1.25}$$\uparrow$ \\
        \toprule
        DPSNet~\cite{im2019dpsnet} & 0.126 & 0.314 & ---  \\
        NAS~\cite{kusupati2020normal} & 0.107& 0.281 & ---  \\
        IIB~\cite{yifan2021input} & 0.116  & 0.281 & 0.908  \\
        \midrule
        \textbf{DeFiNe\xspace} ($128 \times 192$) & 0.093  & 0.246 & 0.911 \\
        \textbf{DeFiNe\xspace} ($240 \times 320$) & \textbf{0.089} & \textbf{0.232} & \textbf{0.915} \\
        \bottomrule
    \end{tabular}
    }
    \caption{Depth estimation results.}
    \label{fig:stereo_results}
\end{subfigure}%
\hfill
\begin{subfigure}[t]{0.37\textwidth}
    \centering
    \includegraphics[width=\textwidth]{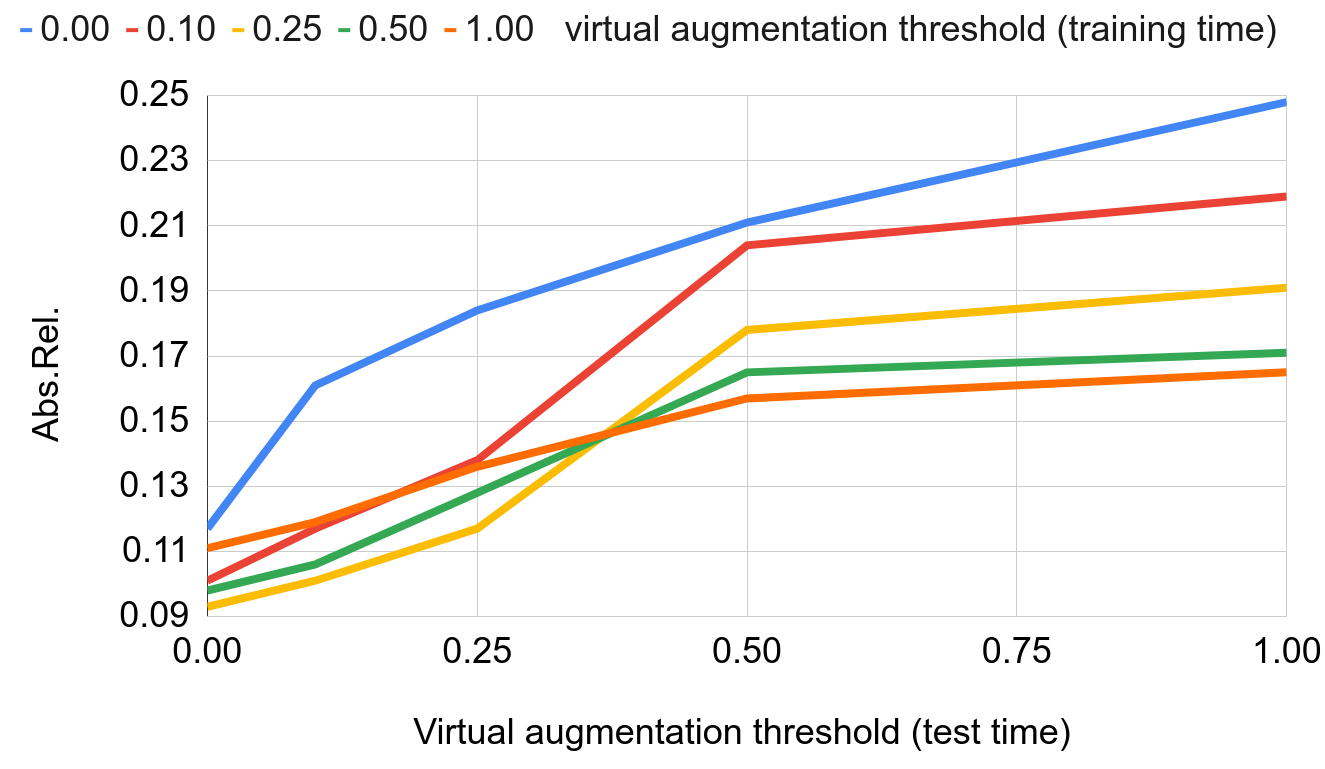}
    \caption{Virtual depth estimation results.}
    \label{fig:virtual_results}
\end{subfigure}
\caption{\textbf{Depth estimation results on ScanNet-Stereo}. (a) We outperform contemporary methods by a large margin. (b) Depth estimation results on virtual cameras using different values for $\sigma_v$ at training and test time. 
}
\label{fig:depth_scannet_stereo}
\end{figure}

%% file: papers/define/tables/ablation.tex
\begin{table*}[t!]
\renewcommand{\arraystretch}{0.92}
\centering
{
\small
\setlength{\tabcolsep}{0.3em}
\begin{tabular}{c|l|ccc|ccc}
\toprule
\multicolumn{2}{c|}{\multirow{2}{*}{\textbf{Variation}}} &
\multicolumn{3}{c|}{Lower is better $\downarrow$} &
\multicolumn{3}{c}{Higher is better $\uparrow$}
 \\
& & 
Abs.\ Rel &
Sq.\ Rel &
RMSE &
$\delta_{1.25}$ &
$\delta_{1.25^2}$ &
$\delta_{1.25^3}$ \\
\toprule
1 & Depth-Only & 0.098 & 0.046 & 0.257 & 0.902 & 0.972 & 0.990 \\
\midrule
2 & w/ Conv. RGB encoder~\cite{jaegle2021perceiverio} & 0.114 & 0.058 & 0.294 & 0.866 & 0.961 & 0.982  \\
3 & w/ 64-dim R18 RGB encoder & 0.104 & 0.049 & 0.270 & 0.883 & 0.966 & 0.985  \\
\midrule
4 & w/o camera information & 0.229 & 0.157 & 0.473 & 0.661 & 0.874 & 0.955  \\
5 & w/o global rays encoding & 0.097 & 0.047 & 0.261 & 0.897 & 0.962 & 0.988  \\
6 & w/ equal loss weights & 0.095 & 0.047 & 0.259 & 0.908 & 0.968 & 0.990  \\
7 & w/ epipolar cues~\cite{yifan2021input} & 0.094 & 0.048 & 0.254 & 0.905 & 0.972 & 0.990 \\
\midrule
8 & w/o Augmentations & 0.117 & 0.060 & 0.291 & 0.870 & 0.959 & 0.981  \\
9 & w/o Virtual Cameras & 0.104 & 0.058 & 0.268 & 0.891 & 0.965 & 0.986  \\
10 & w/o Canonical Jittering & 0.099 & 0.046 & 0.261 & 0.897 & 0.970 & 0.988  \\
11 & w/o Canonical Randomization & 0.096 & 0.044 & 0.253 & 0.905 & 0.971 & 0.989 \\
\midrule
& \textbf{DeFiNe\xspace} & \textbf{0.093} & \textbf{0.042} & \textbf{0.246} & \textbf{0.911} & \textbf{0.974} & \textbf{0.991} \\
\bottomrule
\end{tabular}
}
\caption{\textbf{Ablation study for ScanNet-Stereo}, using different variations. } 
\label{tab:ablation}
\end{table*}

%% file: papers/define/figures/pointclouds.tex
\begin{figure}[t!]
\centering
\rotatebox{90}{\hspace{6mm}\tiny{GT}} \!
\subfloat{\includegraphics[width=0.22\textwidth,height=1.6cm,trim={20cm 0 0 0},clip]{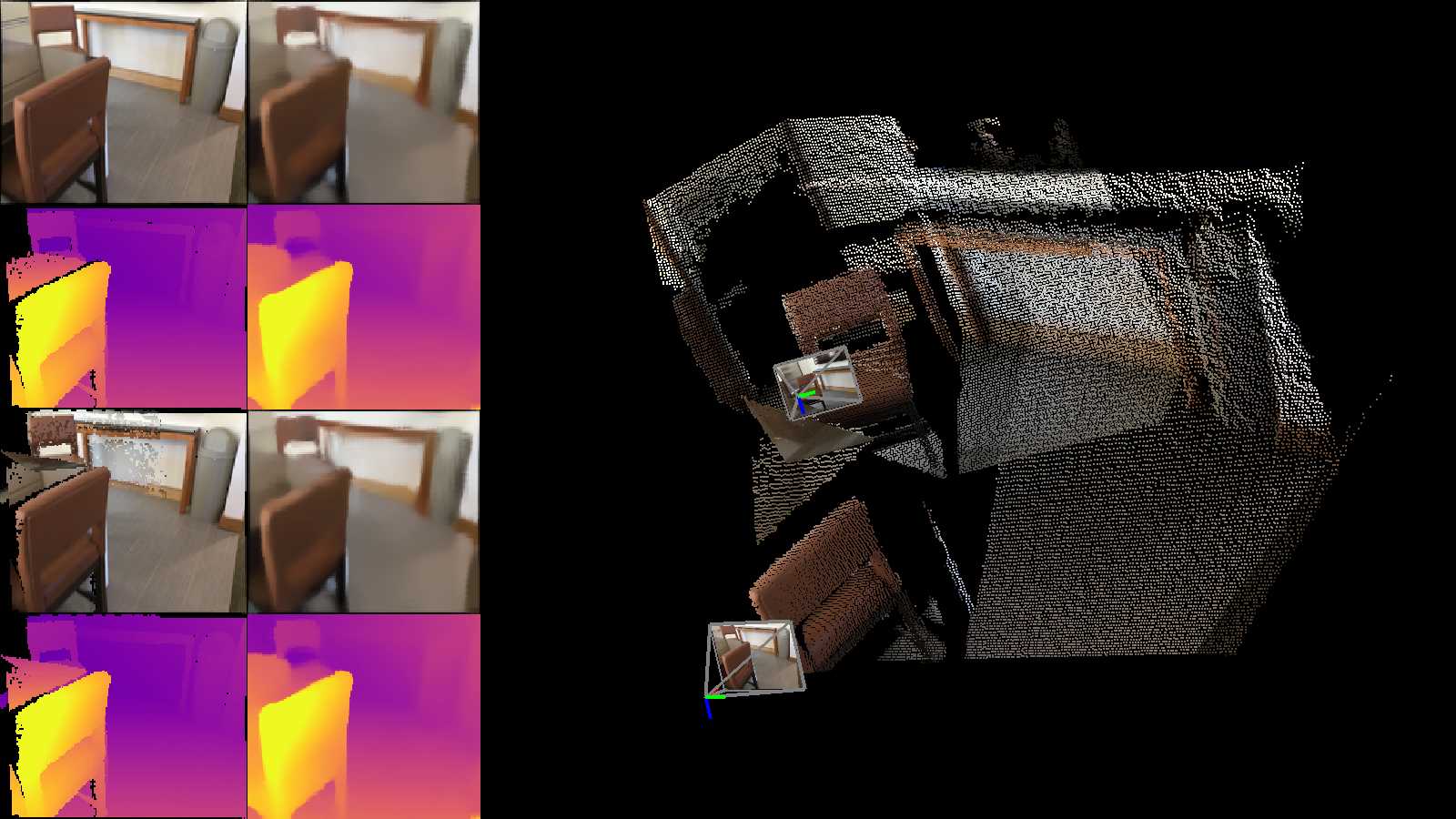}} \hspace{1mm}
\subfloat{\includegraphics[width=0.22\textwidth,height=1.6cm,trim={20cm 0 0 0},clip]{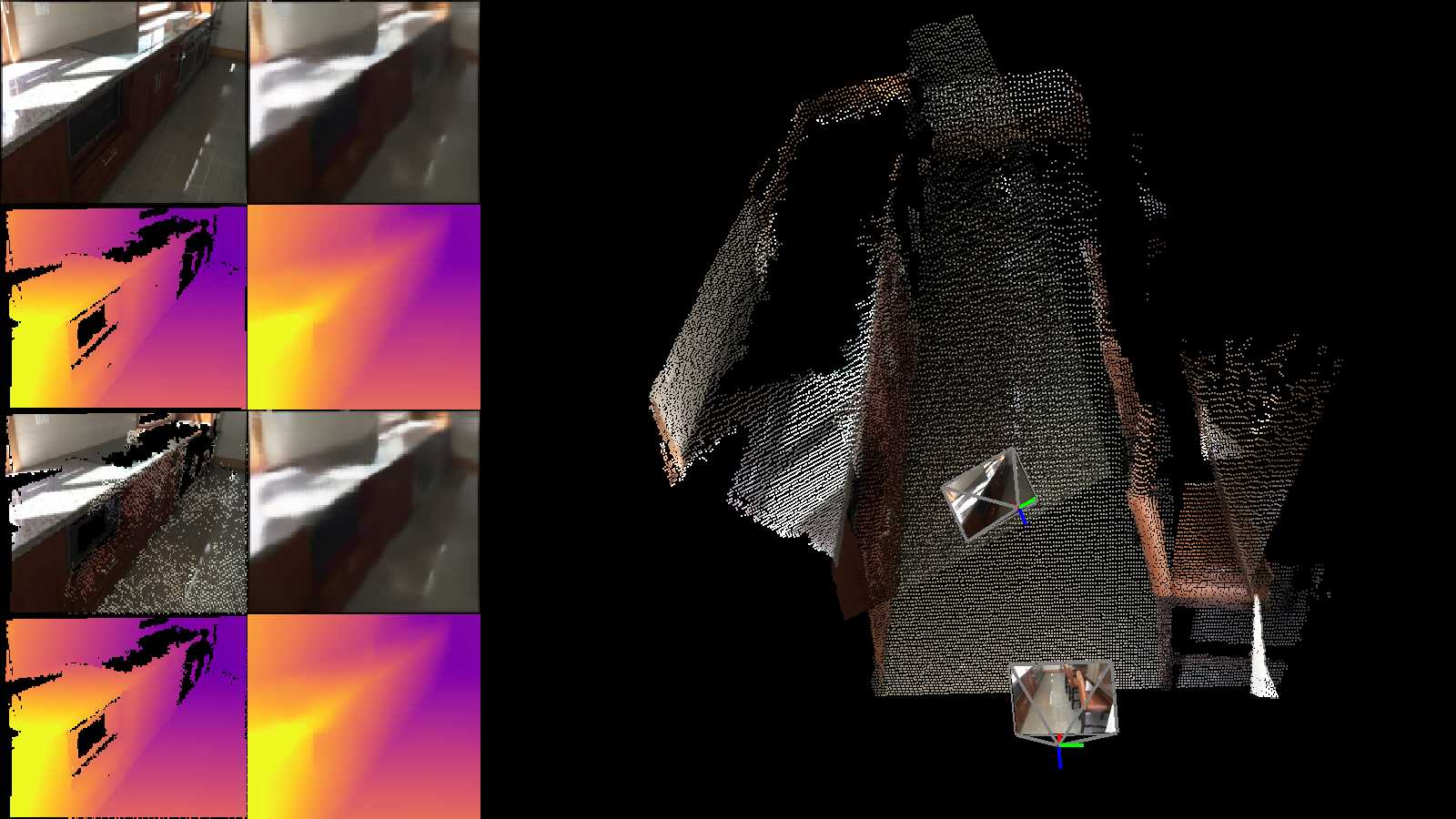}} \hspace{1mm}
\subfloat{\includegraphics[width=0.22\textwidth,height=1.6cm,trim={20cm 0 0 0},clip]{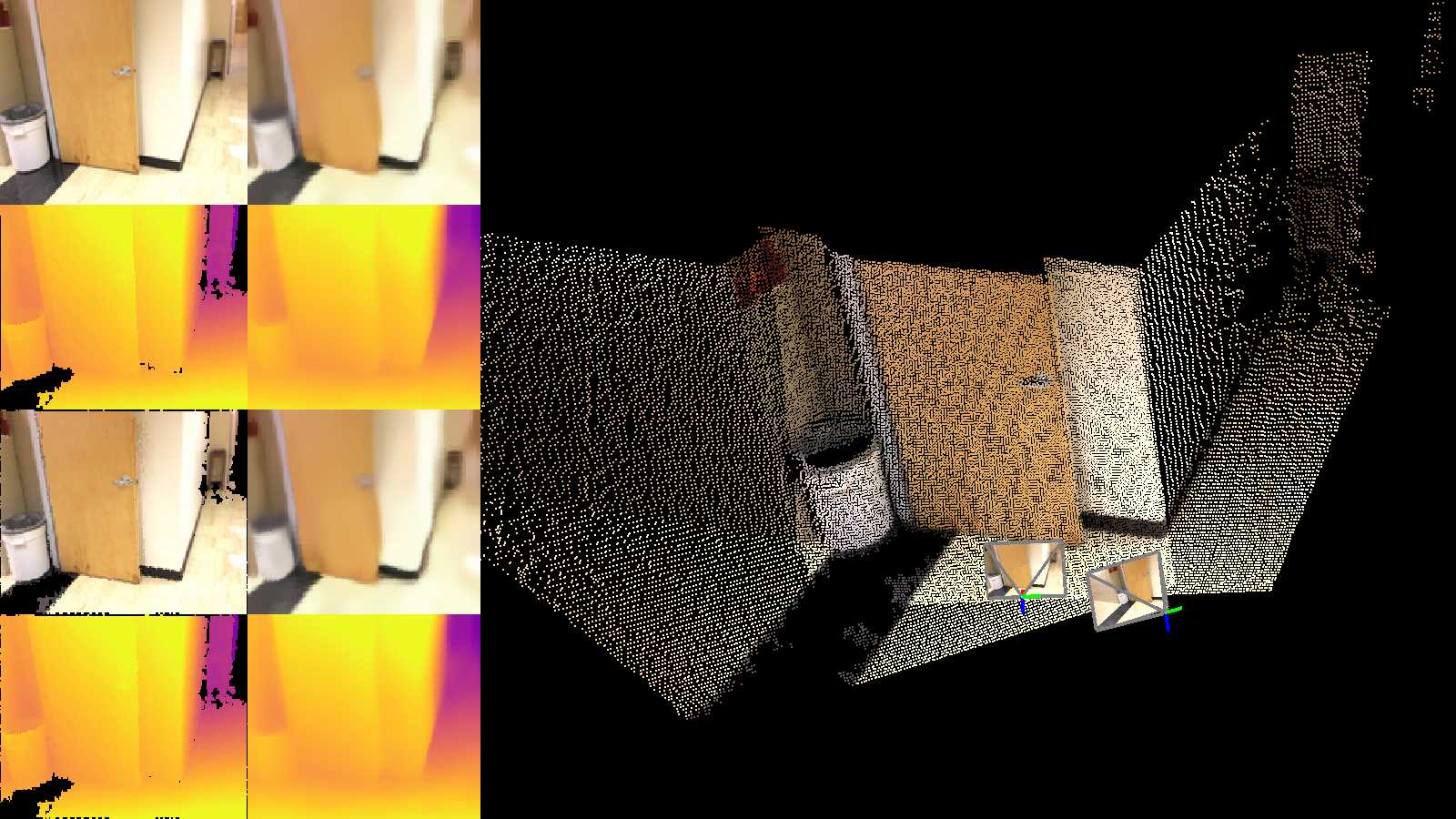}} \hspace{1mm}
\subfloat{\includegraphics[width=0.22\textwidth,height=1.6cm,trim={20cm 0 0 0},clip]{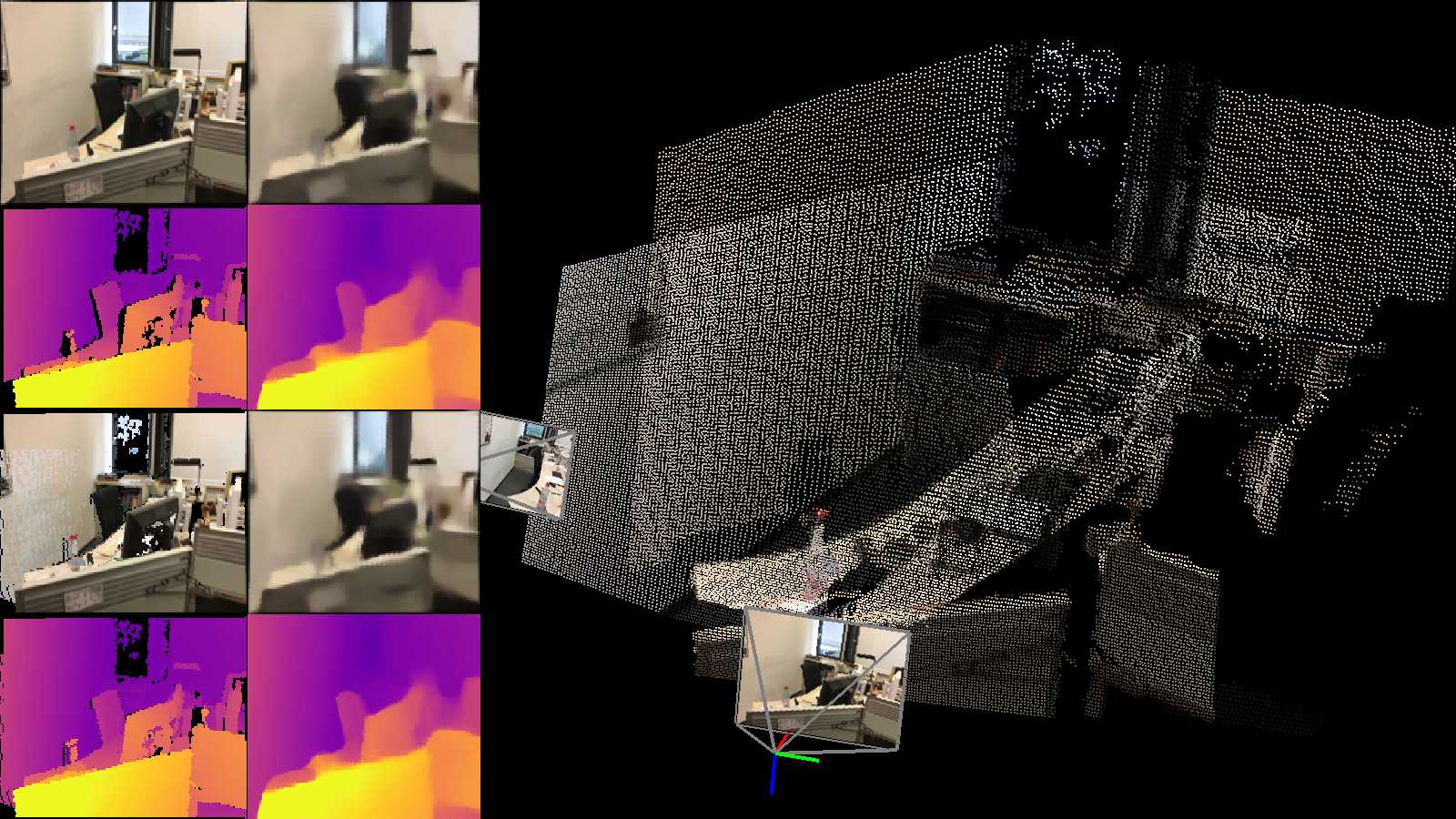}}
\\ %
\rotatebox{90}{\hspace{4mm}\tiny{DeFiNe\xspace}} \!
\subfloat{\includegraphics[width=0.22\textwidth,height=1.6cm,trim={20cm 0 0 0},clip]{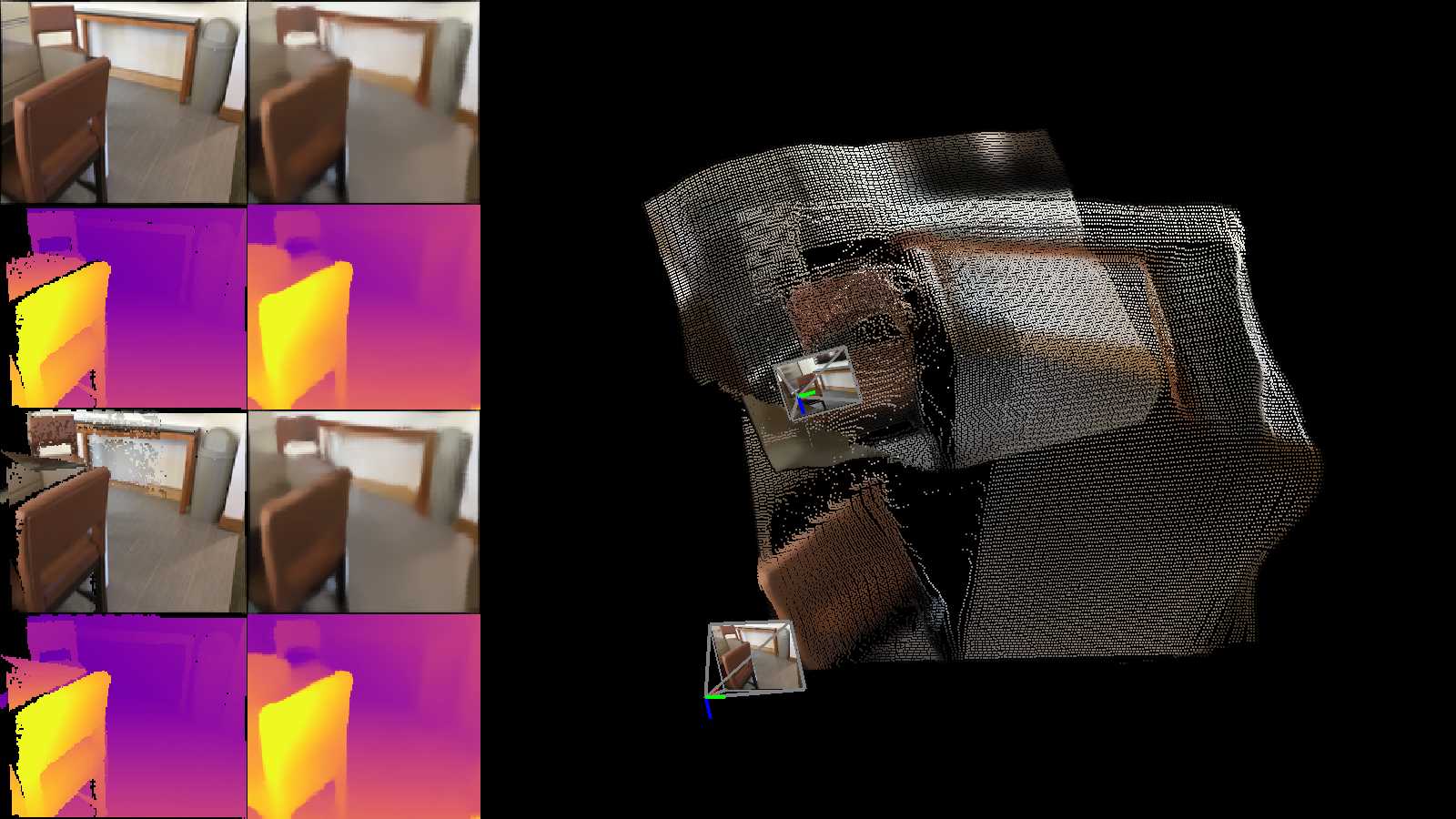}} \hspace{1mm}
\subfloat{\includegraphics[width=0.22\textwidth,height=1.6cm,trim={20cm 0 0 0},clip]{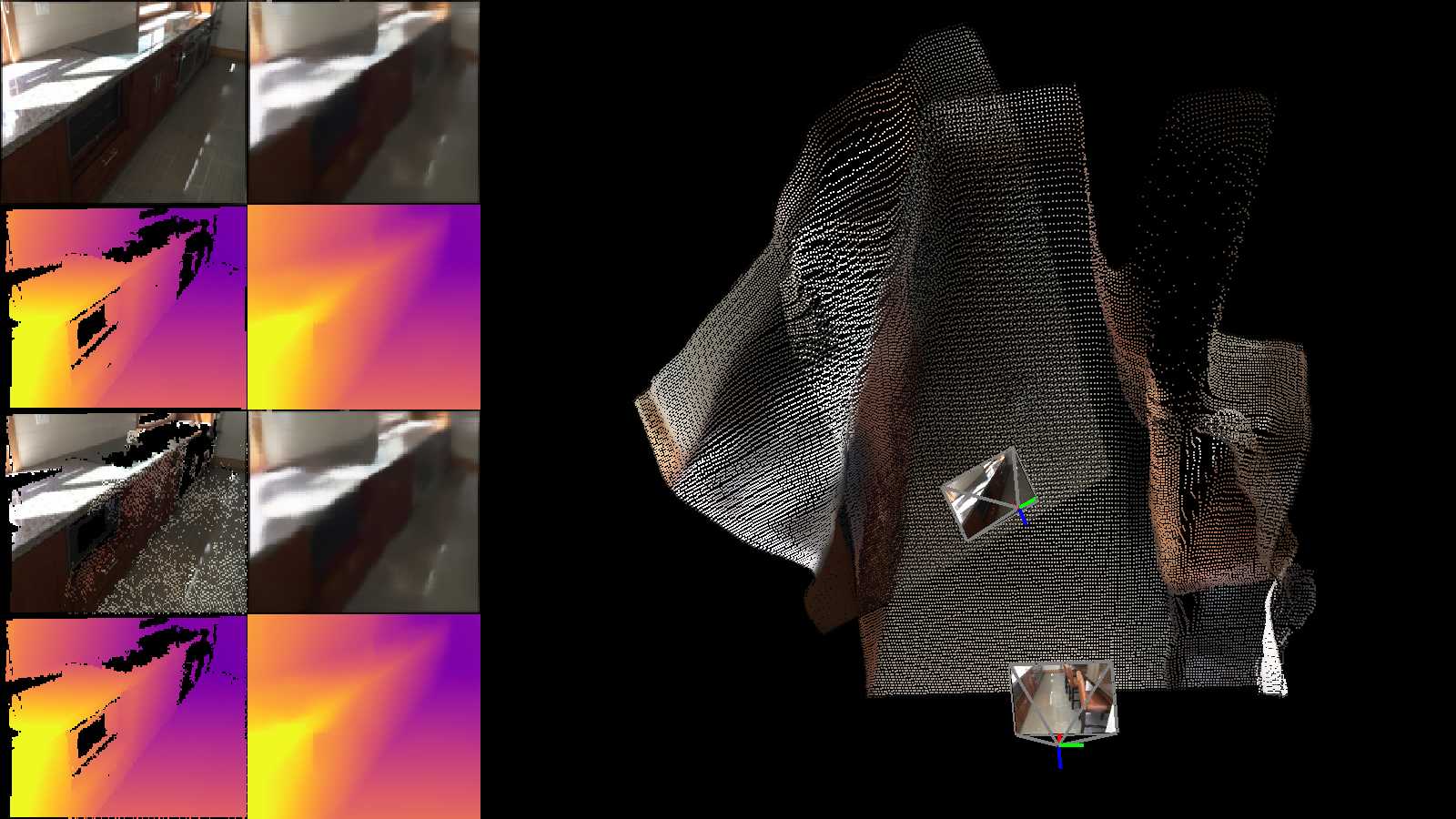}} \hspace{1mm}
\subfloat{\includegraphics[width=0.22\textwidth,height=1.6cm,trim={20cm 0 0 0},clip]{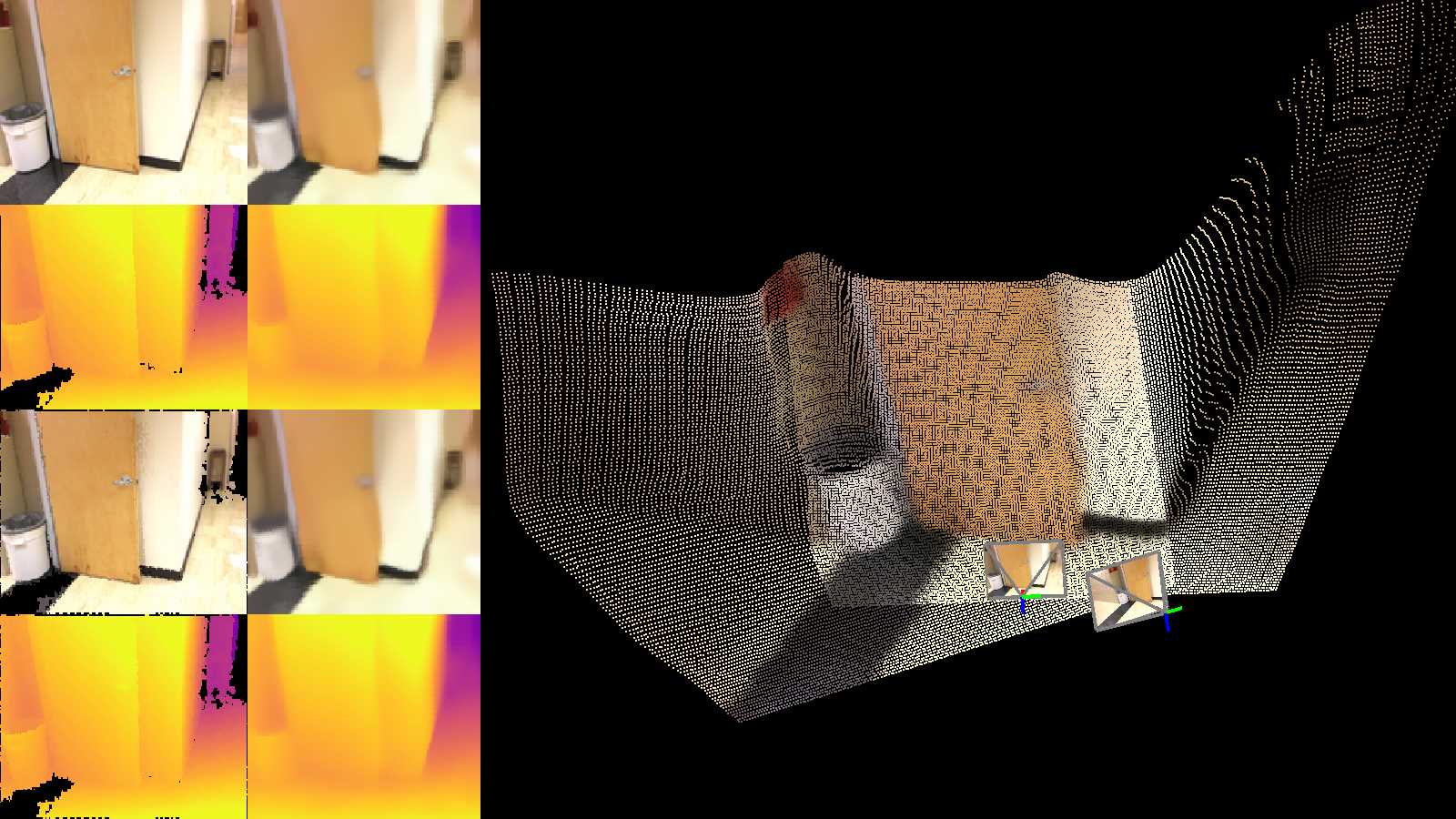}} \hspace{1mm}
\subfloat{\includegraphics[width=0.22\textwidth,height=1.6cm,trim={20cm 0 0 0},clip]{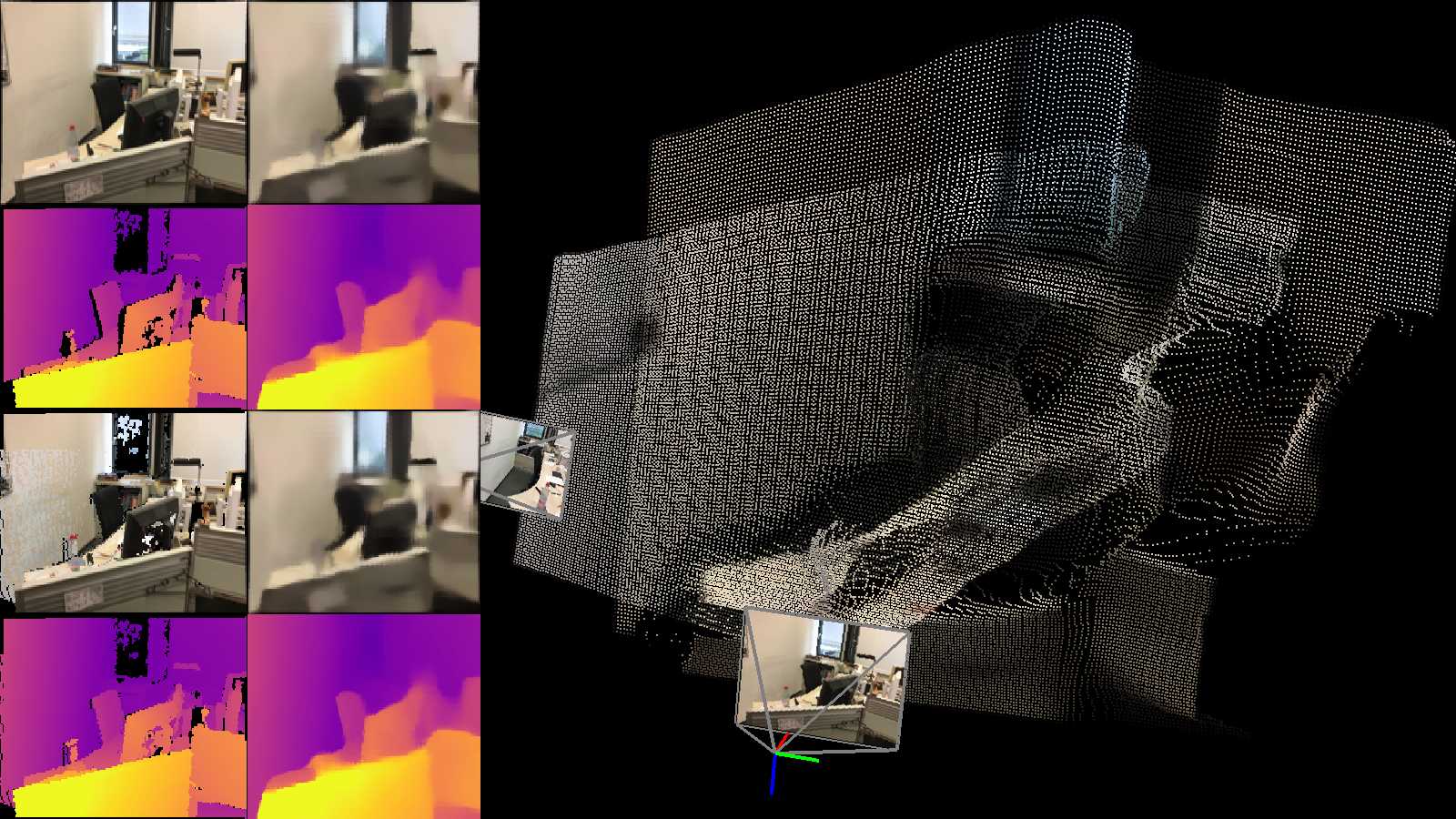}}
\caption{\textbf{Reconstructed two-view pointclouds}, from ScanNet-Stereo. DeFiNe\xspace pointclouds are generated using both depth maps and RGB images queried from our learned latent representation.}
\label{fig:pointclouds}
\end{figure}

%% file: papers/define/figures/temporal_qualitative.tex
\begin{figure}[t!]
\centering
\rotatebox{90}{\hspace{2.5mm} \tiny{GT}} \!
\subfloat{\includegraphics[width=0.13\textwidth,height=1.1cm]{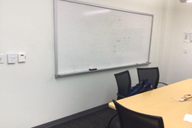}}
\hspace{0.05mm}
\subfloat{\includegraphics[width=0.13\textwidth,height=1.1cm]{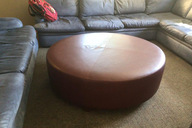}}
\hspace{0.05mm}
\subfloat{\includegraphics[width=0.13\textwidth,height=1.1cm]{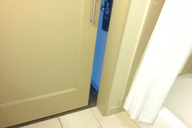}}
\hspace{0.05mm}
\subfloat{\includegraphics[width=0.13\textwidth,height=1.1cm]{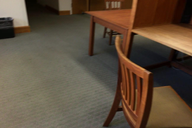}}
\hspace{0.05mm}
\subfloat{\includegraphics[width=0.13\textwidth,height=1.1cm]{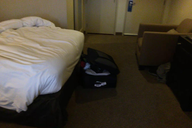}}
\hspace{0.05mm}
\subfloat{\includegraphics[width=0.13\textwidth,height=1.1cm]{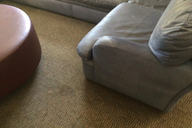}}
\hspace{0.05mm}
\subfloat{\includegraphics[width=0.13\textwidth,height=1.1cm]{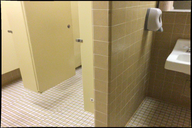}}
\\ \vspace{-1mm}
\rotatebox{90}{\hspace{0.5mm} \tiny{DeFiNe}} \!
\subfloat{\includegraphics[width=0.13\textwidth,height=1.1cm]{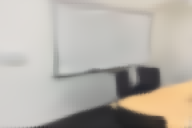}}
\hspace{0.05mm}
\subfloat{\includegraphics[width=0.13\textwidth,height=1.1cm]{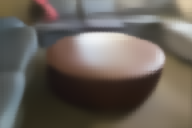}}
\hspace{0.05mm}
\subfloat{\includegraphics[width=0.13\textwidth,height=1.1cm]{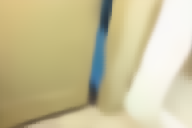}}
\hspace{0.05mm}
\subfloat{\includegraphics[width=0.13\textwidth,height=1.1cm]{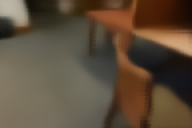}}
\hspace{0.05mm}
\subfloat{\includegraphics[width=0.13\textwidth,height=1.1cm]{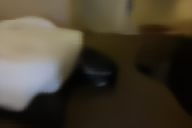}}
\hspace{0.05mm}
\subfloat{\includegraphics[width=0.13\textwidth,height=1.1cm]{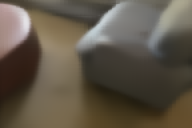}}
\hspace{0.05mm}
\subfloat{\includegraphics[width=0.13\textwidth,height=1.1cm]{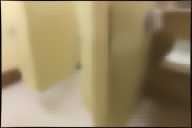}}
\\ \vspace{-1mm}
\rotatebox{90}{\hspace{2.5mm} \tiny{GT}} \!
\subfloat{\includegraphics[width=0.13\textwidth,height=1.1cm]{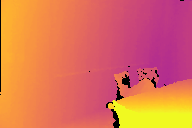}}
\hspace{0.05mm}
\subfloat{\includegraphics[width=0.13\textwidth,height=1.1cm]{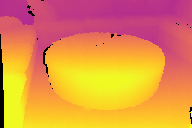}}
\hspace{0.05mm}
\subfloat{\includegraphics[width=0.13\textwidth,height=1.1cm]{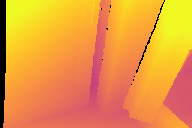}}
\hspace{0.05mm}
\subfloat{\includegraphics[width=0.13\textwidth,height=1.1cm]{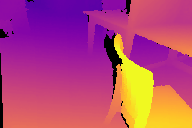}}
\hspace{0.05mm}
\subfloat{\includegraphics[width=0.13\textwidth,height=1.1cm]{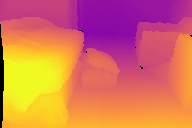}}
\hspace{0.05mm}
\subfloat{\includegraphics[width=0.13\textwidth,height=1.1cm]{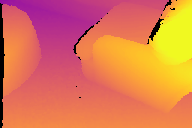}}
\hspace{0.05mm}
\subfloat{\includegraphics[width=0.13\textwidth,height=1.1cm]{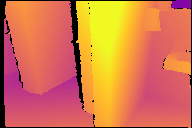}}
\\ \vspace{-1mm}
\rotatebox{90}{\hspace{0.5mm} \tiny{DeFiNe}} \!
\subfloat{\includegraphics[width=0.13\textwidth,height=1.1cm]{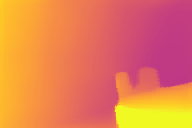}}
\hspace{0.05mm}
\subfloat{\includegraphics[width=0.13\textwidth,height=1.1cm]{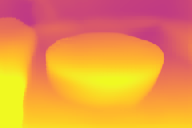}}
\hspace{0.05mm}
\subfloat{\includegraphics[width=0.13\textwidth,height=1.1cm]{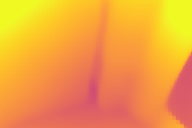}}
\hspace{0.05mm}
\subfloat{\includegraphics[width=0.13\textwidth,height=1.1cm]{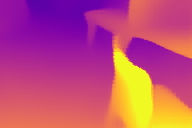}}
\hspace{0.05mm}
\subfloat{\includegraphics[width=0.13\textwidth,height=1.1cm]{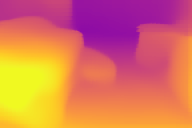}}
\hspace{0.05mm}
\subfloat{\includegraphics[width=0.13\textwidth,height=1.1cm]{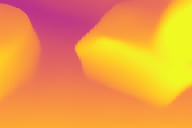}}
\hspace{0.05mm}
\subfloat{\includegraphics[width=0.13\textwidth,height=1.1cm]{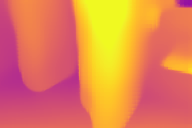}}
\caption{\textbf{Depth estimation and view synthesis results} on ScanNet. Although view synthesis is not our primary goal, it can be achieved with minimal modifications, and we show that it improves depth estimation performance.}
\label{fig:temporal_qualitative}
\end{figure}

%% file: papers/define/figures/scannet_temporal_deepv2d.tex
\begin{table}[t!]
\renewcommand{\arraystretch}{0.95}
\centering
    \begin{tabular}{l|ccc|c}
        \toprule
        \textbf{Method} &
        \small{Abs.\ Rel}$\downarrow$ &
        Sq.\ Rel$\downarrow$ &
        RMSE$\downarrow$ & Speed (ms)$\downarrow$
        \\
\midrule
\midrule
\multicolumn{5}{l}{\textbf{ScanNet test split} \cite{tang2018ba}} \\
\midrule
DeMoN~\cite{ummenhofer2017demon} &
0.231 & 0.520 & 0.761 & 110  \\
MiDas-v2~\cite{ranftl2020towards} & 
0.208 & 0.318 & 0.742 & - \\
BA-Net~\cite{tang2018ba} &
0.091 & 0.058 & 0.223 & 95 \\
CVD~\cite{luo2020consistent} &   
0.073 & 0.037 & 0.217 & 2400 \\
DeepV2D~\cite{deepv2d} &  
0.057 & \textbf{0.010} & \underline{0.168} & 690 \\
NeuralRecon~\cite{Sun_2021_CVPR} & 
\textbf{0.047} & 0.024 & \textbf{0.164} & \textbf{30} \\
\midrule
\textbf{DeFiNe\xspace} ($128 \times 192$)&
0.059 & 0.022 & 0.184 &  \underline{49} \\
\textbf{DeFiNe\xspace} ($240 \times 320$) &
\underline{0.056} & \underline{0.019} & 0.176 &  78 \\
\midrule
\midrule
\multicolumn{5}{l}{\textbf{Zero-shot transfer to 7-Scenes}~\cite{shotton2013scene}} \\
\midrule
DeMoN~\cite{ummenhofer2017demon} &  
0.389 & 0.420 & 0.855 & 110 \\
NeuralRGBD~\cite{liu2019neural} &
0.176 & 0.112 & 0.441 & 202 \\
DPSNet~\cite{im2019dpsnet} &  
0.199 & 0.142 & 0.438  & 322 \\
DeepV2D~\cite{deepv2d} &  
0.437 & 0.553 & 0.869  & 347 \\
CNMNet~\cite{long2020occlusion} &   
0.161 & 0.083 & 0.361 & 80 \\
NeuralRecon~\cite{Sun_2021_CVPR} &  
0.155 & 0.104 & 0.347 & \textbf{30} \\
EST~\cite{yifan2021input} &  
\underline{0.118} & \underline{0.052} & \underline{0.298} & 71 \\
\midrule
\textbf{DeFiNe\xspace} ($128 \times 192$) & 
\textbf{0.100} & \textbf{0.039} & \textbf{0.253} &  \underline{49} \\
\bottomrule
\end{tabular}
\caption{\textbf{Depth estimation results on ScanNet and 7-Scenes}. 
DeFiNe\xspace is competitive with other state-of-the-art methods on ScanNet, and outperforms all published methods in zero-shot transfer to 7-Scenes by a large margin.
}
\label{tab:depth_scannet_temporal}
\end{table}

%% file: papers/define/figures/synthesis.tex
\begin{figure}[t!]
\centering
\begin{subfigure}[t]{0.48\textwidth}
    \centering
    \includegraphics[width=\textwidth, height=3.0cm]{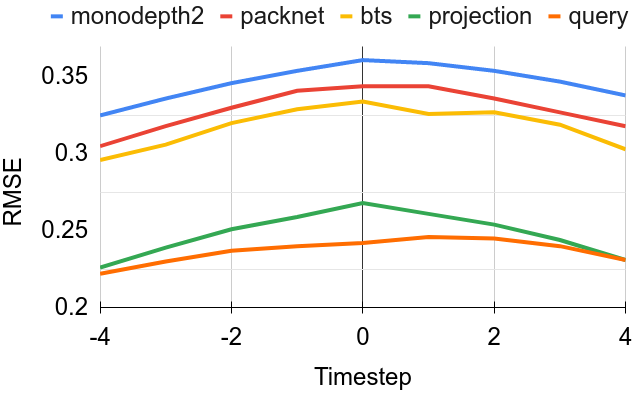}
    \caption{Depth interpolation results.}
    \label{fig:synthesis_interpolation}
\end{subfigure}
\hspace{0.8mm}
\begin{subfigure}[t]{0.48\textwidth}
    \centering
    \includegraphics[width=\textwidth, height=3.0cm]{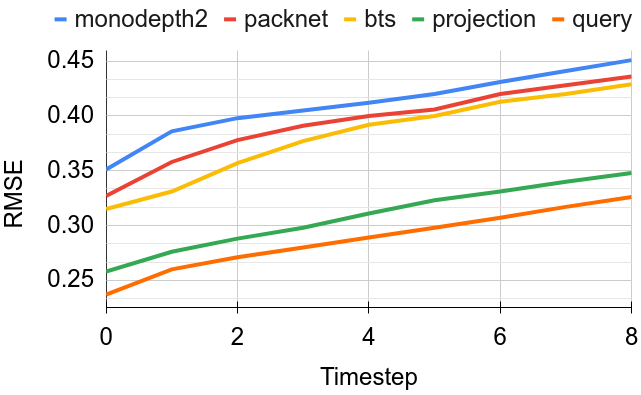}
    \caption{Depth extrapolation results.}
    \label{fig:synthesis_extrapolation}
\end{subfigure}
\\ %
\begin{subfigure}[t]{0.99\textwidth}
    \centering
    \includegraphics[width=\textwidth]{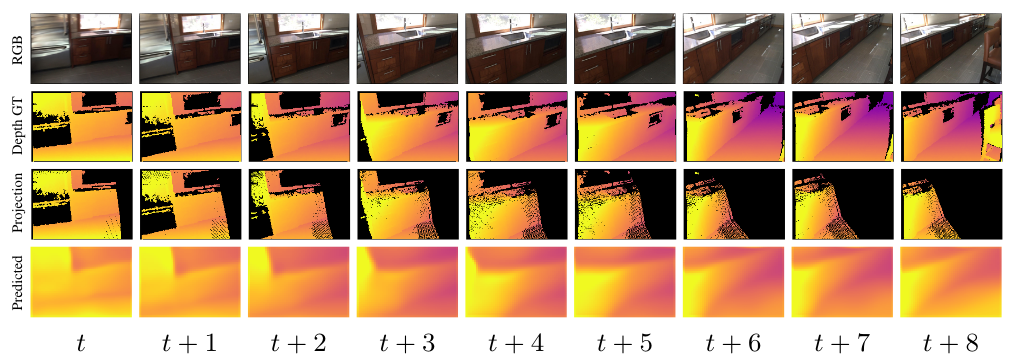}
    \caption{Depth extrapolation to future timesteps. Images and ground-truth depth maps are displayed only for comparison. DeFiNe\xspace can complete unseen portions of the scene in a geometrically-consistent way, generating dense depth maps from novel viewpoints.}
    \label{fig:synthesis_qualitative}
\end{subfigure}
\caption{\textbf{Depth estimation results} from novel viewpoints.}
\label{fig:depth_synthesis}
\end{figure}

%% file: papers/define/sections/05conclusion.tex
We introduced Depth Field Networks\xspace (DeFiNe\xspace), a generalist framework for training multi-view consistent depth estimators. Rather than explicitly enforcing geometric constraints at an architecture or loss level,
we use geometric embeddings to condition network inputs, alongside visual information. To learn a geometrically-consistent latent representation, we propose a series of 3D augmentations designed to promote viewpoint, rotation, and translation equivariance.  
We also show that the introduction of view synthesis as an auxiliary task improves depth estimation without requiring additional ground truth.  
We achieve state-of-the-art results on the popular ScanNet stereo benchmark, and competitive results on the ScanNet video benchmark with no iterative refinement or explicit geometry modeling. We also demonstrate strong generalization properties by achieving state-of-the-art results on zero-shot transfer from ScanNet to 7-Scenes.
The general nature of our framework enables many exciting avenues for future work, including additional tasks such as optical flow, extension to dynamic scenes, spatio-temporal representations, and uncertainty estimation.

%% file: papers/nerfuser/sections/abstract.tex
\begin{figure}[H]
    \centering
    \def\svgwidth{\textwidth}
    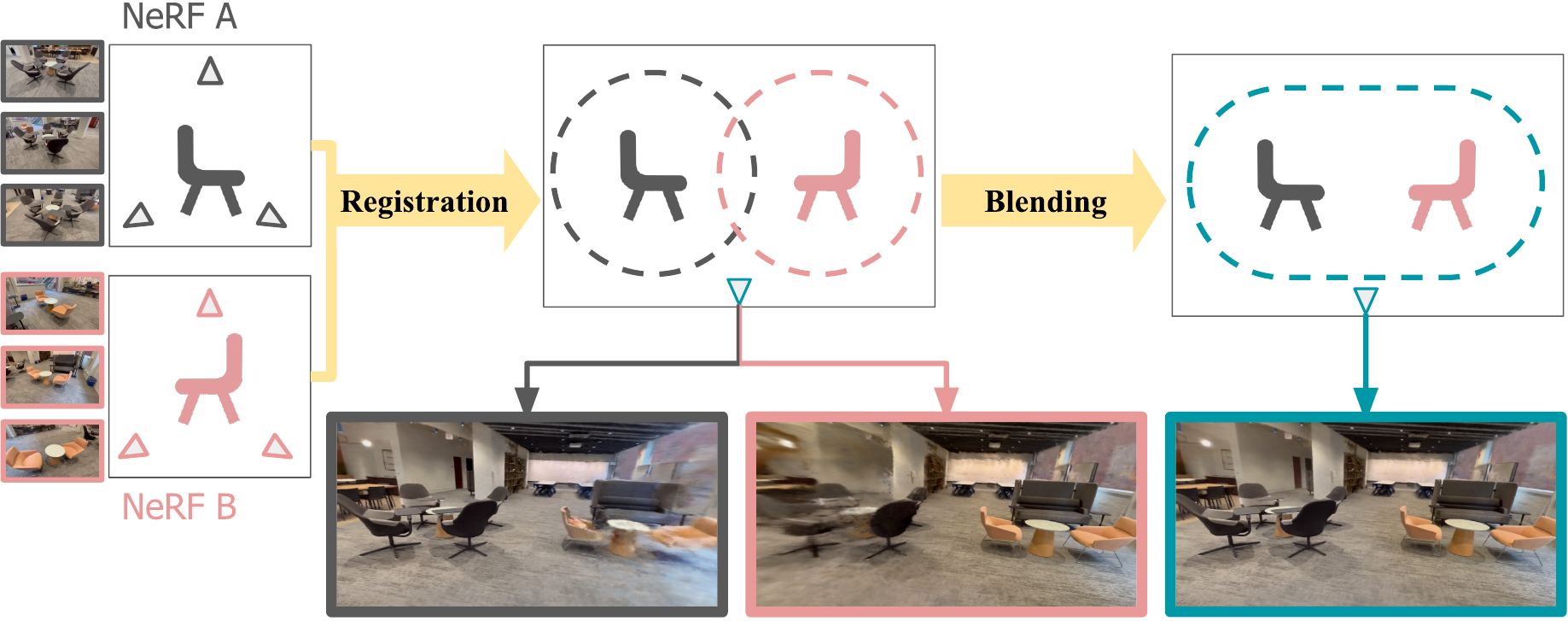
    \caption{Overall framework of NeRFuser}
    \label{fig:nerfuser-teaser}
\end{figure}

We present \textbf{NeRFuser}\space, a novel framework that extends the representational capacity of neural radiance fields (NeRFs) to produce high-fidelity representations of large-scale scenes. %
Integral to our approach is the decomposition of spatially extended environments into a collection of small-scale scenes, each represented by an individual NeRF model (i.e., a sub-map) produced by one or more agents. Critically, we only assume access to these individual NeRFs and \emph{not any} of the original training images or the poses with which they were trained, nor the relative transformations between the NeRFs. 
Towards this goal, NeRFuser/space adopts a two-stage procedure that consists of NeRF registration and NeRF blending.
For registration, we propose \emph{registration from re-rendering}, a technique that infers the transformation between NeRFs based on images synthesized from individual NeRFs, with an effective image quality measure \emph{distant accumulation} for pose filtering.
For blending, we propose \emph{sample-based inverse distance weighting} to blend visual information at the ray sample level. 
We evaluate NeRFuser/space on an indoor dataset and two public benchmarks, showing state-of-the-art results on test views including those that are otherwise challenging to render for individual source NeRFs.

%% file: svg-inkscape/teaser_svg-tex.pdf_tex
\begingroup%
  \makeatletter%
  \providecommand\color[2][]{%
    \errmessage{(Inkscape) Color is used for the text in Inkscape, but the package 'color.sty' is not loaded}%
    \renewcommand\color[2][]{}%
  }%
  \providecommand\transparent[1]{%
    \errmessage{(Inkscape) Transparency is used (non-zero) for the text in Inkscape, but the package 'transparent.sty' is not loaded}%
    \renewcommand\transparent[1]{}%
  }%
  \providecommand\rotatebox[2]{#2}%
  \newcommand*\fsize{\dimexpr\f@size pt\relax}%
  \newcommand*\lineheight[1]{\fontsize{\fsize}{#1\fsize}\selectfont}%
  \ifx\svgwidth\undefined%
    \setlength{\unitlength}{881.5748291bp}%
    \ifx\svgscale\undefined%
      \relax%
    \else%
      \setlength{\unitlength}{\unitlength * \real{\svgscale}}%
    \fi%
  \else%
    \setlength{\unitlength}{\svgwidth}%
  \fi%
  \global\let\svgwidth\undefined%
  \global\let\svgscale\undefined%
  \makeatother%
  \begin{picture}(1,0.39549839)%
    \lineheight{1}%
    \setlength\tabcolsep{0pt}%
    \put(0,0){\includegraphics[width=\unitlength,page=1]{svg-inkscape/teaser_svg-tex.pdf}}%
  \end{picture}%
\endgroup%

%% file: papers/nerfuser/sections/introduction.tex
In order to effectively carry out complex tasks, robots require rich representations of their environments. As such, the robotics community has paid significant attention to the problem of scene reconstruction, particularly with regard to scaling to spatially extended environments~\cite{cadena16}. %
Neural radiance fields (NeRFs)~\cite{neuralfields, Mildenhall2020NeRFRS} provide a powerful means of generating high-fidelity, memory-efficient 3D scene representations from commodity (i.e., low-cost) cameras, though they are typically applied to small table-top scenes. %
In this paper, we seek to extend the representational power of NeRFs to efficiently model large-scale environments. While recent methods have shown some success at modelling large scenes with a single NeRF~\cite{Mller2022InstantNG, Barron2023ZipNeRFAG}, its representation power is fundamentally limited by model capacity when the entire environment is represented by one network, which limits scalability. Instead, our approach achieves scalable NeRF representations by decomposing large environments into a collection of small scenes, each represented by their own individually trained NeRF. Such a decomposition not only improves quality and computational efficiency~\cite{Tancik2022BlockNeRFSL}, but also allows mapping to be distributed among a team of robots and/or human-operated cameras. %

Existing methods that seek to improve the scalability of NeRF models (e.g.,  Block-NeRF~\cite{Tancik2022BlockNeRFSL}) require access to the full set of original training images over which they perform registration to estimate their global poses using a large-scale structure-from-motion (SfM) method (e.g., COLMAP~\cite{schoenberger2016sfm}). Other methods~\cite{Mller2022InstantNG, Barron2023ZipNeRFAG} then perform a lengthy training process to produce a joint NeRF model. Instead, we consider a setting in which a team of people and/or robots %
can individually capture and produce NeRF (or similar) representations of small-scale scenes, and then transfer these models. In addition to their fidelity, NeRFs provide representations that are highly memory efficient. For example, a typical scene may be captured by $300$ images, each about $0.6$\,MB in size. In contrast, NeRF
provides an implicit representation of the scene that takes up approximately $60$\,MB, a $3\times$ reduction from the set of original images. Targeting such use cases, we propose NeRFuser (Fig.~\ref{fig:teaser}), a NeRF fusion framework for the registration and blending of pre-trained NeRFs. NeRFuser requires access only to the originally constructed NeRFs as black boxes, but not the original training images or the training poses, nor the relative transformation between the individual NeRFs.\footnote{We note that training poses are not informative for registration as they are defined with respect to individual NeRF-specific reference frames. Further, while NeRFuser can make use of training poses when available for rendering, we empirically find that images rendered at non-training poses often yield better registration results.} Nonetheless, NeRFuser can register NeRFs (in terms of both pose and scale) and render images from blended NeRFs with greater quality than can be achieved by rendering according to the individual NeRFs.

\begin{figure*}[!t]
    \centering
    \includegraphics[width=\linewidth]{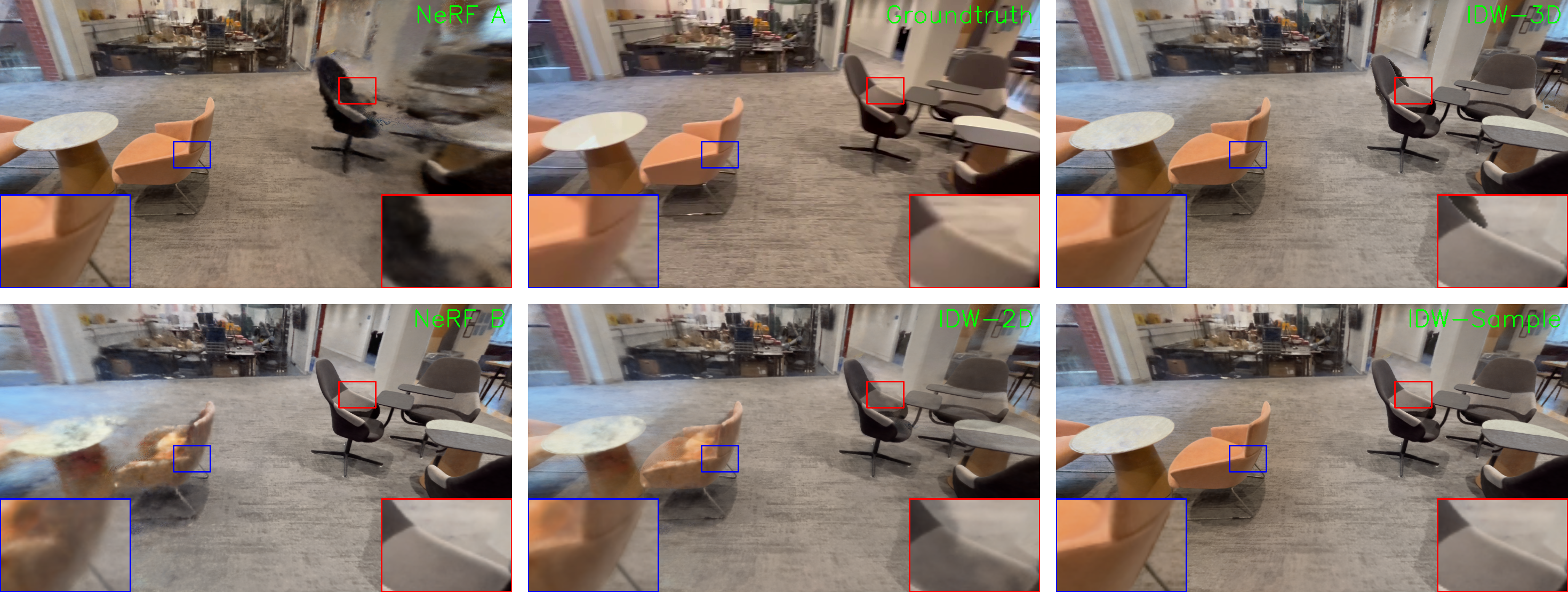}%
    \caption{Qualitative comparison of blending methods. Our proposed \emph{IDW-Sample} produces high-quality blending for both chairs, while baseline methods fail on at least one chair. Notice that the blended results (e.g., \emph{IDW-Sample}) are even sharper than the real test image, which exhibits motion blur, demonstrating an advantage of fusing information from multiple NeRFs.}%
    \label{fig:blending_demo}
\end{figure*}

NeRFuser fuses NeRFs in two steps: \emph{registration} and \emph{blending}. For the first step, we propose \emph{registration from re-rendering}, a technique that takes advantage of the ability of modern NeRFs to synthesize high-quality views to solve for 6-DoF pose and scale (i.e. $\mathrm{SIM3}$) transformation among the individual NeRFs via 2D image matching. A new NeRF image quality measure \emph{distant accumulation} is proposed for pose filtering.
For the second step, we propose a fine-grained sample-based blending technique following inverse-distance-weighting (IDW) principle as proposed in~\cite{Tancik2022BlockNeRFSL}.
In summary, we propose
\begin{enumerate*}[label=(\roman*)]
    \item \textbf{registration from NeRF re-rendering}, a registration method that solves for relative scale and pose to align uncalibrated implicit representations with a new NeRF image quality measure for pose filtering; and
    \item \textbf{sample-based NeRF blending}, a  NeRF blending method to composite predictions at the ray sample level, resulting in image quality better than images rendered by any individual NeRF.
\end{enumerate*}
Code is made publicly available as a Python package.

%% file: papers/nerfuser/sections/related_work.tex
\textbf{Neural Radiance Fields }
A Neural Radiance Field (NeRF)~\cite{Mildenhall2020NeRFRS} is an implicit representation of a 3D scene. It optimizes a neural network composed of MLPs to encode the scene as density and radiance fields, which can be used to synthesize novel views through volumetric rendering. Since its introduction, many follow-up works~\cite{Barron2021MipNeRFAM,Mller2022InstantNG, Barron2021MipNeRF3U, Yu2021PlenoxelsRF, Sun2021DirectVG, Tancik2023NerfstudioAM} have improved over the original implementation.
One line of improvement involves the reconstruction of large-scale NeRFs~\cite{Xiangli2021CityNeRFBN, Zhang2022NeRFusionFR, Tancik2022BlockNeRFSL, Sucar2021iMAPIM, zhu2022nice, Zhu2023NICERSLAMNI, Turki2021MegaNeRFSC, Reiser2021KiloNeRFSU, Tancik2022BlockNeRFSL}. However, most of these works focus on reconstructing the entire scene with a single model. While progressive training~\cite{Xiangli2021CityNeRFBN, Sucar2021iMAPIM} and carefully designed data structures~\cite{Zhang2022NeRFusionFR, zhu2022nice, Zhu2023NICERSLAMNI} have helped to expand the expressivity of a single model, other works~\cite{Reiser2021KiloNeRFSU, Tancik2022BlockNeRFSL} have shown that a collection of many small models can perform better, while maintaining the same number of parameters. Our method provides a novel way to reason over many small models, combining them to improve performance.

\noindent\textbf{NeRF Registration }
NeRFs are optimized from posed images, with poses usually obtained using a structure-from-motion (SfM) method \cite{schoenberger2016sfm, schoenberger2016mvs, sarlin2019coarse, sarlin2020superglue, DeTone2017SuperPointSI, Revaud2019R2D2RA, Dusmanu2019D2NetAT}. Because these methods are scale-agnostic, the resulting coordinate system will have an arbitrary scale specific to each NeRF. Jointly using multiple NeRFs requires \emph{NeRF Registration}, i.e., solving for the relative transformation among their coordinate systems. Note that the setting is different from ``NeRF Inversion''~\cite{yen2021inerf, lin2022parallel} that estimates the 6-DoF camera pose relative to the pre-trained NeRF given an image, a technique that has been used for NeRF-based  localization~\cite{Adamkiewicz2021VisionOnlyRN, Maggio2022LocNeRFMC, Chen2023CATNIPSCA}. However, these tasks can potentially be handled by NeRFuser/space if formulated as NeRF-to-NeRF pose estimation problems. LENS~\cite{moreau2022lens} proposes using NeRF to render extra images to train a per-scene pose regression network, while NeRFuser assumes neither training poses nor training images and does not require training an extra network for each scene. 
Also relevant are works that jointly optimize NeRF representations along with the poses and intrinsics~\cite{lin2021barf, wang2021nerf, jeong2021self}. However, NeRFuser/space only uses SfM on re-rendered images, and does not modify the pre-trained NeRFs themselves.
Of the few works in this emerging field of NeRF registration, most~\cite{Goli2022nerf2nerfPR,Peat2022ZeroNR,Jiang2023RegisteringNR} utilize only geometric cues---extracting surface or density fields from the NeRF representations, and thus do not take full advantage of the rich radiance information. On the other hand, DReg-NeRF~\cite{chen2023dreg} learns to predict 3D correspondences given the input NeRFs, which may suffer from a severe performance drop for out-of-distribution scenes.
Moreover, DReg-NeRF as well as nerf2nerf~\cite{Goli2022nerf2nerfPR}
are only capable of registering two NeRFs at a time, require known scales, and fail catastrophically for unbounded large-scale scenes. nerf2nerf~\cite{Goli2022nerf2nerfPR} further requires a reasonable initialization from human annotations. In contrast, our method can register multiple NeRFs simultaneously including scale recovery, is designed to work on large-scale real-world scenes, and does not involve human annotations.

\noindent\textbf{NeRF Blending}
Given aligned NeRFs from NeRF registration, it remains a problem as how to best merge them, i.e., blending.
While image blending is a highly researched topic in computational photography~\cite{Burt1983TheLP, Brown2003RecognisingP}, there are a few works that discuss blending in terms of NeRFs. Nerflets \cite{Zhang2023NerfletsLR} performs decomposition of the scene with ellipsoids and blend the NeRFs according to their covariance matrices while each NeRF has access to information from all images. Blended-NeRF \cite{Gordon2023BlendedNeRFZO} focuses on generative metrics rather than the reconstruction setting as in ours. More similar to our setting is Block-NeRF~\cite{Tancik2022BlockNeRFSL}, which proposes to blend NeRFs in either image- or pixel-wise manner, similar to traditional 2D image blending techniques. In terms of blending weights, it suggests to use either inverse distance weighting (IDW) or predicted visibility where IDW is eventually adopted because of temporal consistency. Note that the assumption of our paper is that, only the trained NeRFs are provided without their training camera poses, which are required by visibility prediction as they are used to train the visibility network where training poses gives data for calculating ground truth visibility. Thus visibility prediction is not feasible in our setting.

\noindent\textbf{3D Gaussian Splatting}
Compared to NeRF, 3DGS~\cite{kerbl20233d} is a more explicit 3D representation that employs a flexible set of Gaussian blobs to encode the scene geometry and radiance field.
While it is out of scope to compare 3DGS with NeRF, the need for registering and jointly rendering multiple 3DGS similarly applies. Though this work targets the use of NeRFs, some of the proposed techniques may shed inspiration on solving the same problem with this scene representation, especially for the registration.

%% file: papers/nerfuser/sections/methodology.tex
In this section, we describe our NeRF registration method \emph{registration from re-rendering} and our blending technique \emph{IDW-Sample}. Without loss of generality, we discuss the case where two NeRFs $A$ and $B$ are involved, as extension to three or more NeRFs is straightforward, of which the implementation is included in our public code release.

\paragraph{Registration from NeRF Re-rendering}
The first part of our framework is to estimate the relative transformations among the input NeRFs.
We assume that each NeRF is trained with its own set of images, and that individual NeRFs capture different, yet overlapping, areas of a big scene.
Each NeRF may have its own coordinate system that is inconsistent with others. So our goal is to find the transformation $T_{BA}\in\mathrm{SIM}(3)$ that transforms a 3D point $p_B$ in NeRF $B$ to its corresponding point $p_A$ in NeRF $A$ as $p_A = T_{BA} p_B$.

\noindent\textbf{Overview}~
Since NeRFs are known for high quality novel view synthesis, we strategically sample a set of poses and use them as local poses to query each input NeRF to get re-rendered images. We then re-purpose off-the-shelf structure-from-motion methods ~\cite{DeTone2017SuperPointSI,sarlin2020superglue} on the union of re-rendered images. Intuitively, the SfM step tries to recover the poses of all re-rendered images in a shared coordinate system. Note that the pose of each re-rendered image in its source NeRF's local coordinate system is sampled, hence known, so we can solve the transformation from the shared SfM coordinate system to each NeRF's local one. The relative transformation between NeRFs can then be inferred. 
Detailed derivation of how $\mathrm{SIM}(3)$ transformations among coordinate systems can be recovered from $\mathrm{SE}(3)$ poses of individual camera images is presented in the supplementary material.

\noindent\textbf{Sampling strategy of poses for re-rendering}~
Though we do not assume access to the exact poses of images used in generating the input NeRFs, we do assume that these poses are pre-processed in a standardized way~\cite{nerfstudio}:
\begin{enumerate*}[label=(\roman*)]
    \item centered so that the mean translation vector becomes the origin;\label{preprocess:1}
    \item rotated so that the mean up direction is aligned with $z$ axis;\label{preprocess:2}
    \item uniformly scaled so that the largest range of pose locations along an axis is $[-1, 1]$.\label{preprocess:3}
\end{enumerate*}
Accordingly, our strategy is to uniformly sample poses mainly on the upper hemisphere (due to \ref{preprocess:2}) of radius $1$ (due to \ref{preprocess:3}) located at the origin (due to \ref{preprocess:1}). This pose sampling strategy is simple yet effective given no access to NeRFs' training poses.

\begin{figure}[ht]
    \centering
    \begin{subfigure}[b]{0.32\textwidth}
         \centering
         \fbox{\includegraphics[width=\textwidth]{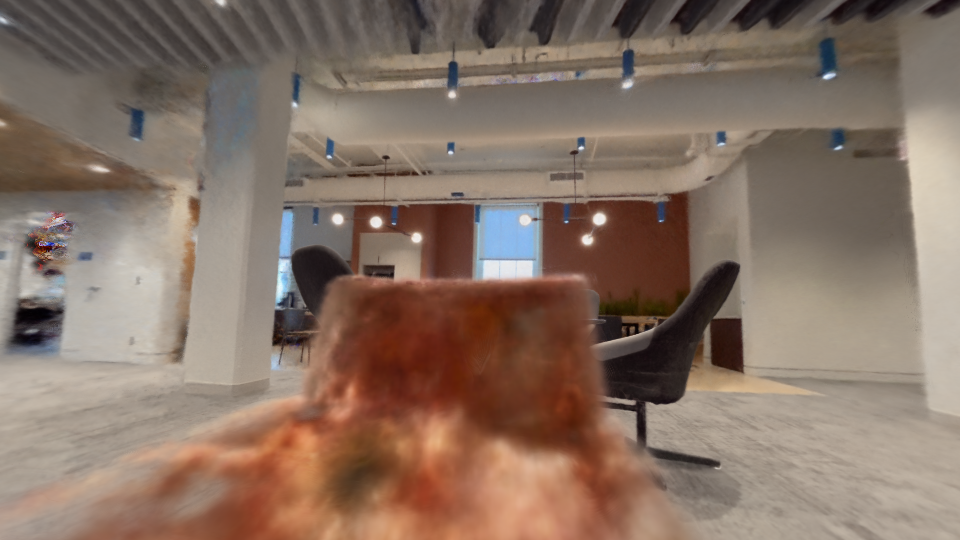}}
     \end{subfigure}
     \hfill
     \begin{subfigure}[b]{0.32\textwidth}
         \centering
         \fbox{\includegraphics[width=\textwidth]{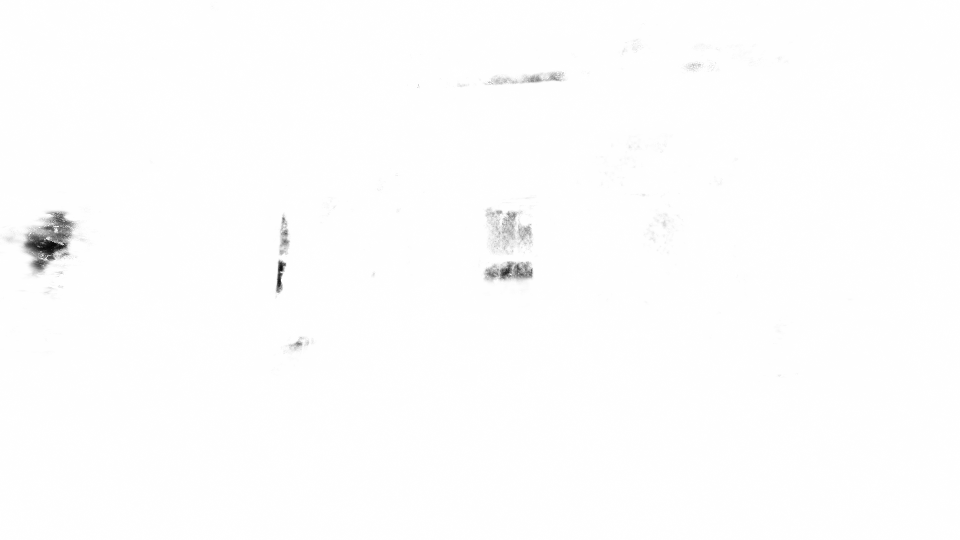}}
     \end{subfigure}
     \hfill
     \begin{subfigure}[b]{0.32\textwidth}
         \centering
         \fbox{\includegraphics[width=\textwidth]{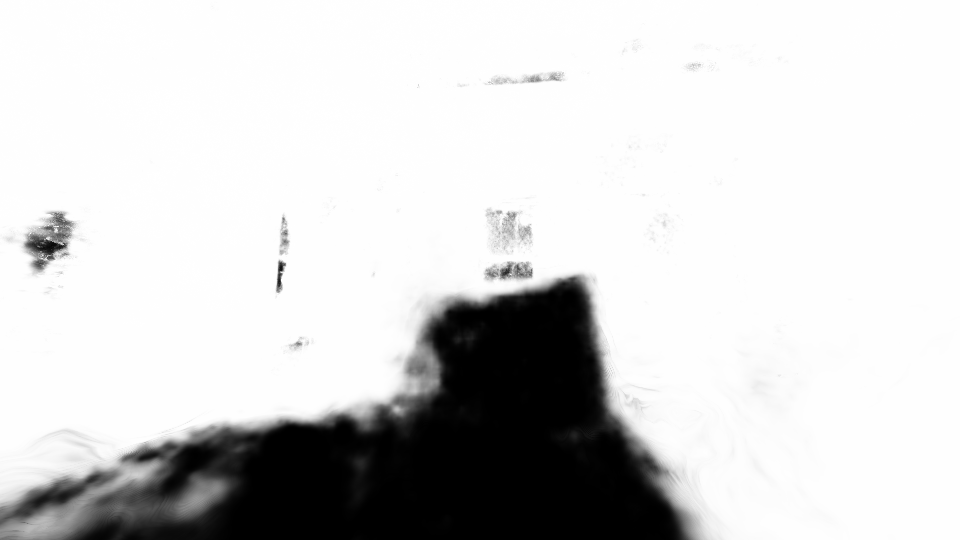}}
     \end{subfigure}
     \\
     \begin{subfigure}[b]{0.32\textwidth}
         \centering
         \fbox{\includegraphics[width=\textwidth]{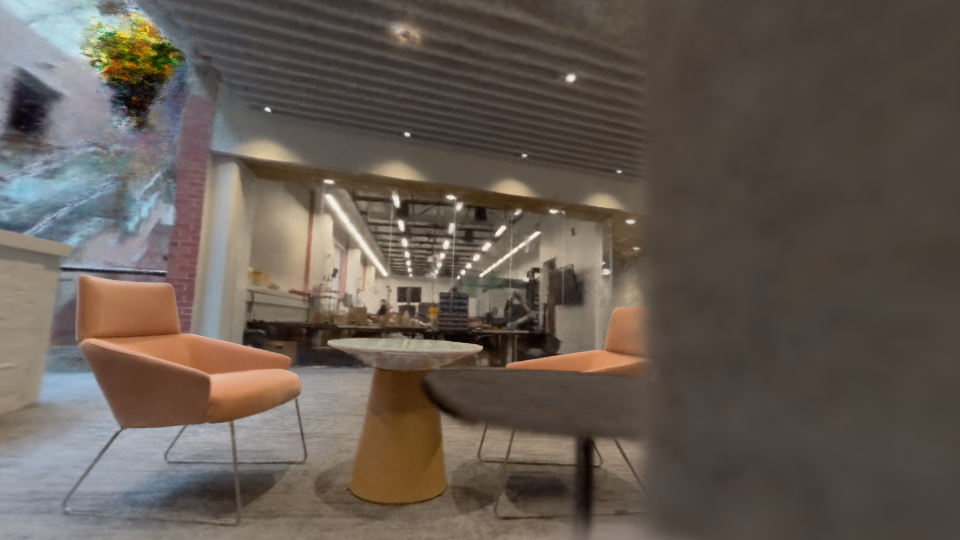}}
     \end{subfigure}
     \hfill
     \begin{subfigure}[b]{0.32\textwidth}
         \centering
         \fbox{\includegraphics[width=\textwidth]{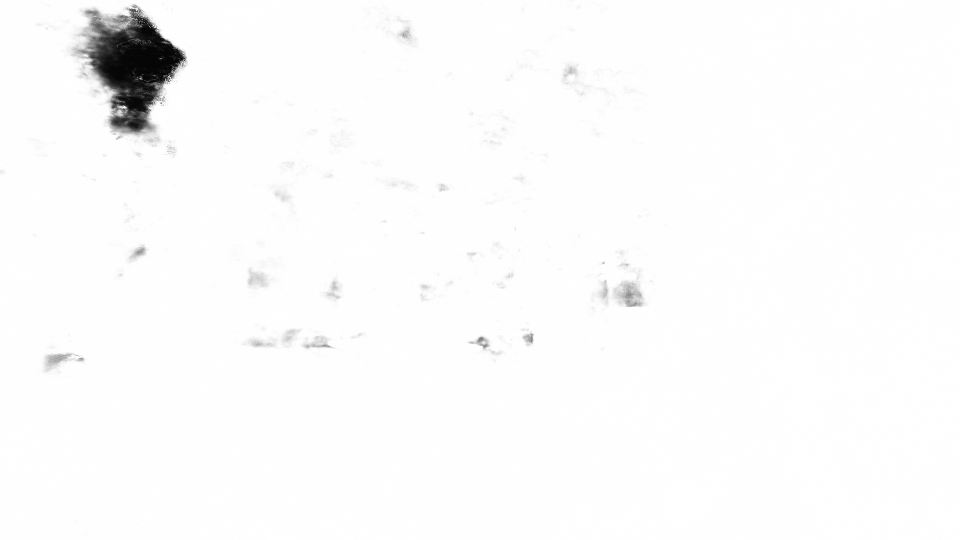}}
     \end{subfigure}
     \hfill
     \begin{subfigure}[b]{0.32\textwidth}
         \centering
        \fbox{\includegraphics[width=\textwidth]{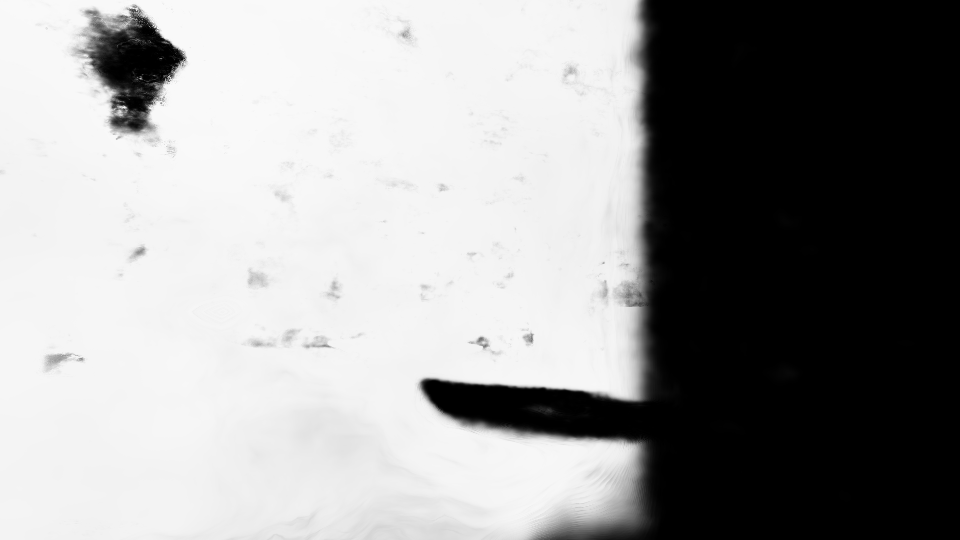}}
     \end{subfigure}
     \\
     \begin{subfigure}[b]{0.32\textwidth}
         \centering
         \fbox{\includegraphics[width=\textwidth]{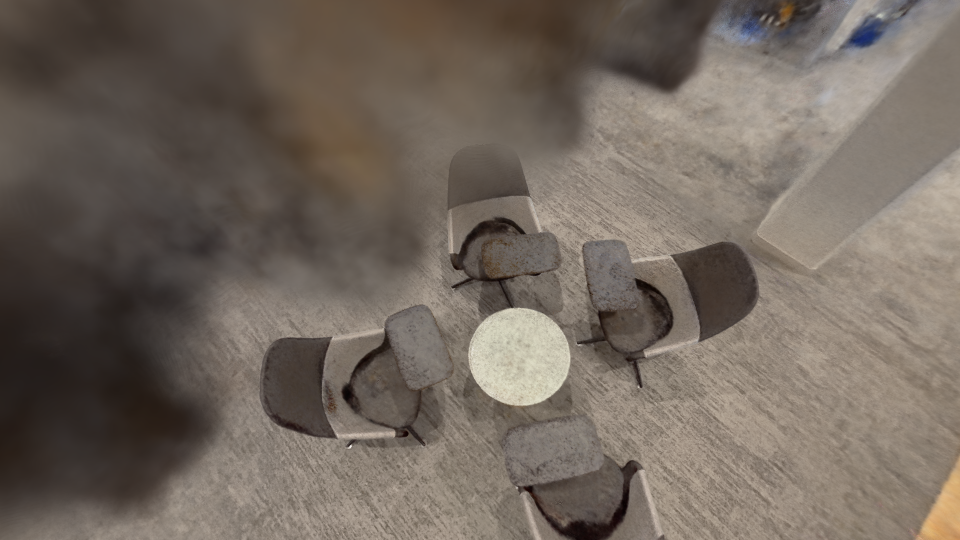}}
         \caption{RGB rendering}
         \label{col:rgb}
     \end{subfigure}
     \hfill
     \begin{subfigure}[b]{0.32\textwidth}
         \centering
         \fbox{\includegraphics[width=\textwidth]{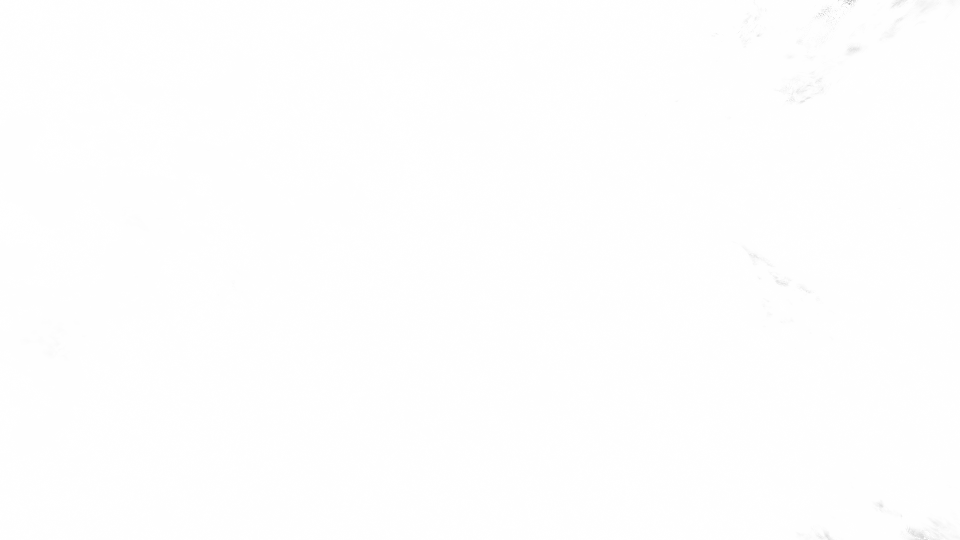}}
         \caption{{\scriptsize regular accumulation}}
         \label{col:regular_accumulation}
     \end{subfigure}
     \hfill
     \begin{subfigure}[b]{0.32\textwidth}
         \centering
        \fbox{\includegraphics[width=\textwidth]{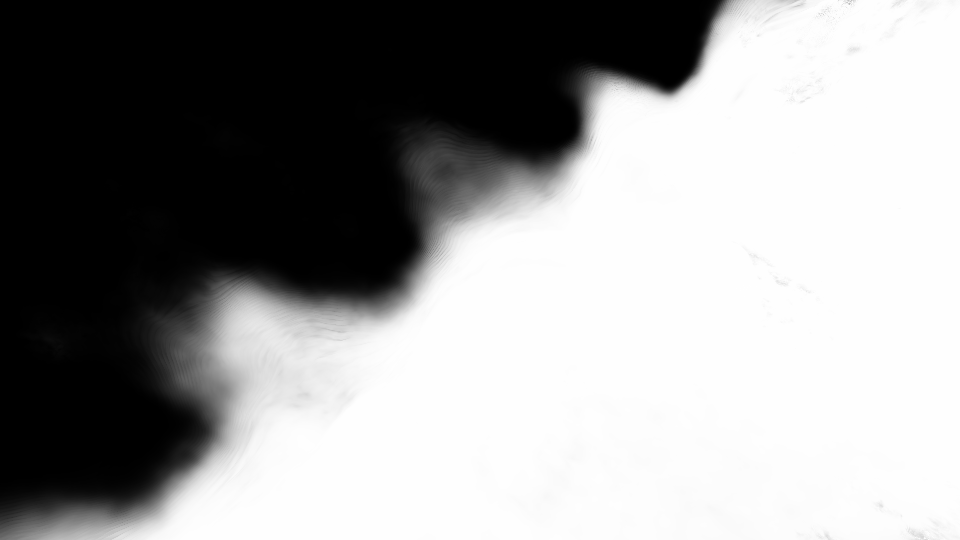}}
        \caption{{\scriptsize distant accumulation}}
        \label{col:distant_accumulation}
     \end{subfigure}
\caption{Illustration of NeRF renderings and their accumulations. Column (a) shows RGB renderings. Columns (b) and (c) show the gray-scale plots for regular and distant accumulation, respectively, where white means high accumulation. The visualizations show that regular accumulations are mostly close to $1.0$ regardless of the various rendering quality, while the distant accumulation clearly discriminates poorly rendered regions.}
\label{fig:distant_accumulation}
\end{figure}

\noindent\textbf{Distant Accumulation for Pose Filtering}~
Some of the renderings using the poses sampled from the above procedure can be of poor quality nonetheless (Figure \ref{col:rgb}). Although the SfM procedure can be robust to poor renderings to some extent (through its feature points extraction and matching), it is beneficial to filter them out to reduce processing time and improve accuracy. Computed from NeRF's volumetric rendering, accumulation can be used to measure NeRF ray uncertainty~\cite{Snderhauf2022DensityawareNE}, since low accumulation values usually correspond to under-trained pixels. However, we find that regular accumulation does not discriminate poor renderings (Figure \ref{col:regular_accumulation}). While it does spot some under-trained pixels, the mean accumulation over the entire image is too close to $1.0$, which is too sensitive to noise for filtering. We further observe that,
\begin{enumerate*}[label=(\roman*)]
    \item when a ray is under-trained, the termination probability mass tends to spread evenly;
    \item cameras inside or very close to an object usually bring poor renderings.
\end{enumerate*}
Based on the observations, we propose a new NeRF ray quality measure \emph{Distant Accumulation}
\begin{equation}
    q_d = \int_d^\infty{}T(t)\sigma(t)\mathrm{d}t,
\end{equation}
where $T(t),\sigma(t)$ denotes transmittance and density respectively, as introduced in~\cite{Mildenhall2020NeRFRS}. Note that distant accumulation $q_d$ is equivalent to regular accumulation when $d=0$. We use distant accumulation $q_d$ averaged over all pixels as the image quality measure, where $d$ is set to be a moderate value within $[0,1]$ (recall that the NeRF's coordinate system is assumed to be normalized in scale; $d=0.3$ is used in Figure~\ref{col:distant_accumulation}). The idea is that,
\begin{enumerate*}[label=(\roman*)]
    \item when a ray is under-trained, evenly spread termination probability results in $q_d$ being substantially smaller than $1.0$;
    \item when a camera is inside or too close to an object, most rays will have high termination probability at a close distance, also resulting in a small $q_d$.
\end{enumerate*}
As shown in Figure \ref{col:distant_accumulation}, our proposed measure clearly discriminates against poor renderings. We use the mean distant accumulation to filter out poor renderings before applying the SfM procedure.

\noindent\textbf{Robustness to SfM errors}~
As our proposed registration method works on top of SfM, we discuss its robustness to SfM errors, which are manifested in two ways:
\begin{enumerate*}[label=(\roman*)]
    \item failure to recover poses for some images\label{sfm_error:1} and
    \item returning erroneous poses for some others.\label{sfm_error:2}
\end{enumerate*}
For \ref{sfm_error:1}, to compute the transformation from the shared SfM coordinate system to a NeRF's local one, our method only requires that the poses of as few as two re-rendered images from that NeRF are correctly recovered by SfM (required when computing the relative scale using distances of pose pairs; see Equation (13) in supplementary material). For \ref{sfm_error:2}, our method recovers a list of candidate results for both scale and transformation (details in supplementary material), so it boils down to solving a constrained optimization problem to obtain the final result. Computing the mean would produce the optimal estimation under Gaussian noise assumption following maximum likelihood estimation principle. In practice, median is computed instead for more robustness to outliers.

\paragraph{Sampled-based NeRF Blending}

NeRF blending aims to combine predictions from the input NeRFs in pursuit of high-quality novel view synthesis. 
There are two key questions to consider:
\begin{enumerate}[label=(\roman*)]
    \item\textbf{What to blend}: At what granularity should we blend the information?\label{itm:blend-what}
    \item\textbf{How to blend}: How to compute the blending weights?\label{itm:blend-how}
\end{enumerate}

Block-NeRF~\cite{Tancik2022BlockNeRFSL} answers \ref{itm:blend-how} with inverse-distance-weighting (IDW) and predicted visibility weighting. IDW determines the contribution of each NeRF according to
$w_i \propto d_i^{-\gamma}$,
where $d_i$ is some notion of distance between the center of NeRF $i$ to the element in question (answered by \textbf{What to blend}), $\gamma\in\mathbb{R}^+$ is a hyper-parameter that modulates the blending rate. 
On the other hand, visibility weighting tries to weight the element by its visibility from the camera views during NeRF training. However, a visibility prediction network is required to be trained jointly with the NeRF and used during inference, which is a requirement that we can not assume for typical NeRFs.
Despite slightly better visual evaluation metrics, due to temporal consistency issues with visibility weighting, Block-NeRF presents their demo with IDW weighting (IDW-2D, explained below).

For question \ref{itm:blend-what}, Block-NeRF answers it with image-wise and pixel-wise blending. For image-wise blending, all pixels in an image have the same blending weights. When combined with IDW (IDW-2D), the weights are computed based on the distance between the camera center and NeRF centers. For pixel-wise blending, each pixel has its own blending weight. When combined with IDW, the weights are computed from the distance between the expected point of ray termination for each pixel and NeRF centers. These techniques naturally inherit from image blending methods. However, as NeRF is inherently a 3D representation whose computing happens at the ray sample level, keep using 2D approaches deems sub-optimal.

\begin{figure}[!t]
    \centering
    \def\svgwidth{0.9\linewidth}
    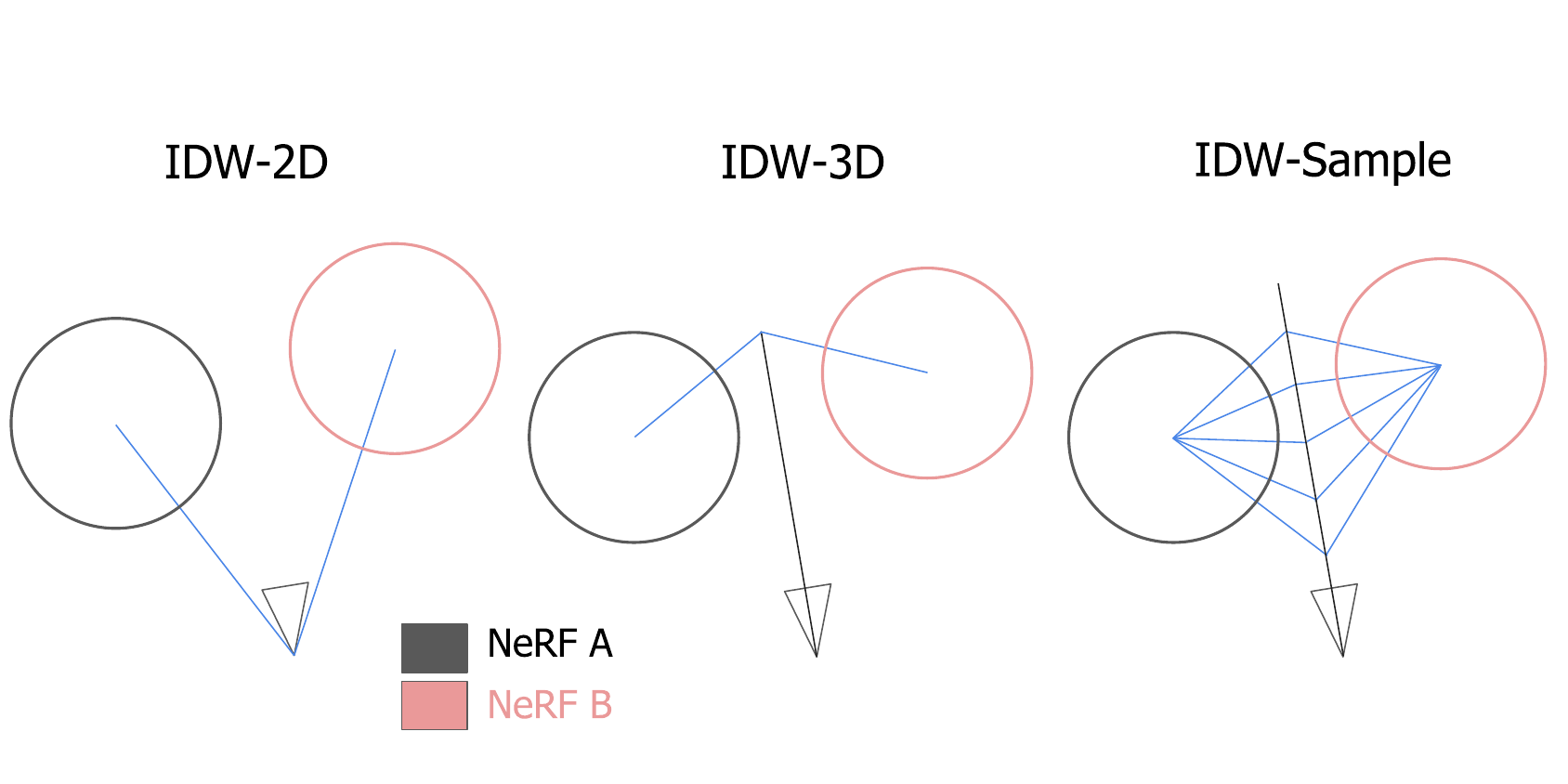
    \caption{Illustration of IDW-based blending methods: IDW-2D, IDW-3D and IDW-Sample.}
    \label{fig:idw}
\end{figure}

In this paper, recognizing the fact that the color of a pixel is computed using samples along the ray in NeRF during volumetric rendering, we answer \ref{itm:blend-what} by proposing a novel sample-based blending method. On the other hand, although IDW may not be the best measure of quality, it is nevertheless a good approximation under the assumption of surrounding camera view distribution. Thus we still follow the IDW principle to answer \ref{itm:blend-how} but adapt it by calculating the distance between the ray sample and NeRF centers, achieving a finer blending result in principle. Moreover, the weighting is irrespective of depth quality, which can be an issue for NeRF and thus the major downside of IDW-3D. Since we use IDW with sample-based blending, we coin our method \emph{IDW-Sample}. Figure~\ref{fig:idw} provides an illustration of the comparison between different methods.

\paragraph{Sample-wise Blending with IDW}

During NeRF's volumetric rendering stage, a pixel's color is computed using samples along the ray. Recognizing this fact, we propose a sample-wise blending method that calculates the blending weights for each ray sample using IDW. We show that the original volumetric rendering methodology can be easily extended to take advantage of these new sample-wise blending weights, resulting in our proposed \emph{IDW-Sample} strategy.\\

\noindent\textbf{Proximity Test}~
The first question to ask before blending is always whether it needs blending, as NeRFs can only render with high quality within their effective range. Using distant NeRFs for rendering, whose quality is poor, can only be harmful. Hence, we introduce a \emph{Proximity Test} to decide which NeRFs are relevant for the given rendering pose.

As we located each NeRF in a shared coordinate frame from the NeRF registration phase, we can calculate the distances between the given rendering camera position to all NeRFs and sort them in an increasing order $d_0<d_1<\dots<d_{N-1}$, assuming there are $N$ NeRFs to blend. We then divide them by the closest distance $d_0$ to have a set of distance ratios $(1=\frac{d_0}{d_0}<\frac{d_1}{d_0}<\dots<\frac{d_{N-1}}{d_0})$. A threshold $\tau$ is set to filter out NeRFs with distance ratio larger than it (meaning that they are too far away). Note that this threshold $\tau$ can be set in an adaptive way, for example keeping only NeRFs with $k$-smallest or $k$-th quantile $\tau$ for blending. So after the filtering, the computational complexity for NeRF blending is $\mathcal{O}(k)$.

\noindent\textbf{Merge Ray Samples}~
Consider a pixel to be rendered, which gets unprojected into a ray. Since ray samples are separately proposed according to the density field of each source NeRF, we need to merge them into a single set.
Given samples $\{(t^A_k,\delta^A_k)\}_k$ and $\{(t^B_k,\delta^B_k)\}_k$ proposed from NeRF $A$ and NeRF $B$, respectively, we merge them into a single set of ray samples $\{(\bar{t}_k, \bar{\delta}_k)\}_k$ by taking the sample location $t$ and length $\delta$, as illustrated by Figure 1 in the supplementary material. We update the termination probability and color of each new sample in the merged set for each source NeRF. Given a ray sample proposed by a NeRF, we assume its termination probability mass is uniformly distributed over its length, while its color is the same for any point within coverage.

\noindent\textbf{Blending Process}~
We use IDW to compute the blending weight for each sample. Specifically, let $\bm{x}_i$ be the origin of $\textrm{NeRF}_i$ for $i\in\{A,B\}$, $\bm{o}$ be the camera's optical center, $\bm{r}=(\bm{o},\bm{d})$ be the ray corresponding to pixel $j$ to be rendered, and $(\bar{t}_k,\bar{\delta}_k)$ be a ray sample from the merged samples set, with updated termination probability mass $\bar{p}_{i,k}$ and color $\bar{\bm{c}}_{i,k}$ for each $\textrm{NeRF}_i$. We compute its blending weight as $w_{i,k}\propto{}{d_{i,k}}^{-\gamma}$, where $d_{i,k}=\|\bm{x}_i-(\bm{o}+\bar{t}_k\bm{d})\|_2$.
The blended pixel $j$ is
\begin{equation}
    I^{(j)}=\sum_k\sum_iw_{i,k}\bar{p}_{i,k} \bar{\bm{c}}_{i,k}
\end{equation}
Weights $w_{i,k}$ are normalized following two steps in order:
\begin{enumerate*}[label=(\roman*)]
    \item$\sum_iw_{i,k}=1,\;\forall k$; and \label{itm:normalize1}
    \item$\sum_k\sum_iw_{i,k}\bar{p}_{i,k}=1$.\label{itm:normalize2}
\end{enumerate*}
Step~\ref{itm:normalize1} means that our method does not change the relative weighting of samples along a given ray, which is already dictated by the termination probability. Step~\ref{itm:normalize2} ensures that the rendered pixel has a valid color. 

\input{papers/nerfuser/tables/ttic_psnr}

%% file: svg-inkscape/idw-compare_svg-tex.pdf_tex
\begingroup%
  \makeatletter%
  \providecommand\color[2][]{%
    \errmessage{(Inkscape) Color is used for the text in Inkscape, but the package 'color.sty' is not loaded}%
    \renewcommand\color[2][]{}%
  }%
  \providecommand\transparent[1]{%
    \errmessage{(Inkscape) Transparency is used (non-zero) for the text in Inkscape, but the package 'transparent.sty' is not loaded}%
    \renewcommand\transparent[1]{}%
  }%
  \providecommand\rotatebox[2]{#2}%
  \newcommand*\fsize{\dimexpr\f@size pt\relax}%
  \newcommand*\lineheight[1]{\fontsize{\fsize}{#1\fsize}\selectfont}%
  \ifx\svgwidth\undefined%
    \setlength{\unitlength}{810.14172363bp}%
    \ifx\svgscale\undefined%
      \relax%
    \else%
      \setlength{\unitlength}{\unitlength * \real{\svgscale}}%
    \fi%
  \else%
    \setlength{\unitlength}{\svgwidth}%
  \fi%
  \global\let\svgwidth\undefined%
  \global\let\svgscale\undefined%
  \makeatother%
  \begin{picture}(1,0.49991253)%
    \lineheight{1}%
    \setlength\tabcolsep{0pt}%
    \put(0,0){\includegraphics[width=\unitlength,page=1]{svg-inkscape/idw-compare_svg-tex.pdf}}%
  \end{picture}%
\endgroup%

%% file: papers/nerfuser/tables/ttic_psnr.tex
\begin{table*}[h!]
    \centering
    \footnotesize
    \begin{tabularx}{\linewidth}{Xcccccc} 
        \toprule
        \multirow{2}{*}{Blending} & \multicolumn{3}{c}{Ground-truth $T_{BA}$} & \multicolumn{3}{c}{Estimated $\hat{T}_{BA}$} \\
        \cmidrule{2-7}
        & PSNR $\uparrow$ & SSIM $\uparrow$ & LPIPS $\downarrow$ & PSNR $\uparrow$ & SSIM $\uparrow$ & LPIPS $\downarrow$ \\ 
        \midrule
        NeRF & 20.92 & 0.716 & 0.369 & 20.90 & 0.714 & 0.370 \\
        Nearest (Block-NeRF) & 23.81 & 0.779 & 0.283 & 23.68 & 0.774 & 0.287 \\ 
        IDW-2D (Block-NeRF) & 24.70 & 0.795 & 0.267 & 24.64 & 0.792 & 0.267 \\
        IDW-3D (Block-NeRF) & 23.48 & 0.776 & 0.279 & 23.45 & 0.772 & 0.280 \\
        IDW-Sample (Ours) & \textbf{24.91} & \textbf{0.813} & \textbf{0.228} & \textbf{24.83} & \textbf{0.810} & \textbf{0.229} \\
        \bottomrule
    \end{tabularx}
    \caption{Blending results on Object-Centric Indoor Scenes. \emph{IDW-Sample} works the best for all metrics with both ground-truth and estimated transformations. Results with estimated $\hat{T}_{BA}$ are only marginally worse than those with ground-truth $T_{BA}$, which demonstrates that our proposed NeRF registration is accurate enough for the downstream blending task.}
    \label{tab:nf_bld}
\end{table*}

%% file: papers/nerfuser/sections/experiments.tex
In this section, we describe our registration and blending experiments on our self-collected object-centric indoor dataset as well as two public benchmarks: ScanNet \cite{Dai2017ScanNetR3} and MissionBay \cite{Tancik2022BlockNeRFSL}. While our framework is agnostic to base NeRF implementation, in our experiments, we use NeRFacto \cite{nerfstudio} as the base NeRF model. All NeRF training hyper-parameters follow the default settings except that camera optimization is turned off to avoid affecting registration evaluation. Detailed settings of the datasets can be found in supplementary results.

\paragraph{Datasets}
\noindent\textbf{Object-Centric Indoor Scenes}~
\label{subsec:ttic-data}
We created a dataset consisting of three indoor scenes, using an iPhone $13$ mini in video mode. Each scene consists of three video clips---we choose two objects in each scene, and collect two (overlapping) video sequences that focus on each object. We collect a third (test) sequence that observes the entire scene as the test set.
We test NeRFuser on this dataset and report results of both registration and blending in Section~\ref{subsec:nerf-fusion}.

\begin{figure*}[!t]
    \centering
    \begin{tabular}{ccc}
        \includegraphics[scale=0.08]{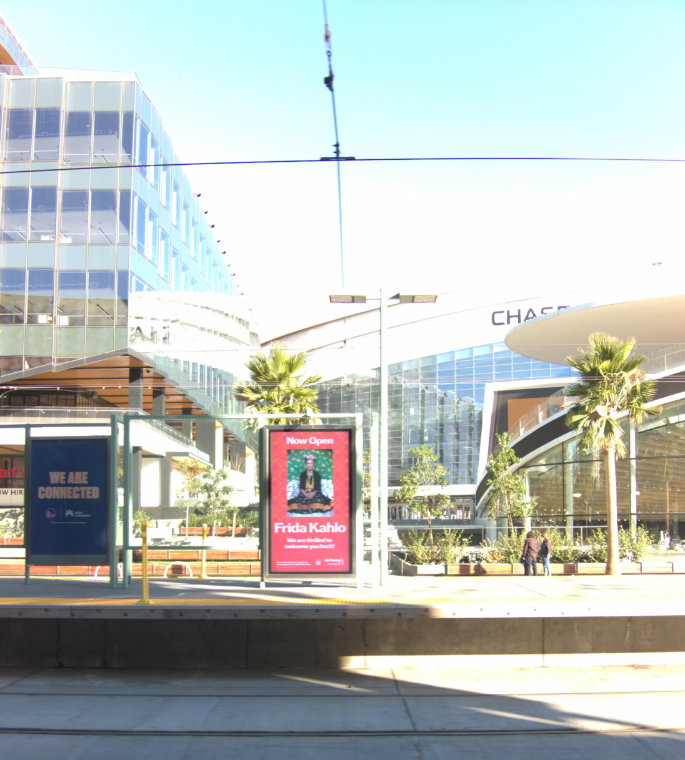} & \includegraphics[scale=0.08]{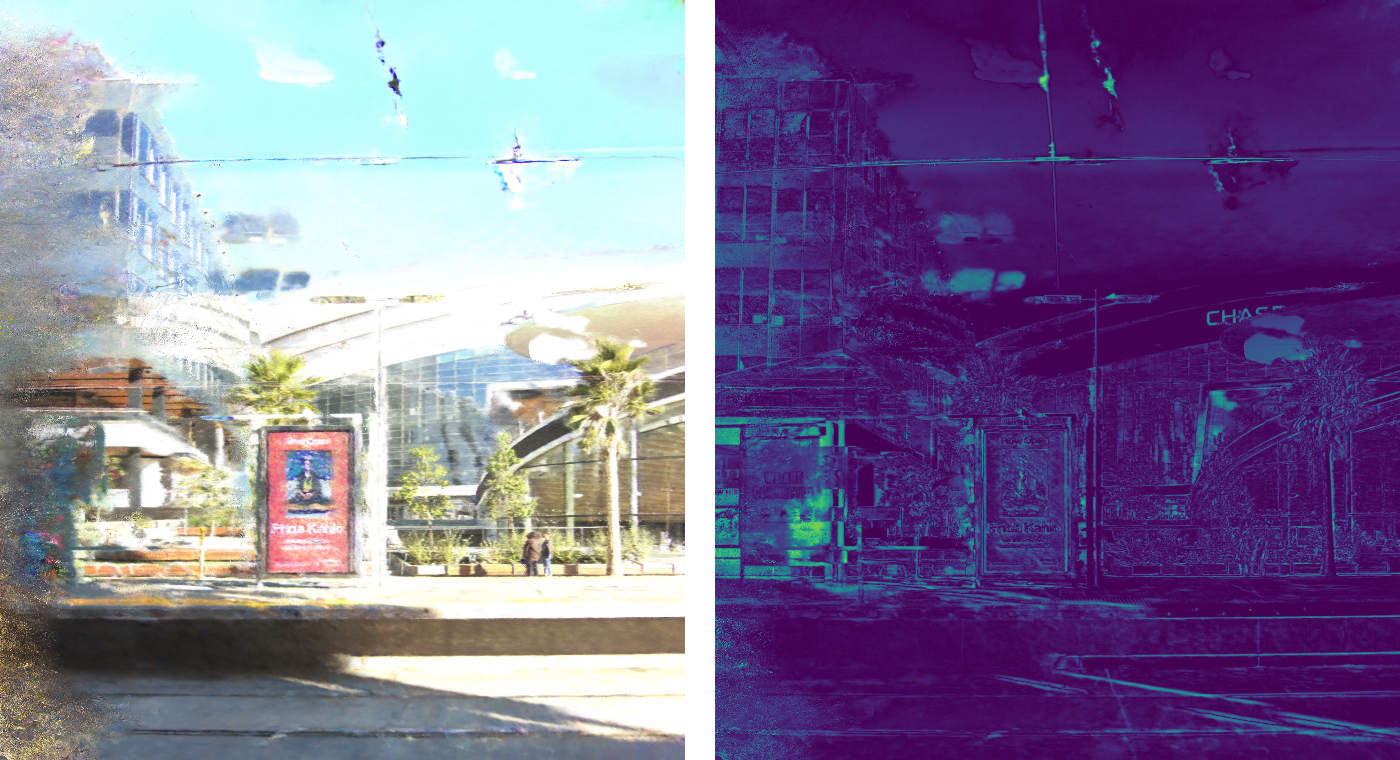} & \includegraphics[scale=0.08]{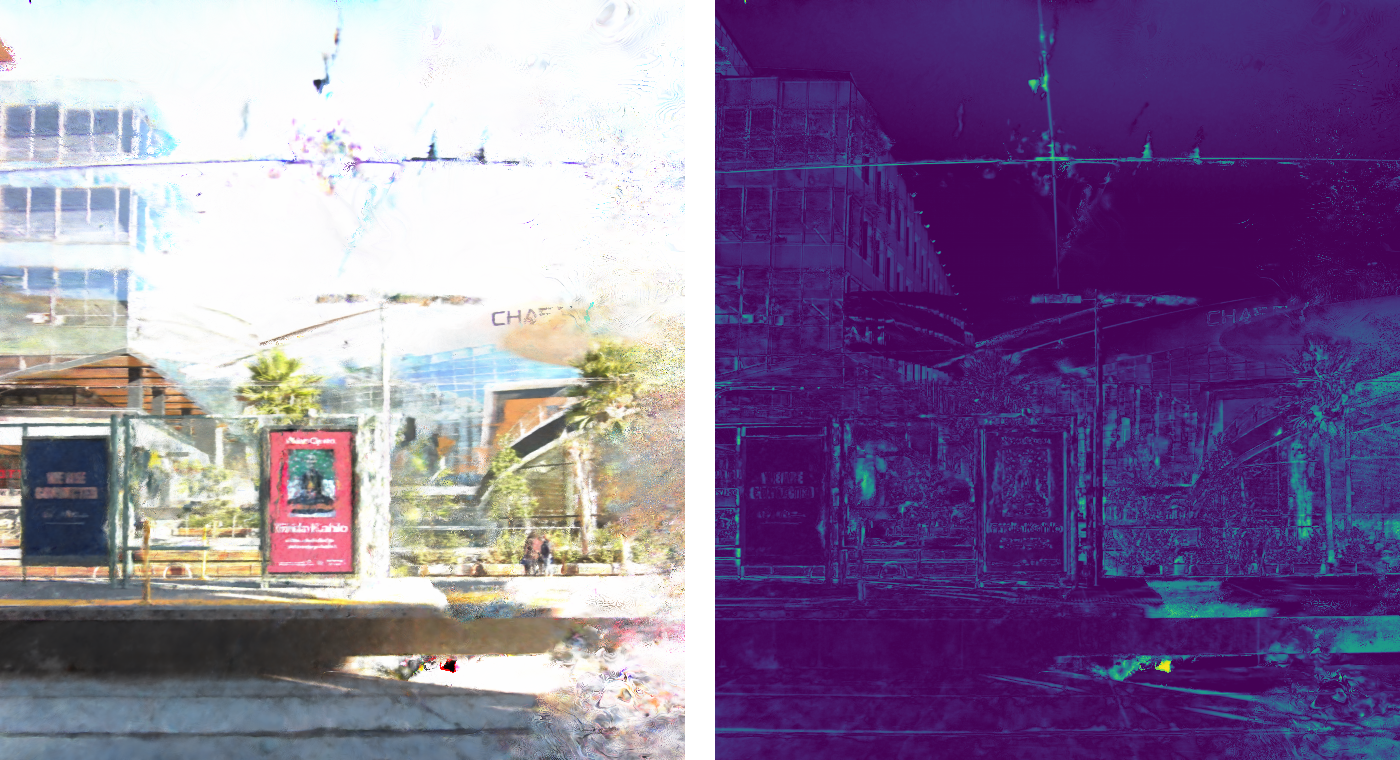}\\
        \cmidrule(lr){1-1}\cmidrule(lr){2-2}\cmidrule(lr){3-3}
        Ground-truth & NeRF A & NeRF B \\
        \includegraphics[scale=0.08]{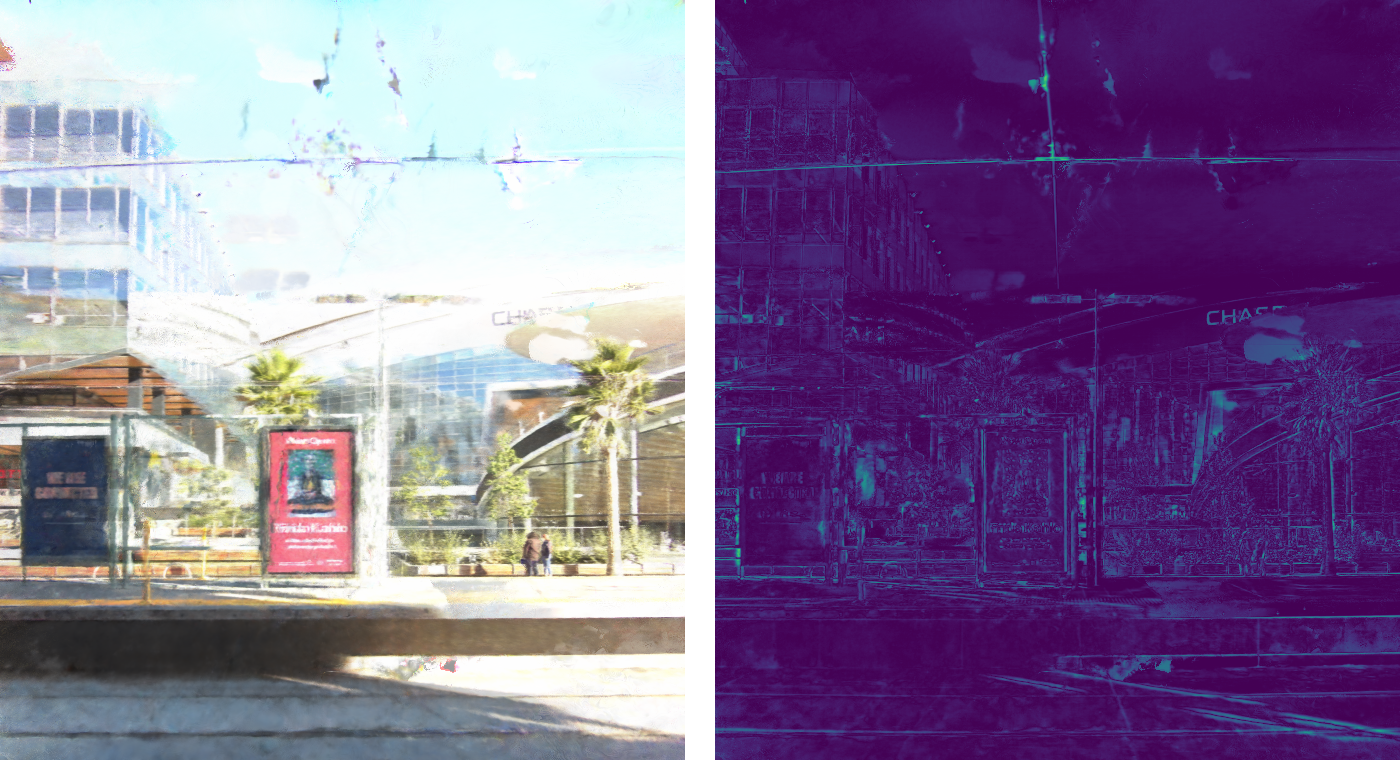} & \includegraphics[scale=0.08]{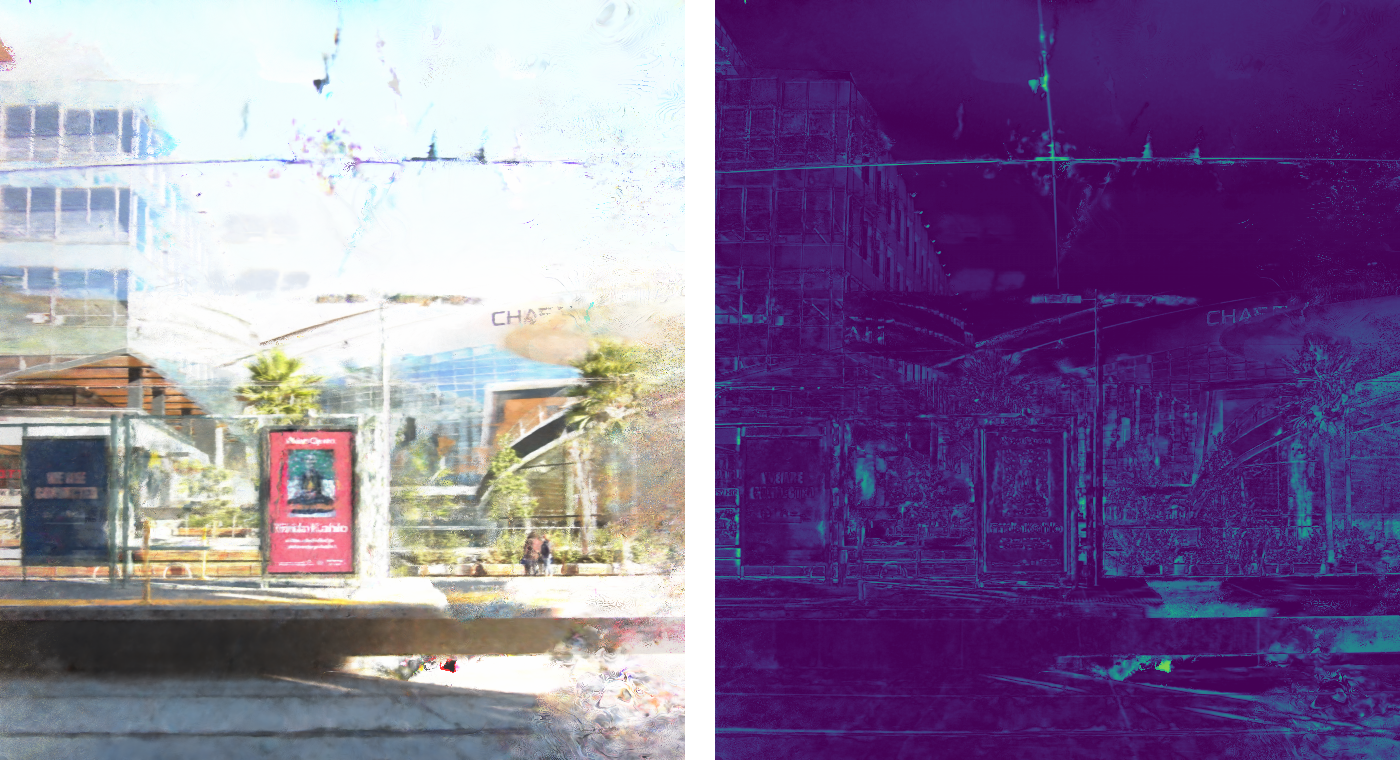} & \includegraphics[scale=0.08]{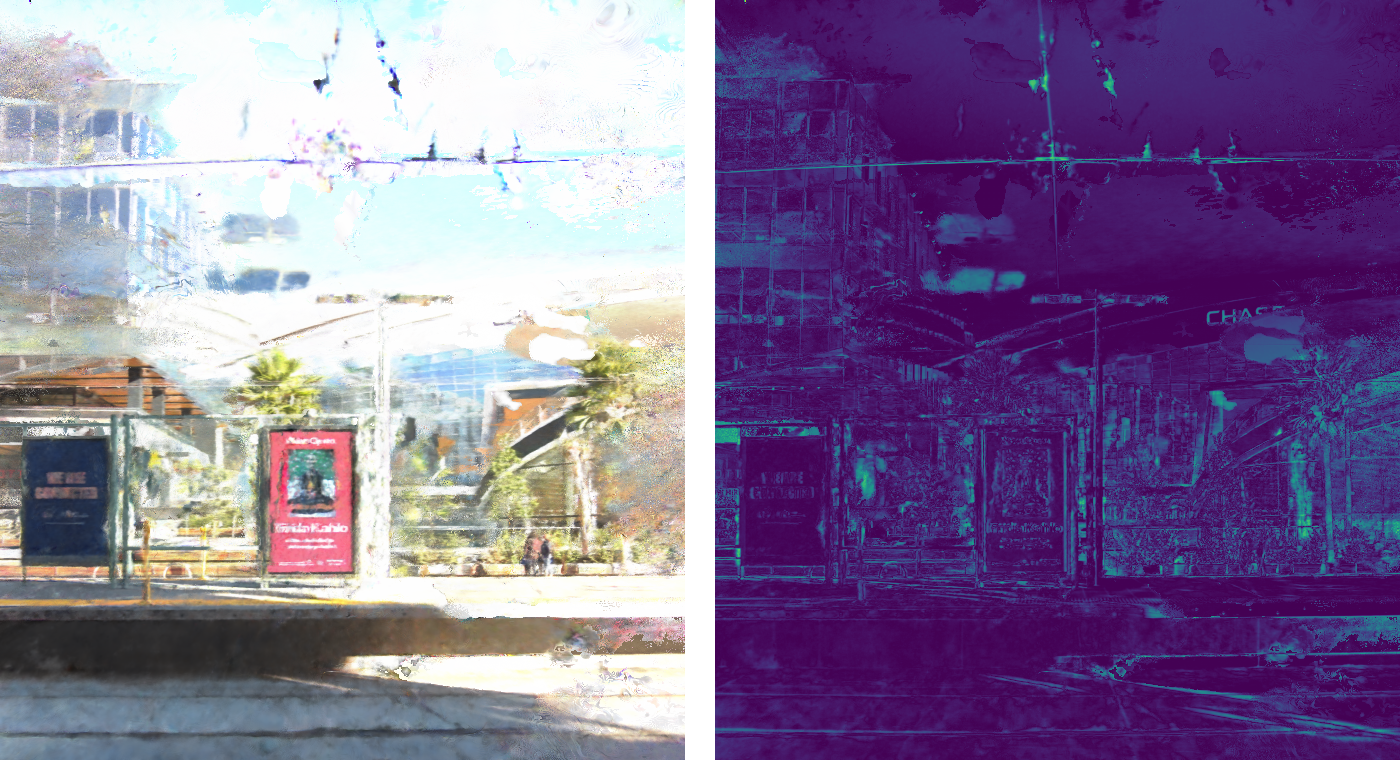}\\
        \cmidrule(lr){1-1}\cmidrule(lr){2-2}\cmidrule(lr){3-3}
        IDW-Sample & IDW-2D & IDW-3D \\
    \end{tabular}
    \caption{NeRF blending with IDW-based methods on the Mission Bay dataset. Per-pixel errors are visualized as heat maps. Individual NeRFs renderings have large artifacts on either side, which are best resolved by \emph{IDW-Sample} blending.}
    \label{fig:blocknerf-blending}
\end{figure*}

\noindent\textbf{ScanNet Dataset}~
\label{subsubsec:scannet-data}
The ScanNet dataset provides a total of $1513$ RGB-D scenes with annotated camera poses, from which we use the first $218$ scenes. We downsample the frames so that roughly $200$ posed RGB-D images are kept from each scene. We then split the images into three sets: two for training NeRFs and one for testing. We test NeRF registration on this dataset and compare with point-cloud registration in Section~\ref{subsec:ner-reg}.

\noindent\textbf{Mission Bay Dataset}
\label{sec:blocknerf-data}
To further test the rendering quality of different blending methods, we run experiments on the Mission Bay dataset from Block-NeRF~\cite{Tancik2022BlockNeRFSL}, which features a street scene from a single capture. We report results in Section~\ref{subsec:nerf-blending}.

\paragraph{NeRF Fusion (Registration plus Blending)}\label{subsec:nerf-fusion}
We test NeRFuser including both NeRF registration and NeRF blending on Object-Centric Indoor Scenes. 
We report numbers averaged over test images of all three scenes from our dataset in Table~\ref{tab:nf_bld}.
Detailed experimental setup can be found in supplementary materials.

\noindent\paragraph{NeRF Registration}\label{subsec:ner-reg}
To further test the registration performance on a large-scale dataset, we use the ScanNet dataset~\cite{Dai2017ScanNetR3} as prepared according to Section~\ref{subsubsec:scannet-data}. We repeat the same registration procedure as in Section~\ref{subsec:nerf-fusion}, except that $60$ hemispheric poses are sampled instead of $30$. During experiments, we notice failure cases due to NaN or outlier values. To report more meaningful numbers, we treat cases that meet any of the following conditions as failure:
\begin{enumerate*}[label=(\roman*)]
    \item is NaN or
    \item $r_{\textrm{err}}>5^\circ$ or
    \item $t_{\textrm{err}}>0.2$ or
    \item $s_{\textrm{err}}>0.1$.
\end{enumerate*}
We also compare our method against various point-cloud registration (PCR) baselines using both
\begin{enumerate*}[label=(\roman*)]
    \item point-clouds extracted from NeRFs and\label{itm:pcr1}
    \item point-clouds fused from ground-truth posed RGB-D images.\label{itm:pcr2}
\end{enumerate*}
We report in Table~\ref{table:scannet_registration_filenames_segmentation_nerf_extraction} the results of our registration method and various PCR baselines averaged over all successfully registered scenes, as well as the success rate. Detailed experimental results can be found in supplementary results.
\input{papers/nerfuser/tables/registration_scannet}

\paragraph{NeRF Blending}\label{subsec:nerf-blending}
\begin{table}[h!]
    \centering
    \renewcommand{\arraystretch}{1.0}
    \begin{tabularx}{0.7\linewidth}{Xccc}
        \toprule
        Blending & PSNR $\uparrow$ & SSIM $\uparrow$ & LPIPS $\downarrow$ \\ 
        \midrule
        NeRF & 17.306 & 0.571 & 0.502 \\
        Nearest & 19.070 & 0.657 & 0.398 \\ 
        IDW-2D & 19.692 & 0.659 & 0.413 \\
        IDW-3D & 18.806 & 0.636 & 0.433 \\
        IDW-Sample & \textbf{19.986} & \textbf{0.678} & \textbf{0.388} \\
        \bottomrule
    \end{tabularx}
    \caption{Blending results on Mission Bay dataset. \emph{IDW-Sample} performs the best for all metrics.}
    \label{table:blocknerf_blending}
\end{table}
To further test our blending performance, we use the outdoor Mission Bay dataset as described in Section~\ref{sec:blocknerf-data}, with ground-truth transformations. We set the distance test ratio $\tau=1.2$, and the blending rate $\gamma=10$. Quantitative results averaged over test images of all scenes are reported in Table~\ref{table:blocknerf_blending}. Qualitative results are visualized in Figure~\ref{fig:blocknerf-blending}.

\begin{figure}[h!]
    \centering
    \def\svgwidth{\linewidth}
    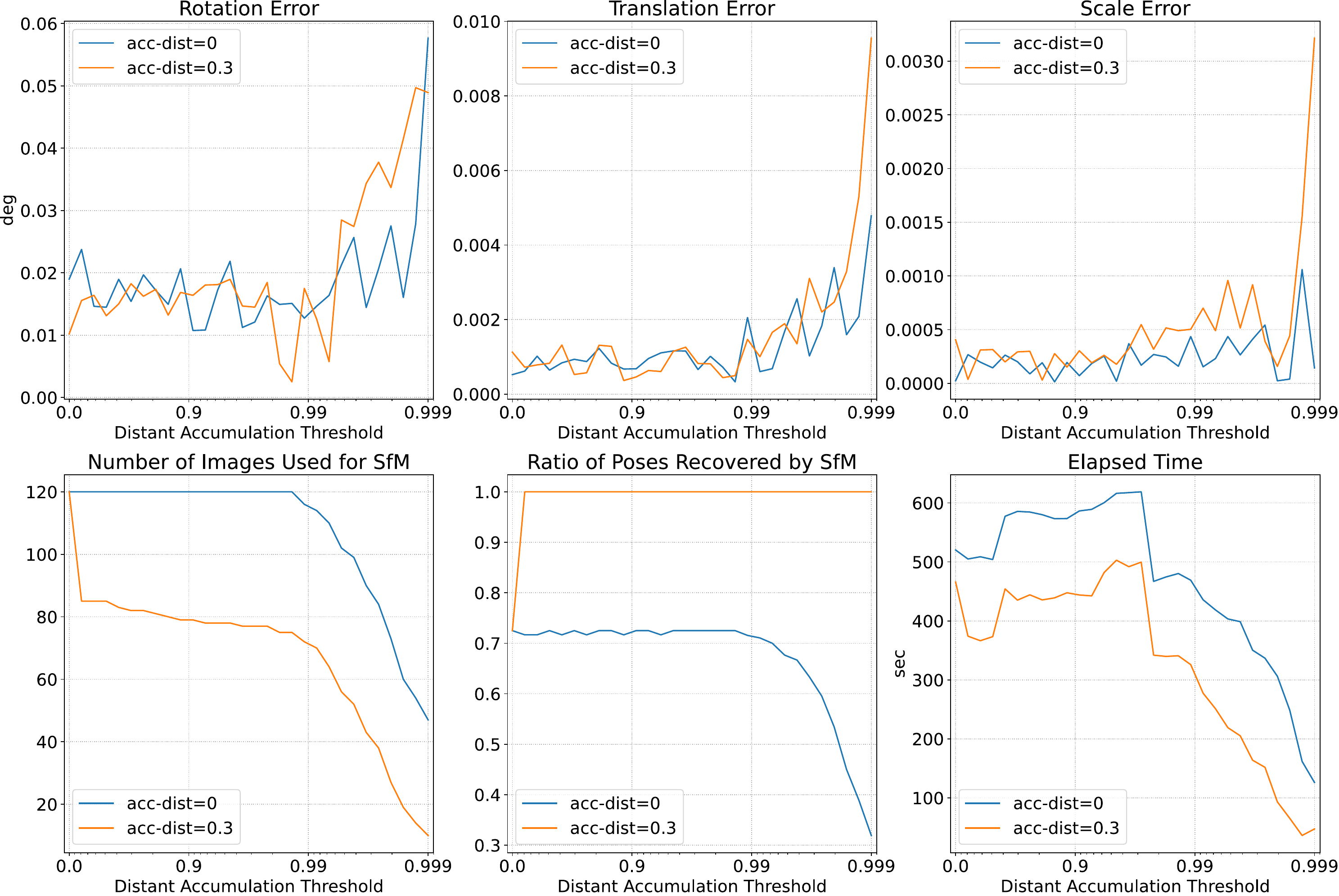
    \caption{Registration error and time consumption against varying accumulation-based pose filtering thresholds with distant accumulation $q_d$ ($d=0.3$) and regular accumulation ($d=0.0$).}
    \label{fig:distant_accumulation_0.3}
\end{figure}

\paragraph{Ablation Studies}
\noindent\textbf{Ablation of distant accumulation for filtering poses in NeRF registration}
We ran NeRF registration experiments using different thresholds of the proposed mean distant accumulation for filtering poses. As shown in Figure~\ref{fig:distant_accumulation_0.3}, distant accumulation-based filtering removes mostly bad images when compared to the regular accumulation, leading to a significant decrease in registration time, while yielding similar registration accuracy.

\noindent\textbf{Ablation on re-rendering poses for NeRF registration}
\begin{figure}[!tbp]
    \centering
    \def\svgwidth{\linewidth}
    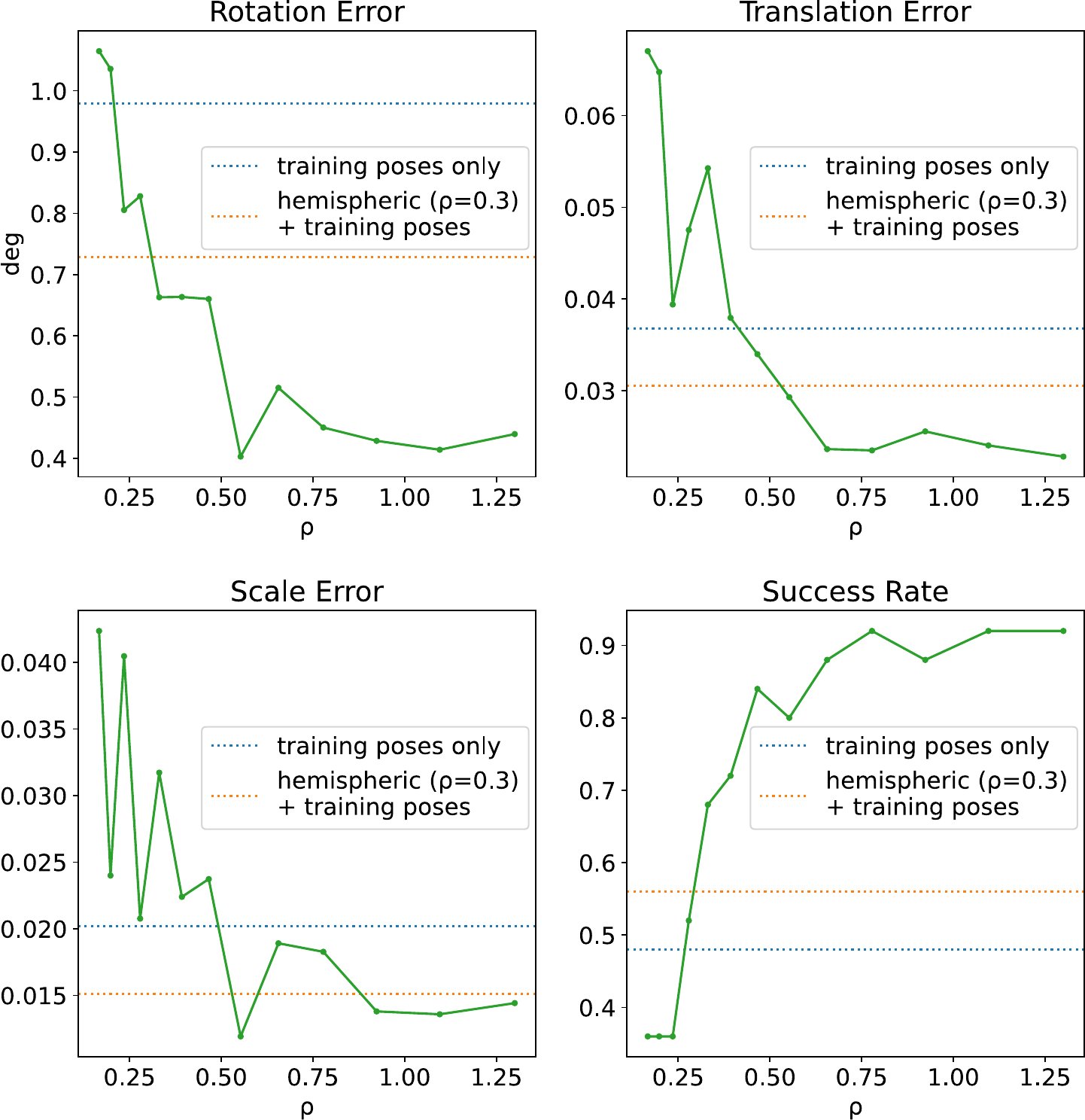
    \caption{Effect of re-rendering poses on NeRF registration.}
    \label{fig:ablation-reg-r}
\end{figure}
We study the registration performance on ScanNet dataset w.r.t. the number of sampled poses. To account for the fact that each scene may be of a different scale, we introduce $\rho$ as the ratio of the number of sampled poses over the number of training views. We geometrically sample $\rho\in[0.167, 1.3]$, and generate the hemispheric poses accordingly. We evaluate the performance of NeRF registration w.r.t. $\rho$ averaged over all ScanNet scenes. In addition, we include 2 more settings.
\begin{enumerate*}[label=(\roman*)]
    \item\emph{training poses only}: instead of hemispheric poses, use NeRF's training poses for re-rendering;
    \item\emph{hemispheric $+$ training poses}: use NeRF's training poses together with hemispherically sampled ones ($\rho=0.3$) for re-rendering.
\end{enumerate*}

Results are reported in Figure~\ref{fig:ablation-reg-r}. With more sampled poses, the registration errors go down while success rate improves (\textcolor{teal}{green curve}). Using additional hemispheric poses besides the training ones also proves helpful (\textcolor{orange}{orange line} vs. \textcolor{blue}{blue line}). Interestingly, with a large enough ratio $\rho$, registration with hemispherically sampled poses outperforms training poses when using the same number or fewer poses in total. It shows that it is beneficial to have a larger spatial span of re-rendering poses for registration.

\begin{figure}[H]
    \centering
    \def\svgwidth{\linewidth}
    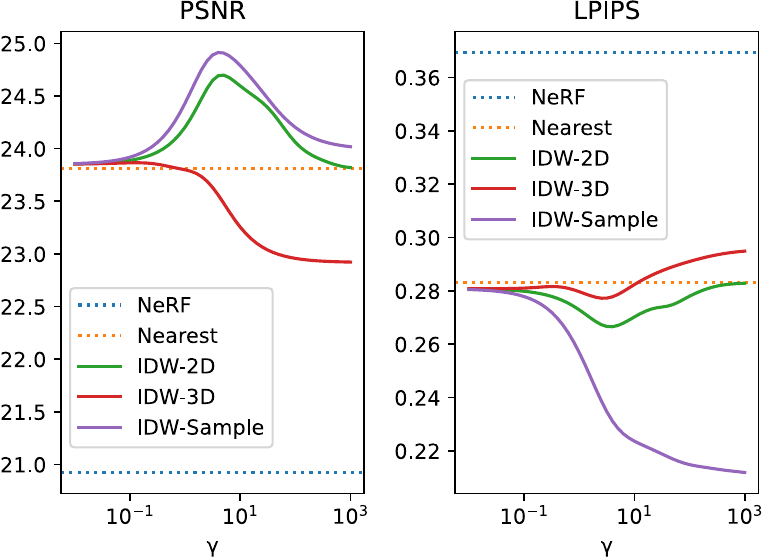
    \caption{Effect of blending rate $\gamma$ in IDW-based blending.}
    \label{fig:gamma-ablation}
\end{figure}

\noindent\textbf{Ablation of $\gamma$ in IDW-based blending}
\label{sec:gamma-ablation}
We study the effect of blending rate $\gamma$ in IDW-based blending on Object-Centric Indoor Scenes. Specifically, we use ground-truth transformations and set distance test ratio $\tau=1.8$. We geometrically sample $\gamma$ in $[10^{-2},10^3]$. For each sampled $\gamma$, we blend NeRFs with all IDW-based methods and report the results averaged over test images of all 3 scenes from Object-Centric Indoor Scenes. Since \emph{Nearest} and \emph{NeRF} are not affected by $\gamma$, we draw dotted horizontal lines for comparison. The results are shown in Figure~\ref{fig:gamma-ablation}.  For all blending methods, quality initially increases with $\gamma$, but then decreases as $\gamma$ increases further. In $\gamma\rightarrow{}0$ case, all IDW-based methods become the same as using the mean image. In $\gamma\rightarrow{}\infty$ case, \emph{IDW-2D} becomes the same as \emph{Nearest}, while \emph{IDW-Sample} becomes analogous to KiloNeRF~\cite{Reiser2021KiloNeRFSU}.  For details, please see the supplementary. We find the optimal $\gamma$ in between the extremes for any IDW-based method. Moreover, our proposed \emph{IDW-Sample} almost always performs the best for any given $\gamma$.

%% file: papers/nerfuser/tables/registration_scannet.tex
\begin{table}[h!]
    \centering
    \begin{tabularx}{0.8\linewidth}{Xcccc} 
        \toprule
        Registration & $r_{\textrm{err}}~(^\circ)$ & $t_{\textrm{err}}$ & $s_{\textrm{err}}$ & Success\\
        \midrule
        \multicolumn{5}{c}{NeRF-extracted point-cloud} \\
        \midrule[0.1pt]
        ICP \cite{Rusinkiewicz2001EfficientVO} & 3.027 & 0.1151 & N/A & 0.13 \\ 
        FGR \cite{Zhou2016FastGR} & 4.549 & 0.1844 & N/A & 0.04 \\
        FPFH \cite{Rusu2009FastPF} & 2.805 & 0.0381 & N/A & 0.17 \\
        \midrule
        \multicolumn{5}{c}{RGB-D-fused point-cloud} \\
        \midrule[0.1pt]
        ICP \cite{Rusinkiewicz2001EfficientVO} & 1.598 & 0.0816 & N/A & 0.17 \\ 
        FGR \cite{Zhou2016FastGR} & 1.330 & 0.0372 & N/A & 0.71 \\
        FPFH \cite{Rusu2009FastPF} & \textbf{0.049} & \textbf{0.0205} & N/A & \textit{0.79} \\
        \midrule[0.1pt]
        NeRFuser & \textit{0.588} & \textit{0.0315} & \textbf{0.0211} & \textbf{0.84}\\
        \bottomrule
    \end{tabularx}
    \caption{\textbf{Registration results on ScanNet.} We compare to point-cloud registration methods on both NeRF-extracted point-cloud and ground-truth RGB-D-fusion point-cloud. Due to the noisy geometry of NeRF reconstructions, registration performance on NeRF-extracted point-clouds is inferior. However, NeRFuser is comparable to the registration performance on RGB-D-fused methods in terms of $r_{\textrm{err}}$ and $t_{\textrm{err}}$, while having the highest success rate. Unlike point-cloud baselines, our method also recovers the relative scale. %
    $\textbf{Bold}$ numbers are the best, $\textit{italic}$ numbers are second best.}
    \label{table:scannet_registration_filenames_segmentation_nerf_extraction}
\end{table}

%% file: svg-inkscape/ablation_dist_acc_svg-tex.pdf_tex
\begingroup%
  \makeatletter%
  \providecommand\color[2][]{%
    \errmessage{(Inkscape) Color is used for the text in Inkscape, but the package 'color.sty' is not loaded}%
    \renewcommand\color[2][]{}%
  }%
  \providecommand\transparent[1]{%
    \errmessage{(Inkscape) Transparency is used (non-zero) for the text in Inkscape, but the package 'transparent.sty' is not loaded}%
    \renewcommand\transparent[1]{}%
  }%
  \providecommand\rotatebox[2]{#2}%
  \newcommand*\fsize{\dimexpr\f@size pt\relax}%
  \newcommand*\lineheight[1]{\fontsize{\fsize}{#1\fsize}\selectfont}%
  \ifx\svgwidth\undefined%
    \setlength{\unitlength}{1427.62597656bp}%
    \ifx\svgscale\undefined%
      \relax%
    \else%
      \setlength{\unitlength}{\unitlength * \real{\svgscale}}%
    \fi%
  \else%
    \setlength{\unitlength}{\svgwidth}%
  \fi%
  \global\let\svgwidth\undefined%
  \global\let\svgscale\undefined%
  \makeatother%
  \begin{picture}(1,0.66914857)%
    \lineheight{1}%
    \setlength\tabcolsep{0pt}%
    \put(0,0){\includegraphics[width=\unitlength,page=1]{svg-inkscape/ablation_dist_acc_svg-tex.pdf}}%
  \end{picture}%
\endgroup%

%% file: svg-inkscape/ablation_registration_rho_svg-tex.pdf_tex
\begingroup%
  \makeatletter%
  \providecommand\color[2][]{%
    \errmessage{(Inkscape) Color is used for the text in Inkscape, but the package 'color.sty' is not loaded}%
    \renewcommand\color[2][]{}%
  }%
  \providecommand\transparent[1]{%
    \errmessage{(Inkscape) Transparency is used (non-zero) for the text in Inkscape, but the package 'transparent.sty' is not loaded}%
    \renewcommand\transparent[1]{}%
  }%
  \providecommand\rotatebox[2]{#2}%
  \newcommand*\fsize{\dimexpr\f@size pt\relax}%
  \newcommand*\lineheight[1]{\fontsize{\fsize}{#1\fsize}\selectfont}%
  \ifx\svgwidth\undefined%
    \setlength{\unitlength}{692.04217529bp}%
    \ifx\svgscale\undefined%
      \relax%
    \else%
      \setlength{\unitlength}{\unitlength * \real{\svgscale}}%
    \fi%
  \else%
    \setlength{\unitlength}{\svgwidth}%
  \fi%
  \global\let\svgwidth\undefined%
  \global\let\svgscale\undefined%
  \makeatother%
  \begin{picture}(1,1.03158968)%
    \lineheight{1}%
    \setlength\tabcolsep{0pt}%
    \put(0,0){\includegraphics[width=\unitlength,page=1]{svg-inkscape/ablation_registration_rho_svg-tex.pdf}}%
  \end{picture}%
\endgroup%

%% file: svg-inkscape/ablation_blending_gamma_svg-tex.pdf_tex
\begingroup%
  \makeatletter%
  \providecommand\color[2][]{%
    \errmessage{(Inkscape) Color is used for the text in Inkscape, but the package 'color.sty' is not loaded}%
    \renewcommand\color[2][]{}%
  }%
  \providecommand\transparent[1]{%
    \errmessage{(Inkscape) Transparency is used (non-zero) for the text in Inkscape, but the package 'transparent.sty' is not loaded}%
    \renewcommand\transparent[1]{}%
  }%
  \providecommand\rotatebox[2]{#2}%
  \newcommand*\fsize{\dimexpr\f@size pt\relax}%
  \newcommand*\lineheight[1]{\fontsize{\fsize}{#1\fsize}\selectfont}%
  \ifx\svgwidth\undefined%
    \setlength{\unitlength}{366.24902344bp}%
    \ifx\svgscale\undefined%
      \relax%
    \else%
      \setlength{\unitlength}{\unitlength * \real{\svgscale}}%
    \fi%
  \else%
    \setlength{\unitlength}{\svgwidth}%
  \fi%
  \global\let\svgwidth\undefined%
  \global\let\svgscale\undefined%
  \makeatother%
  \begin{picture}(1,0.72965212)%
    \lineheight{1}%
    \setlength\tabcolsep{0pt}%
    \put(0,0){\includegraphics[width=\unitlength,page=1]{svg-inkscape/ablation_blending_gamma_svg-tex.pdf}}%
  \end{picture}%
\endgroup%

%% file: papers/nerfuser/sections/conclusion.tex
We have introduced NeRFuser/space, a NeRF fusion pipeline that registers and blends arbitrary input NeRFs, extending a standard image-based processing pipeline to treat NeRFs as input data. To address NeRF registration, we propose \emph{registration from re-rendering}, taking advantage of NeRFs' ability to synthesize high quality novel views with \emph{distant accumulation},including an effective image quality measure \emph{distant accumulation} for pose filtering. To address NeRF blending, we propose \emph{IDW-Sample}, leveraging the ray sampling nature in NeRF volumetric rendering. 
We believe this tool will help towards the proliferation of implicit representations as raw data for future 3D robotic applications.

%% file: contents/chap_reason.tex
\chapter{Embodied Spatial Reasoning with Hybrid Systems}\label{chap:reason}
Once scene representations are constructed, robotic agents must reason about their environment based on natural language instructions to execute appropriate actions. While earlier systems were limited to structured languages \cite{tellex2011understanding}, the advent of Large Language Models (LLMs) \cite{brown2020language} has significantly enhanced the ability to process natural language. This breakthrough has created new opportunities for robots to perform complex spatial reasoning directly from human commands.
However, emerging research suggests that LLMs primarily exhibit "System 1" thinking—a fast, automatic, and intuitive mode of cognition \cite{yao2024tree}. This stands in contrast to the slow, deliberate, and conscious "System 2" thinking, which is essential for solving the complex, long-horizon tasks prevalent in robotics, as described in foundational psychology studies~\cite{kahneman2011thinking}.
My own work \cite{ding2024mango} corroborates this limitation. We systematically evaluated the navigation and mapping capabilities of several top-performing LLMs using data from text-based game interactions. Our results revealed a significant performance gap between even the leading model, GPT-4, and human benchmarks. This indicates that direct generation from LLMs reflects "System 1" thinking, struggling with compositional reasoning tasks such as path-finding or destination identification.
To bridge this gap and enable "System 2" thinking, I have focused on developing hybrid systems that augment LLMs with external tools and structured processes. For the task of 3D referring resolution—which requires understanding spatial relationships from natural language expressions—I proposed Transcrib3D~\cite{fang2024transcrib3d}. Recognizing the scarcity of large-scale 3D training data, Transcrib3D diverges from methods that operate directly on 3D representations \cite{Jain2021BottomUT, Zhu20233DVisTAPT}. Instead, it transcribes 3D scenes into textual descriptions. This transformation allows the LLM to offload complex spatial calculations to external tools, such as code execution, while focusing on high-level decision-making. This hybrid approach achieved state-of-the-art results on 3D referring resolution benchmarks.
Long-horizon planning presents another significant challenge for purely generative LLMs, which often struggle with maintaining context and consistency over extended interactions~\cite{liu2024lost}. This capability is critical for robotics, where tasks typically involve executing a long sequence of actions. To address this, my work on Statler~\cite{yoneda2023statler} introduces the concept of an explicit \textit{state} for LLM-based reasoning. In this framework, the state serves as a concise summary of the task's history, which is iteratively updated by one LLM based on the previous state and new observations. A second LLM then uses this state to plan the next action, effectively grounding its decisions in the full context of the task. This two-LLM, state-centric system demonstrably outperforms single-LLM inference, even those enhanced with prompting techniques like Chain-of-Thought~\cite{Wei2022ChainOT}.
In summary, these works demonstrate a powerful paradigm: using language as a unified interface for both scene representation and task specification. This approach reduces the complexity of raw 3D scenes and enables joint planning directly within the language space, unlocking new possibilities for human-robot interaction. However, this strategy is not without its trade-offs. On the representation side, transcribing 3D scenes into language is a form of lossy compression, potentially discarding geometric details crucial for certain tasks. On the reasoning side, even with architectural improvements, LLMs continue to face limitations in long-term state tracking and compositional logic—skills that are foundational for robust robotic operation.
Furthermore, human cognition for embodied tasks relies on more than just abstract language~\cite{Dove2018, Tomasello2014, Pinker1994, GordonEtAl2021, GoldinMeadow2001}. The path forward likely involves a co-evolution of both representations and model capabilities. I envision a future with more expressive, multi-modal latent representations that retain nuanced information while remaining language-decodable, coupled with advanced models capable of seamlessly integrating both "System 1" and "System 2" thinking from multi-modal inputs.

%% file: papers/mango/sections/abstract.tex
\begin{figure}[H]
     \centering
    \includegraphics[width=\linewidth]{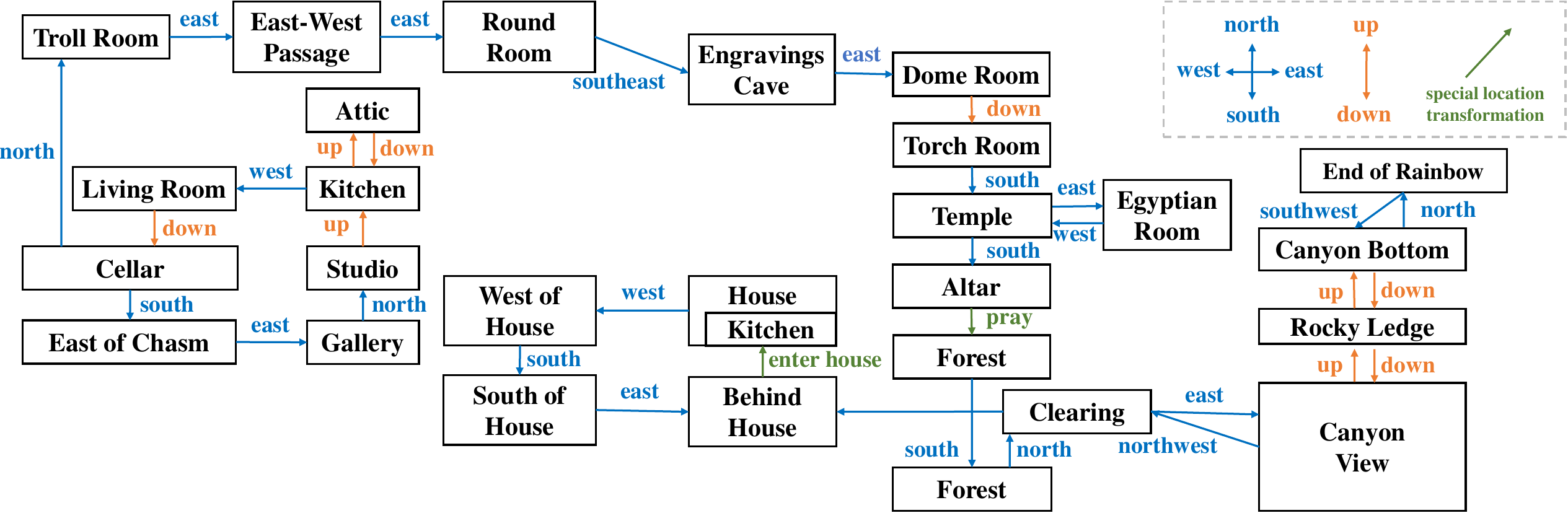}
    \caption{Map of Zork-I. Arrows denote the direction of travel during the walkthrough, while the reverse direction is unseen but may be possible. Note that it is a 3D map projected onto a 2D plane so \code{up} may not point upward in the 2D visualization (e.g., \code{Rocky Ledge} to \code{Canyon View}).}
    \label{fig:map}
\end{figure}

Large language models such as ChatGPT and GPT-4 have recently achieved astonishing performance on a variety of natural language processing tasks. 
In this paper, we propose MANGO, a benchmark to evaluate their ability to perform text-based mapping and navigation.
Our benchmark includes $53$ mazes taken from a suite of textgames: each maze is paired with a walkthrough that visits every location but does \emph{not} cover all possible paths. 
The task is question-answering: for each maze, a large language model reads the walkthrough and answers hundreds of mapping and navigation questions such as ``How should you go to \code{Attic} from \code{West of House}?'' and ``Where are we if we go \code{north} and \code{east} from \code{Cellar}?''.
Although these questions are easy for humans, it turns out that even GPT-4, the best-to-date language model, performs poorly when answering them. 
Further, our experiments suggest that a strong mapping and navigation ability would benefit the performance of large language models on relevant downstream tasks, such as playing textgames. %
Our MANGO benchmark will facilitate future research on methods that improve the mapping and navigation capabilities of LLMs. 
We host our leaderboard, data, code, and evaluation program at 
{\small \url{http://mango.ttic.edu/}}. 

%% file: papers/mango/sections/introduction.tex
 Mapping and navigation are fundamental abilities of 
human intelligence~\citep{spiers06, epstein17}. 
Humans are able to construct maps---in their minds~\citep{epstein17} or on physical media like paper---as they explore unknown environments. 
Following these maps, humans can navigate through complex environments~\citep{spiers06,spiers15,javadi17}, making informed decisions, and interact with their surroundings.
Such abilities empower humans to explore, adapt, and thrive in diverse environments. 
An example is remote (e.g., deep-sea) exploration for which humans have drawn upon their intuition to develop algorithms that enable robots to autonomously navigate and map their surroundings based only on onboard sensing. %

Do large language models (LLMs) possess such abilities? %
In this paper, we investigate this research question by creating a benchmark and evaluating several widely used LLMs. 
Our MANGO benchmark is the \emph{first} to measure the \underline{ma}pping and \underline{n}avi\underline{g}ati\underline{o}n abilities of LLMs. 
It includes $53$ complex mazes, such as the one visualized in \cref{fig:map}. %
It pairs each maze with hundreds of destination-finding questions (e.g., ``Where will you be if you go \code{north}, \code{north}, and then \code{up} from \code{Altar}?'') and route-finding questions (e.g., ``How do you reach \code{Dome Room} from \code{Altar}?'').
For each maze, the language model has to answer these questions after reading a walkthrough of the maze. 
Many questions involve possible routes that are not traced during the walkthrough, making the benchmark challenging. %
In our experiments, GPT-4 only correctly answered half of the route-finding questions, performing disastrously on the difficult questions (e.g., those involving long and unseen routes).
MANGO will facilitate future research in improving the mapping and navigation abilities of LLMs.

Another contribution of MANGO is to draw a novel connection between natural language processing and robotics. 
There has been significant interest in employing LLMs %
to endow intelligent agents (including robots) with complex reasoning~\citep{yang2023foundation}. 
Aligning with this interest, MANGO enables the investigation of the LLMs' capabilities in simultaneous localization and mapping (SLAM) within text-based worlds. 
Focusing on this aspect, our work stands out and complements previous SLAM-related research, 
which predominantly relies on richer sensory inputs (e.g., vision and LiDAR).

%% file: papers/mango/sections/related_work.tex
Over the past few years, the field of natural language processing has experienced remarkable advancements with the emergence of large language models. 
This progress has spurred a multitude of research endeavors that propose benchmarks challenging the limits of these models. 
Those benchmarks assess the capacities of LLMs in linguistics~\citep{wang2018glue,superglue}, 
reading comprehension~\citep{richardson2013mctest,lai2017race}, 
commonsense reasoning~\citep{zellers2019hellaswag,bisk2020piqa,huang2019cosmos,talmor2018commonsenseqa}, 
arithmetic reasoning~\citep{miao2021diverse,cobbe2021training,patel2021nlp}, 
and knowledge memorization and understanding~\citep{clark2018think,mihaylov2018can,khot2020qasc,clark2020f,hendrycks2020measuring,srivastava2022beyond}. 
Recent models have achieved remarkable performance not only on these benchmarks, but also across a diversity of human-oriented academic and professional exams~\citep{gpt4} as well as general tasks~\citep{bubeck2023sparks}. %
Our benchmark presents a unique challenge to large language models, evaluating their capacity to acquire spatial knowledge about new environments and answering complex navigation questions; it is a dimension orthogonal to the aforementioned reasoning abilities. 

The advances of LLMs have sparked a recent wave of endeavors that integrate these models into embodied agents~\citep{huang2022inner,yang2023foundation,vemprala2023chatgpt,Wang2023VoyagerAO}. 
Generally, they utilize language models as a means to understand human instructions and plan executable actions~\citep{driess2023palm,liang2022code,huang2022language,pmlr-v205-ichter23a}. 
This includes instructions related to object manipulation and tool operation~\citep{wang2023programmatically,ren2023leveraging} as well as localization and navigation~\citep{majumdar2020improving,gadre2023cows,shah2023lm,Huang2022VisualLM}. 
Our MANGO benchmark aligns with the growing trend to deploy LLMs in embodied agents and provides a comprehensive investigation of their capacities in mapping and navigation. 
Our benchmark operates in text-based environments, distinguishing itself from previous benchmarks~\citep{puig2018virtualhome,shridhar2020alfred,fan2022minedojo} that allow agents to utilize visual signals. 
This ``text-only'' design enables us to conduct controlled experiments that investigate the capacity of language models to acquire knowledge about environments solely from textual inputs and answer navigation questions based on that knowledge. 
It complements the existing benchmark and methodological research in vision-language navigation~\citep{duvallet14, mei2016listen, Anderson2017VisionandLanguageNI, fried18, zhu20, Min2021FILMFI}. Our work is related to recent studies that demonstrate the emergence of maps with learned neural representations as a consequence of navigation~\citep{huynh20, wijmans2023emergence} with the key distinction that our agents are provided with textual descriptions of their environments.

Given our focus on mapping and navigation, it is worth noting the work on simultaneous localization and mapping (SLAM), a classic problem in which a mobile agent (e.g., a robot or hand-held camera) is tasked with mapping an a priori unknown environment while concurrently using its estimated map to localize itself in the environment~\citep{MurArtal2015ORBSlamaV, cadena16}. Particularly relevant are the methods that maintain spatial-semantic maps of the environments based on natural language descriptions~\citep{Walter2013LearningSM, hemachandra15a}, however they rely on non-linguistic observations (e.g., vision) to ground these descriptions.

%% file: papers/mango/sections/benchmark.tex
Our MANGO benchmark measures the mapping and navigation capabilities of LLMs.
It leverages a suite of text-based adventure games that offer expert-designed complex environments but only require simple actions. 
\cref{fig:map} is an example: 
it was drawn according to the first 70 steps of the walkthrough of Zork-I, which can be found in \cref{lst:walkthrough}. 
This map is imperfect: the annotator had to draw the only \code{Kitchen} twice to avoid a cluttered visualization; the \code{Living Room} was incorrectly placed outside the \code{House}. 
However, equipped with this map, one could correctly answer questions about any route in the maze such as ``How do you reach \code{Dome Room} from \code{Altar}?'' and ``Where will you be if you go \code{north}, \code{north}, and \code{up} from \code{Altar}?''. 
The walkthrough has not traced a route from \code{Altar} to \code{Dome Room}, but humans possess the remarkable capacity to %
plan a route by identifying the three individual steps---which the walkthrough has covered---from \code{Dome Room} to \code{Altar} and retracing those steps. 
MANGO tests whether a large language model can perform the same kind of reasoning. 
Particularly, when evaluating a language model, we first let it read a walkthrough like \cref{lst:walkthrough} and then ask it %
questions like those in \cref{lst:dfq,lst:rfq}. 
A question like \cref{lst:dfq} is a \emph{destination-finding} (DF) question, and a question like \cref{lst:rfq} is a \emph{route-finding} (RF) question. 
Users of MANGO have the flexibility to phrase the DF and RF questions in their own ways: as shown in \cref{lst:dfqraw,lst:rfqraw}, we provide the \emph{skeletons} of these questions, which users can plug into their own templates.  
\begin{figure}[t]
\begin{minipage}[t]{0.5\linewidth}
\begin{lstlisting}[caption={Zork-I walkthrough.},label={lst:walkthrough}]
STEP NUM: 0
ACT: Init
OBSERVATION: West of House
You are standing in an open field west of a white house, with a boarded front door. There is a small mailbox here.

STEP NUM: 1
ACT: south
OBSERVATION: South of House
You are facing the south side of a white house.

STEP NUM: 2
ACT: east
OBSERVATION: Behind House
You are behind the white house. A path leads into the forest to the east. In one corner of the house there is a small window which is slightly ajar.@\progvdots@

STEP NUM: 70
ACT: east
OBSERVATION: Gallery
This is an art gallery. Most of the paintings have been stolen by vandals with exceptional taste. The vandals left through either the north or west exits. Fortunately, there is still one chance for you to be a vandal, for on the far wall is a painting of unparalleled beauty.
\end{lstlisting}
\end{minipage}
\begin{minipage}[t]{0.5\linewidth}

\begin{lstlisting}[caption={A destination-finding question.},label={lst:dfq}]
Starting from Altar, perform actions [north, north, up], where are you now? 
\end{lstlisting}

\vspace{4pt}
\begin{lstlisting}[caption={A route-finding question.},label={lst:rfq}]
How can you go from Altar to Dome Room? 
\end{lstlisting}

\vspace{6pt}
\begin{lstlisting}[caption={Skeleton of DF question in \cref{lst:dfq}.},label={lst:dfqraw}]
@\mvar{S}@: Altar            @\textcolor{commentgray}{\# starting location}@
@\mvar{A}@: north, north, up   @\textcolor{commentgray}{\# list of actions}@
\end{lstlisting}

\vspace{4pt}
\begin{lstlisting}[caption={Skeleton of RF question in \cref{lst:rfq}.},label={lst:rfqraw}]
@\mvar{S}@: Altar            @\textcolor{commentgray}{\# starting location}@
@\mvar{D}@: Dome Roomm             @\textcolor{commentgray}{\# destination}@
\end{lstlisting}

\vspace{6pt}
\begin{lstlisting}[caption={Full route of \cref{lst:dfq,lst:dfqraw}.},label={lst:route}]
@\mvar{S}@: Altar              
@\mvar{A}@: north
@\mvar{D}@: Temple

@\mvar{S}@: Temple              
@\mvar{A}@: north
@\mvar{D}@: Torch Room

@\mvar{S}@: Torch Room              
@\mvar{A}@: up
@\mvar{D}@: Dome Room
\end{lstlisting}

\end{minipage}
\vspace{-8pt}
\end{figure}

\paragraph{Maze Collection: From Game Walkthroughs to Mazes}
\label{sec:data}
Our mazes are taken from the textgames in the Jericho game suite~\citep{hausknecht2020interactive}. 
The main release of Jericho includes $57$ popular textgames as well as a program that can generate walkthroughs for 56 of them.
The original walkthrough of a game is a list of actions (such as \code{east}, \code{north}, and \code{open door}) that one could execute to efficiently complete the game.
We enhanced %
each walkthrough by executing the sequence of actions and augmenting each step with the new observation (i.e., the text feedback that the game engine provides after the action is executed).
Unless explicitly specified, the word ``walkthrough'' refers to the enhanced, but not original, walkthroughs (such as \cref{lst:walkthrough}) throughout the paper. 

In a walkthrough, not every action triggers a location change: it may update the inventory (such as \code{take lamp} and \code{drop pen}) or time (such as \code{wait}). 
For each game, we read the walkthrough, labeled the actions (such as \code{east} and \code{up}) that change the locations, and made note of the names of the locations (such as \code{Temple} and \code{Altar}). 
This annotation is nontrivial and can not be automated. 
We had to pay extra attention to appropriately handle the tricky cases including: 
\circone the name of a location may be mentioned in a rich, but distracting context (e.g., the context may have ten paragraphs and hundreds of words with the name briefly mentioned in the middle); 
\circtwo a location may be visited multiple times, so we need to assign the same name to all its mentions; 
\circthree different locations may be referred to with the same name in the textual feedback, so we need to rename them in a sensible way.

The location name resolution results in a maze for each game. 
Three of the games have no location change, and so we left them out, resulting in $53$ mazes. %
We store each maze as a directed graph: each node is a named location (e.g., \code{Altar}); each directed edge is a movement (e.g., \code{north}); and each node-edge-node combination is a location-changing step that was followed in the walkthrough.
Note that a graph may be cyclic since the walkthrough may trace back-and-forth between locations (e.g., \code{Temple} and \code{Egyptian Room} in \cref{fig:map}). 

\paragraph{Generation of Question Skeletons: Traversing Mazes and Imputing Edges}
\label{sec:qg}
To generate DF and RF skeletons for a maze, a naive approach is to perform brute-force traversal. 
First, we collect all the possible \mvar{S}-\mvar{P}-\mvar{D} tuples, where \mvar{S} and \mvar{D} are locations and \mvar{P} is a simple path from \mvar{S} to \mvar{D}. 
A simple path %
is a directed path that does not visit any location more than once. 
This ``simple'' restriction ensures that we will have a finite number of \mvar{S}-\mvar{P}-\mvar{D} tuples. 
\cref{lst:route} is a simple path of 3 \mvar{S}-\mvar{A}-\mvar{D} edges from \code{Altar} to \code{Dome Room}.
Each unique \mvar{S}-\mvar{P}-\mvar{D} tuple gives a unique DF skeleton: e.g., \cref{lst:dfqraw} is obtained from \cref{lst:route}. 
Each unique \mvar{S}-\mvar{P}-\mvar{D} tuple gives an RF skeleton as well, such as \cref{lst:rfqraw} obtained from \cref{lst:route}. 
However, the same RF skeleton may be obtained from other tuples since there may be multiple possible simple paths between the same pair of locations \mvar{S} and \mvar{D}. 
As a consequence, we may end up with fewer RF questions than DF questions for a given maze. 

The particular DF and RF questions in \cref{lst:dfq,lst:rfq} are challenging to large language models, since they involve actions---such as going \code{north} from \code{Altar} to \code{Temple}---that are not covered in the walkthrough. 
Answering such hard questions requires a deeper understanding of the spatial relationships between locations. 
However, also because these steps are not in the walkthrough, the skeletons in \cref{lst:dfqraw,lst:rfqraw} can not be obtained through a naive traversal of the directed graph in \cref{fig:map}. 
That is, we have to traverse an extended graph that includes \emph{imputed} edges. 
An imputed edge denotes a valid step that is not explicitly mentioned in the walkthrough, such as going \code{north} from \code{Altar} to \code{Temple} (i.e., \code{Altar}-\code{north}-\code{Temple}). 
Most mentioned edges involve directional moves (e.g., \code{up}, \code{east}), so reversing them is a straightforward way to impute new edges. 
We manually examined other edges: for some of them, we proposed intuitive reverses (such as \code{exit} for \code{enter}); for the others (e.g., \code{pray}), no reverse could be found. 
We then examined the imputed edges through real game play and discarded those failing to cause the expected location changes.

After extending all the mazes in our benchmark, we collected 21046 DF skeletons and 14698 RF skeletons by traversing the extended graphs. 
Being evaluated on a maze, the LLM may not be able to consume the entire walkthrough in its context window. 
That is, we may only feed it an appropriate prefix of the walkthrough (e.g., the first $70$ steps for Zork-I as shown in \cref{lst:walkthrough}), leaving some of the DF and RF skeletons \emph{unanswerable} given that prefix. 
Therefore, our benchmark provides the ANSWERABLE label (an integer) for each skeleton such that this skeleton is only answerable if the maximum STEP NUM in that prefix (e.g., $70$ in \cref{lst:walkthrough}) is greater than or equal to its ANSWERABLE label. 
Furthermore, given a walkthrough prefix, an answerable skeleton may be easy or hard, depending on whether it involves edges that are not covered in the prefix. 
Precisely, a DF skeleton is considered to be easy if all the \mvar{S}-\mvar{A}-\mvar{D} edges in its corresponding simple path are covered in the walkthrough prefix; an RF skeleton is easy if the shortest simple path from its starting location to its destination only involves the \mvar{S}-\mvar{A}-\mvar{D} steps covered in the prefix.
When a longer walkthrough prefix is used, more answerable questions tend to become easy. 
Our benchmark provides the EASY label (also an integer) for each skeleton: a skeleton is easy if the maximum STEP NUM in the walkthrough prefix is no smaller than its EASY label; otherwise, it is a hard skeleton. 

\paragraph{Evaluation Program}
\label{sec:eval}

The evaluation program in our benchmark implements a range of evaluation and analysis methods.
Reading the model-generated answers, it can return a set of evaluation scores together with rich analysis. 
In this section, we introduce the most important scores used in our main experiments.

For DF questions, the most straightforward evaluation is the success rate: 
i.e., the fraction of questions that the language model answers correctly. 
What answers will be considered to be correct? 
A strict criteria is that the model answer is correct if and only if it exactly matches the ground-truth location name. 
However, due to the variability of natural language, a correct answer may not exactly match the ground-truth. 
For example, the model may yield \code{The House} or \code{That House} when the ground-truth location name is just \code{House}. 
To account for such cases, we generalize the success rate to allow partial matches. 
Given a model answer $\hat{A}$ and the ground-truth answer $A$, we compute their (character-level) edit-distance $d$ and define a correctness score $c \defeq 1-d/\ell$ where $\ell$ is the length of the longer answer. 
The score is $\in [0,1]$: when the answer exactly matches the ground-truth, we have $c=1$; if they have no character overlap at all, then $c=0$. 
We then define the success rate to be the sum of the correctness scores over all the questions, divided by the number of questions. 

For RF questions, the main metric is still the success rate, but the definition of ``success'' is different from that for DF questions. 
Note that an answer to an RF question is a sequence of moves. 
We consider an answer to be correct if and only if it can reach the destination after our evaluation program executes it in the maze. 
A correct answer to an RF question may not be a good path: it doesn't have to be the shortest; it doesn't even have to be a simple path. 
It is possible that an LLM-generated move is meaningful but doesn't exactly match any valid move in the graph: e.g., the LLM may give \code{walk south}, which means the same as \code{south}.   
Therefore, when executing a model-generated move, our evaluation program will select the closest (i.e., smallest edit-distance) valid move. 

%% file: papers/mango/sections/experiments.tex
\label{sec:exp}
In this section, we present the results our evaluation of several widely used LLMs. 

\paragraph{Experiment Setup}
\label{sec:setup}
The evaluated models are: GPT-3.5-turbo~\citep{brown-2020-gpt,stiennon2020learning,gao2022scaling}, GPT-4~\citep{gpt4}, Claude-instant-1~\citep{claude1}, Claude-2~\citep{claude2}, Llama-2 with 13B parameters~\citep{touvron2023llama2}, %
and RWKV with 14B parameters~\citep{peng2023rwkv}. %
For GPTs and Claudes, we used the prompt templates in \cref{lst:maptemplate,lst:navtemplate}, converting the DF and RF skeletons like \cref{lst:dfqraw,lst:rfqraw} into LLM-friendly questions like \cref{lst:dfq,lst:rfq}. 
The templates were carefully designed and examined through pilot experiments, in order to ensure that we do not underestimate the models on our benchmark. 
In our templates, each question starts with a list of legal actions, followed by a list of reachable locations; these lists help mitigate the hallucination of language models. 
The templates ask the model to spell out the entire trajectory including all the intermediate locations. 
This design is inspired by Chain-of-Thought prompting~\citep{wei2022chain}: eliciting an LLM to give its entire reasoning process tends to improve its overall performance on downstream tasks. 
In addition, it allows us to conduct a deeper evaluation and analysis, such as the reasoning accuracies of the models.
Note that our templates request the model to form its answer as a list of Python dictionaries with specific key names. 
We found that this restriction encourages the model to generate structured answers---which are easy to parse and analyze---as well as improves its performance. %
For Llama-2 and RWKV, we made moderate revisions to the prompts in order to generate well-structured answers as well as optimize for their performance. %
\begin{figure}[t]%
\begin{minipage}[t]{0.5\linewidth}
\begin{lstlisting}[caption={Our DF template.},label={lst:maptemplate}]
The allowed actions are: @\ldots@
The list of places are: @\ldots@
Starting from @\mvar{S}@, perform a list of 
actions [@\mvar{A}@], where are you now? 
Describe the trajectory in a Python list of Python dictionaries with keys 'prev_node', 'node' and 'action'. 
Start your response with '['.
\end{lstlisting}
\end{minipage}
\begin{minipage}[t]{0.5\linewidth}
\begin{lstlisting}[caption={Our RF template.},label={lst:navtemplate}]
The allowed actions are: @\ldots@
The list of places are: @\ldots@
How can you go from @\mvar{S}@ to @\mvar{D}@? 

Describe the trajectory in a Python list of Python dictionaries with keys 'prev_node', 'node' and 'action'. 
Start your response with '['.
\end{lstlisting}
\end{minipage}
\vspace{-8pt}
\end{figure}

For GPT-3.5, we experimented with the 4K version, which can consume 4096 tokens in its context window. 
This context limit restricts the length of the walkthrough that it can read, and the number of DF and RF questions that it can answer. 
For GPT-4, Claude-1 and Claude-2, we used the same walkthrough prefixes and questions as GPT-3.5 for a fair comparison. we used the same walkthrough prefixes and questions as GPT-3.5 for a fair comparison. 
Llama-2 has a $4096$ context window as well. But its tokenizer is different from GPTs' so we evaluated it on a slightly different set of questions. 
RWKV is capable of handling infinite context. 
For each maze, we experimented it with the $70$-step prefix of the walkthrough so that its set of answerable questions includes all the questions answered by all the other models. 
We also evaluated Llama-2 and RWKV in a simplified setting, where the observation at each step of the walkthrough only includes the location name but nothing else. 
For example, at STEP 1 of the simplified \cref{lst:walkthrough}, OBSERVATION only has \code{South of House} and everything else (i.e., \code{Your are...}) is omitted. 
We refer to Llama-2 and RWKV with the simplified walkthroughs as Llama-2-S and RWKV-S, respectively. %

\begin{figure*}[t]
	\begin{center}
	    \begin{subfigure}[t]{0.48\linewidth}
                \includegraphics[width=0.49\linewidth]{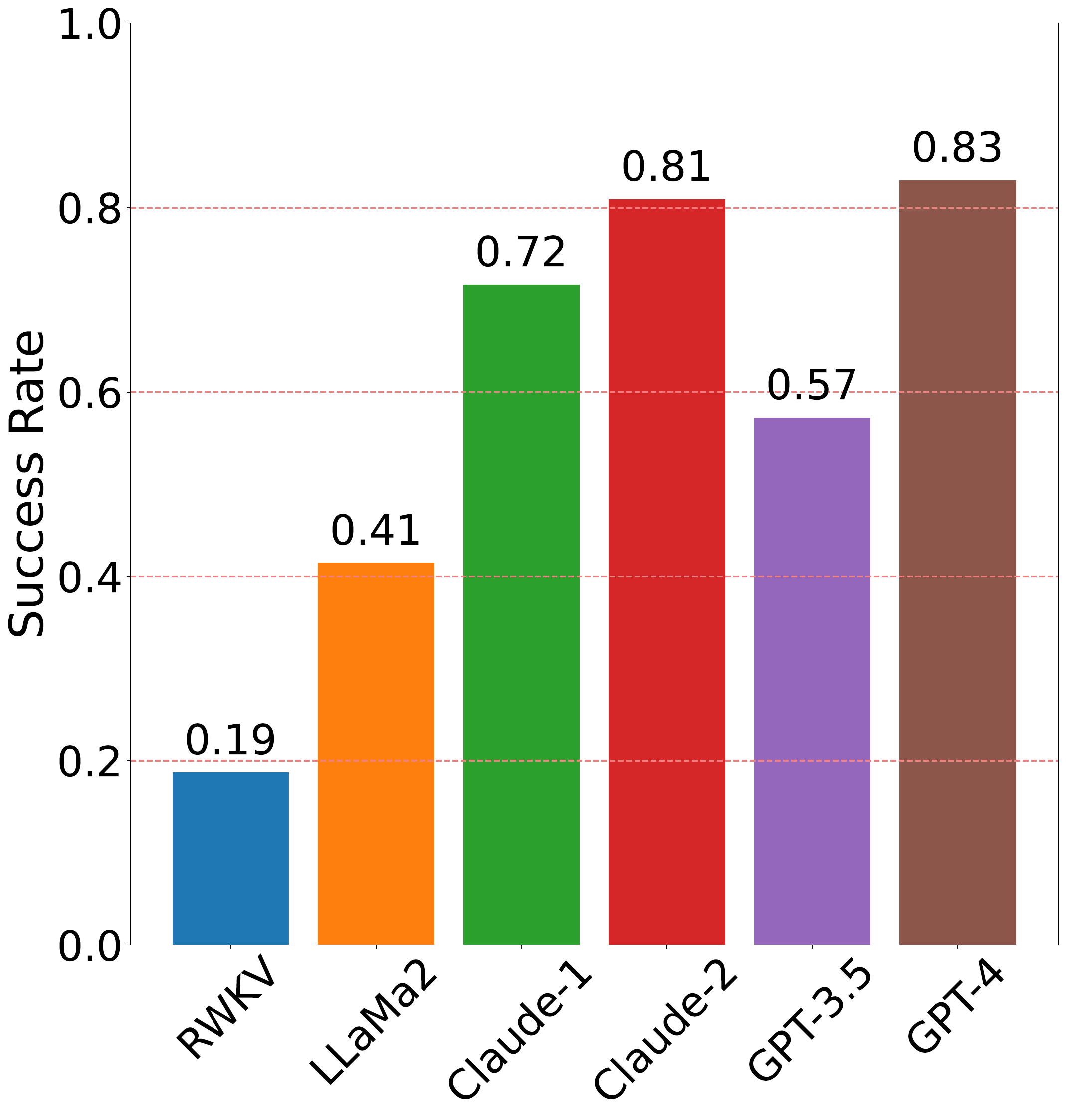}
                \hfill
                \includegraphics[width=0.49\linewidth]{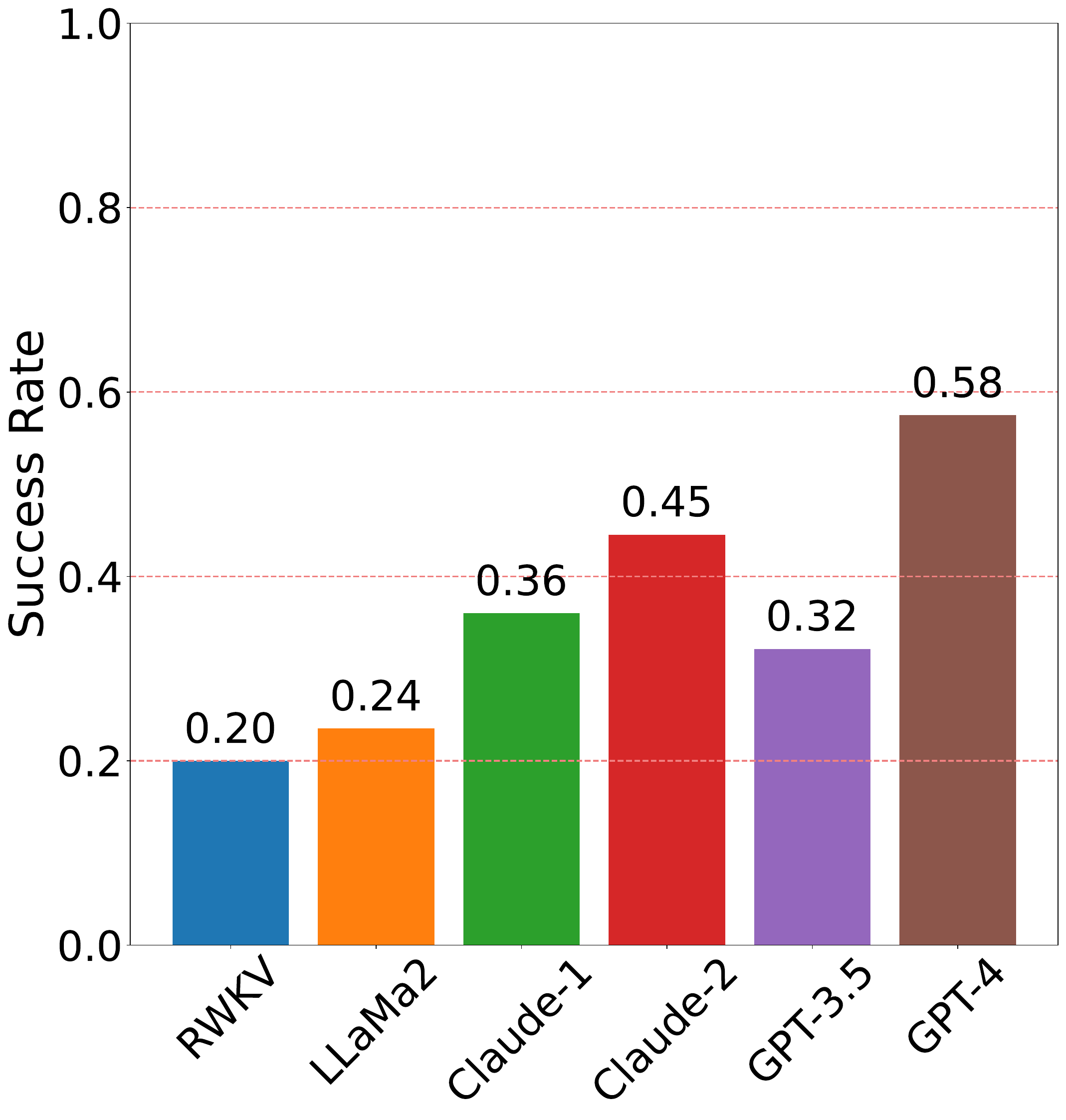}
		    \vspace{-16pt}
			\caption{On easy (left) and hard (right) DF questions.}\label{fig:main_df}
		\end{subfigure}
		\hfill
            \begin{subfigure}[t]{0.48\linewidth}
			\includegraphics[width=0.49\linewidth]{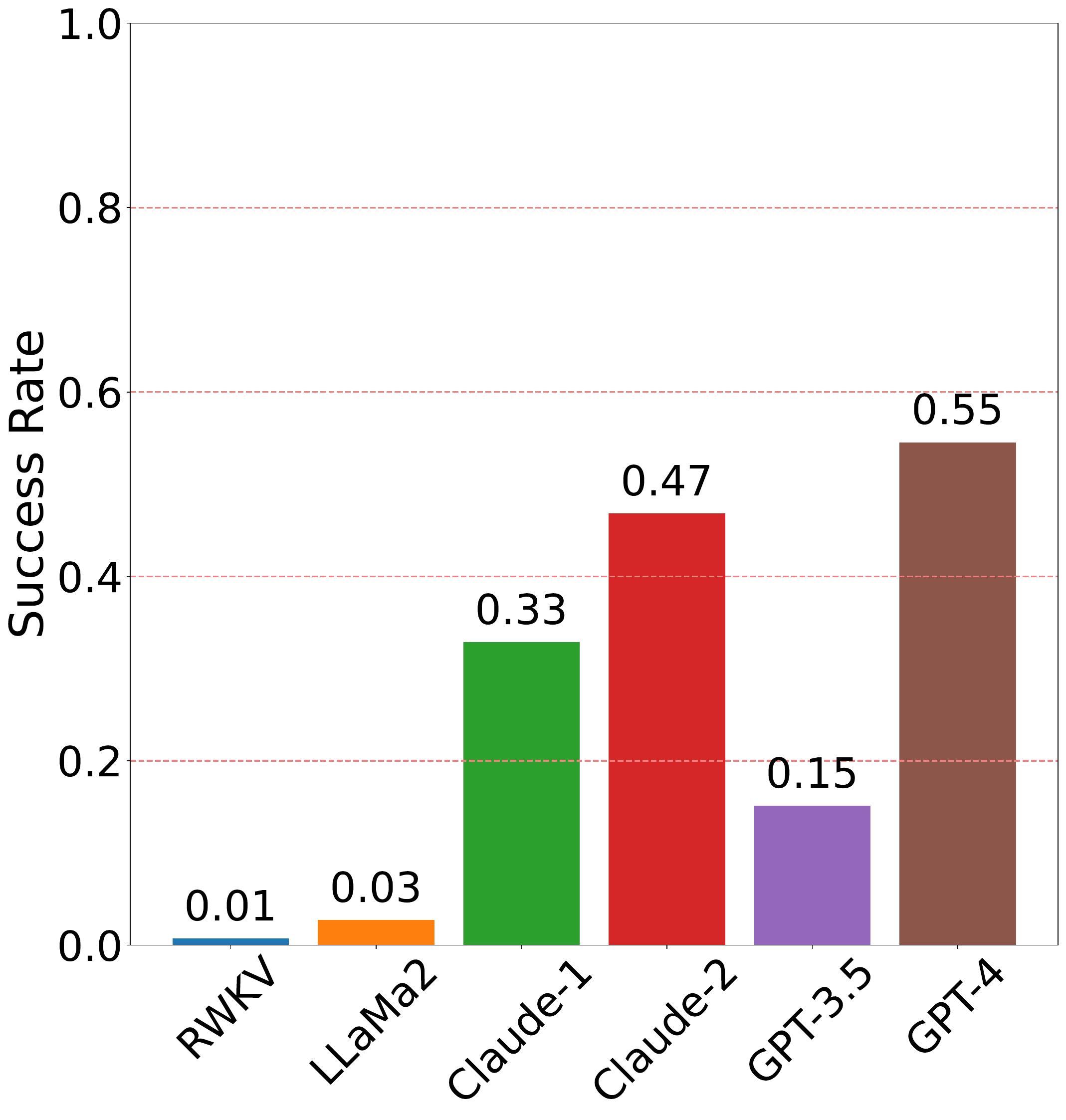}
                \hfill
                \includegraphics[width=0.49\linewidth]{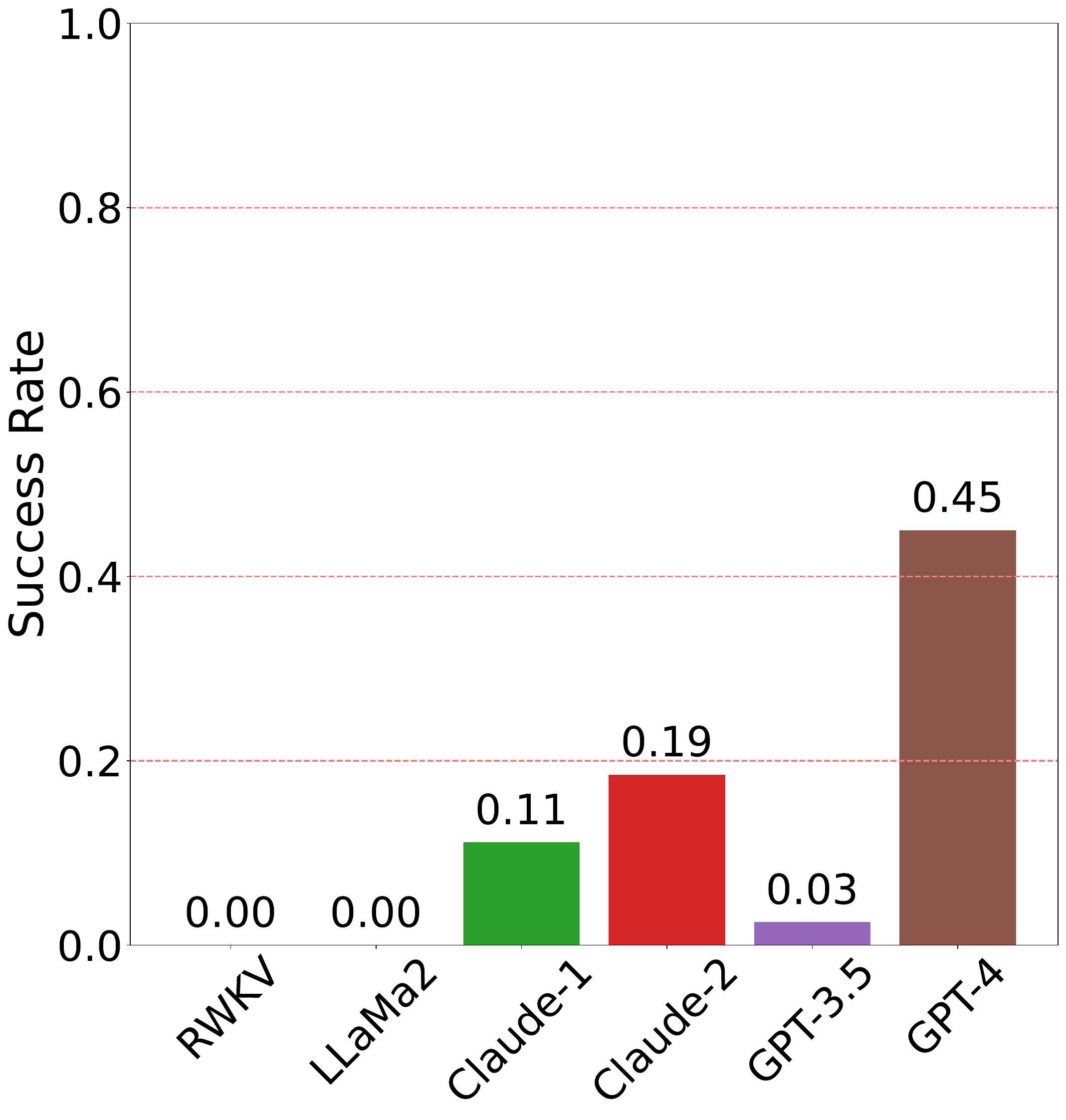}
			\vspace{-16pt}
			\caption{On easy (left) and hard (right) RF questions.}\label{fig:main_rf}
		\end{subfigure}
		\vspace{-8pt}
		\caption{Success rates of the examined models on (\subref{fig:main_df}) DF and (\subref{fig:main_rf}) RF questions, averaged over all $53$ mazes. 
            }
            \label{fig:main}
	\end{center}
\end{figure*}

\paragraph{Main Results}
\label{sec:mainresult}
\cref{fig:main} presents the success rates of all models. 
For each kind of question (i.e., DF or RF), we show the results on easy and hard questions separately. 
As we can see, GPT-4 significantly outperforms all the other models on all kinds of questions. 
However, it only correctly answers half of the RF questions, far worse than what a human could do: in our experiments, humans perfectly answered a randomly sampled set of questions.
\begin{table}[t]
\scriptsize
\setlength{\tabcolsep}{4pt} %
\begin{sc}
\begin{subtable}{1.0\linewidth}
\begin{center}

\begin{tabular}{lccccccc}
\toprule
Method & RWKV & Llama-2 & Claude-1 & Claude-2 & GPT-3.5 & GPT-4 & $\overline{{\rm HARD}}|$ \\
\midrule
RWKV & * & 0.20 | 0.24 & 0.19 | 0.41 & 0.19 | 0.51 & 0.19 | 0.33 & 0.19 | 0.62 & * \\
Llama-2 & 0.43 | 0.20 & * & 0.24 | 0.41 & 0.24 | 0.45 & 0.24 | 0.31 & 0.24 | 0.66 & * \\
Claude-1 & 0.74 | 0.19 & 0.78 | 0.41 & * & 0.36 | 0.44 & 0.38 | 0.32 & 0.36 | 0.57 & * \\
Claude-2 & 0.82 | 0.19 & 0.85 | 0.41 & 0.81 | 0.72 & * & 0.44 | 0.32 & 0.44 | 0.58 & * \\
GPT-3.5 & 0.59 | 0.19 & 0.61 | 0.42 & 0.57 | 0.74 & 0.57 | 0.83 & * & 0.32 | 0.59 & * \\
GPT-4 & 0.86 | 0.19 & 0.90 | 0.42 & 0.84 | 0.72 & 0.83 | 0.81 & 0.86 | 0.57 & * & * \\
|$\underline{{\rm EASY}}$ & * & * & * & * & * & * & *\\
\bottomrule
\end{tabular}
\caption{Pairwise comparison on easy (lower left) and hard (higher right) DF questions.}
\begin{tabular}{lccccccc}
\toprule
Method & RWKV & LlaMa-2 & Claude-1 & Claude-2 & GPT-3.5 & GPT-4 & $\overline{{\rm HARD}}|$ \\
\midrule
RWKV & * & 0.00 | 0.00 & 0.00 | 0.13 & 0.00 | 0.20 & 0.00 | 0.03 & 0.00 | 0.54 & * \\
Llama-2 & 0.02 | 0.02 & * & 0.00 | 0.16 & 0.00 | 0.21 & 0.00 | 0.05 & 0.00 | 0.46 & * \\
Claude-1 & 0.36 | 0.01 & 0.34 | 0.03 & * & 0.11 | 0.19 & 0.13 | 0.03 & 0.11 | 0.45 & * \\
Claude-2 & 0.49 | 0.01 & 0.46 | 0.03 & 0.47 | 0.33 & * & 0.20 | 0.03 & 0.19 | 0.46 & * \\
GPT-3.5 & 0.16 | 0.01 & 0.17 | 0.03 & 0.15 | 0.36 & 0.15 | 0.50 & * & 0.03 | 0.48 & * \\
GPT-4 & 0.57 | 0.01 & 0.56 | 0.03 & 0.55 | 0.33 & 0.55 | 0.47 & 0.58 | 0.15 & * & * \\
|$\underline{{\rm EASY}}$ & * & * & * & * & * & * & *\\
\bottomrule
\end{tabular}

\caption{Pairwise comparison on easy (lower left) and hard (higher right) RF questions.}
\end{center}
\end{subtable}%
\end{sc}
\caption{Success rates on DF and RF questions broken down into pairwise comparisons. In each table, the cell of row-A and col-B contains the success rates of the models---in the format of A | B---on the intersection of the questions that A and B answered individually. The lower left triangle displays the results on easy questions, while the upper right triangle shows the results on hard questions.}
\label{tab:successratepartialmatch}
\end{table}
Note that each model was evaluated on its specific set of questions determined by the length and format of the walkthrough it read. 
To be fair, we also compared each pair of models on the intersection of the questions that they answered. 
The results are presented in \cref{tab:successratepartialmatch}: as we can see, GPT-4 and GPT-3.5 consistently outperform the other models and GPT-4 significantly outperforms GPT-3.5.

\paragraph{Analysis of GPTs}
\label{sec:analysis}
Now, we focus our analysis on the best model, namely GPT-4.
Particularly, we would like to understand the improvements of GPT-4 over GPT-3.5 as well as its current bottlenecks, shedding light on opportunities for future improvements. 

\begin{figure*}[t]
	\begin{center}
	    \begin{subfigure}[t]{0.24\linewidth}
                \begin{center}
                \includegraphics[width=1.00\linewidth]{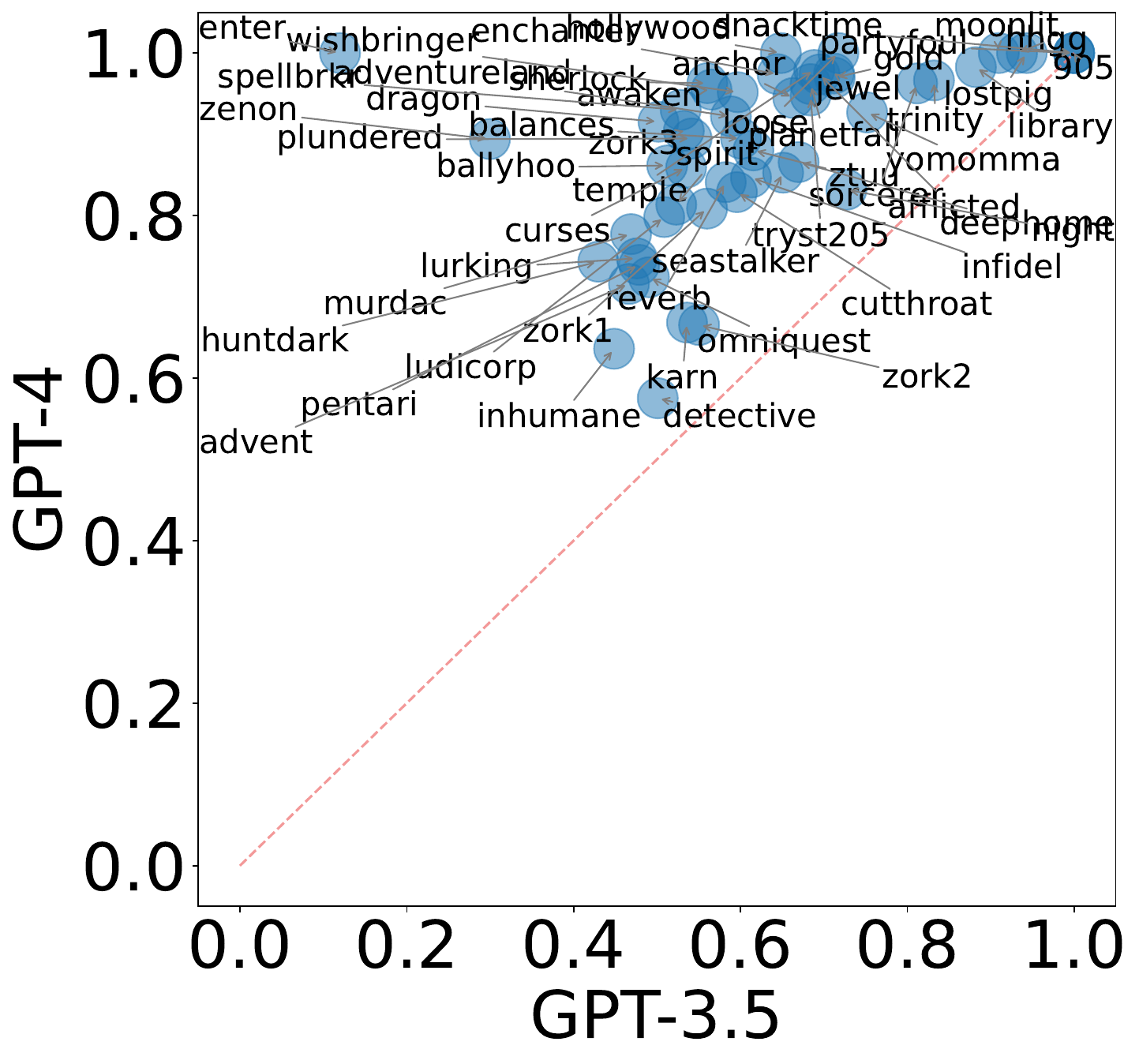}
                \end{center}
                \vspace{-8pt}
               \caption{Easy DF.}\label{fig:gpt3vs4_df_easy}
		\end{subfigure}
            \hfill
            \begin{subfigure}[t]{0.24\linewidth}
                \begin{center}
                \includegraphics[width=1.00\linewidth]{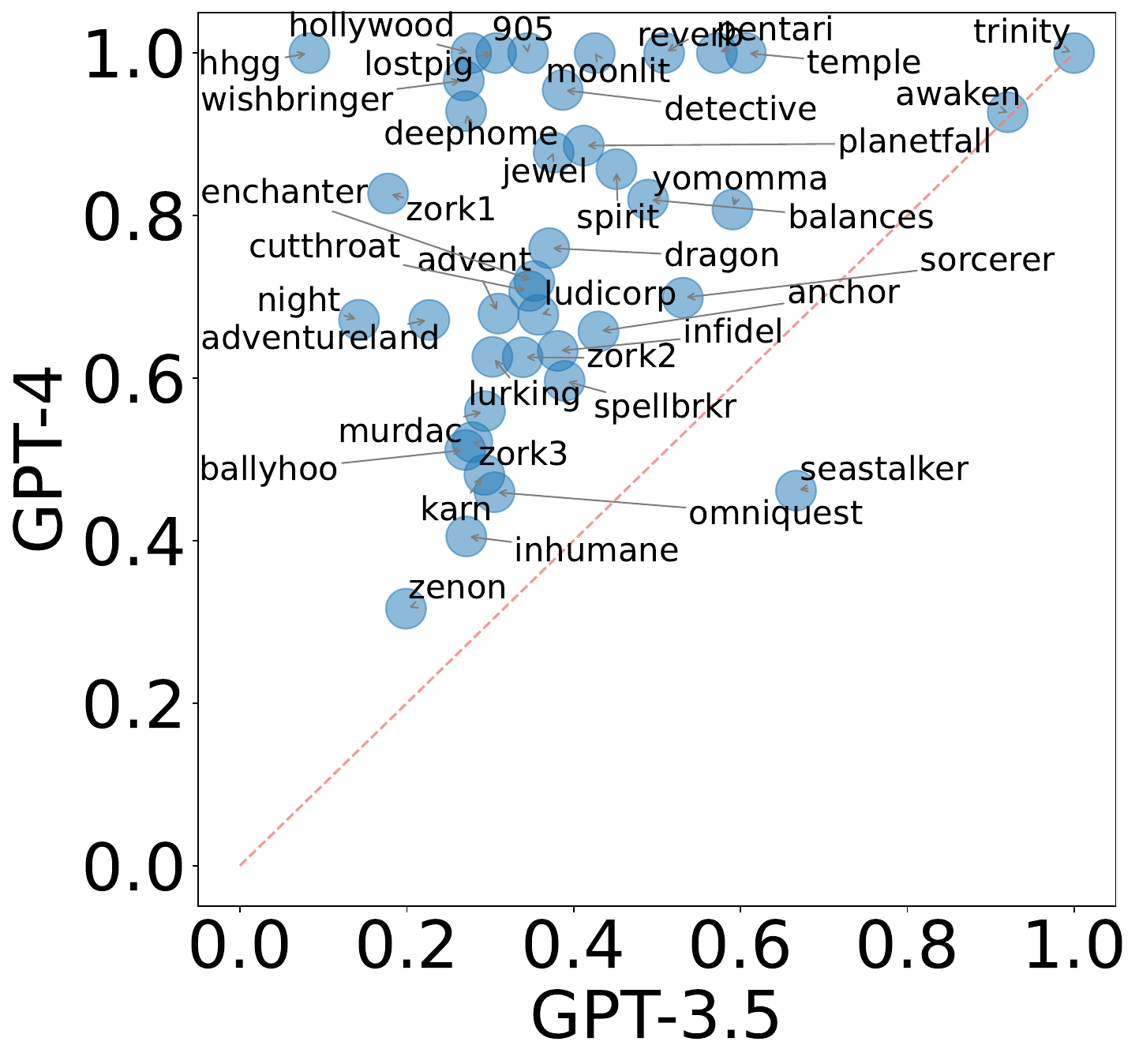}
                \end{center}
                \vspace{-8pt}
               \caption{Hard DF.}\label{fig:gpt3vs4_df_hard}
		\end{subfigure}
		\hfill
            \begin{subfigure}[t]{0.24\linewidth}
                \begin{center}
                \includegraphics[width=1.00\linewidth]{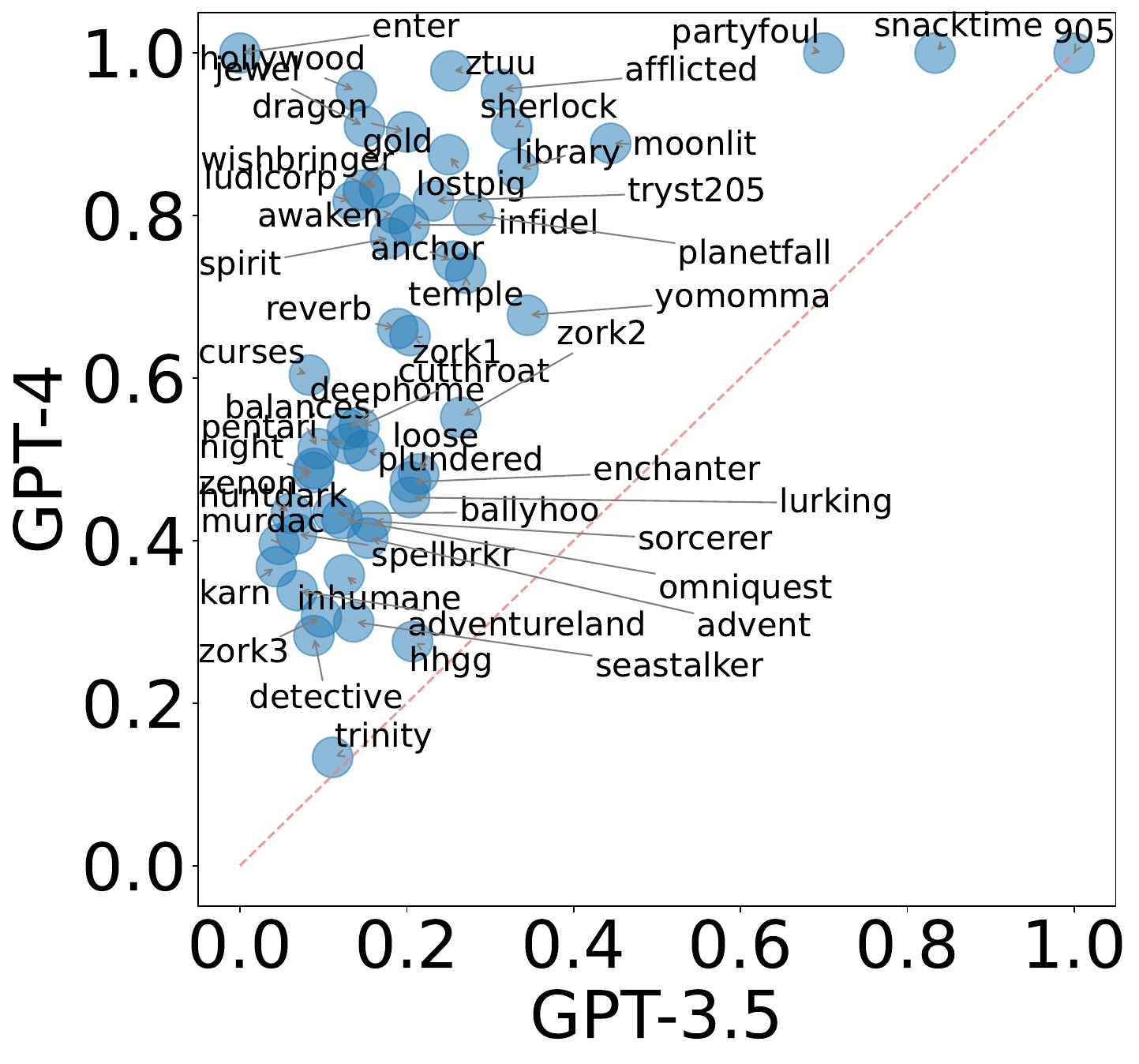}
                \end{center}
                \vspace{-8pt}
			\caption{Easy RF.}\label{fig:gpt3vs4_rf_easy}
		\end{subfigure}
            \hfill
            \begin{subfigure}[t]{0.24\linewidth}
                \begin{center}
                \includegraphics[width=1.00\linewidth]{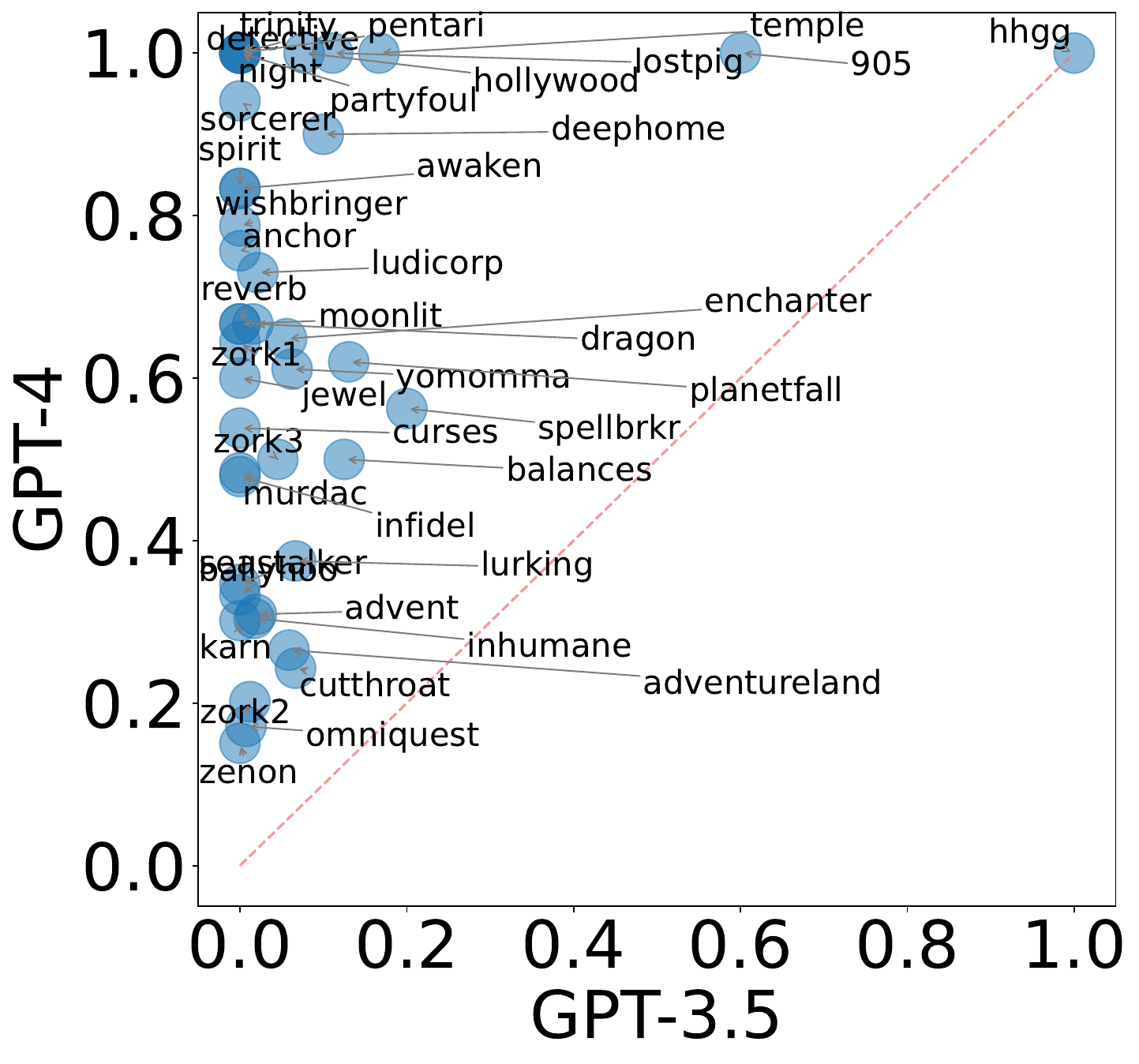}
                \end{center}
                \vspace{-8pt}
			\caption{Hard RF.}\label{fig:gpt3vs4_rf_hard}
		\end{subfigure}
		\vspace{-4pt}
		\caption{
            Success rates of GPT-3.5 and GPT-4 broken down into individual games. 
            }\label{fig:gpt3vs4}
	\end{center}
	\vspace{-8pt}
\end{figure*}

By analyzing the errors of GPT-3.5 and GPT-4, we discovered that these models occasionally hallucinate nonexistent locations or edges. 
Once they made such a mistake at any step of reasoning, they would be misled and deviate from the correct path towards the correct answer. 
Furthermore, we found that the mazes are not equally difficult for the models. 
\cref{fig:gpt3vs4} displays the success rates of the GPT models broken down into their per-game results.
In \cref{fig:gpt3vs4}, each dot is a maze: the $x$-axis coefficient is the performance of GPT-3.5 on this maze while the $y$-axis is that of GPT-4. 
As we can see, the success rates of the models vary across different mazes as well as across different kinds of questions. 
GPT-4 consistently outperforms GPT-3.5 across nearly all the mazes. 
The only exception is Seastalker: there are too few hard DF questions for this maze, and thus it is a noisy outlier. 
Apparently, both GPTs tend to work better on easy questions than on hard questions. 
However, some mazes seem to be particularly challenging to GPT-4, such as Zenon and OMNIQuest.\\

\noindent\textbf{What makes those mazes challenging?}
We collected some important statistics about the mazes and analyzed their correlation with the success rates of the models. 
To understand the success rates on the easy questions, it is interesting to investigate: 
\begin{itemize}
    \item number of locations (\# locations) and number of explicit edges (\# exp edges). They directly measure the size of a maze, which may be a key indicator of its difficulty.
    \item number of potentially confusingly named locations (\# conf locations). Recall from \cref{sec:data} that different locations may have similar or related names, which may confuse a language model. 
    To quantify the number of confusingly named locations, we compute a confusion score for each location, and then sum the scores across all the locations. 
    For a location name $A$, the confusion score is defined to be the maximum word-level edit distance between $A$ and any other location name in the maze, divided by the maximum word-level length of the pair of location names being compared. Technically, it is $\max_{B} \left( \text{edit-distance}(A,B) / \max(\text{len}(A),\text{len}(B)) \right)$, and it is $\in [0,1]$. 
    \item average length of the easy simple paths (avg len easy), i.e., the simple paths that do not include any imputed edges. A longer path may tend to be more difficult for models. 
    \item average number of words in the scene descriptions (avg len scene). The walkthroughs exhibit very diverse styles: for some of them, the text description for each scene is very concise and the name of each location is appropriately highlighted; for others, each description may be verbose (e.g., ten paragraphs and hundreds of words) and the location names are often not obvious from the contexts. 
    It is useful to analyze whether a long scene description poses a challenge for the models. 
\end{itemize}
In order to understand the models' performance on hard questions, we analyze the effects of the variables above (except avg len easy) as well as the following: 
\begin{itemize}
    \item number of imputed edges (\# imp edges); 
    \item average length of hard---i.e., involving imputed edges---simple routes (avg len hard); 
    \item average number of imputed edges in the hard simple routes (avg \# imp in hard). 
\end{itemize}

We use regression analysis to understand the effects of these variables on model performance. 
In particular, for each model on each type of question (DF or RF, easy or hard), we ran single-variable linear regression to understand how the success rate varies with each variable. 
\begin{itemize}
    \item on easy questions, GPTs are significantly influenced by the size of the maze, the confusion level of location name, and the path length. The $p$-values are extremely small. 
    \item on easy questions, the average length of the scene descriptions does not have a significant effect on the performance of GPT-3.5, but interestingly has a significant positive effect on GPT-4's performance. 
    It is perhaps because GPT-4 possesses a strong capability to understand texts and can leverage the rich contexts in each description. 
    This allows it to better distinguish confusingly named locations and establish a better internal representation of the map. %
    However, this richness seems to confuse GPT-3.5 and impede its ability to create a good internal representation of the maze, possibly due to GPT-3.5's weaker overall language understanding capabilities.    
    \item on hard questions, the variables do not significantly affect the performance of GPT-3.5. 
    Note that GPT-3.5 yields very low success rates when answering the hard DF and RF questions. 
    GPT-3.5 seems to struggle when it has to reason about a path with any number of imputed edges, making the effect of other factors less important to its performance.
    \item on hard questions, GPT-4 exhibits a stronger ability to handle paths with imputed edges, compared to GPT-3.5. 
    However, it will experience difficulties when the challenge of inferring imputed edges is amplified by other factors such as the size of the maze or the length of the path. 
    As a result, nearly all the variables have significant effects on GPT-4. 
\end{itemize}

The results of our regression analysis are consistent with the plots in \cref{fig:gpt3vs4}. 
For example, both Zenon and OMNIQuest stay at the lower-left corners of the hard-question plots in \cref{fig:gpt3vs4} since their mazes are particularly challenging to both GPT-3.5 and GPT-4: they both have substantially larger numbers of imputed edges than the other mazes; OMNIQuest also has more locations. 
Wishbringer and Lost Pig have several imputed edges, but their paths are short, so they fall in the upper-left corners of the hard-question plots in \cref{fig:gpt3vs4}.

\paragraph{Does Mapping and Navigation Ability Matter in Downstream Tasks?}
\label{sec:downstream}

Now we present a case study showing that a strong mapping and navigation ability of an LLM would benefit it in downstream tasks. 
In particular, we selected 284 \emph{minigames} in the Jericho game suite, and investigated how the map knowledge may improve the performance of an LLM in playing these minigames.
Each minigame is a selected step from one of 53 textgames; the selection criterion is that the best action to take at this step is a movement. 
Through evaluating an LLM on a minigame, we synthesize a scenario in which the LLM has to figure out the best action to take given its previous actions and observations (i.e., the prefix of walkthrough up to the current step). 
Note that this task is different and more challenging than answering the DF and RF questions: the LLM is not explicitly given a route (as in DF questions) or a destination (as in RF questions), but has to spontaneously figure out which action may contribute to its long-term goal.
\begin{figure}[t]%
\begin{minipage}[t]{0.62\linewidth}
\vspace{0pt}
\begin{lstlisting}[caption={A prompt of the playing game experiments},label={lst:playgameprompt}]
@\ldots@           @\textcolor{commentgray}{\# previous actions and observations}@
Small local map info: if you want to go to North of House, you should go south; if you want to go to Up a Tree, you should go up; if you want to go to Altar, you should go west. 
Consider what you should do next, and choose one appropriate action from the valid actions list: [up, take on egg, put down egg, go around forest, throw egg at tree, open egg with all, north, south, west, east]
Please just tell me the selected action without any extra words.
\end{lstlisting}
\end{minipage}
\begin{minipage}[t]{0.37 \linewidth}
\vspace{0pt}
\includegraphics[width=0.9\linewidth]{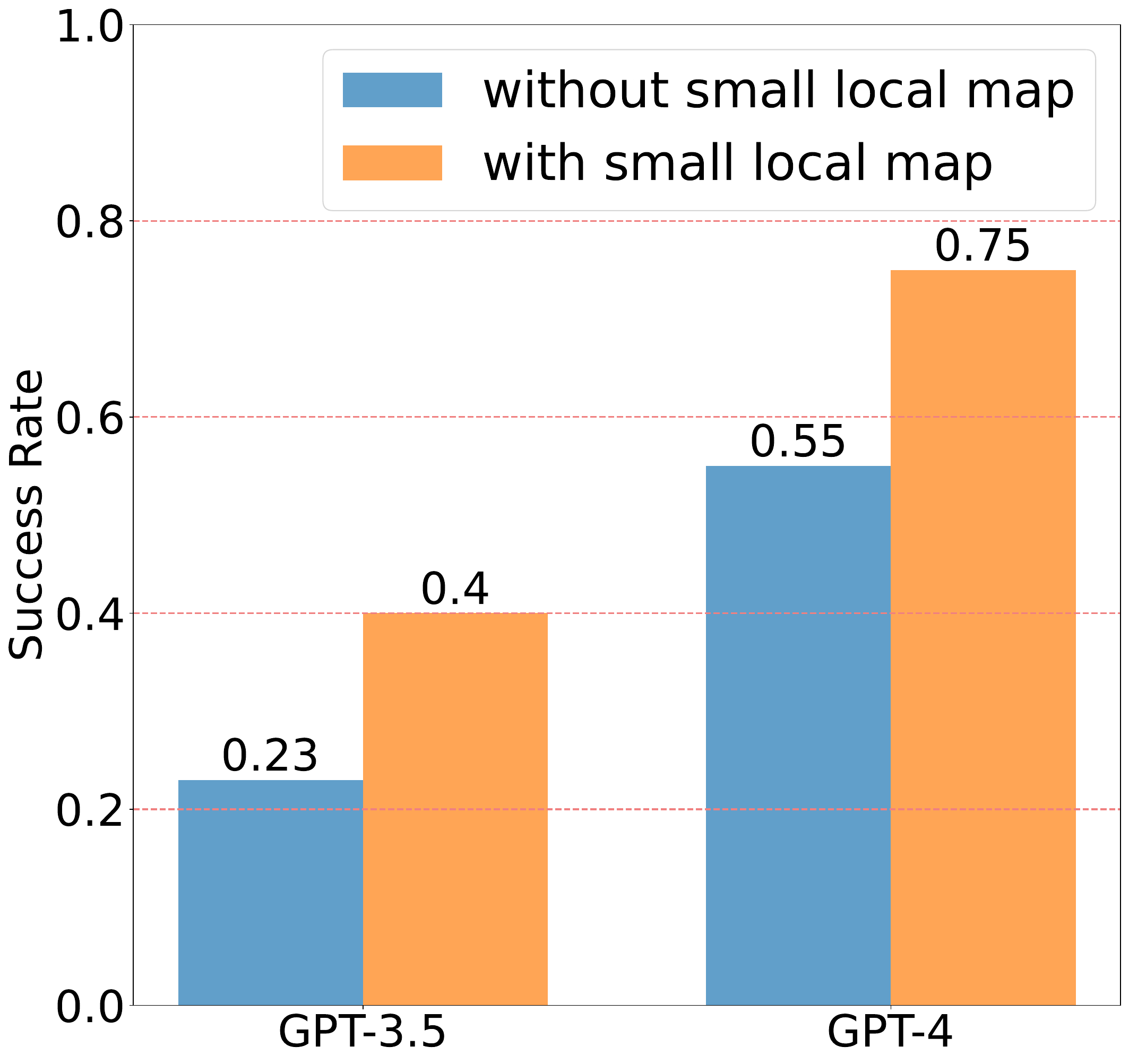}
\vspace{-2pt}
\caption{Playing minigames.}\label{fig:game_results}
\end{minipage}
\vspace{-8pt}
\end{figure}

For this task, we evaluated GPT-3.5 and GPT-4. 
For each model, we tried two settings: the first is to condition the LLM on the walkthrough like \cref{lst:walkthrough}; the second is to include in the prompt the information about the nearby locations, and an example of the full prompt is given in \cref{lst:playgameprompt}. 
The information about nearby locations is the ground-truth information that the LLM, in principle, should have learned from the walkthrough prefix. 
If the LLM had a perfect mapping and navigation ability, it would be able to perfectly spell it out and use that information to guide its decision making.
\cref{fig:game_results} presents the results of this experiment. 
GPT-4 significantly outperforms GPT-3.5 in playing these minigames, consistent with their relative performance when answering the DF and RF questions of our MANGO benchmark. 
For each of the GPT models, having access to nearby location information significantly improves its performance, demonstrating that a strong mapping and navigation ability is essential to succeeding at relevant downstream tasks.

%% file: papers/mango/sections/conclusion.tex
We present MANGO, a benchmark that evaluates the mapping and navigation abilities of large language models.
Our benchmark covers a diversity of mazes as well as a variety of evaluation and analysis programs, offering a comprehensive testbed in a great breadth and depth.
In our experiments, the current best model still performs poorly on the benchmark, with a sharp degradation on the more difficult questions. 
We release our benchmark---along with the source code for data generation and evaluation---to track the advances of the mapping and navigation capabilities of future LLMs as well as to facilitate future research in related areas. 

%% file: papers/transcrib3d/sections/abstract.tex
\begin{figure}[H]
    \centering
    \includegraphics[width=\linewidth]{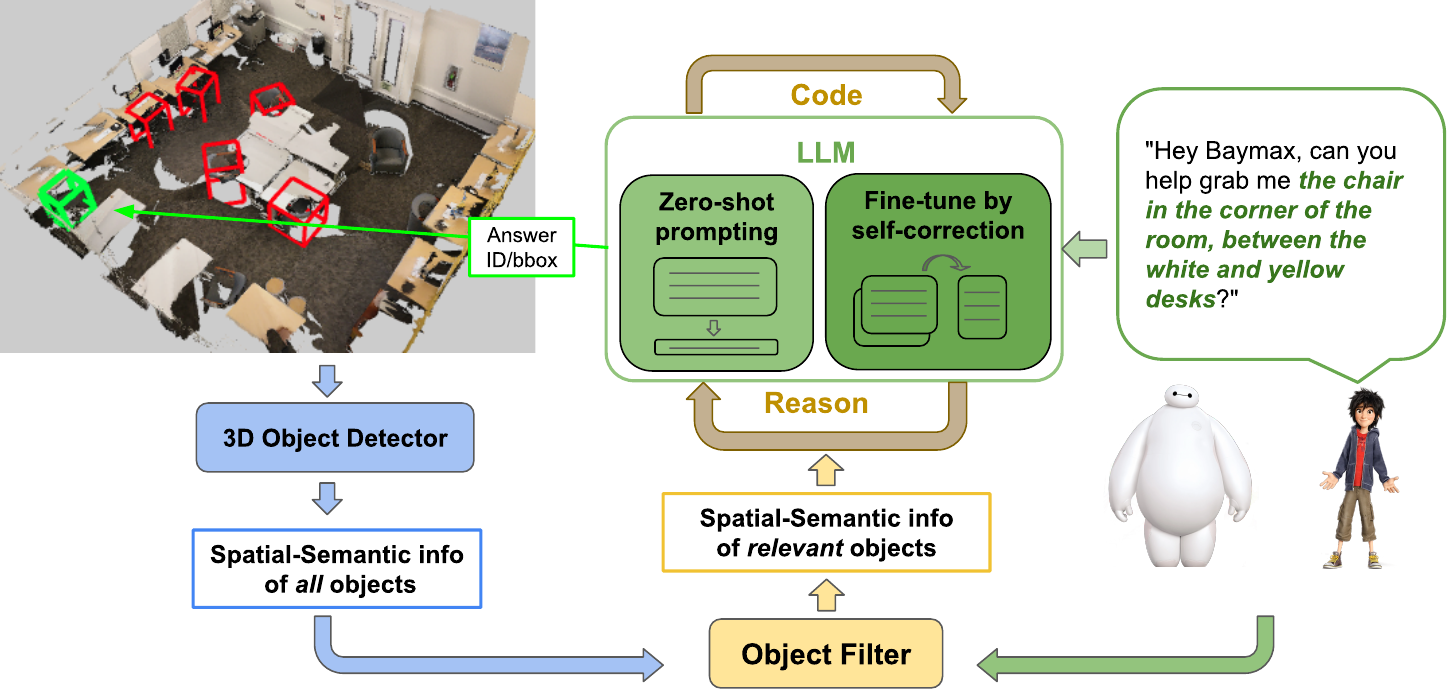}
    \caption{The overall \textbf{Transcrib3D\xspace} framework, which takes as input the colored point-cloud and referring expression (in \textcolor{mygreen}{green}), and outputs the ID or bounding box of the referent object. To resolve the referring expression \emph{``the chair in the corner of the room, between the white and yellow desks''}, the framework needs to locate the pillow in the \textcolor{green}{green} box, while all other pillows in \textcolor{red}{red} boxes are distractors. 
    }
    \label{fig:pipeline}
\end{figure}

If robots are to work effectively alongside people, they must be able to interpret natural language references to objects in their 3D environment. Understanding 3D referring expressions is challenging---it requires the ability to both parse the 3D structure of the scene and correctly ground free-form language in the presence of distraction and clutter. 
We introduce Transcrib3D\xspace, an approach that brings together 3D detection methods and the emergent reasoning capabilities of large language models (LLMs). Transcrib3D\xspace uses \emph{text} as the unifying medium, which allows us to sidestep the need to learn shared representations connecting multi-modal inputs, which would require massive amounts of annotated 3D data.
As a demonstration of its effectiveness, Transcrib3D\xspace achieves state-of-the-art results on 3D reference resolution benchmarks, with a great leap in performance from previous multi-modality baselines.
To improve upon zero-shot performance and facilitate local deployment on edge computers and robots, we propose self-correction for fine-tuning that trains smaller models, resulting in performance close to that of large models.
We show that our method enables a real robot to perform pick-and-place tasks given queries that contain challenging referring expressions. Code is available at \url{https://ripl.github.io/Transcrib3D}.

%% file: papers/transcrib3d/sections/intro.tex
Comprehending a natural language expression that mentions an object within a given environment is a routine activity for humans. It often occurs as part of a question (e.g., ``Whose jacket is hanging on the black chair?'') or an instruction (e.g., ``Pass me the smaller yellow mug.'').
This capability is important for embodied agents that work with humans to accomplish tasks. While humans excel in this task with over 90\% accuracy on existing benchmarks~\cite{Achlioptas2020ReferIt3DNL}, contemporary methods %
only achieve mediocre accuracy. They typically rely on supervised learning, e.g., training a Transformer module to obtain contextualized embeddings of various modalities (i.e., text, image, and point-cloud), from which the final prediction is made by a small decoder.

However, bridging different modalities in a latent space is challenging. For instance, state-of-the-art 2D vision-language understanding models like CLIP~\cite{Radford2021LearningTV} require billions of image-text pairs for training, yet still exhibiting a limited grasp of compositional and relational concepts~\cite{liu2022compositional}. The challenge intensifies in 3D domains, where annotated data is much more scarce~\cite{Deitke2023ObjaverseXLAU}. As a result, the capacity of existing models to perform 3D reference resolution is limited.

On the contrary, philosophers such as Ludwig Wittgenstein argue that our understanding of reality is confined by the language we use, who famously stated, ``The limits of my language mean the limits of my world.''
This concept underpins our approach, wherein we propose to employ \emph{text} as the unifying medium to bridge the gap between 3D scene parsing and referential reasoning.
This approach is grounded in realizing that the challenge of resolving 3D referring expressions can be fundamentally divided into two components: \emph{detection} (identifying objects in the scene) and \emph{reasoning} (associating one of the candidates with the referring expression). 
By harnessing text as a cohesive bridge, we can capitalize on the recent advancements in 3D detection and the enhanced natural language reasoning abilities offered by large language models (LLMs)~\cite{OpenAI2023GPT4TR}.

\begin{figure}[t]
    \centering
    \includegraphics[width=0.7\linewidth]{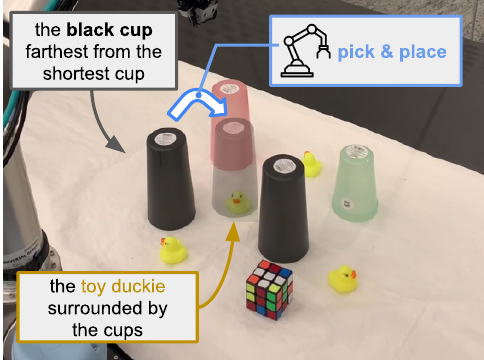}
    
    \caption{Transcrib3D\xspace enables a robot to resolve complex 3D referring expressions necessary to to follow pick-and-place instructions. In this example, the robot is tasked with a natural language instruction that includes challenging referring expressions \emph{``cover \textcolor{yellow}{\textbf{the toy duckie surrounded by the cups}} with \textbf{the black cup farthest from the shortest cup}''}.}
    \label{fig:real-robot-demo}
\end{figure}
Specifically, from the results of an off-the-shelf 3D detector~\cite{Schult23ICRA}, Transcrib3D\xspace first converts the detected spatial and semantic 3D scene information---the category, location, spatial extent, and color of objects---into texts, thereby creating an object-centric 3D scene \emph{transcript} (hence the name of our method, \textbf{Transcrib3D\xspace}). We then filter out non-relevant objects in regard to the query (e.g., information about a trash can is not relevant to the expression ``the white pillow on top of the chair and next to a blue pillow''). Subsequently, we compose a prompt that incorporates the filtered 3D transcript and the referring expression, and process it through an LLM-based reasoning mechanism. The reasoning module incorporates three key elements to make LLMs more effective and generalizable for the task:
\begin{enumerate*}[label=(\arabic*)]
    \item Iterative code generation and reasoning;
    \item Principle-guided zero-shot prompting; and
    \item Fine-tuning from self-reasoned correction.
\end{enumerate*}

We evaluate our method on standard 3D referring expression resolution benchmarks, ReferIt3D~\cite{Achlioptas2020ReferIt3DNL} and ScanRefer~\cite{Chen2019ScanRefer3O}, achieving state-of-the-art performance on both. We also perform real robot experiments (Fig.~\ref{fig:real-robot-demo}) that task a robot manipulator with following natural language commands that require sophisticated 3D spatial-semantic reasoning, demonstrating the practicality of the method.

%% file: papers/transcrib3d/sections/related_works.tex
The problem of resolving 3D referring expressions has garnered significant attention of late, in large part due to the introduction of the ReferIt3D~\cite{Achlioptas2020ReferIt3DNL} and ScanRefer~\cite{Chen2019ScanRefer3O} benchmarks. ReferIt3D contains two subsets: SR3D, which consists of template-based utterances, and NR3D, which consists of human-sourced free-form utterances.
Contemporary methods~\cite{Huang2022MultiViewTF,Jain2021BottomUT,Yang2021SAT2S,Chen2022LearningPH} perform 3D referring expression resolution by aggregating different input modalities into contextualized embeddings using Transformer architectures~\cite{Vaswani2017AttentionIA} in an end-to-end fashion.
MVT~\cite{Huang2022MultiViewTF} projects 3D information into 2D to achieve better feature encoding. BUTD-DETR~\cite{Jain2021BottomUT} fine-tunes detected 3D bounding boxes within the Transformer. SAT~\cite{Yang2021SAT2S} uses 2D semantics during training to learn a mapping from the query to its 3D grounding. 
HAM~\cite{Chen2022LearningPH} presents a hierarchical alignment model that learns multi-granularity visual and linguistic representations. Different from these methods, NS3D~\cite{Hsu2023NS3DNG} proposes a neuro-symbolic framework that utilizes a language-to-code model to generate programs, where each module is represented by neural networks. D3Net~\cite{Chen2021D3NetAU} and 3DJCG~\cite{Cai20223DJCGAU} jointly learn 3D captioning and grounding together, where D3Net proposes self-critical training while 3DJCG proposes task-agnostic shared modules and separate task-specific heads. ViL3DRel~\cite{Chen2022LanguageCS} designs a spatial self-attention layer to account for relative distances and orientations between 3D objects. 3D-VisTA~\cite{Zhu20233DVisTAPT} performs pre-training on a dataset of $278$k 3D scene-text pairs, and fine-tunes the model on specific tasks.

In contrast, we propose to connect 3D detection and the LLM reasoning module via a textual representation, which spares us from learning the joint representation of different input modalities from limited 3D annotated data.

\subsection{Grounding Large Language Models}
LLMs trained on Internet-scale text data have shown dominant performance across various NLP tasks~\cite{OpenAI2023GPT4TR}. However, LLMs have to be grounded to answer questions or execute actions in the physical world.
VisProg~\cite{Gupta2022VisualPC} uses in-context learning for LLMs to generate code for 2D image processing tasks.
SayCan~\cite{Ahn2022DoAI} and SayPlan~\cite{Rana2023SayPlanGL}
instead modulate the LLM outputs with a model of the perceived environment (i.e., ``affordances'' for SayCan and scene graphs for SayPlan). 

When it comes to explicitly reason over 3D inputs, one branch of works train multi-modal models that directly incorporate 3D representations into the token library. 3D-LLM~\cite{Hong20233DLLMIT} incorporates distilled 3D features~\cite{Kobayashi2022DecomposingNF} with language tokens. 3D-VisTA~\cite{Zhu20233DVisTAPT} performs multi-modal fusion of language and PointNet++ features~\cite{Qi2017PointNetDH} with self-supervised masked encoding training.

Instead of multi-modal approaches, studies have investigated the use of textual 3D information to enhance downstream tasks. For instance, \citet{Feng2023LayoutGPTCV} employ textual 3D data in conjunction with LLMs to generate 3D indoor scenes. Similarly, \citet{Yu2023CoNavGPTMC} use LLMs for multi-robot navigation, leveraging extracted 3D scene information. We see this body of work as indicative of the promise of textual 3D scene information, and a motivation for our approach. Concurrent with our work, \citet{Yang2023LLMGrounderO3} also use LLMs to reason over textual data, however, their methodology differs in its use of a three-step, task-specific reasoning process as opposed to our general, flexible approach.

\subsection{LLM Reasoning}
There are many techniques proposed to enhance the reasoning capabilities of LLMs~\cite{Wei2022ChainOT, Yao2023TreeOT, Yoneda2023StatlerSL}. The reasoning module of our method is similar in spirit to the general framework of ReAct~\cite{yao2022react}, where each round of code generation is followed by a round of LLM analysis to proceed with the reasoning or debugging, although our context is 3D-specific. Self-correction is also studied in LLM community. In contrast to Reflexion \cite{shinn2024reflexion}, which uses test-time environmental feedback, we instead collect feedback from training samples, and use the corrected samples to fine-tune models.

%% file: papers/transcrib3d/sections/methodology.tex
Figure~\ref{fig:pipeline} illustrates our proposed Transcrib3D\xspace framework.
Given the input colored point-cloud, Transcrib3D\xspace first applies a 3D object detector to generate an exhaustive list of objects in the scene transcribed as text (Section ~\ref{sec:detect_and_transcribe}). The list is then filtered to identify objects relevant to the provided referring expression (Section~\ref{sec:object_filter}). The resulting filtered object information along with the referring expression serve as input to the LLM-based reasoning module for inference. To reach the final answer, the reasoning module includes a ``code interpreter'' mode in which the LLM iterates between code generation and reasoning over outputs from code execution (Section~\ref{sec:code_interpreter}). We consider two options for interfacing with the LLM:
\begin{enumerate*}[label=\arabic*)]
    \item zero-shot with principles-guided prompting, and
    \item fine-tuning from self-reasoned correction
\end{enumerate*}
(Section~\ref{sec:self_correction}).

\paragraph{Detect and Transcribe 3D Information}
\label{sec:detect_and_transcribe}

Given a colored point-cloud of the scene, we first perform semantic segmentation of objects using Mask3D~\cite{Schult23ICRA}. We associate with each detected object its category based on the semantics, center location, spatial extent according to the 3D bounding box, as well as its mean color. The 3D orientation of an object can also be incorporated using PartNet~\cite{Mo2018PartNetAL}. Transcrib3D\xspace compiles the information associated with all detected objects as a list to form an object-centric scene description. The following shows an example of such a scene transcript (exact numeric values and duplicate object categories are replaced by \dots).
\begin{prompt}[h]
\begin{tcolorbox}[
    enhanced,
    attach boxed title to top left={xshift=6mm,yshift=-3mm},
    colback=light_cornflower_blue_2!20,
    colframe=light_cornflower_blue_2,
    colbacktitle=light_cornflower_blue_2,
    title=Spatial Semantic Scene Description for All Objects,
    fonttitle=\bfseries\color{black},
    boxed title style={size=small,colframe=light_cornflower_blue_2,},
]
scene0592: ... Scene center: [...] ... objs list:\\
monitor, id=0, ctr=[...], size=[...], rgb=[...];\\
...\\
box, id=5, ctr=[...], size=[...], rgb=[...];\\
copier, id=6, ctr=[...], size=[...], rgb=[...];\\
floor, id=7, ctr=[...], size=[...], rgb=[...]; \\
wall, id=8, ctr=[...], size=[...], rgb=[...];\\
...\\
armchair, id=15, ctr=[...], size=[...], rgb=[...];\\
...\\
chair, id=19, ctr=[...], size=[...], rgb=[...];\\
...
\end{tcolorbox}
\label{prompt:scene_description}
\vspace{-3ex}
\end{prompt}

\paragraph{Pre-Filtering Relevant Objects for Utterance}
\label{sec:object_filter}

The aforementioned procedure results in a representation of every object detected in the scene. However, only a small fraction of these objects will typically be relevant to the given referring expression. 
\noindent Given the utterance,
``\texttt{chair in the corner of the room, between white and yello desks}'',
the model identifies the following objects as relevant,
\begin{tcolorbox}[
    enhanced,
    attach boxed title to top left={xshift=6mm,yshift=-3mm},
    colback=light_yellow_2!20,
    colframe=light_yellow_2,
    colbacktitle=light_yellow_2,
    title=Spatial Semantic Scene Description for Relevant Objects,
    fonttitle=\bfseries\color{black},
    boxed title style={size=small,colframe=light_yellow_2,},
]
scene0592: ... Scene center: [...] ... objs list:\\
    wall, id=8, ctr=[...], size=[...], rgb=[...];\\
    ...\\
    armchair, id=15, ctr=[...], size=[...], rgb=[...];\\
    ...\\
    chair, id=19, ctr=[...], size=[...], rgb=[...];\\
    ...
\end{tcolorbox}
\noindent and the remainder as irrelevant.
Simplifying the object list by filtering out irrelevant items (taking into account synonyms and hypernyms) not only reduces processing time and the number of tokens for LLMs, but also facilitates reasoning by reducing content that may potentially be distracting. Such an approach has been shown to improve the efficiency of language grounding~\cite{patki19}.

\paragraph{Iterative Code Generation and Reasoning}
\label{sec:code_interpreter}

\begin{figure}[!t]
    \centering
    \includegraphics[width=0.8\linewidth]{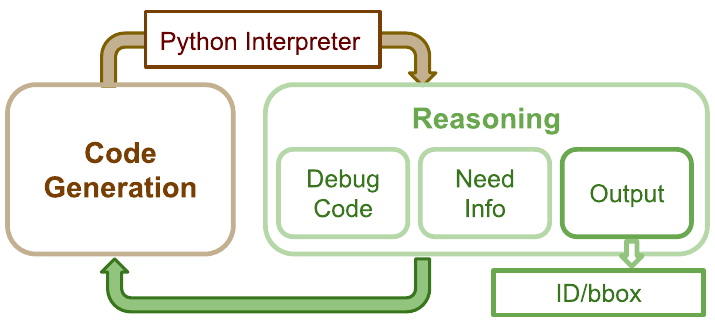}
    \caption{Illustration of the iterative code generation and reasoning process. After the code generation, execution results from a local Python interpreter is fed back to the LLM for further reasoning. The LLM then proceeds to either 1) fix code errors when encountered, 2) generate additional code to obtain more information, or 3) output the referred objects if it has all the needed information. This process continues until the LLM believes the reasoning to be complete.}
    \label{fig:code-gen}
\end{figure}
Compositional reasoning that involves arithmetic calculations, which is crucial for spatial reasoning, is well known to be a weakness of Transformers and LLMs~\cite{Dziri2023FaithAF}. However, most contemporary approaches to 3D referring expression resolution rely on a single forward pass of a Transformer to reach the final answer, which can be limiting in terms of reasoning power. In order to avoid this weakness, Transcrib3D\xspace equips the LLM with a Python interpreter and directs the LLM to generate code whenever quantitative evaluations are necessary. Transcrib3D\xspace then locally executes the generated code using a Python interpreter and appends the output to the original conversation. The resulting prompt is sent back to the LLM to generate the next response. If there is any error in executing the generated code, Transcrib3D\xspace also feeds information about the error back to the LLM for re-generation. This process continues until the LLM believes that the reasoning is complete. Figure~\ref{fig:code-gen} provides an illustration of this process.

We want to emphasize that our design does not distinguish between code generation and chat reasoning steps. Such a framework simplifies the design, allows for flexible reasoning, and makes it easy for fine-tuning. But in turn, it requires the LLM to have the ability to do both, so language models that are specifically trained for only coding or only chatting will not perform the best within our framework.

\paragraph{Principles-Guided Zero-Shot Prompting}
\label{sec:prompt_principles}

Spatial reasoning can require complex compositional logic that is challenging for LLMs~\cite{Dziri2023FaithAF}. For example, when handling left/right spatial relations, which should factor in the observer's viewpoint, LLMs often naively ground them with smaller/larger values in the $x$-coordinates. In order to overcome these deficiencies, Transcrib3D\xspace employs a few general principles to guide LLM reasoning in a zero-shot fashion, including how to 1) use HSL space to match colors; 2) perform vector operations to resolve directional relations; 3) calculate the point-to-plane distance to determine who's closer to a wall, to name just a few.
We find that this set of guiding principles works across 3D referring benchmarks. 
The full prompt will be included in the released code.

\paragraph{Fine-tuning from Self-Reasoned Correction}
\label{sec:self_correction}

\begin{prompt}[!t]
\begin{tcolorbox}[
    enhanced,
    attach boxed title to top left={xshift=6mm,yshift=-3mm},
    colback=dark_green_1!20,
    colframe=dark_green_1,
    colbacktitle=dark_green_1,
    title=Self-Correction for fine-tuning,
    fonttitle=\bfseries\color{black},
    boxed title style={size=small,colframe=dark_green_1},
]
\colorbox{light_green_1}{Assistant:}\\
Based on the calculated metrics, we can see that chair 49 has the smallest distance to the nearest two walls, which means it is most likely to be in the corner of the room. It also has the smallest distance to the nearest white desk and a relatively small distance to the nearest yellow desk, which means it is likely to be between a white desk and a yellow desk. Therefore, chair 49 is the most likely to be the chair described in the description. Now the answer is complete -- {'ID':49}

\colorbox{light_green_1}{User:}
The correct answer is object 18. Can you double check the information of object 18 and the given prompt and see where you got wrong?

\colorbox{light_green_1}{Assistant:}
Let's revisit the information provided for object 18 and the constraints given in the prompt to identify where the mistake might have occurred.

Object 18 is described as follows:
- Type: Chair
- Center: [-2.98, -3.31, 0.39]
- Size: [0.53, 0.61, 0.81]
- RGB: [60, 58, 50]

The constraints from the prompt are:
1. The chair is in the corner of the room.
2. The chair is between a white and a yellow desk.

Upon reviewing the code execution results, it was noted that chair 18 has a relatively small distance to the nearest two walls, indicating it could be in the corner of the room. Additionally, chair 18 has a very small distance to the nearest white desk (0.66) and a reasonable distance to the nearest yellow desk (2.58), which aligns with the constraint of being between a white and a yellow desk.

\colorbox{light_green_2}{The mistake in the initial analysis was not properly considering the combination of} \colorbox{light_green_2}{both constraints together.} While chair 49 had the smallest distance to the nearest two walls and a very small distance to the nearest white desk, it did not optimally satisfy the condition of being between a white and a yellow desk as well as chair 18 did. Chair 18's proximity to both a white and a yellow desk, along with its corner location, makes it the correct answer according to the given constraints.

\end{tcolorbox}
\caption{An example of self-reasoned correction for fine-tuning. During self-correction, the LLM realizes its mistake during the initial analysis.}\label{prompt:self_correction}
\end{prompt}

Rule-based systems~\cite{Winograd1971ProceduresAA} are effective for relatively simple domains that involve a limited set of spatial-semantic concepts and structured language, but struggle to generalize, particularly to open-world domains~\cite{tellex2020robots}. %
Motivated by the now well known benefits of data-driven alternatives to rule-based methods, we adopt a novel fine-tuning method for LLMs that enables learning beyond the given set of rules (i.e., the general guiding principles) by enabling the model to learn from its own mistakes. In effect, we seek to endow LLMs with introspection capabilities. We do so via the following procedure:
\begin{enumerate}%
    \item Use the prompt with general principles on the training set for the LLM to generate an initial set of answers with elaborated reasoning process.
    \item For any incorrect answer, augment the original prompt with the correct object ID and request the LLM to reflect on why the original answer is incorrect (``What went wrong?''). This is followed by a request to output ``clean'' reasoning for the correct answer.
    \item Gather the reasoning processes of the correct examples, and the re-generated ones of the initially incorrect examples to produce the dataset for LLM fine-tuning.
\end{enumerate}
Prompt~\ref{prompt:self_correction} shows an example of the self-reasoned correction step. Note that after fine-tuning, we no longer include general principles in the prompt. In this way, the system not only incorporates the guiding principles, but further improves itself by learning from self-reasoned correction.

%% file: papers/transcrib3d/sections/experiments.tex
We evaluate the effectiveness of Transcrib3D\xspace using the ReferIt3D~\cite{Achlioptas2020ReferIt3DNL} and ScanRefer~\cite{Chen2019ScanRefer3O} benchmarks. ReferIt3D formulates 3D referring expression understanding as the multiple-choice problem: given a set of segmented objects in a 3D scene along with a corresponding referring expression, identify the unique referent object from the set, typically containing several instances of the same fine-grained category. 
There are five different types of relations in SR3D, namely ``horizontal'', ``vertical'', ``support'', ``between'' and ``allocentric'', which make up approximately $81\%$, $4\%$, $2\%$, $8\%$, and $5\%$ of the data, respectively. ReferIt3D measures performance in terms of accuracy.  Both Sr3D and Nr3D are
split by “Easy”/“Hard” and “ViewDep”/“ViewIndep”. "Easy" samples has one or none distractors in a scene, while "Hard" samples have two or more. 
The view-dependent samples contain language descriptions
that rely on viewing directions. Unlike ReferIt3D, ScanRefer does not provide object segmentation information, and instead tasks methods with returning the 3D bounding box of the target object, given the query utterance and a colored point-cloud. ScanRefer measures performance in terms of accuracy conditioned on the intersection-over-union (IoU) of the ground-truth 3D bounding box over the predicted one, with thresholds at $25\%$ (acc@0.25IoU) or $50\%$ (acc@0.5IoU).

\begin{figure}[!t]
    \centering
    \begin{subfigure}[b]{0.48\linewidth}
         \centering
         \includegraphics[width=\textwidth]{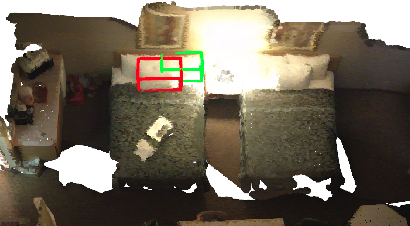}
         \caption{Utterance: ``On the bed with the laptop its the pillow by the light.'' This query requires understanding three constraints together: ``On the bed'', ``with the laptop'', and ``by the light''.} \label{fig:comp_example_1}
     \end{subfigure}
     \hfill
     \begin{subfigure}[b]{0.48\linewidth}
         \centering
         \includegraphics[width=\textwidth]{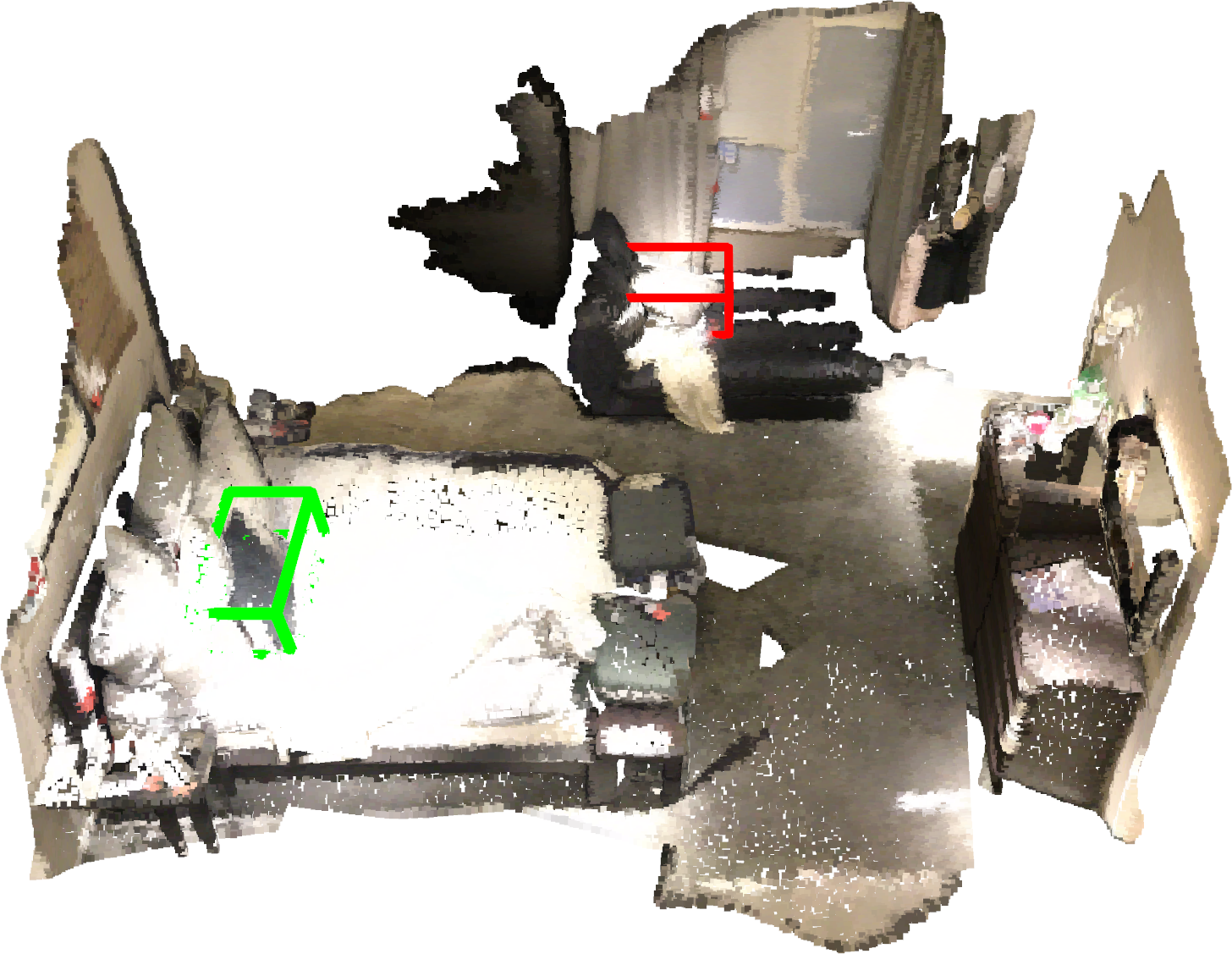}
         \caption{Utterance: ``The foremost pillow on the bed of the group of pillows.'' This query requires understanding that the pillow needs to be among a group.}
         \label{fig:comp_example_2}
     \end{subfigure}
    \caption{Qualitative comparisons between Transcrib3D (ours, in \textcolor{green}{green}) and 3D-VisTA (in \textcolor{red}{red}) on the NR3D dataset.}
    \label{fig:qual_comp_nr3d}
\end{figure}

\paragraph{Grounding Accuracy on ReferIt3D}
We test different variations of our method against contemporary baselines on the SR3D and NR3D subsets of ReferIt3D. On NR3D, we evaluate our best model, Transcrib3D\xspace (GPT4-P), on the full test set, while other variants of Transcrib3D\xspace are evaluated on a subset of $500$ randomly sampled data points. On SR3D, we evaluate all Transcrib3D\xspace models on the same subset of $500$ random samples from the test set.\footnote{We use a subset of the test set due to the cost of evaluating all variations of Transcrib3D\xspace on the full test set, however we believe that the results would be similar on the full test set due to the templated nature of the utterances.}
Table~\ref{tab:ReferIt3D} presents the results. Figure~\ref{fig:qual_comp_nr3d} also provides a qualitative comparison between 3D-VisTA~\cite{Zhu20233DVisTAPT} and Transcrib3D\xspace that highlights the strength of our method with regards to complex reasoning.
\begin{table*}[!ht]
    \centering
    \begin{tabularx}{1.0\linewidth}{Xcccccc}
        \toprule
        & \multicolumn{3}{c}{Nr3D} & \multicolumn{3}{c}{Sr3D} \\
        \cmidrule(lr){2-4}\cmidrule(lr){5-7}
        Method & Overall & Easy & Hard & Overall & Easy & Hard \\
        \midrule
        SAT$^\dagger$~\cite{Yang2021SAT2S} & 49.2 & 56.3 & 42.4 & 57.9 & 61.2 & 50.0 \\
        BUTD-DETR$^\dagger$~\cite{Jain2021BottomUT} & 54.6 & 60.7 & 48.4 & 67.0 & 68.6 & 63.2 \\
        MVT$^\dagger$~\cite{Huang2022MultiViewTF} & 59.5 & 67.4 & 52.7 & 64.5 & 66.9 & 58.8 \\
        \midrule
        ViL3DRel$^\mathsection$~\cite{Chen2022LanguageCS} & 64.4 & 70.2 & 57.4 & 72.8 & 74.9 & 67.9 \\ 
        3D-VisTA$^\mathsection$~\cite{Zhu20233DVisTAPT} & 64.2 & 72.1 & 56.7 & 76.4 & 78.8 & 71.3 \\
        \midrule
        Transcrib3D\xspace \scriptsize{(GPT-3.5-NP)} & 33.8 & 42.6 & 25.0 & 79.3 & 82.8 & 70.7 \\
        Transcrib3D\xspace \scriptsize{(GPT-3.5-P)} & 46.6 & 56.0 & 37.1 & 80.0 & 80.8 & 78.1 \\
        Transcrib3D\xspace \scriptsize{(GPT-4-NP)} & 64.5 & 71.8 & 57.1 & 97.4 & 98.4 & 94.7 \\
        Transcrib3D\xspace \scriptsize{(GPT-4-P)} & \textbf{70.2} & \textbf{79.7} & \textbf{60.3} & \textbf{98.4} & \textbf{99.2} & \textbf{96.2} \\
        \bottomrule
    \end{tabularx}
    \caption{Grounding accuracy (\%) on Nr3D and Sr3D. $^\dagger$denotes results from the official benchmarks while $^\mathsection$denotes results reported in the respective papers. ``P'': ``with principles'', ``NP'': ``no principles''. All our models are equipped with interactive code generation. Transcrib3D\xspace with GPT-4 and general principles surpasses all baselines by a large margin. 
    } 
    \label{tab:ReferIt3D}
\end{table*}

\paragraph{Grounding Accuracy on ScanRefer}
We test the best variant (GPT4-P) of our method with different detection modules against baselines on the ScanRefer benchmark, which unlike ReferIt3D, does not provide methods with the ground-truth bounding boxes during inference. %
In addition to comparing to baseline results reported in their respective papers, we also re-run the checkpoints from 3D-VisTA~\cite{Zhu20233DVisTAPT}, the best baseline method, on the same $500$ random samples from the ScanRefer validation set. Table~\ref{tab:scanrefer} presents the results. We achieve state-of-the-art performance on ScanRefer with both detected or ground-truth bounding boxes. It is worth noting that the performance gain over the baseline is greater with ground-truth compared to detected bounding boxes. This is partly due to the fact that the baseline method is trained with the lower-quality detected bounding boxes, which ironically leads to the ground-truth ones being out of distribution. 

\begin{table}[!ht]
    \centering
    \begin{tabularx}{0.8\linewidth}{Xcccc}
        \toprule
        \multirow{2}{*}{Method} & \multirow{2}{*}{Data} & \multirow{2}{*}{Det.} & \multicolumn{2}{c}{Overall} \\
                                 & & & acc@0.25 & acc@0.5 \\
        \midrule
        ViL3DRel~\cite{Chen2022LanguageCS} & Full & PG & 47.9 & 37.7 \\
        3D-VisTA~\cite{Zhu20233DVisTAPT} & Full & M3D & 50.6 & 45.8 \\
        3D-VisTA~\scriptsize{(re-run)} & Full & M3D & 50.7 & 45.9 \\
        \midrule
        3D-VisTA~\scriptsize{(re-run)} & Part. & M3D & 50.6 & 44.6 \\
        Transcrib3D\xspace & Part. & M3D & 51.2 & 44.4 \\
        Transcrib3D\xspace + Cam & Part. & M3D & 51.3 & 45.5 \\
        \midrule
        3D-VisTA~\scriptsize{(re-run)} & Part. & GT & 55.6 & 55.6 \\
        Transcrib3D\xspace & Part. & GT & 62.0 & 62.0 \\
        Transcrib3D\xspace + Cam & Part. & GT & 64.2 & 64.2 \\
        \bottomrule
    \end{tabularx}
    \caption{Grounding accuracy (\%) on ScanRefer. ``Full'': the full validation set of ScanRefer consisting of 5410 samples; ``Part.'': a subset of 500 random samples from the validation set, which is the same for all methods; ``Det.'': the 3D object detection module used in the model;
    ``PG'' stands for  PointGroup~\cite{jiang2020pointgroup}, while ``M3D'' stands for Mask3D~\cite{Schult23ICRA} (where the detection accuracy is 56.7 for PG
    and 73.7 for M3D on the ScanNet dataset for the mAP@0.5 metric~\cite{Schult23ICRA}), and ``GT'' for ground-truth bounding boxes. We test our method and re-run 3D-VisTA on the same subset of 500 samples with the Mask3D detector, GT bounding boxes, and an additional ``+ Cam'' setting, where camera view information provided by the ScanRefer dataset is also included in the scene transcript. Note that ScanRefer allows the use of all provided data modalities and ranks methods on the same benchmark regardless. The zero-shot nature of our method allows ease use of this extra information.} \label{tab:scanrefer}
\end{table}

\paragraph{Effects of Fine-tuning Methods}
We study the effects of our proposed approach that involves fine-tuning from self-reasoned corrections. Following Section~\ref{sec:self_correction}, we run Transcrib3D\xspace with GPT-4 on 500 samples from the NR3D training set and collect both correct and incorrect examples. For the incorrect examples, we let the LLM re-generate the self-corrected reasoning. We use the combination of correct and self-corrected examples to fine-tune the smaller model (i.e., \texttt{gpt-3.5-turbo}) for 3 epochs. We then evaluate all models on the same subset of 500 random samples from the NR3D test set. Notably, we remove all rule-based prompts from the fine-tuning data so that the fine-tuned models are not constrained to human-designed rules, but are instead adaptive to new rules learned via self-correction.

Note that inference is typically faster with a smaller model. %
A fine-tuned small model with comparable performance to that of a large model on this task would be more desirable for local deployment on edge computers or robots, with additional privacy benefits.
Table~\ref{tab:fine-tuning} presents the results.

\begin{table}[h]
    \centering
    \begin{tabularx}{0.8\linewidth}{Xccc}
        \toprule
        Models & Total & Easy & Hard \\
        \midrule
        GPT-3.5 (zero-shot) & 33.8 & 42.6 & 24.0 \\
        GPT-3.5 (correct-only fine-tuning) & 60.7 & 52.8 & 41.9  \\
        GPT-3.5 (self-correct fine-tuning) & 61.5 & 54.6 & 80.0  \\
        GPT-4 (zero-shot) & 69.4 & 78.7 & 60.0 \\
        \bottomrule
    \end{tabularx}
    \caption{Performance of fine-tuning models on NR3D. Fine-tuned GPT-3.5 models demonstrate a significant improvement in performance when compared to zero-shot models, closely approaching the capabilities of GPT-4. The model fine-tuned on self-corrected examples sees an increase in performance compared to that fine-tuned on only correct examples, particularly for hard queries. Notably, the fine-tuned models are not provided with rule-based prompts during both fine-tuning and inference time, suggesting that implicit decision rules are learned from examples.}
    \label{tab:fine-tuning}
\end{table}

\paragraph{Referring Expressions for Robot Manipulation}

We demonstrate how Transcrib3D\xspace supports a robot's ability to follow natural-language instructions for pick-and-place manipulation, particularly when complicated referring expressions are involved. %
As a core capability of robot manipulation, language-guided pick-and-place involves (i) breaking down the language instruction into two referring expressions, one each for the ``pick'' and ``place'' identities; (ii) resolving the referring expressions in the context of the robot's surrounding environment; and (iii) performing the pick-and-place actions. We conduct the pick-and-place task with a Universal Robots UR5 arm equipped with a Robotiq 3-Finger Adaptive Robot Gripper placed in a table-top setting. Figure~\ref{fig:real-robot-demo} visualizes the execution of one utterance using our method. 

To parse the given instruction, we employ a language model-generated program (LMP) from few-shot prompting as in Code-as-Policies~\cite{Liang2022CodeAP}, which is instructed to call the \mintinline{python}{put_first_on_second(arg1, arg2)} function with desired arguments \mintinline{python}{arg1} and \mintinline{python}{arg2}. This approach allows free-form texts as input, which is more flexible and natural than template parsing. The following is an example that is used in the prompt for the LMP:
\begin{minted}[fontsize=\footnotesize]{python}
'# query: Pick up the orange between the apples and place it in the bowl
# with a banana in it.
put_first_on_second("orange between the apples", "bowl with a banana in it")'
\end{minted}

The \mintinline{python}{put_first_on_second(arg1, arg2)} function first composes an exhaustive list of objects in the environment along with their spatial-semantic details. To do so, the function employs MDETR~\cite{kamath2021mdetr}, an open-vocabulary object segmentation method, that takes as input an RGB image and an object category, and outputs the 2D spatial attributes of all objects in the scene of that category. The 2D attributes are then lifted to 3D using depth information from a Realsense RGB-D camera. Repeating the process for all candidate object categories results in a 3D spatial-semantic transcript 
of all objects in the scene. Transcrib3D\xspace then reasons over this transcript along with the referring expression to identify the object in question. Finally, we compute the pick and place poses accordingly to control the robot's end effector. We provide the pseudo-code below.
\begin{minted}[fontsize=\footnotesize]{python}
def put_first_on_second(self, arg1, arg2):
    # obtain the objects list in the environment
    objs = self.env.get_objs()
    pick_id = get_obj_id(objs, arg1) # Referring
    place_id = get_obj_id(objs, arg2) # Referring
    pick_pose, place_pose = self.get_obj_pose(objs, pick_id, place_id)
    self.env.step({'pick': pick_pose, 'place': place_pose}) # robot manipulation
\end{minted}

We apply our method to robot manipulation by integrating Transcrib3D\xspace as the perception process in the Code-as-Policies (CaP)~\cite{Liang2022CodeAP} framework (``CaP+Transcrib3D\xspace''). As a baseline, we compare to the standard implementation of Code-as-Policies, which uses MDETR~\cite{kamath2021mdetr} for perception (``CaP\xspace''). When grounding an object in the scene, CaP\xspace assumes that the object can be uniquely identified by MDETR according to its associated noun phrase (e.g., ``the duckie''). CaP\xspace then simply selects the output grounding with the highest score. In order to make the comparison fair, we extend CaP's prompt so that it is able to use the entire referring expression as the object identifier, which is indeed unique. Otherwise, CaP would perform poorly in our setting, since it has never seen complicated referring expressions in the original prompts.

\begin{figure}[!t]
     \centering
     \begin{subfigure}[b]{0.49\linewidth}
         \centering
         \includegraphics[width=\textwidth]{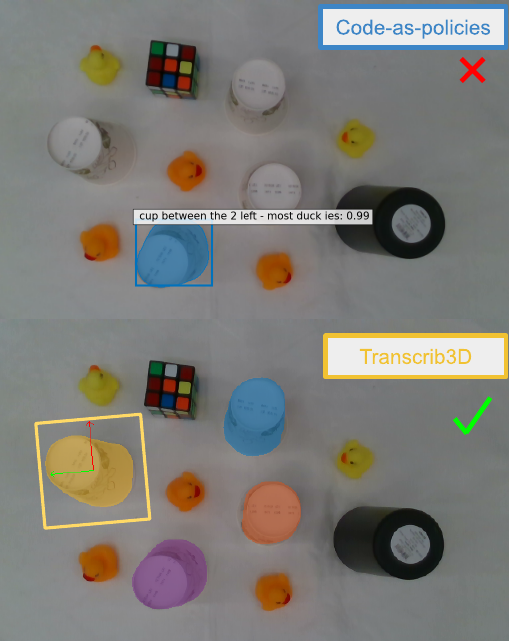}
         \caption{``the cup between the two left-most duckies''}
         \label{fig:cup between the 2 left-most duckies}
     \end{subfigure}
     \begin{subfigure}[b]{0.49\linewidth}
         \centering
         \includegraphics[width=\textwidth]{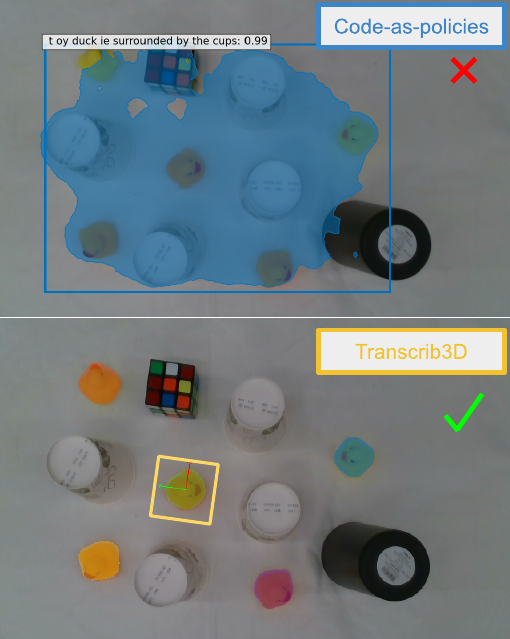}
         \caption{``the toy duckie surrounded by the white cups''}
         \label{fig:toy duckie surrounded by white cups}
     \end{subfigure}
        \caption{Qualitative comparison of the grounding performance of (top) CaP\xspace and (bottom) CaP+Transcrib3D\space on a real robot.}
        \label{fig:real_robot_comparison}
\end{figure}
We evaluate the performance of both methods on five different natural language instructions (each involving two referring expressions). Figure~\ref{fig:real_robot_comparison} shows a qualitative comparison for one of the test cases.
Our results reveal that when assessed at both the referring expression and instruction levels, CaP+Transcrib3D\xspace significantly outperforms CaP\xspace. CaP+Transcrib3D\xspace achieves an instruction success rate of 80\% (4 out of 5) compared to 20\% (1 out of 5) for CaP\xspace, and correctly resolves 90\% of the referring expressions (9 out of 10) compared to 40\% (4 out of 10) for CaP\xspace.
This underscores a pronounced benefit in employing Transcrib3D\xspace for resolving referring expressions within robot pick-and-place tasks.

%% file: papers/transcrib3d/sections/discussion.tex
\begin{figure}[!t]
    \centering
    \begin{subfigure}[b]{0.48\linewidth}
         \centering
         \includegraphics[width=\textwidth]{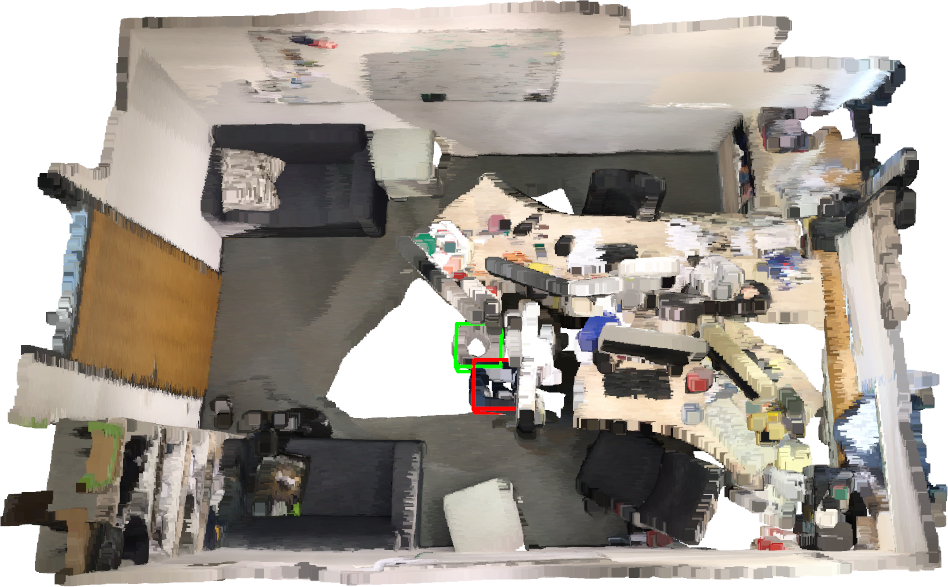}
         \caption{Utterance: ``The cylinder shaped trash can.'' Our model is unable to ground the ``cylinder'' shape, leading to failure.}
         \label{fig:cylinder_failure}
     \end{subfigure}
     \hfill
     \begin{subfigure}[b]{0.48\linewidth}
         \centering
         \includegraphics[width=\linewidth]{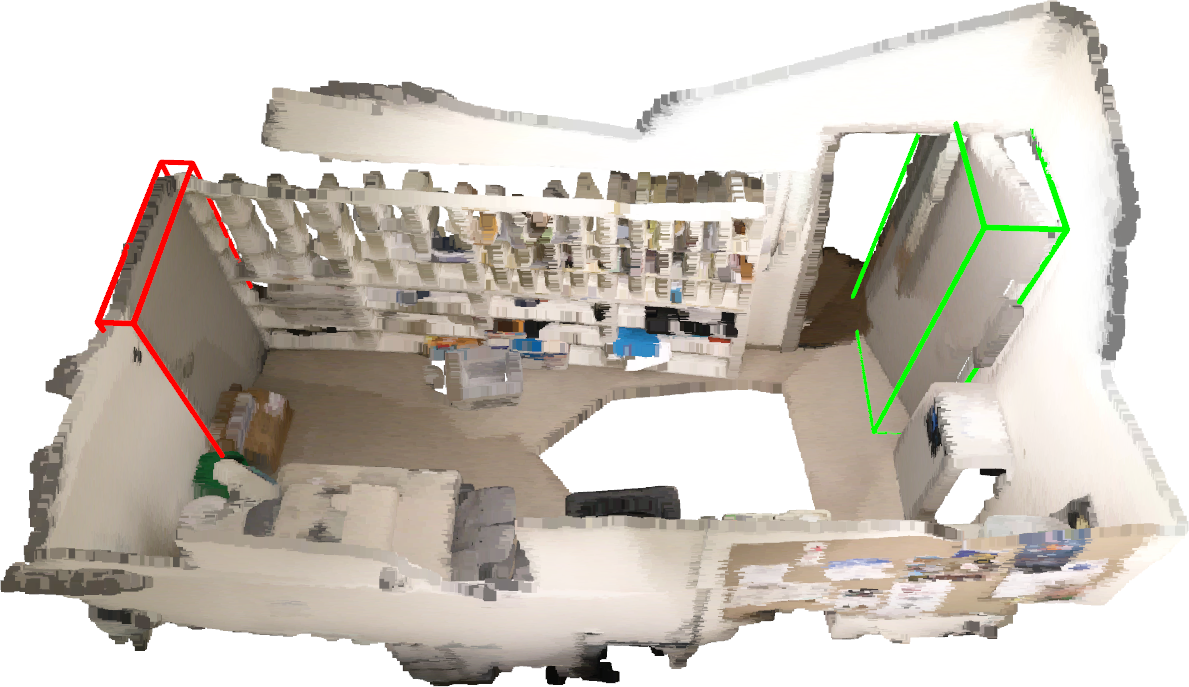}
         \caption{Utterance: ``This door is open and has nothing blocking it.'' This query requires understanding the concept of ``open'', which is hard to infer from a bounding box.}
         \label{fig:failure_2}
     \end{subfigure}
    \caption{Typical failure cases in Transcrib3D\xspace, with ground-truth bounding boxes shown in \textcolor{green}{green} and ours in \textcolor{red}{red}.}
    \label{fig:failure_cases}
\end{figure}

We acknowledge that Transcrib3D\xspace is not without limitations.
First, our scene transcript is object-centric. Although object-level details such as bounding boxes are sufficient for numerous 3D spatial reasoning tasks, there exist scenarios that necessitate a finer level of object details. We show examples of two such cases in Figure~\ref{fig:failure_cases}.
Second, our reliance on existing 3D detectors introduces a constraint: the quality of 3D detection itself. In our experiments, we observed that even state-of-the-art 3D detection methods~\cite{Liu2021GroupFree3O, Schult23ICRA} yield sub-optimal results, highlighting the room for improvement in 3D detection. %
Third, our method involves manual specification of the desired information to be extracted from 3D detections (e.g., each object's center, size, and orientation), which has its limitations. An adaptive feature selection strategy could potentially yield better results.

However, even with those limitations, our method surpasses all current multi-modality baselines. This achievement leads us to propose two critical insights: firstly, the connection module facilitating interaction between different modalities may not be as effective since our experiment can be roughly regarded as a controlled test that replaces a typical multi-modal cross-attention module with just \emph{text}, and demonstrates that it works better; and secondly, the reasoning module might be too simplistic to ground the complex logic inherent in natural language.
We believe that the observed limitations in current multi-modality methods can largely be attributed to the scarcity of annotated 3D %
data, which is orders-of-magnitude smaller than its 2D counterpart, primarily due to the higher cost in its collection. We call for more efficient data collection pipelines in 3D and robotics, potentially leveraging generative methods.

%% file: papers/transcrib3d/sections/conclusion.tex
We introduce Transcrib3D\xspace, a simple and effective method for the 3D grounding of natural language referring expressions that requires no training, yet delivers state-of-the-art performance across leading 3D reference resolution benchmarks. The idea of using text as a unifying medium to connect the 3D scene and LLM reasoning not only achieves great results, but also provides critical insights into the bottleneck of current multi-modal methods.

%% file: papers/statler/sections/abstract.tex
\begin{figure}[H]
    \centering
    \includegraphics[width=\textwidth]{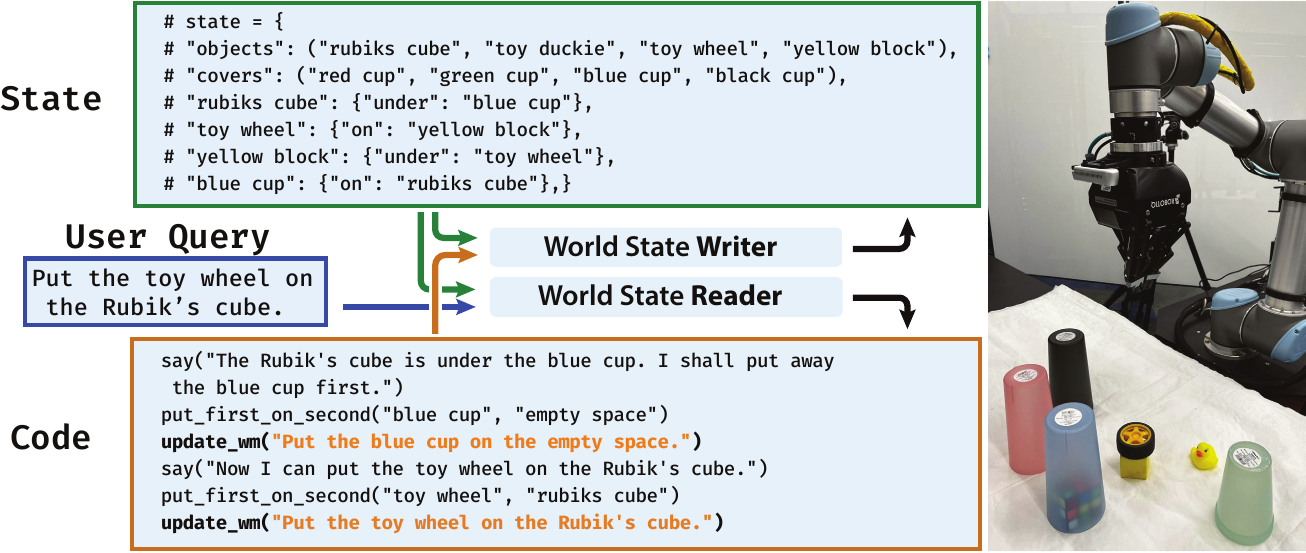}
    \caption{Overall Framework of Statler}
    \label{fig:statler-framework}
\end{figure}

There has been a significant research interest in employing large language models to empower intelligent robots with complex reasoning. 
Existing work focuses on harnessing their abilities to reason about the histories of their actions and observations. 
In this paper, we explore a new dimension in which large language models may benefit robotics planning. 
In particular, we propose Statler\xspace, a framework in which large language models are prompted to maintain an estimate of the world state, which are often {unobservable}, and track its transition as new actions are taken. %
Our framework then conditions each action on the estimate of the current world state.
Despite being conceptually simple, our Statler\xspace framework significantly outperforms strong competing methods (e.g., Code-as-Policies) on several robot planning tasks. 
Additionally, it has the potential advantage of scaling up to more challenging long-horizon planning tasks.

%% file: papers/statler/sections/introduction.tex
Large language models (LLMs) exhibit strong reasoning capabilities that are harnessed to perform a wide range of downstream tasks such as dialogue and code generation~\cite{Kojima2022LargeLM,
DBLP:journals/corr/abs-2107-03374,DBLP:journals/corr/abs-2303-08774}. 
The robotics community has recently seen a significant interest in empowering robots with LLMs, enabling them to understand natural language commands and perform tasks that require sophisticated reasoning~\cite{Ahn2022DoAI,Liang2022CodeAP, Huang2022InnerME,yang2023foundation}. 
However, existing methods are model-free: they use LLMs as policy functions that generate future actions only conditioned on previous actions and observations. %

In this paper, we propose a simple yet effective model-based approach. 
Our framework---named Statler---maintains a running estimate of the world \underline{stat}e by prompting large \underline{l}anguage models and performs multistep \underline{e}mbodied \underline{r}easoning conditioned on the estimated state. 
\cref{fig:diagram} illustrates this framework. 
In particular, Statler utilizes a pair of prompted LLMs: instructed by a few demonstrations, the \textbf{world-state reader} takes as input the user query, reads the estimated world state, and generates an executable action (e.g, a code snippet); instructed by another set of demonstrations, the \textbf{world-state writer} updates the world state estimate based on the action. 
This mechanism resembles how a domain-specific formal language tracks a symbolic world state~\cite{Nordmann2014ASO}, but enjoys greater flexibility since pretrained large language models are known to be domain-agnostic. As we will see soon in \cref{sec:experiments}, the prompts in our experiments are generic and users of our framework will have minimal design workload. 

Our Statler framework is primarily inspired by classical model-based reinforcement learning. 
In a model-based approach, an environment (or world) model learns to capture the dynamics of the environment (e.g., possible outcomes of an action) so that the policy conditioned on the model state will take more informed actions~\citep{sutton98}.
In our framework, the LLMs have acquired massive amounts of commonsense knowledge from pretraining, and they are elicited---by a few demonstrations---to behave as an environment model, estimating the world state and facilitating decision making. 

Another motivation of our design is to handle missing data. 
In robotics tasks, we often have to cope with latent world dynamics that are not directly observable.
In such scenarios, explicitly maintaining an estimated world state improves decision making, although the estimates might be imperfect. 
This is analogous to graphical models with latent variables: spelling out latent variables and imputing their values is often helpful for reasoning about the target variables, although the imputation may not be perfect~\citep{koller2009probabilistic}. 

The final motivation of our state-maintaining design is its potential to scale up to long-horizon planning tasks. 
In multistep reasoning and planning, an LLM has to implicitly maintain the world state in its internal representation, which has been demonstrated to be difficult in previous work~\citep{Anthropic,nelson23,Sun2021DoLL,valmeekam2023planning,guan2023leveraging,liu2023llm+, silver2023generalized}. 
By explicitly maintaining an estimated world state, our framework makes it easier to track and consult the world state at any step of reasoning and planning, thus carrying a higher chance of success in long-horizon tasks. 

In the following sections, we will show that our framework performs as expected: 
in \cref{sec:motivation}, we demonstrate the concept with a pedagogical example; 
in \cref{sec:method}, we introduce the Statler framework; in \cref{sec:experiments}, we present the experiments, in which our framework significantly outperforms strong competing methods such as Code-as-Policies~\cite{Liang2022CodeAP}.

\begin{figure*}[t]
\begin{minipage}[t]{\linewidth}
\begin{minipage}[t]{0.32\linewidth}
\begin{lstlisting}[caption={The prompt and \sethlcolor{aigreen}\hl{desired output} of a vanilla LLM.},label={lst:cup-ball-vanilla-prompt},firstnumber=auto]
@
\textcolor{commentgray}{\# Initial state}
@
cups = [False, True, False] @\label{line:init-cup-state}@
Swapping cup 1 with cup 2
Swapping cup 0 with cup 2
Swapping cup 1 with cup 2
cups = @\sethlcolor{aigreen}\hl{[True, False, False]}@ @\label{line:gptout}@  
@ @
@ @
\end{lstlisting}
\end{minipage}
\hfill
\begin{minipage}[t]{0.32\linewidth}
\begin{lstlisting}[caption={The prompt and \sethlcolor{aigreen}\hl{desired output} of LLM+CoT.},label={lst:cot-cup-ball-prompt},firstnumber=auto]
@
\textcolor{commentgray}{\# Initial state}
@
cups = [False, True, False] @\label{line:init-cup-state}@
Swapping cup 1 with cup 2
Swapping cup 0 with cup 2
Swapping cup 1 with cup 2
cups = @\sethlcolor{aigreen}\hl{[False, False, True]}@
cups = @\sethlcolor{aigreen}\hl{[True, False, False]}@
cups = @\sethlcolor{aigreen}\hl{[True, False, False]}@ @\label{line:gptout}@ 
\end{lstlisting}
\end{minipage}
\hfill
\begin{minipage}[t]{0.32\linewidth}
\begin{lstlisting}[caption={The prompt and \sethlcolor{aigreen}\hl{desired output} of LLM+State.},label={lst:state-maintaining-cup-ball-prompt},firstnumber=auto]
@
\textcolor{commentgray}{\# Initial state}
@
cups = [False, True, False] @\label{line:init-cup-state}@
Swapping cup 1 with cup 2
cups = @\sethlcolor{aigreen}\hl{[False, False, True]}@
Swapping cup 0 with cup 2
cups = @\sethlcolor{aigreen}\hl{[True, False, False]}@
Swapping cup 1 with cup 2
cups = @\sethlcolor{aigreen}\hl{[True, False, False]}@  
\end{lstlisting}
\end{minipage}
\end{minipage}
\vspace{-16pt}
\end{figure*}

%% file: papers/statler/sections/related_work.tex
\textbf{Language Understanding for Robotics } A common approach for language understanding for robotic agents involves symbol grounding~\cite{harnad90}, whereby phrases are mapped to symbols in the robot's world model. Early work~\cite{winograd71,macmahon06} relies upon hand-engineered rules to perform this mapping. More recent methods replace these rules with statistical models the parameters of which are trained on annotated corpora~\cite{kollar10, matuszek10,
chen11, tellex11, matuszek12a, thomason15, howard14,
misra16, thomason16, thomason18, shridhar18, paul18}. Other methods use neural network-based architectures to jointly reason over natural language utterances and the agent's (visual) observations of the scene~\cite{mei2016listen, Anderson2017VisionandLanguageNI, fried18, zhu20, Min2021FILMFI}.

\noindent\textbf{LLMs for Robotics } Since LLMs are trained with Internet-scale corpora, their infused common sense have shown to help in the domain of robotics in terms of high-level planning from natural language instructions~\cite{Ahn2022DoAI,Liang2022CodeAP,huang2022language} for both object manipulation~\cite{wang2023programmatically,ren2023leveraging} and navigation tasks~\cite{majumdar2020improving,gadre2023cows,shah2023lm,Huang2022VisualLM}.
Combining LLMs with expressive visual-language embeddings also enables impressive capabilities \cite{Shridhar2021CLIPortWA}. This has led to efforts to push for general multi-modality embodied models \cite{Zeng2022SocraticMC, driess2023palm}. %

\noindent\textbf{Code Generation with LLMs } Code generation has been one of the most successful use cases for LLMs~\cite{DBLP:journals/corr/abs-2107-03374, Hendrycks2021MeasuringCC, Li2022CompetitionlevelCG, Chen2022CodeTCG, gpt3, DBLP:journals/corr/abs-2303-08774}. 
Since code can connect with executable APIs for tasks including computation, vision and manipulation, a large chunk of work has focused on code generation with different tools \cite{Schick2023ToolformerLM, Suris2023ViperGPTVI, Patil2023GorillaLL}. In particular, Code-as-policies~\cite{Liang2022CodeAP} is one of the first to use code generation within a robotics context.

\noindent\textbf{State Representation in Reasoning } 
The use of state representations have been shown to help in algorithmic reasoning tasks \cite{Nye2021ShowYW, Nam2022LearningTR}. Instead of using one forward pass to predict the execution result for the entire code snippet,  \citet{Nye2021ShowYW} proposes to spell out step-by-step intermediate outputs to help infer the final execution results. 
Also relevant are research efforts that aim to enhance language modeling by rolling out possible future tokens~\cite{du2023autoregressive}. 

\noindent\textbf{Language Models and Planning } Recent work shows that vanilla and instruction-tuned LLMs plan poorly \cite{valmeekam2023planning, silver2023generalized, liu2023llm+}. Some works propose using the LLM as an intermediary between natural language and a domain-specific programming language, and then uses a traditional planner~\cite{guan2023leveraging, liu2023llm+, wong2023word}. \citet{silver2023generalized} employ Chain-of-Thought and iterative reprompting with feedback on generated plans, but require GPT-4 for good performance. \citet{xiang2023language} use parameter-efficient finetuning of LLMs on top of traces from a world-model and show improved performance on related tasks.

%% file: papers/statler/sections/motivating_example.tex
\label{sec:motivation}
We use \textit{three-cups-and-a-ball}, a simple shell game, to demonstrate the effectiveness of our state-maintaining idea. 
In this game, a ball is covered under one of three identical cups and the initial position of the ball is known to the player. 
In each of $K$ rounds, we randomly swap two cups' positions. Finally, we ask the player to guess the position of the ball. 

We present three separate cases of using LLMs to play this game using GPT-3 (precisely, text-davinci-003). \cref{lst:cup-ball-vanilla-prompt} demonstrates the simplest case: We represent the state with Boolean variables with \texttt{True} indicating ``ball is here''. We feed the initial state and the $K$ rounds of swaps into GPT-3, instructing it to complete the final state.
\cref{lst:cot-cup-ball-prompt} is an improved way: after reading $K$ rounds of swaps, GPT-3 is asked to give all the intermediate states over the game. 
This version is inspired by Chain-of-Thought prompting~\citep{DBLP:conf/nips/Wei0SBIXCLZ22}, which improves the performance of an LLM by requesting it to spell out its intermediate reasoning steps. Finally, 
\cref{lst:state-maintaining-cup-ball-prompt} is a simple instantiation of our state-maintaining idea: we ask GPT-3 to return the current state immediately after reading each round of swaps, stimulating the model to track and update the state as the game progresses. 

\begin{figure}[!t]
    \centering
    \includegraphics[width=0.8\linewidth]{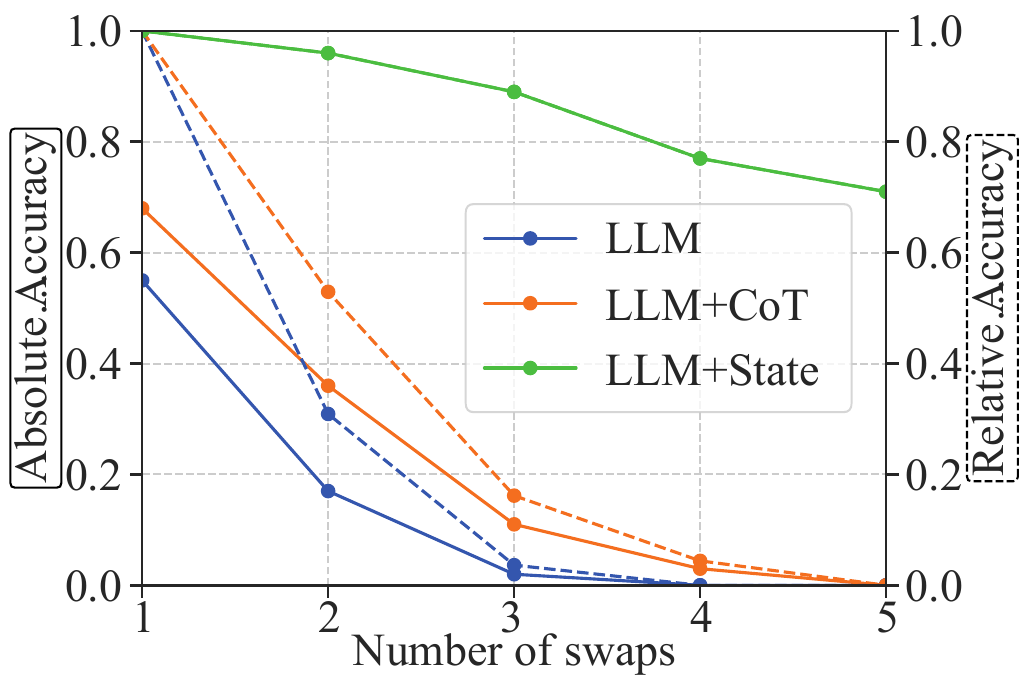}
    \caption{Model accuracies on the \emph{three-cups-and-a-ball} shell game. LLM+State is a simplified version of our proposed Statler framework. 
    For each method, the solid line shows how its accuracy $a(n)$ changes with the number of swaps $n$. 
    The dashed line is the \emph{relative} accuracy: $r(n) = a(n)/a(1)$. Intuitively, it measures how fast the performance decreases from a \emph{hypothetically perfect} one-swap performance. Note that LLM+State indeed achieves $a(1)=100\%$
    }\label{fig:shell-game-plot}
\end{figure}

We evaluate these methods with a range of $K$; for each $K$ and each method, 
we feed $30$ demonstrations with various numbers of swaps to the model, and repeat the experiment $100$ times. 
\cref{fig:shell-game-plot} visualizes the average accuracies. The state-maintaining method significantly outperforms the other methods, with the performance gap increasing with $K$.

%% file: papers/statler/sections/methodology.tex
\label{sec:method}
\begin{figure}[!t]
    \centering
    \begin{subfigure}{0.75\textwidth}
        \centering
        \begin{tikzpicture}[utterance/.style={rectangle, rounded corners, draw=black, font=\scriptsize, minimum width=5.75cm, minimum height=0.5cm, text width = 5.8cm, fill=black!10, outer sep=0pt},
            empty/.style={outer sep=1pt, inner sep=0pt}]
            \node [empty, label=below:{\footnotesize Initial State}] (before) at (0,0) {\includegraphics[width=0.32\linewidth]{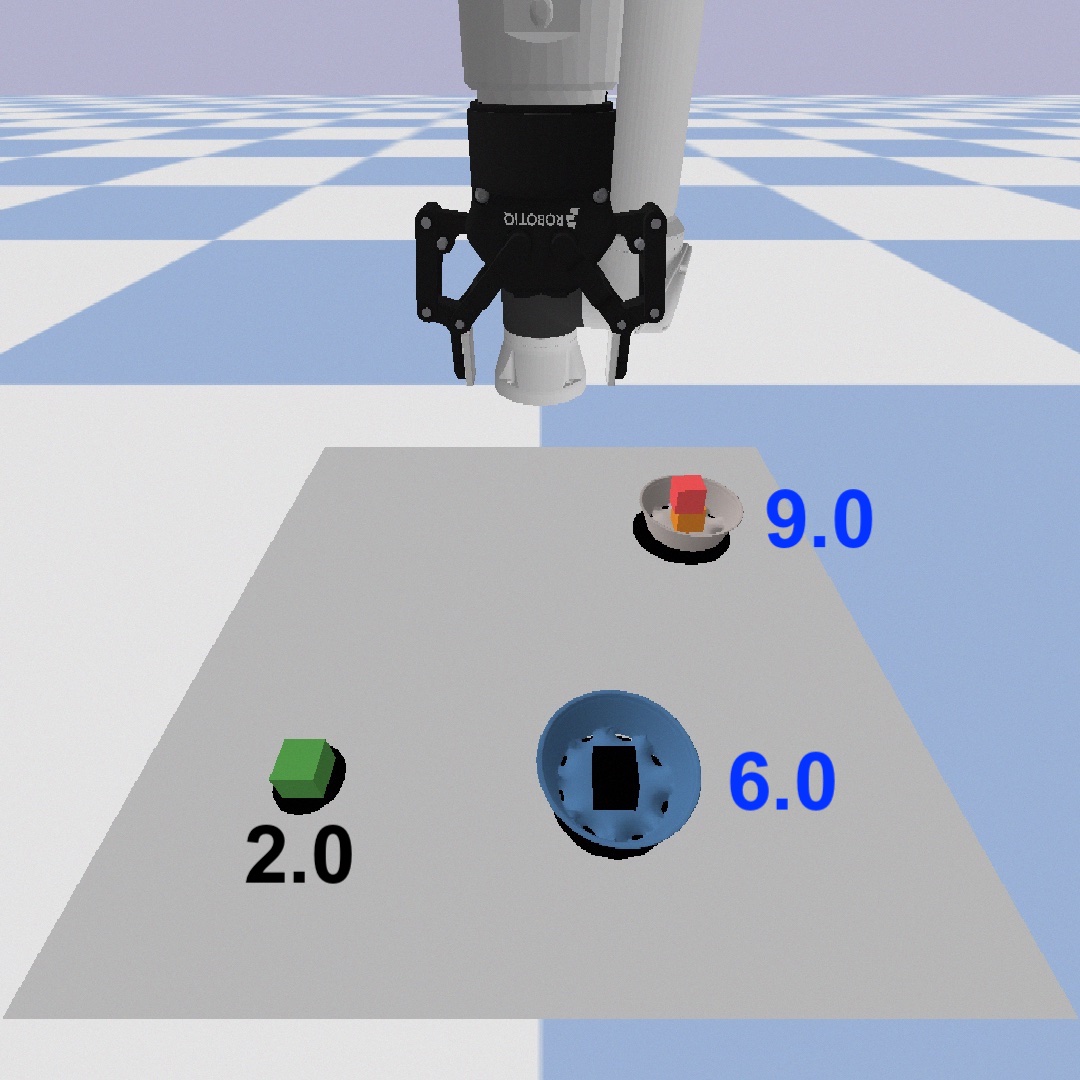}};
            \node [empty, right=0 of before, label=below:{\footnotesize Code-as-Policies\xspace}] (after-cap) {\includegraphics[width=0.32\linewidth]{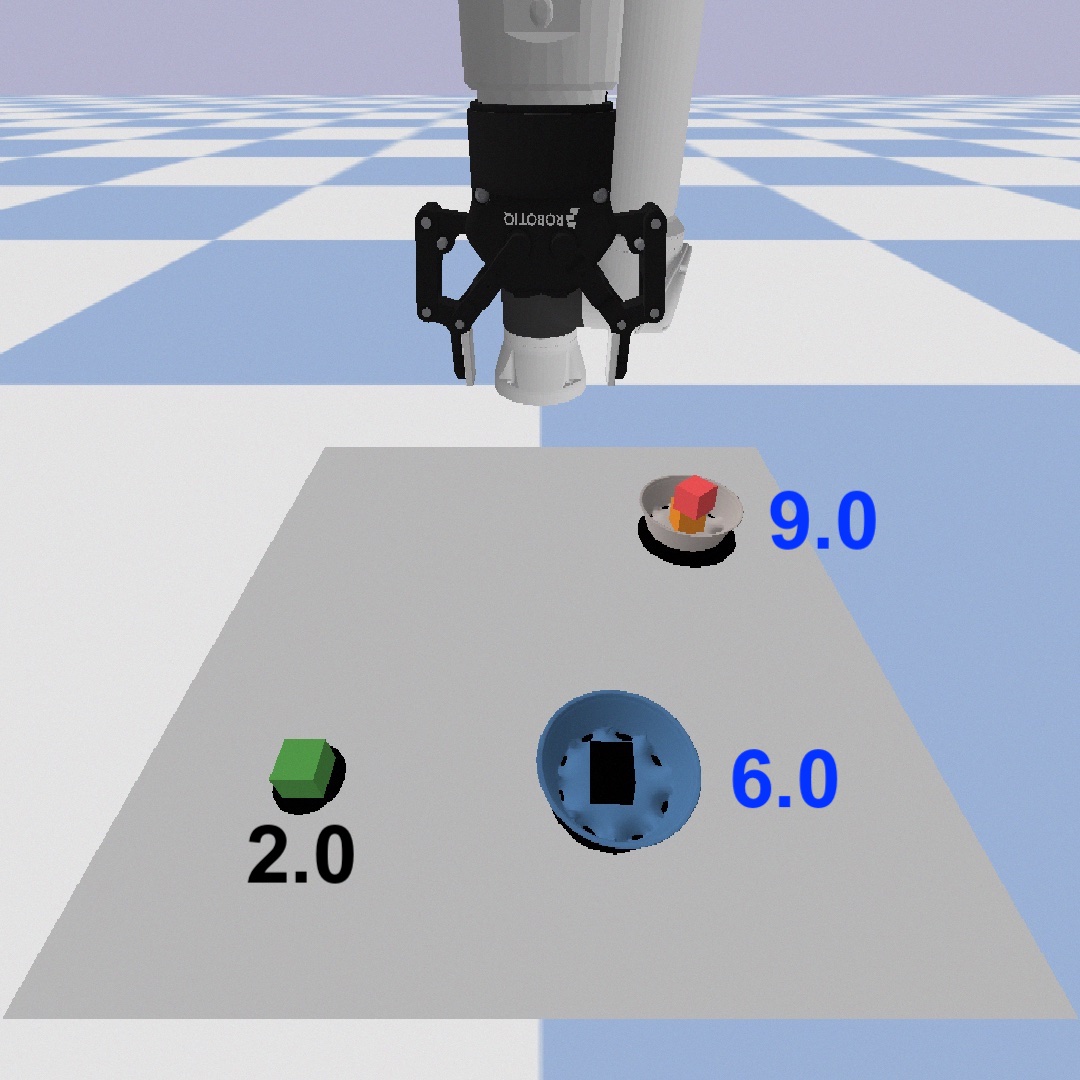}};
            \node [empty, right=0 of after-cap, label=below:{\footnotesize Statler\xspace (ours)}] (after-ours) {\includegraphics[width=0.32\linewidth]{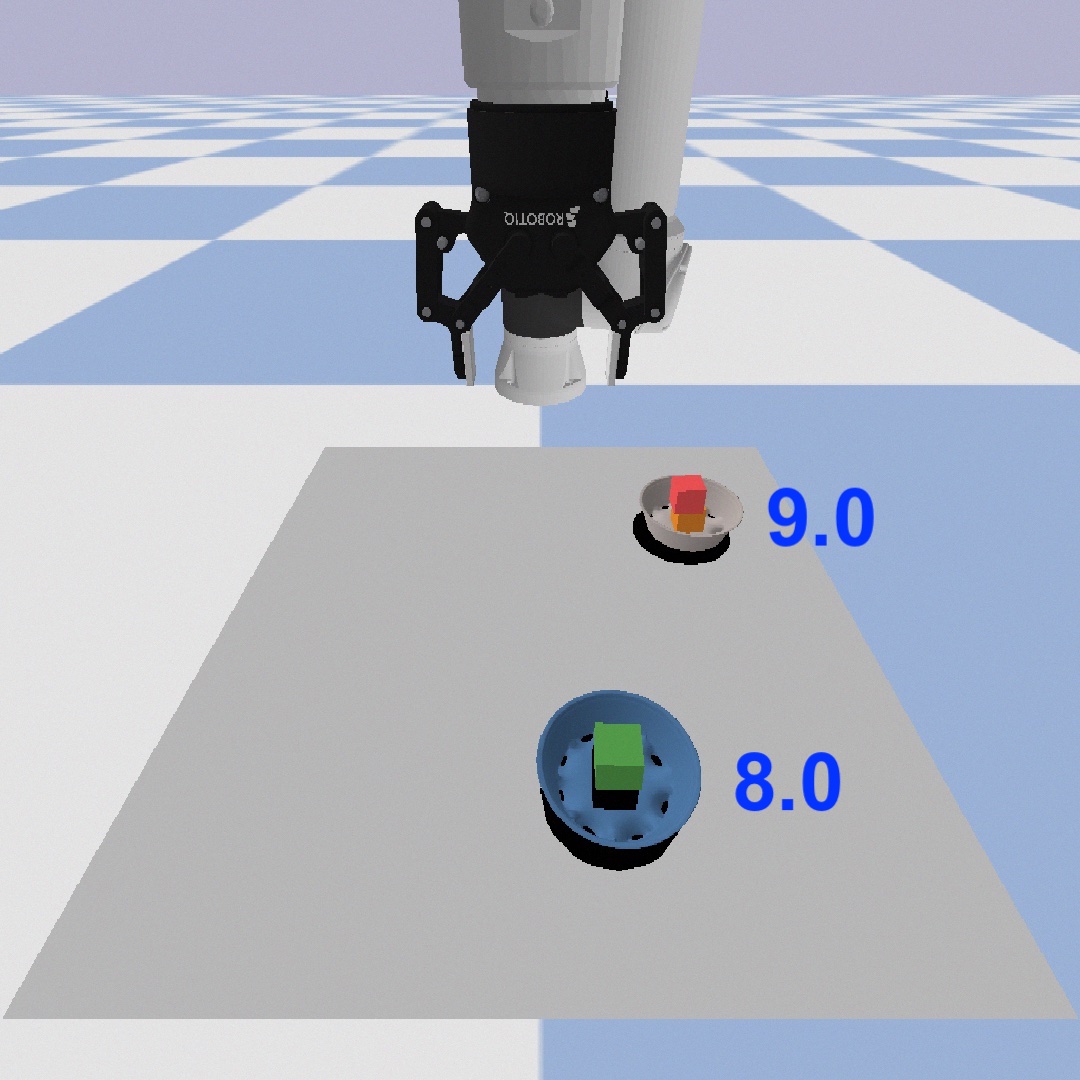}};
            \node[empty, above=0 and 0 of before.north west,anchor=south west] (speak) {\includegraphics[height=5mm]{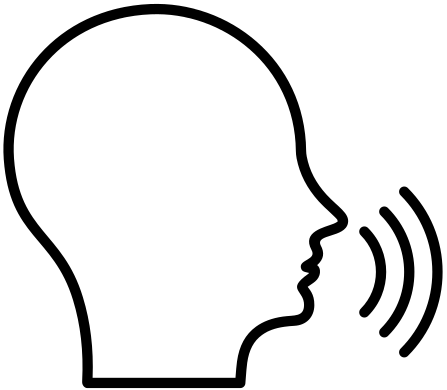}};
            \node[utterance, right= 1pt and 0 of speak.south east, anchor=south west, rounded corners] (utterance) {\textls[-28]{put a block in the blue bowl so that the total weight of blocks in the blue bowl is less than what is in the gray bowl}};
            \node [circle,draw=black, fill=white, inner sep=1pt, minimum size=5pt,  above right=5pt of after-ours.south west, anchor=south west] (check) {\greencheck};
            \node [circle,draw=black, fill=white, inner sep=1pt, minimum size=5pt, above right=5pt of after-cap.south west, anchor=south west] (check) {\redx};
        \end{tikzpicture}
    \end{subfigure}
    \\
    \begin{subfigure}{0.75\textwidth}
        \centering
        \begin{tikzpicture}[utterance/.style={rectangle, rounded corners, draw=black, font=\scriptsize, minimum width=5.75cm, minimum height=0.5cm, text width = 5.75cm, fill=black!10, outer sep=0pt},
            empty/.style={outer sep=1pt, inner sep=0pt}]
            \node [empty, label=below:{\footnotesize Initial State}] (before) at (0,0) {\includegraphics[width=0.32\linewidth]{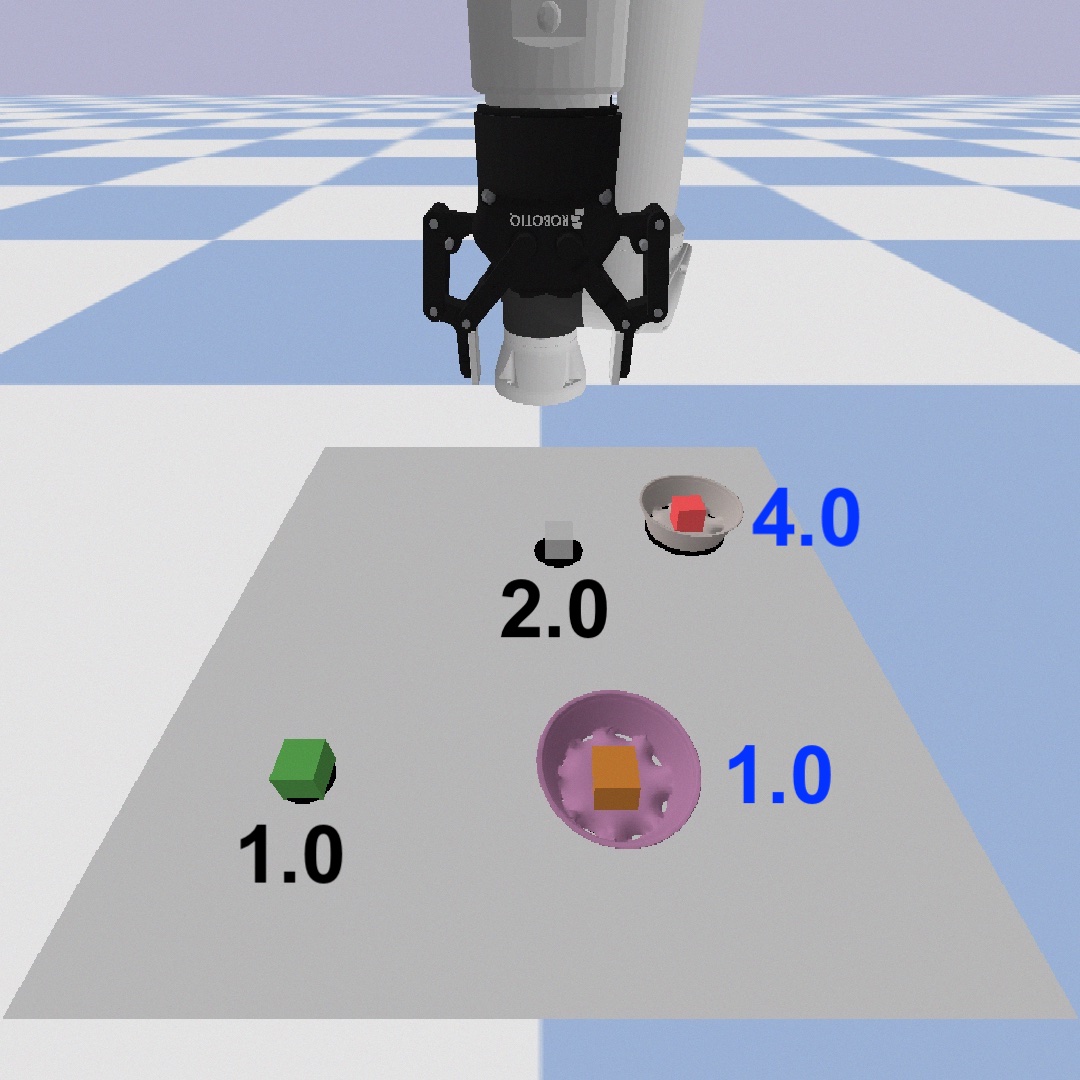}};
            \node [empty, right=0 of before, label=below:{\footnotesize Code-as-Policies\xspace}] (after-cap) {\includegraphics[width=0.32\linewidth]{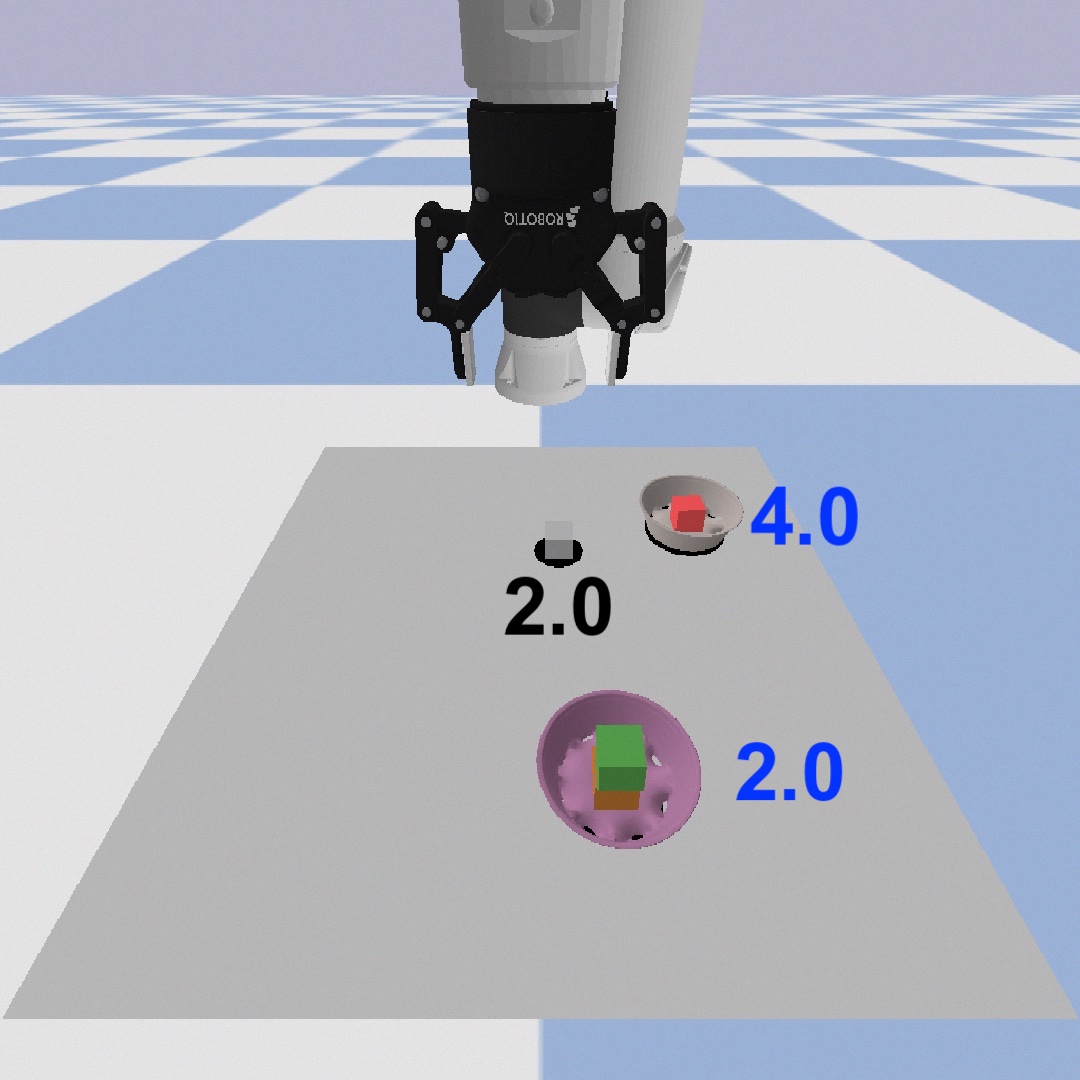}};
            \node [empty, right=0 of after-cap, label=below:{\footnotesize Statler\xspace (ours)}] (after-ours) {\includegraphics[width=0.32\linewidth]{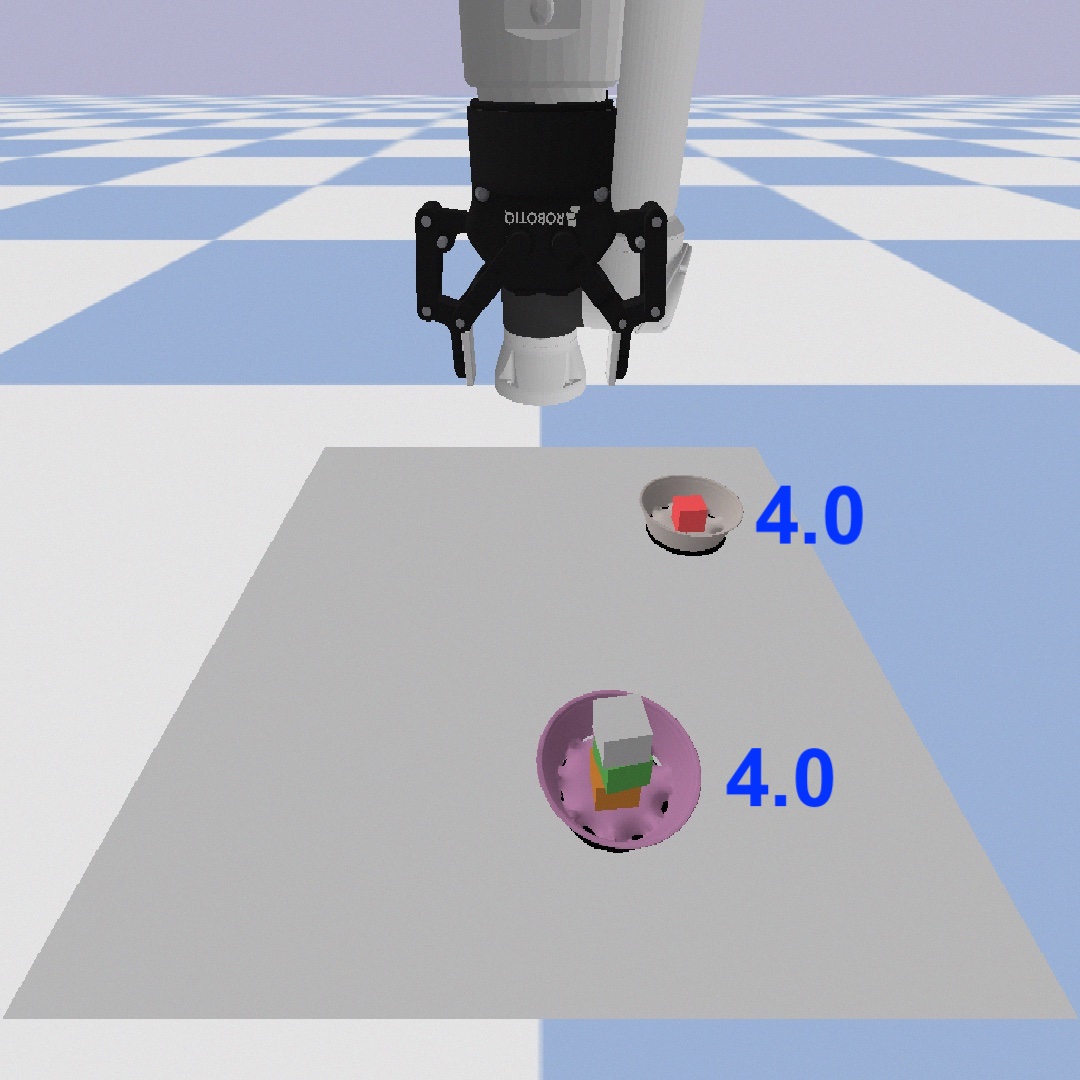}};
            \node[empty, above=0 and 0 of before.north west,anchor=south west] (speak) {\includegraphics[height=5mm]{papers/statler/figures/speak-icon}};
            \node[utterance, right= 1pt and 0 of speak.south east, anchor=south west, rounded corners] (utterance) {put blocks in the purple bowl so that their total weight becomes identical to what is in the gray bowl};
            \node [circle,draw=black, fill=white, inner sep=1pt, minimum size=5pt,  above right=5pt of after-ours.south west, anchor=south west] (check) {\greencheck};
            \node [circle,draw=black, fill=white, inner sep=1pt, minimum size=5pt, above right=5pt of after-cap.south west, anchor=south west] (check) {\redx};
        \end{tikzpicture}
    \end{subfigure} \hfil
    \begin{subfigure}{0.75\textwidth}
        \centering
        \begin{tikzpicture}[utterance/.style={rectangle, rounded corners, draw=black, font=\scriptsize, minimum width=5.75cm, minimum height=0.5cm, text width = 5.75cm, fill=black!10, outer sep=0pt},
            empty/.style={outer sep=1pt, inner sep=0pt}]
            \node [empty, label=below:{\footnotesize Initial State}] (before) at (0,0) {\includegraphics[width=0.32\linewidth]{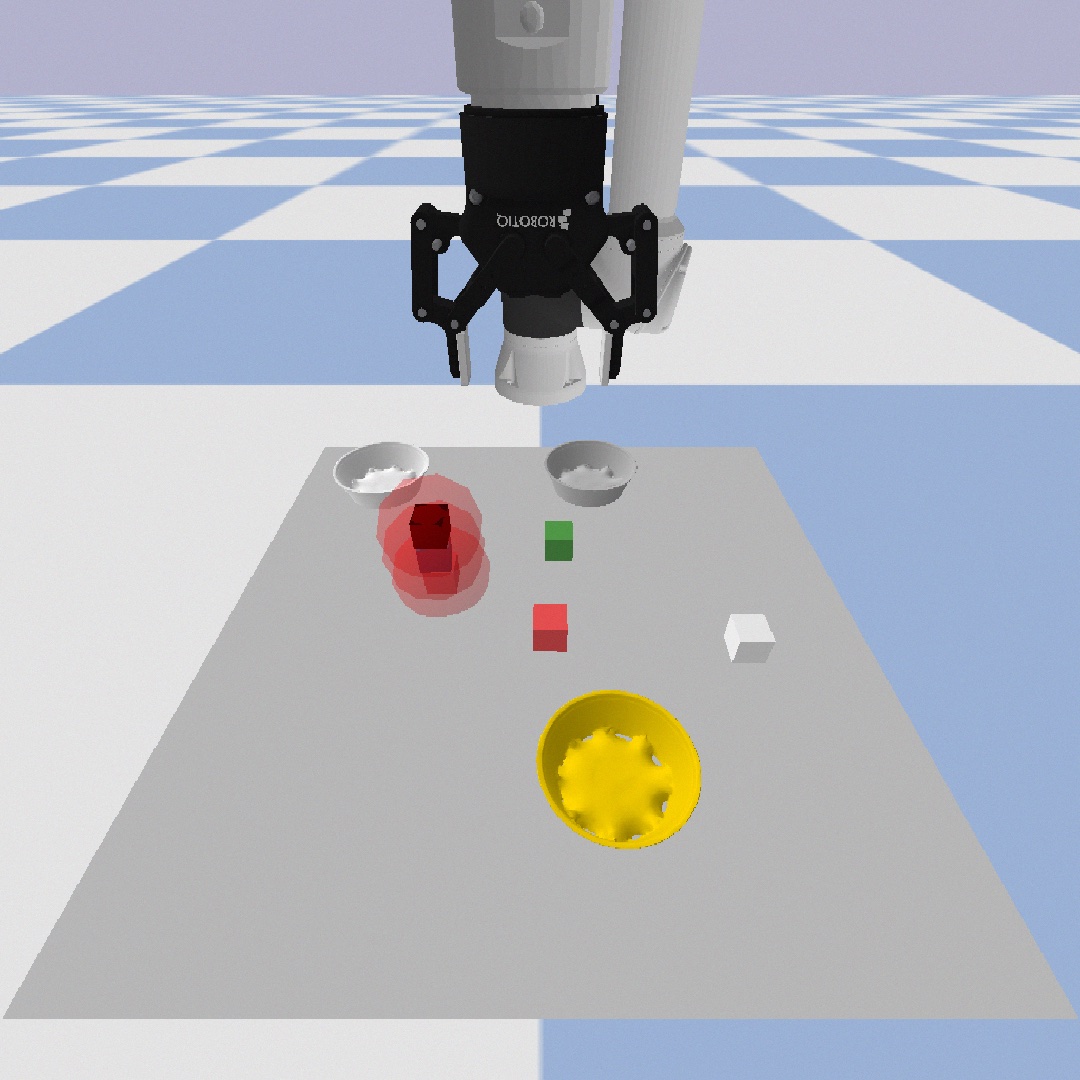}};
            \node [empty, right=0 of before, label=below:{\footnotesize Code-as-Policies\xspace}] (after-cap) {\includegraphics[width=0.32\linewidth]{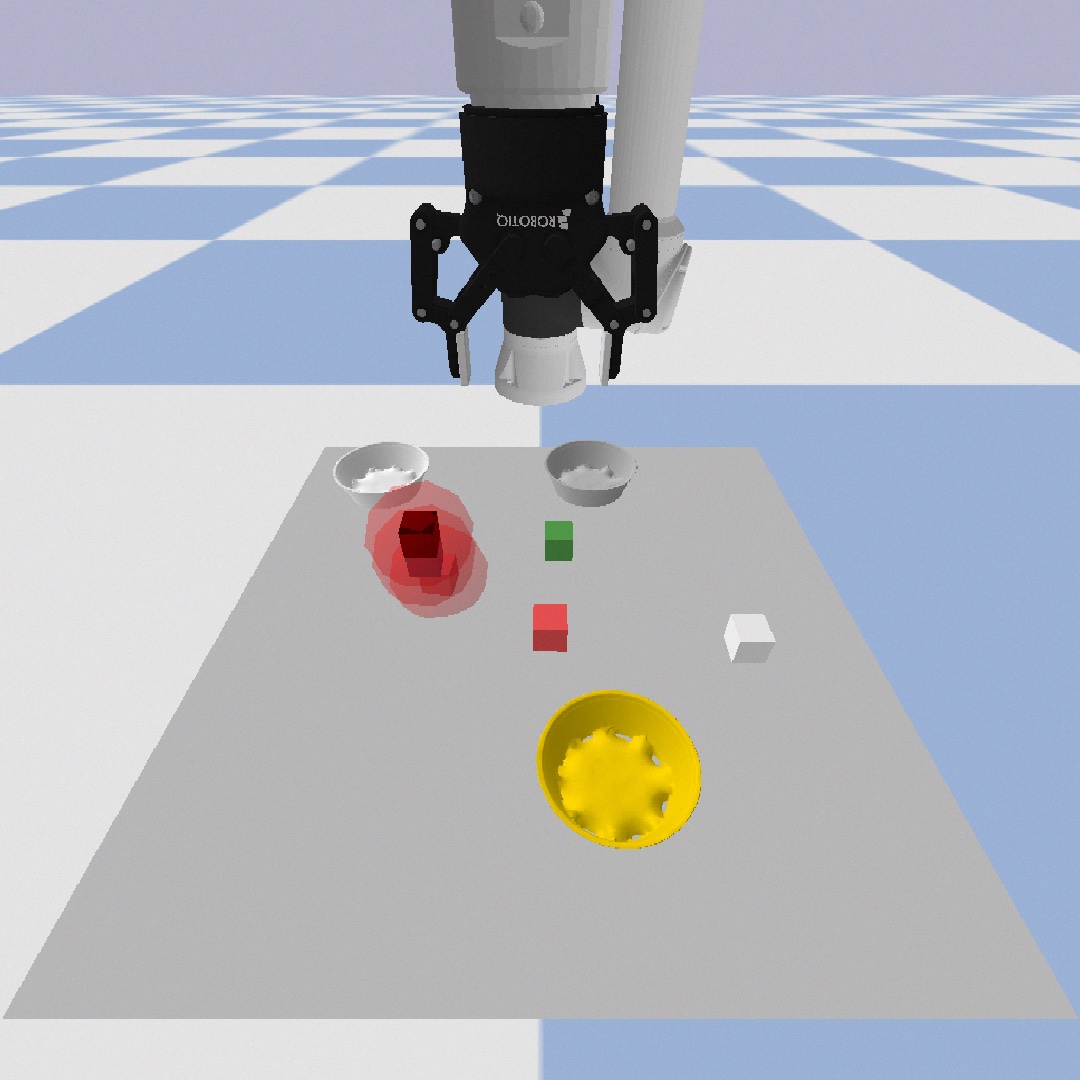}};
            \node [empty, right=0 of after-cap, label=below:{\footnotesize Statler\xspace (ours)}] (after-ours) {\includegraphics[width=0.32\linewidth]{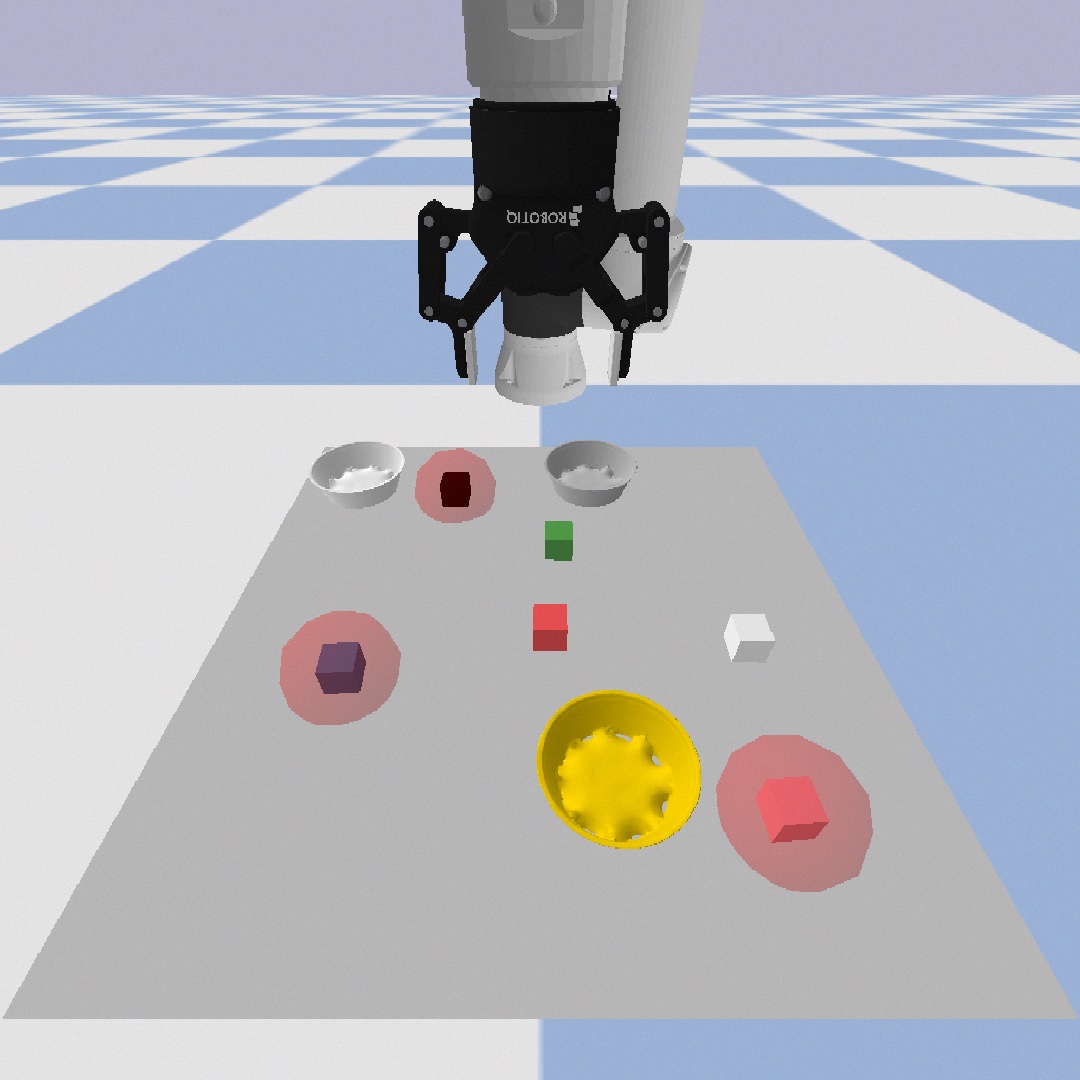}};
            \node[empty, above=0 and 0 of before.north west,anchor=south west] (speak) {\includegraphics[height=5mm]{papers/statler/figures/speak-icon}};
            \node[utterance, right= 1pt and 0 of speak.south east, anchor=south west, rounded corners] (utterance) {put all the dirty blocks on the table};
            \node [circle,draw=black, fill=white, inner sep=1pt, minimum size=5pt,  above right=5pt of after-ours.south west, anchor=south west] (check) {\greencheck};
            \node [circle,draw=black, fill=white, inner sep=1pt, minimum size=5pt, above right=5pt of after-cap.south west, anchor=south west] (check) {\redx};
        \end{tikzpicture}
    \end{subfigure}
    \caption{Examples of simulations that show the result of executing different natural language instructions using Code-as-Policies\xspace and our state-maintaining Statler\xspace algorithm.}\label{fig:example-instructions-simulated}
\end{figure}
\begin{codefloat}
\begin{minipage}[t]{\linewidth}
\begin{minipage}[t]{\linewidth}
\begin{lstlisting}[caption={World-state reader. The text \sethlcolor{aigreen}\hl{highlighted in green} represents the part that the model is expected to generate.},label={lst:wm-reader},firstnumber=auto]
# @\label{line:init-metal-state}@state = {
#     "objects": ["cyan block", "yellow block", "brown block", "purple block", "blue block", "green bowl", "red bowl", "disinfector"],
#     "relations": [],
#     "disinfector": {"contains": []},
#     "cyan block": {"is": ["dirty"]},
#     "yellow block": {"is": ["clean"]},
#     "brown block": {"is": ["clean"]},
#     "purple block": {"is": ["dirty"]},
#     "blue block": {"is": ["clean"]},
#     "green bowl": {},
#     "red bowl": {}
# }@\label{line:end-metal-state}@
@
\label{line:metal-query}\textcolor{commentgray}{\# query: Put the cyan block on the yellow block}
@
@\sethlcolor{aigreen}\hl{put\_first\_on\_second("cyan block", "yellow block")}@
@\sethlcolor{aigreen}\hl{update\_wm("Put the cyan block on the yellow block")}@
\end{lstlisting}
\end{minipage}%
\hfill
\begin{minipage}[t]{\linewidth}
\begin{lstlisting}[caption={World-state writer. The text \textcolor{blue}{rendered in blue} highlights the updated part of the state.},label={lst:wm-writer},firstnumber=auto]
# state = {
#     "objects": ["cyan block", "yellow block", "brown block", "purple block", "blue block", "green bowl", "red bowl", "disinfector"],
#     "relations": [],
#     "disinfector": {"contains": []},
#     "cyan block": {"is": ["dirty"]},
#     "yellow block": {"is": ["clean"]},
#     "brown block": {"is": ["clean"]},
#     "purple block": {"is": ["dirty"]},
#     "blue block": {"is": ["clean"]},
#     "green bowl": {},
#     "red bowl": {}
# }
@\textcolor{commentgray}{\# query: Put the cyan block on the yellow block.}@
@\sethlcolor{aigreen}\hl{\#~state = \{}@
@\sethlcolor{aigreen}\hl{\#~~~~~"objects": ["cyan block", "yellow block", "brown block", "purple block", "blue block", "green bowl", "red bowl", "disinfector"],}@
@\textcolor{blue}{\sethlcolor{aigreen}{\sethlcolor{aigreen}\hl{\#~~~~~"relations": [["cyan block is on yellow block"]],}}}@
@\sethlcolor{aigreen}\hl{\#~~~~~"disinfector": \{"contains": []\},}@
@\sethlcolor{aigreen}\hl{\#~~~~~"cyan block": \{"is": ["dirty"]\},}@
@\textcolor{blue}{\sethlcolor{aigreen}\hl{\#~~~~~"yellow block": \{"is": ["dirty"]\},}}@
@\sethlcolor{aigreen}\hl{\#~~~~~"brown block": \{"is": ["clean"]\},}@
@\sethlcolor{aigreen}\hl{\#~~~~~"purple block": \{"is": ["dirty"]\},}@
@\sethlcolor{aigreen}\hl{\#~~~~~"blue block": \{"is": ["clean"]\},}@
@\sethlcolor{aigreen}\hl{\#~~~~~"green bowl": \{\},}@
@\sethlcolor{aigreen}\hl{\#~~~~~"red bowl": \{\},}@
@\sethlcolor{aigreen}\hl{\# \}}@
\end{lstlisting}
\end{minipage}
\hfill
\end{minipage}
\end{codefloat}
As exemplified in \cref{sec:motivation}, the key to our approach is to allow the LLM to describe the next state while responding to each user query. The motivating example is simple in that the next state \textit{is} the response. Instead, we now consider more general scenarios where there is a significant burden on the LLM to track the state updates as well as generate responses. (Fig.~\ref{fig:example-instructions-simulated}). %
For the general cases,
we propose to \emph{split} the burden across multiple different prompted LLMs. 
Precisely, we maintain a separate prompt that includes instructions and demonstrations for each subtask (i.e., state-tracking or query-responding) and then use the prompt to elicit an LLM to perform the particular subtask. 
As we discuss shortly, our framework includes a \textbf{world-state reader} that responds to the user query and a \textbf{world-state writer} that is responsible for updating the state representation.
Our framework (Fig.~\ref{fig:diagram}) does not pose any fundamental limitation on which domain it can be applied to.
Our approach can be regarded as a model-based extension of Code-as-Policies (CaP) in the sense that it keeps the core capabilities of CaP (e.g., hierarchical code generation) and incorporates a means to explicitly maintain an estimated world state.

It is useful to consider example prompts to understand the operation of the reader and writer models.
Prompt~\ref{lst:wm-reader} is an example of the input passed to the world-state reader. 
Initially, we initialize a JSON-formatted state with a reference to object-oriented principles.
Given a user query ``Put the cyan block on the yellow block'' (Line \ref{line:metal-query}) and the current state representation (Lines \ref{line:init-metal-state}--\ref{line:end-metal-state}),
the world-state reader should generate the code that responds to the query, taking into account the current state. The expected code to be generated is \sethlcolor{aigreen}\hl{highlighted in green}.
After generating the code, our model executes it to complete the query. When the state needs to be updated, the generated code will contain an \texttt{update\_wm} function that triggers the world-state writer with the query specified in its argument.
In Prompt~\ref{lst:wm-writer}, we show the corresponding example for the world-state writer. Similar to the reader, we prepend the current state representation before the user query and the model generates the updated state representation (\hl{highlighted in green}).

%% file: papers/statler/sections/experiments.tex
\label{sec:experiments}

\begin{figure}[!t]
    \centering
    \begin{subfigure}{0.32\textwidth}
        \centering
        \includegraphics[width=\linewidth]{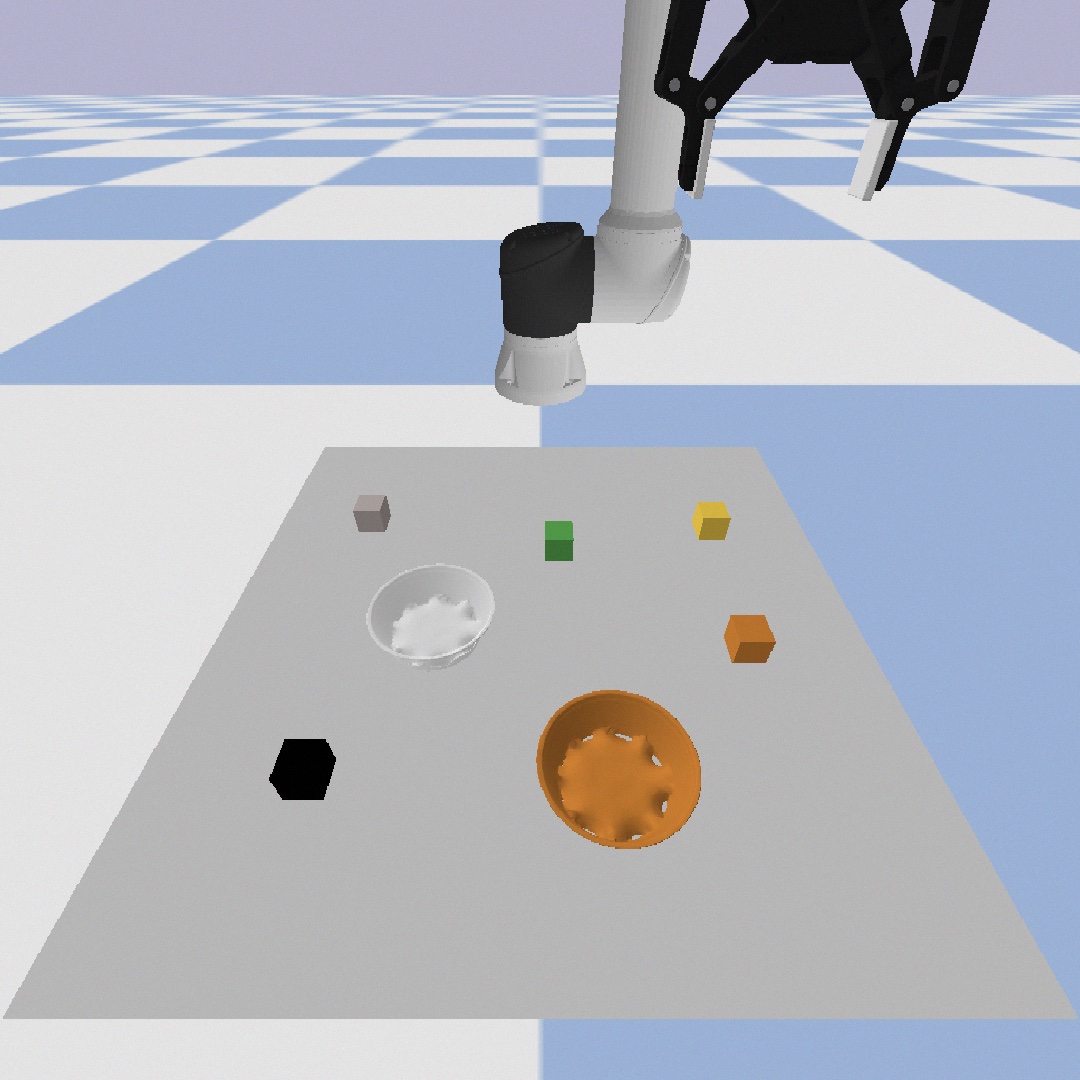}
        \caption{Pick-and-place}\label{fig:pick-and-place}
    \end{subfigure} \hfil
    \begin{subfigure}{0.32\textwidth}
        \centering
        \includegraphics[width=\linewidth]{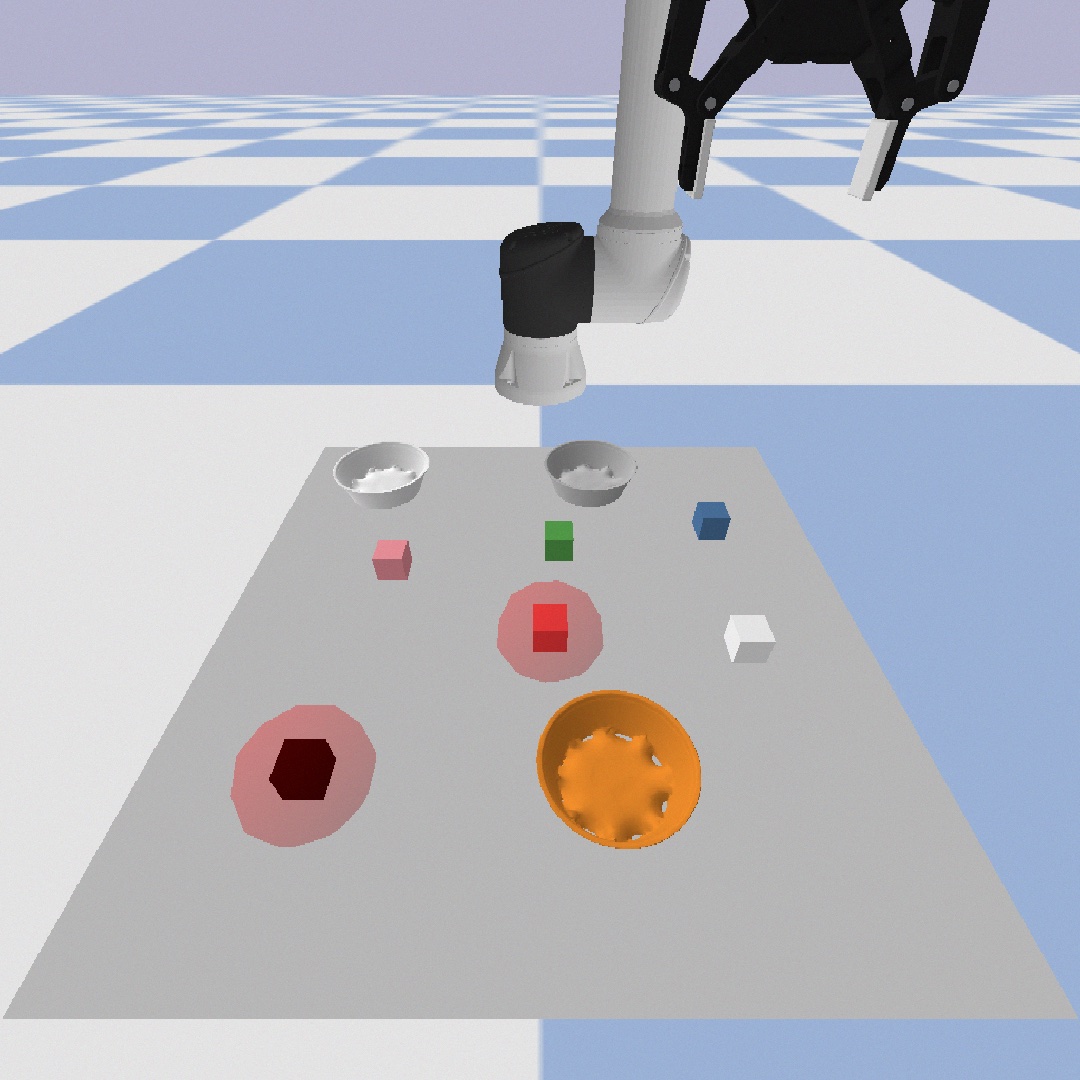}
        \caption{Disinfection}\label{fig:disinfection}
    \end{subfigure} \hfil
    \begin{subfigure}{0.32\textwidth}
        \centering
        \includegraphics[width=\linewidth]{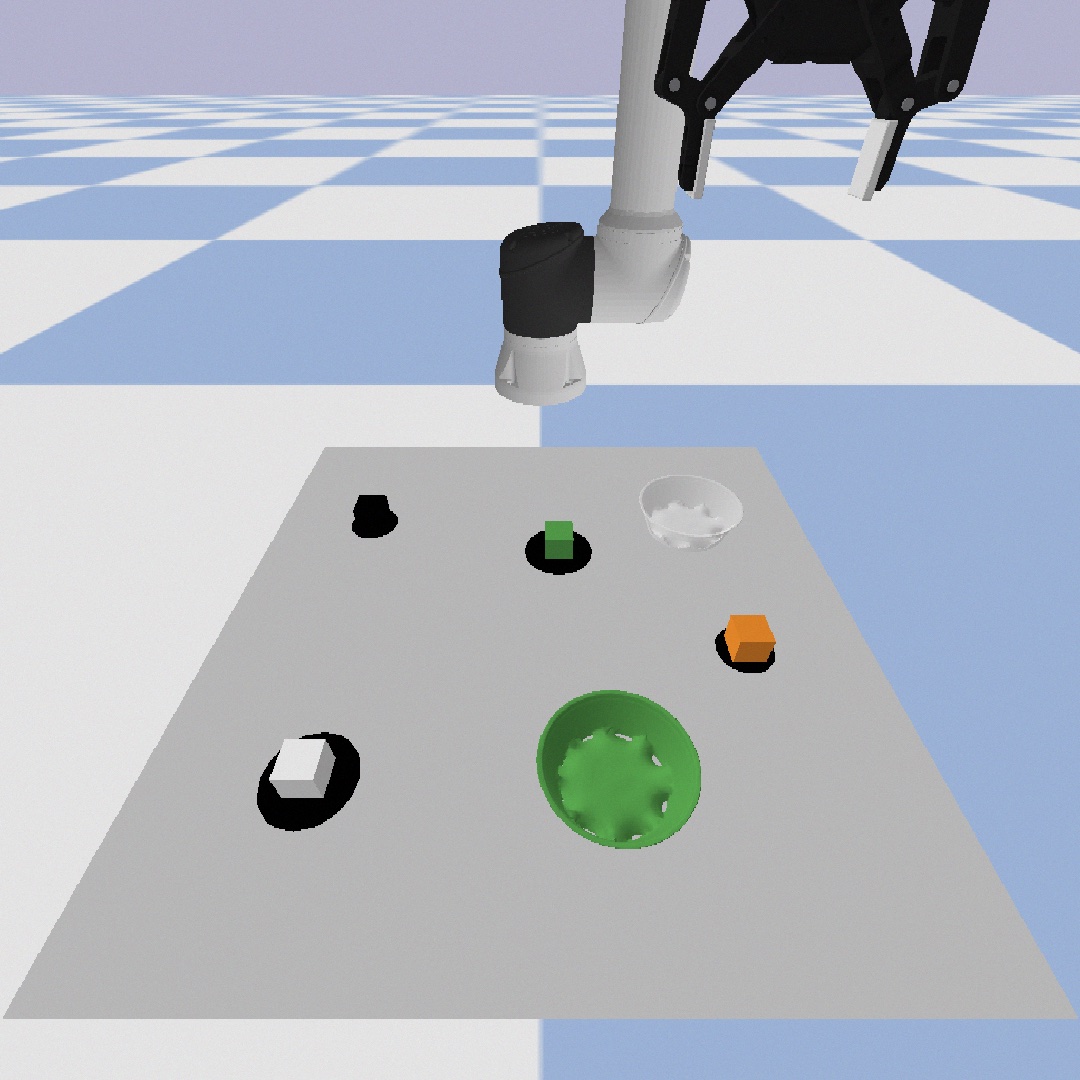}
        \caption{Weight reasoning}\label{fig:weight-reasoning}
    \end{subfigure}
    \caption{The simulated domains we consider include \subref{fig:pick-and-place} Pick-and-Place; \subref{fig:disinfection} Block Disinfection, where the translucent sphere around a block represents its dirtiness (this is not visible to the robot); and \subref{fig:weight-reasoning} Relative Weight Reasoning, where the radius of the disk under each block indicates its weight (this is not visible to the robot).}\label{fig:bp-assisted}
\end{figure}

We evaluate the capabilities of Statler\xspace alongside state-of-the-art LLM models on three tabletop manipulation domains (Fig.~\ref{fig:bp-assisted}): pick-and-place, block disinfection, and relative weight reasoning. For each domain, we design in-context examples and consider $20$ evaluation episodes each of which consists of $5$--$16$ consecutive steps of user queries. Every episode contains at least one query that requires reasoning over the interaction history (i.e., requires ``memory'' across steps), which makes the task significantly  challenging.

\subsection{Simulated Tabletop Manipulation Domains}
The \textbf{Pick-and-Place} domain involves scenarios that require a robot arm to sequentially pick up and place a block onto another block, bowl, or the table. The model needs to remember and reason over the block locations. The example user queries are ``Put the green block in the red bowl.'', ``What is the color of the block under the pink block?'', and ``How many blocks are in the green bowl?''.

In the \textbf{Block Disinfection} domain, we consider a scenario in which a block can be either \textit{dirty} or \textit{clean}, the state of which is not observable by the robot. When a clean block touches a dirty block (e.g., as a result of stacking one block on another), the clean block becomes dirty. There is a \textit{disinfector} on the table that cleans any block placed inside it. 
This scenario emulates a clean-up task in which you might ask a robot to put dirty dishes in a dishwasher or dirty clothes in a washing machine. The user query contains pick-and-place commands similar to those in the pick-and-place domain as well as textual utterances that require reasoning over which blocks are clean and dirty, such as ``Put all the clean blocks in the green bowl.'' This domain presents a particular challenge as the model must track the cleanliness of each block and accurately capture the state mutations that happen when a dirty block comes into contact with another clean block.

\textbf{Relative Weight Reasoning} involves memorizing and reasoning over the relative weights of the blocks. User queries provide information about the weight of blocks (e.g., ``The red block is twice the weight of the bronze block.''), which are followed by queries that require reasoning over the weights (e.g., ``Put blocks in the purple bowl so that their total weight becomes identical to what is in the gray bowl.'').

We compare our proposed approach, Statler\xspace, to two strong competing methods: Code-as-Policies~\cite{Liang2022CodeAP} (CaP) and CaP with Chain-of-Thought prompting~\cite{DBLP:conf/nips/Wei0SBIXCLZ22} (CaP+CoT).
CaP generates code for the current question at each step based on the past actions, but it does not maintain a state.
Following the CoT framework, \textit{at every step}, CaP+CoT deduces the intermediate states based on an \textit{initial state} and past actions, which are considered as its thoughts, to generate the current code. 
But it leads to redundant reasoning and increases the length of the prompt, which may then exceed the LLM's context window size limitations.
Furthermore, longer reasoning also demands longer, more intricate demo example prompts, contributing to increased developer effort.
We ensure that the demonstrations (i.e., in-context examples) given to each of the models are equivalent. Namely, we use the same sequence of user queries and code snippets, except for necessary differences due to their designs such as state representation.

\begin{figure}[!t]
    \centering
        \begin{tikzpicture}[utterance/.style={rectangle, rounded corners, draw=black, font=\scriptsize, minimum width=5.75cm, minimum height=0.5cm, text width = 5.75cm, fill=black!10, outer sep=0pt},
        response/.style={rectangle, draw=black, font=\scriptsize, text width=4.05cm, outer sep=0pt}, empty/.style={outer sep=0pt, inner sep=0pt}]
            \node [empty] (before) at (0,0) {\includegraphics[width=0.24\linewidth]{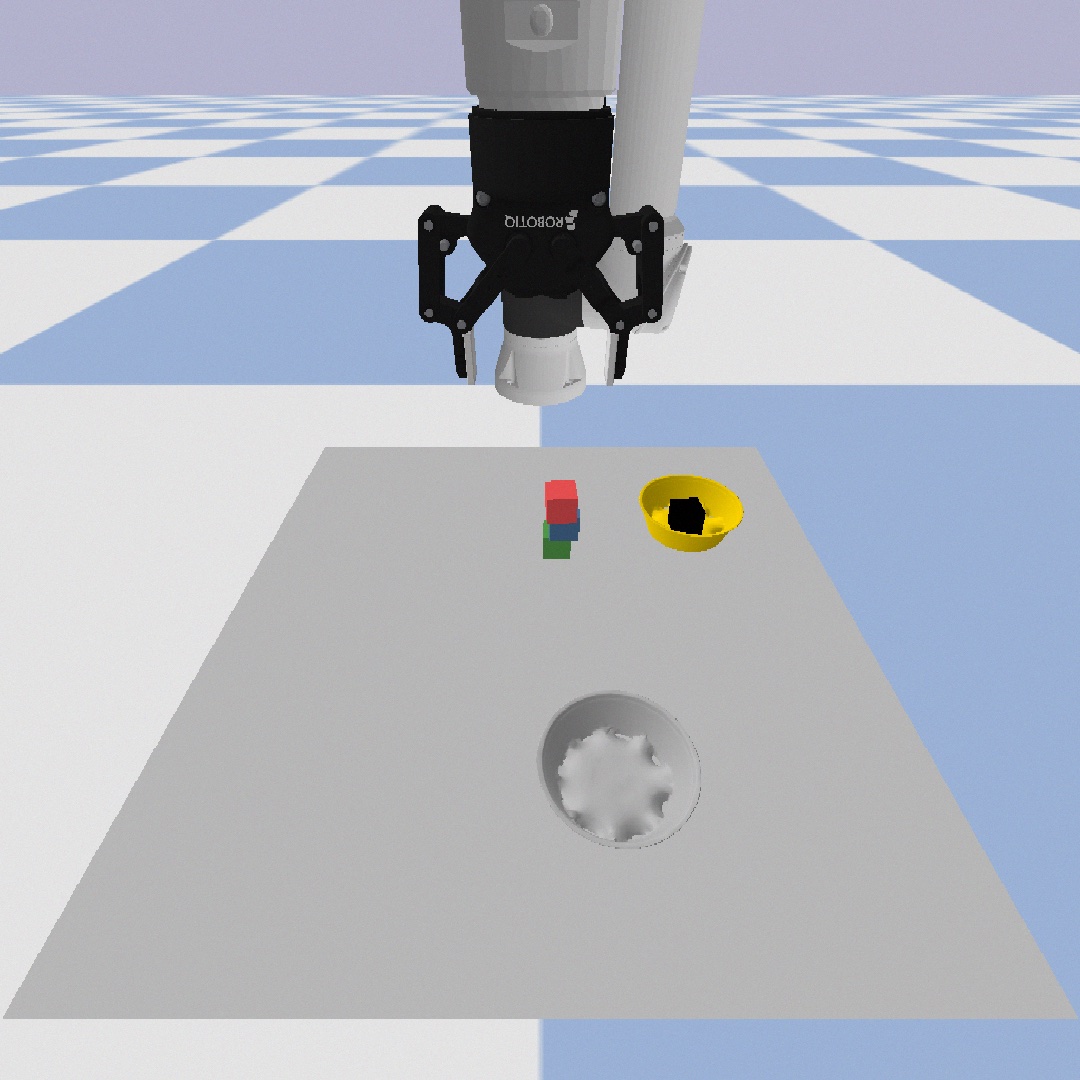}};
            \node[empty, above=0 and 0 of before.north west,anchor=south west] (speak) {\includegraphics[height=5mm]{papers/statler/figures/speak-icon}};
            \node[utterance, right= 1pt and 0 of speak.south east, anchor=south west, rounded corners] (utterance) {\textls[-20]{What is the color of the block right above the blue block?}};
            \path   let \p1 = ($(before.north east)-(utterance.north east)$),
                        \p2 = ($(before.north east)-(before.east)$),
                        \n1 = {veclen(\x1,\y1)},
                        \n2 = {veclen(\x2,\y2)}, in
                    node[response, minimum width=\n1, minimum height=\n2, right=1pt of before.north east,anchor=north west] (input) {\scriptsize \textbf{Code-as-Policies\xspace}: \textit{fails to generate anything}};
            \path   let \p1 = ($(before.north east)-(utterance.north east)$),
                        \p2 = ($(before.north east)-(before.east)$),
                        \n1 = {veclen(\x1,\y1)},
                        \n2 = {veclen(\x2,\y2)}, in
                    node[response, minimum width=\n1, minimum height=\n2, right=1pt of before.east,anchor=north west] (input) {\scriptsize \textbf{Statler\xspace (ours)}: ``red''};
        \end{tikzpicture}
        \begin{tikzpicture}[utterance/.style={rectangle, rounded corners, draw=black, font=\scriptsize, minimum width=5.75cm, minimum height=0.5cm, text width = 5.75cm, fill=black!10, outer sep=0pt},
            response/.style={rectangle, draw=black, font=\scriptsize, outer sep=0pt}, empty/.style={outer sep=0pt, inner sep=0pt}]
                \node [empty] (before) at (0,0) {\includegraphics[width=0.24\linewidth]{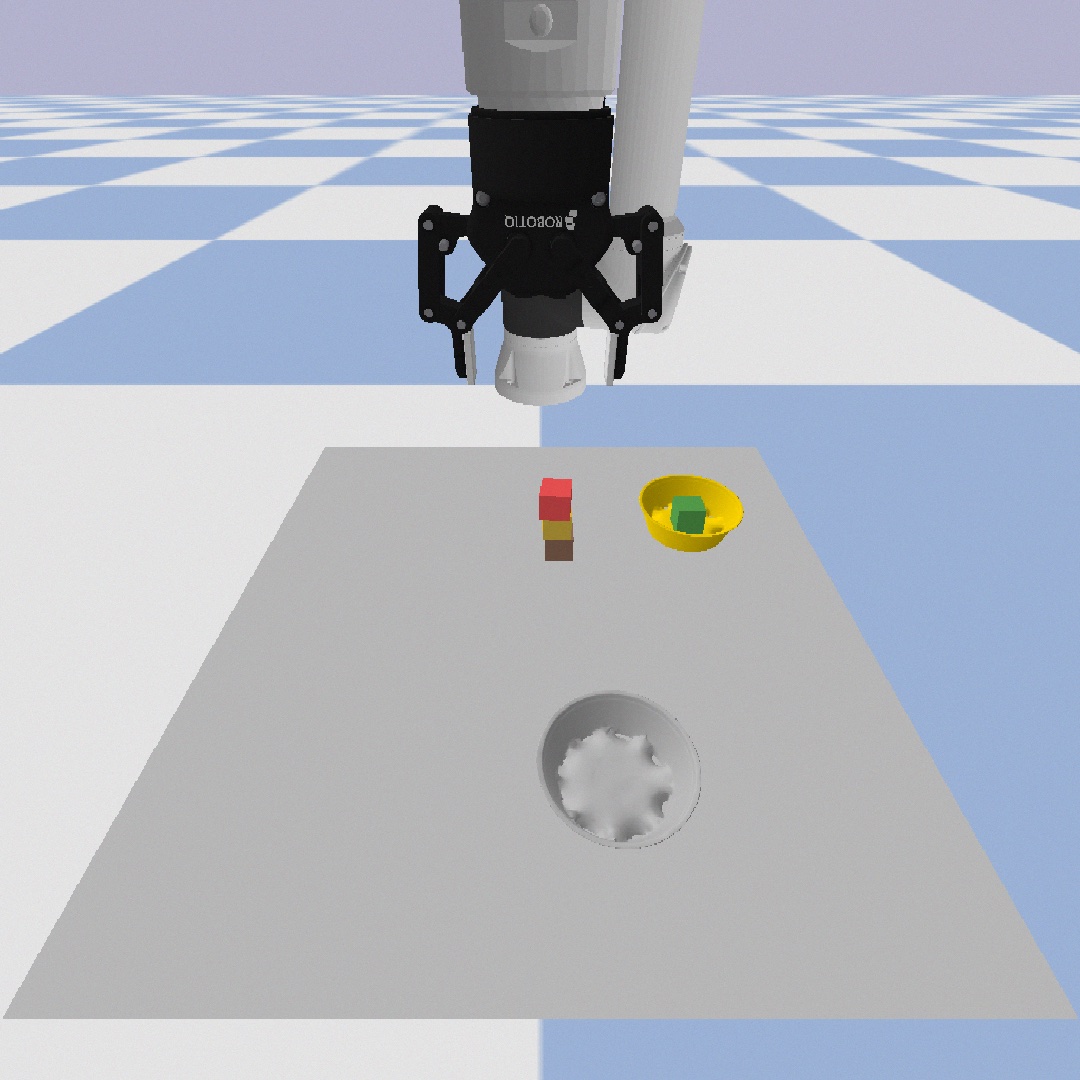}};
                \node[empty, above=0 and 0 of before.north west,anchor=south west] (speak) {\includegraphics[height=5mm]{papers/statler/figures/speak-icon}};
                \node[utterance, right= 1pt and 0 of speak.south east, anchor=south west, rounded corners] (utterance) {How many blocks are not in the bowls?};
                \path   let \p1 = ($(before.north east)-(utterance.north east)$),
                            \p2 = ($(before.north east)-(before.east)$),
                            \n1 = {veclen(\x1,\y1)},
                            \n2 = {veclen(\x2,\y2)}, in
                        node[response, minimum width=\n1, text width=4.0cm, minimum height=\n2, right=1pt of before.north east,anchor=north west] (input) {\textbf{Code-as-Policies\xspace}: ``There are two blocks not in the bowls: brown block and yellow block.''};
                \path   let \p1 = ($(before.north east)-(utterance.north east)$),
                            \p2 = ($(before.north east)-(before.east)$),
                            \n1 = {veclen(\x1,\y1)},
                            \n2 = {veclen(\x2,\y2)}, in
                        node[response, minimum width=\n1, text width=4.0cm, minimum height=\n2, right=1pt of before.east,anchor=north west] (input) {\scriptsize \textbf{Statler\xspace (ours)}: ``three blocks''};
            \end{tikzpicture}
    \caption{Examples that show the result of querying LLMs with and without maintained state. In the first scenario, CaP fails to produce an answer, while our Statler\xspace model produces the correct response. In the second example, one block is not visible and CaP incorrectly identifies two blocks as not being a bowl. By maintaining a persistent world state, our method is aware of the third block and correctly answers the query.}\label{fig:example-queries}
\end{figure}

\begin{table}[!th]
    \centering
    \caption{Episode success rates and individual step success rates (in parentheses) for each sequential task. $^\dagger$indicates that the context limit was often exceeded.} \label{tb:main-result}
    \setlength{\tabcolsep}{3pt}
    \begin{tabularx}{1.0\linewidth}{lYYY}%
        \toprule
         & {Pick \& Place} & {Disinfection} & {Weight}\\
        \midrule
        CaP & 
        \nfrac{0.00}{(0.54)} &%
        \nfrac{0.00}{(0.68)}\hphantom{${^\dagger}$} & %
        \nfrac{0.00}{(0.84)}\\
       CaP+CoT &
        \nfrac{0.25}{(0.76)} & %
        \nfrac{0.00}{(0.20)}${^\dagger}$ & %
        \nfrac{0.30}{(0.88)}\\
       Statler\xspace (ours) &
        \nfracb{0.50}{(0.88)} &%
        \nfracb{0.40}{(0.82)}\hphantom{${^\dagger}$} &%
        \nfracb{0.55}{(0.93)} \\
       \bottomrule
    \end{tabularx}
\end{table}
Table~\ref{tb:main-result} reports the episode success rates of each method along with the the success rate for individual steps. An episode is considered to be a failure if a model fails to respond to one of the user queries in the episode. While the CaP baseline correctly processes more than half of the individual steps in each domain, it fails to successfully complete any of the episodes. As we show later, CaP correctly processes most queries that do not require reasoning over previous steps (e.g.,``Put the red block on the blue block.''), but tends to generate incorrect (or no) code in response to queries that require reasoning over the history (e.g., ``Put all the dirty blocks in the pink bowl.'' and ``What is the color of the block under the purple block?'') (see \cref{fig:example-queries} (top)). CaP+CoT fares slightly better in the Pick-and-Place and Relative Weight Reasoning, but still fails in most episodes. In contrast, Statler\xspace successfully handles the majority of these queries, demonstrating strong improvement over the others. It should be noted we explicitly chose queries that were challenging for LLM-based models, which partially accounts for why our model's scores show room for improvement.

\begin{table}[!th]
    \centering
    \caption{Success rates of Code-as-Policies\xspace (CaP) and Statler\xspace for non-temporal and temporal queries.}%
    \label{tb:main-result-wo-individual-truncation}
    \setlength{\tabcolsep}{3pt}
    {\footnotesize%
    \begin{tabularx}{1.0\linewidth}{lYYYY}%
        \toprule
        & \multicolumn{2}{c}{Non-temporal} & \multicolumn{2}{c}{Temporal}\\
        \midrule
        & CaP & Statler\xspace (ours)& CaP & Statler\xspace (ours)\\
        \midrule
        Pick \& Place &
        \nfrac{1.00}{\hphantom{0}(62/62)\hphantom{0}} &
        \nfrac{1.00}{\hphantom{0}(68/68)\hphantom{0}} &%
        \nfrac{0.31}{(9/29)} &
        \nfracb{0.83}{(48/58)}\\
        Disinfection &
        \nfrac{0.99}{(148/149)} &
        \nfrac{0.98}{(164/168)} &
        \nfrac{0.05}{(1/20)} &
        \nfracb{0.65}{(15/23)}\\
        Weight &
        \nfrac{1.00}{(107/107)} &
        \nfrac{1.00}{(107/107)} &
        \nfrac{0.00}{(0 / 20)} &
        \nfracb{0.55}{(11/20)}\\
        \bottomrule
    \end{tabularx}}
\end{table}
In order to better understand the behavior of Statler\xspace in comparison to Code-as-Policies, we analyze the success rates %
based on the type of user queries. Specifically, we categorize each query as either \emph{temporal} or \emph{non-temporal} depending on whether responding to the query necessitates temporal reasoning. We emphasize that contemporary methods, including the baselines that we consider, use non-temporal queries for evaluation. Table~\ref{tb:main-result-wo-individual-truncation} summarizes the resulting accuracy. 
The models often fail at different steps in an episode. We note that, when calculating accuracy we only consider the sequence of steps until %
the model fails to generate the correct code, which explains the mismatch in the denominators.

We see that both models achieve near-perfect performance on commands that do not require temporal reasoning. However, the performance of CaP noticeably decreases for tasks that require reasoning over the past. In contrast, Statler\xspace achieves success rates of $83\%$ (vs.\ $31\%$ for CaP) on Pick-and-Place, $65\%$ (vs.\ $5\%$ for CaP) on Block Disinfection, and $55\%$ (vs. $0\%$ for CaP) on Relative Weight Reasoning.

\vspace{-1pt}

Although our method enjoys a better robustness than the baseline methods, it inherits some issues of large language models, which hinders its performance. For example, it hallucinates block conditions (e.g., clean or dirty) or locations when the cleanliness of the block is never explicitly described. Moreover, the model's reasoning strategy seems to predominantly treat weight as an abstract concept, e.g. light vs. heavy, rather than executing mathematical computations. This weakness is evident when asking the model to accumulate blocks in a bowl until their total weight surpasses that of another bowl, yet the model underfills the bowl. 
In the disinfection domain, our model struggles to comprehend ambiguous terms like ``other'' in queries such as ``the other blocks are clean.'' It can also wrongly infer from the training prompt that a block \emph{at the bottom} becomes dirty when a block is placed on top of it, irrespective of the latter's cleanliness.

\vspace{-1.5pt}

\subsection{Real Robot Experiments}

\vspace{-1.5pt}

\begin{figure}[!t]
    \centering
    \newlength{\mywidth}
    \setlength{\mywidth}{0.19\linewidth}%
    \begin{tikzpicture}[
        utterance/.style={rectangle, rounded corners, draw=black, font=\scriptsize, minimum width=4.25cm, minimum height=0.5cm, text width = 4.25cm, fill=black!10, outer sep=0pt},empty/.style={outer sep=1pt, inner sep=0pt}]
        \node[empty,rotate=90] at (0,0) (captext) {\bf\tiny Code-as-Policies};
        \node[empty, above=7pt and 5pt of captext.north east,anchor=south west] (speak) {\includegraphics[height=5mm]{papers/statler/figures/speak-icon}};
        \node[utterance, right= 1pt and 0 of speak.south east, anchor=south west, rounded corners] (utterance) {\textls[-20]{Put the black cup on the yellow block.\\ Put the yellow block on the Rubik's cube.}};
        \node[empty, right=0pt of captext.south] (cap0) {\includegraphics[width=\mywidth]{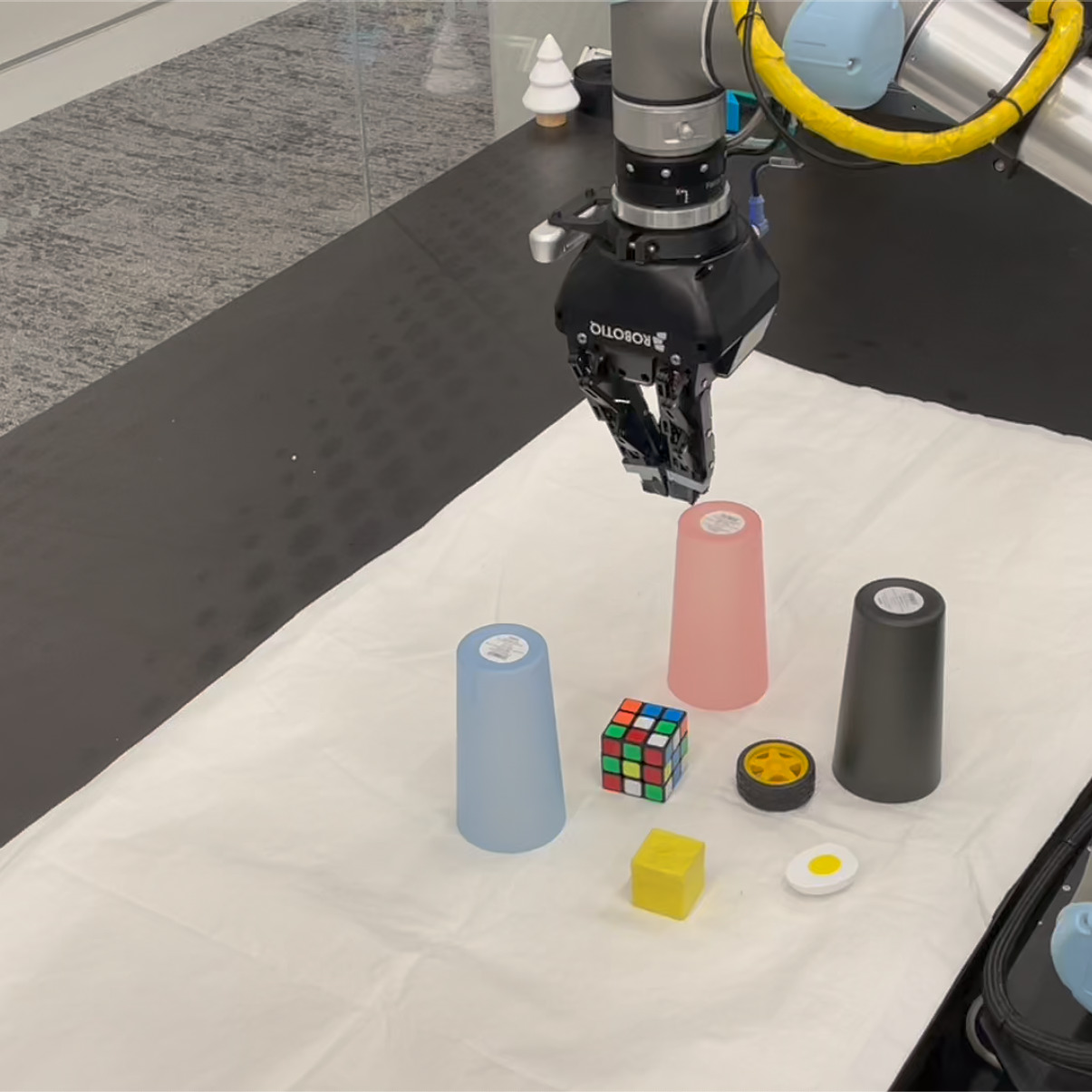}};
        \node[empty,right = -1pt of cap0, draw=green, ultra thick] (cap1) {\includegraphics[width=\mywidth]{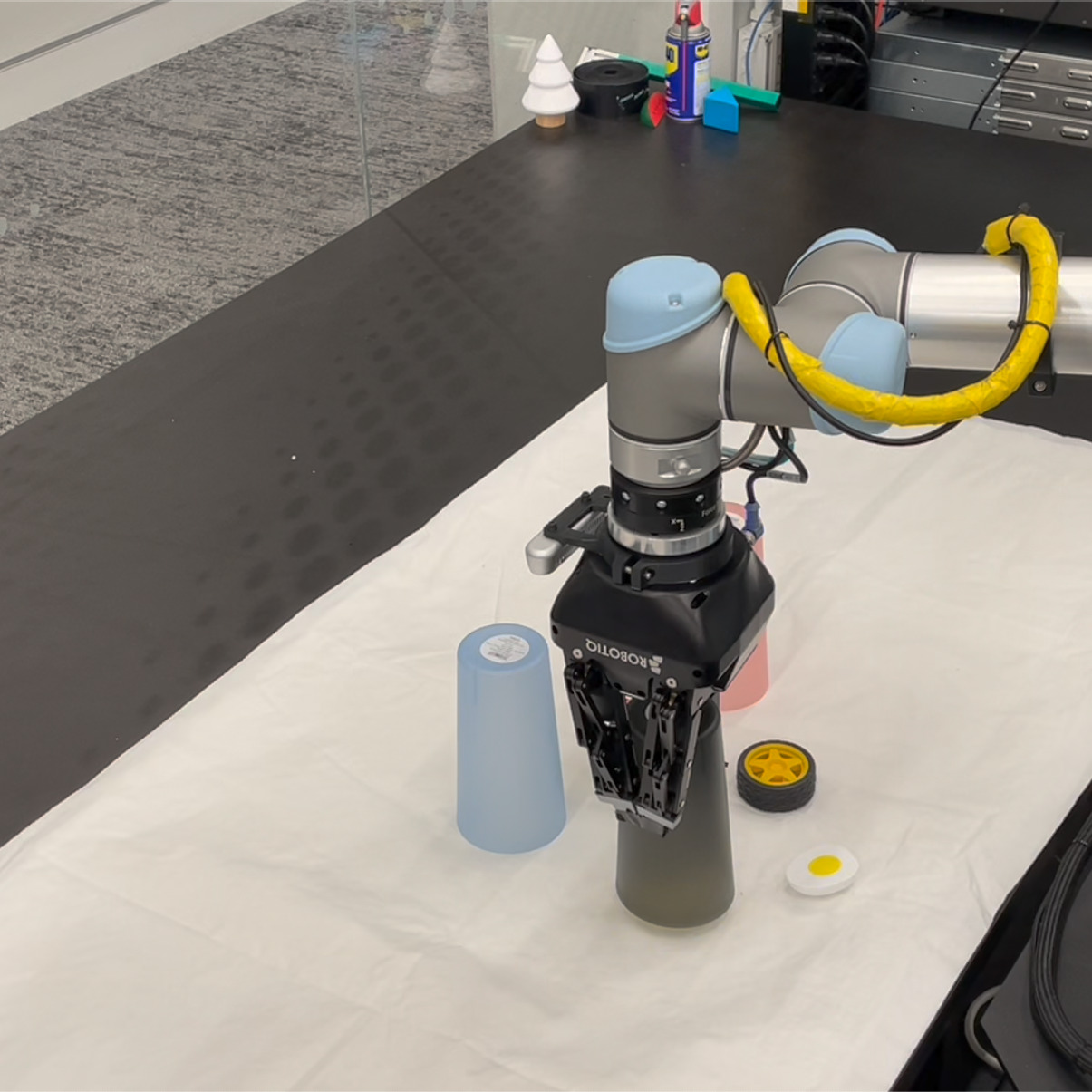}};
        \node[empty,right = -1pt of cap1] (cap2) {\includegraphics[width=\mywidth]{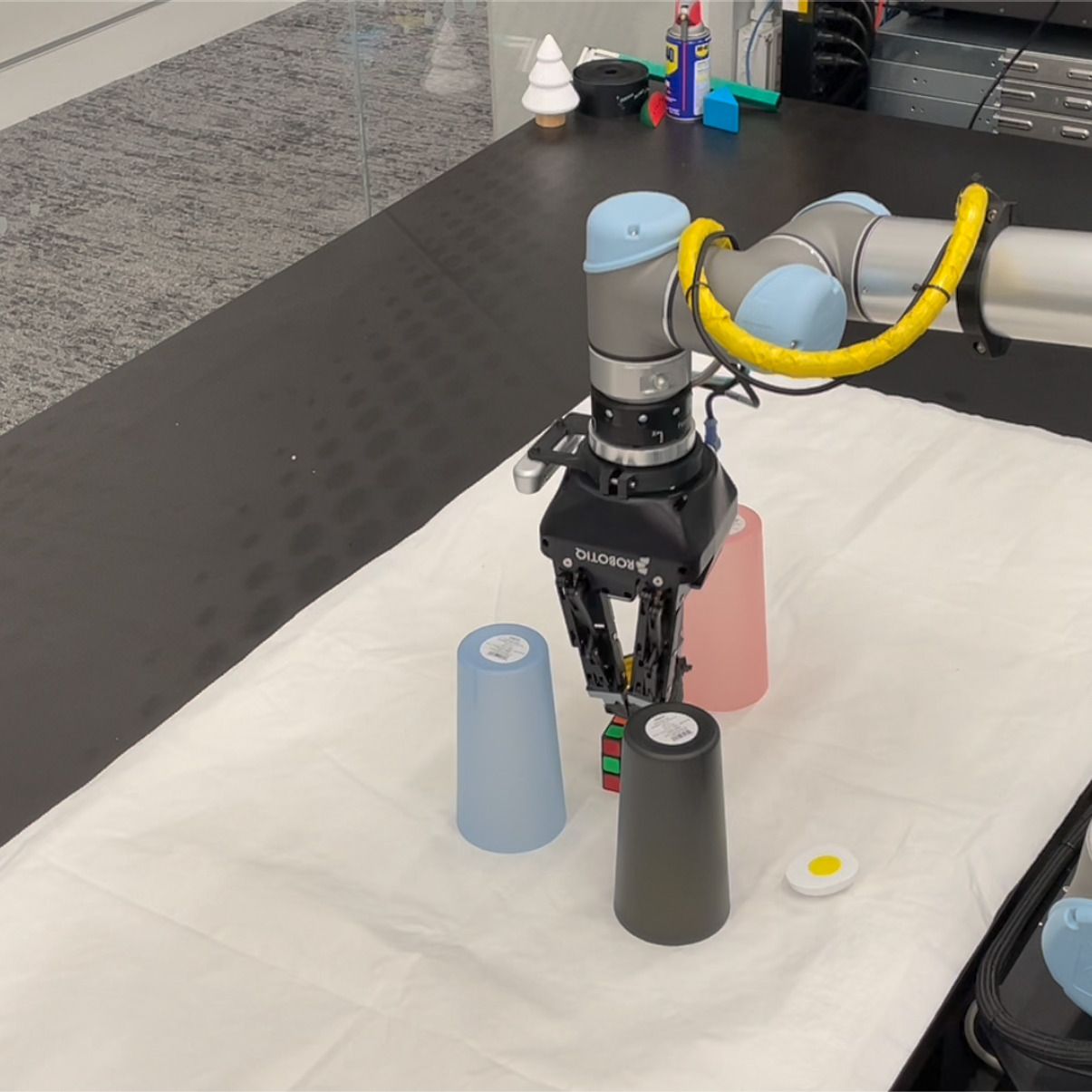}};
        \node[empty,right = -1pt of cap2,draw=red,ultra thick] (cap3) {\includegraphics[width=\mywidth]{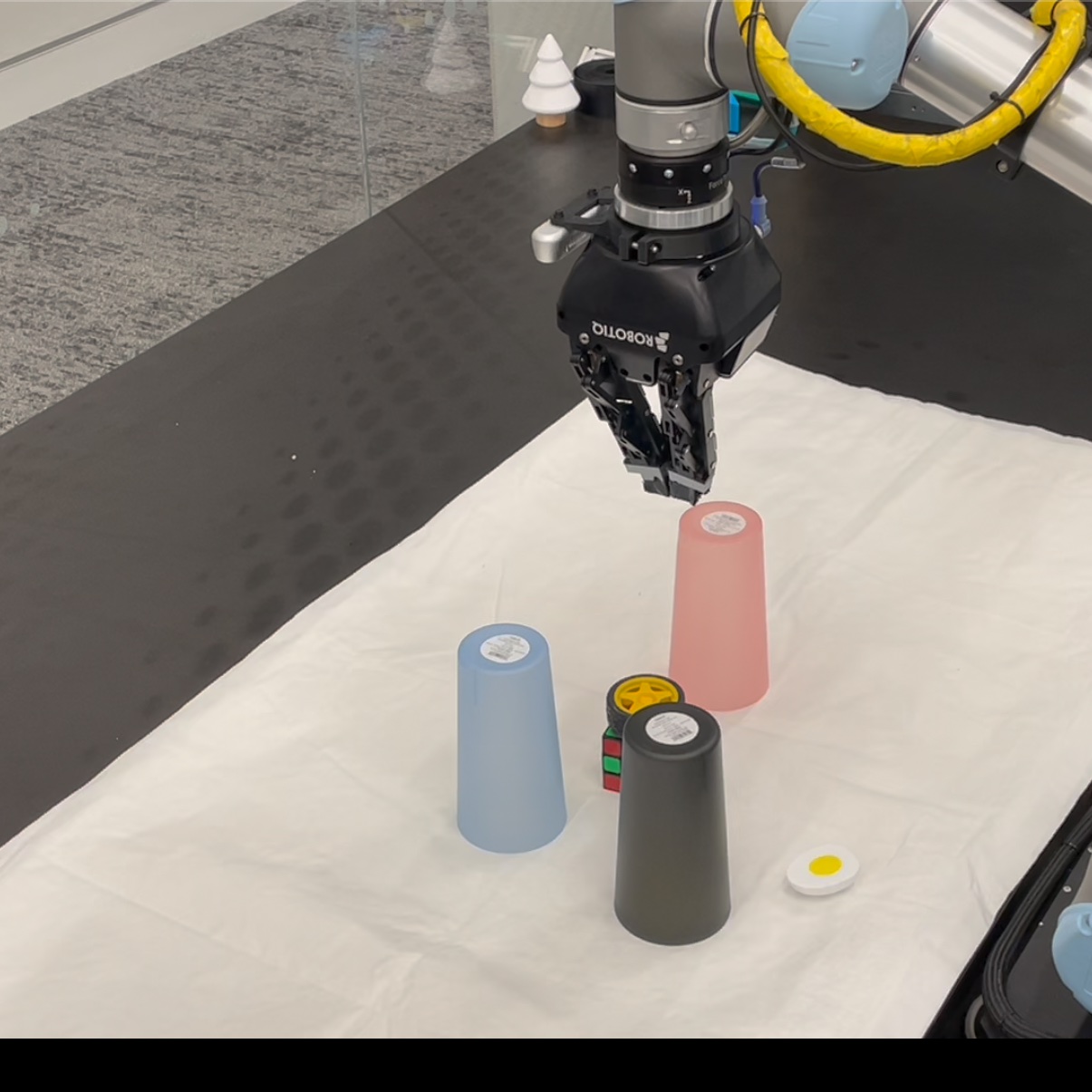}};
        \node [circle,draw=black, fill=white, inner sep=1pt, minimum size=5pt, above right=5pt of cap3.south west, anchor=south west] (check) {\redx};
        \node[empty,below = 1pt of cap0] (statler0) {\includegraphics[width=\mywidth]{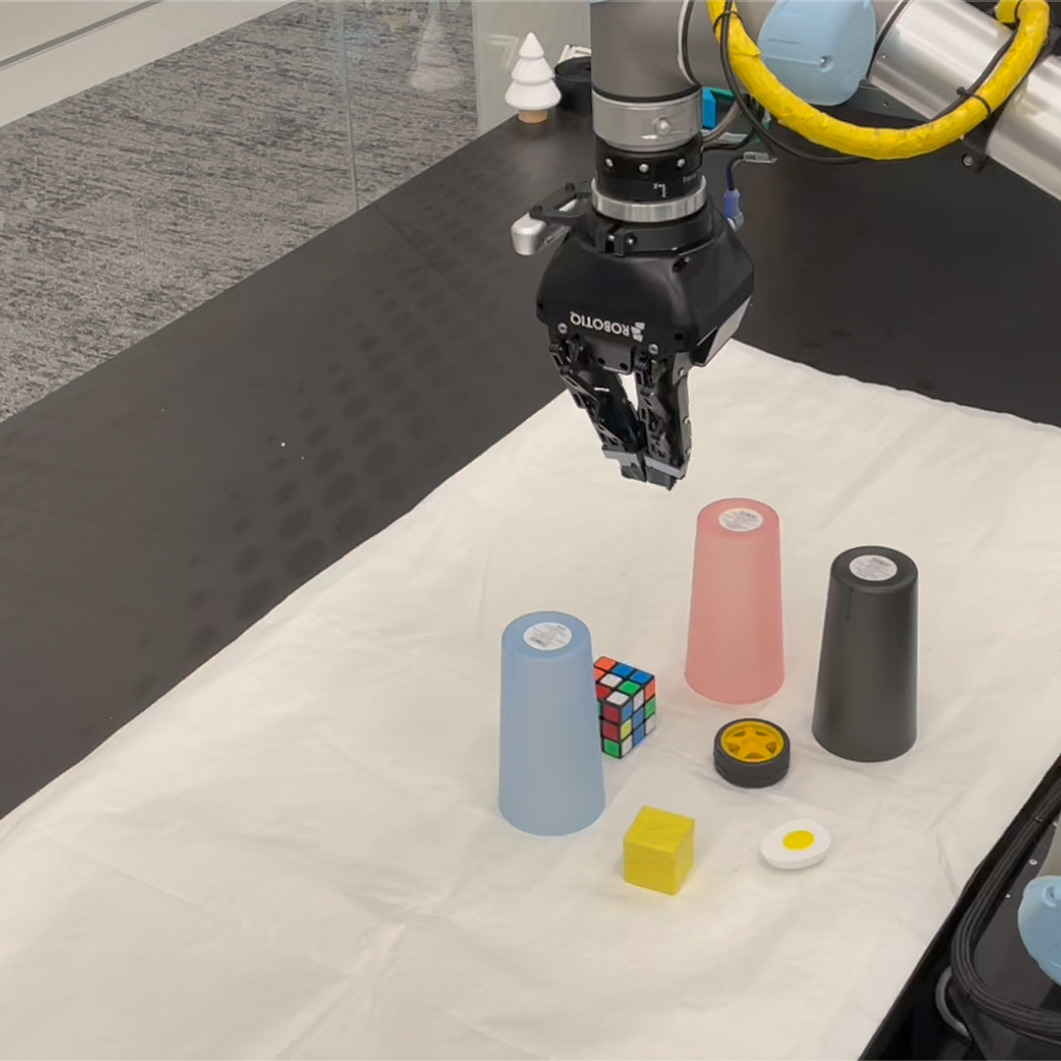}};%
        \node[empty,left = 0pt of statler0, rotate=90, anchor=south] () {\bf\tiny Statler\xspace};
        \node[empty,right = -1pt of statler0, draw=green, ultra thick] (statler1) {\includegraphics[width=\mywidth]{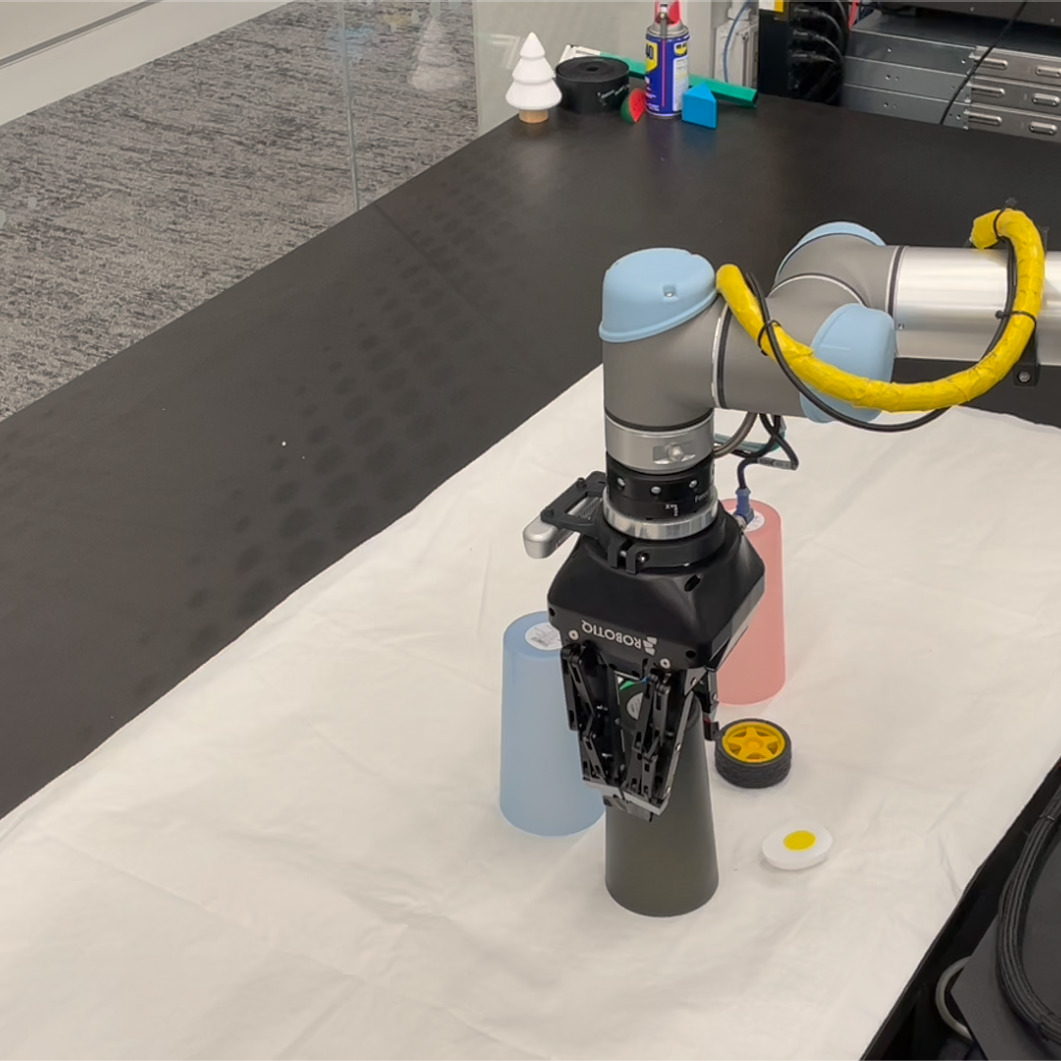}};%
        \node[empty,right = -1pt of statler1] (statler2) {\includegraphics[width=\mywidth]{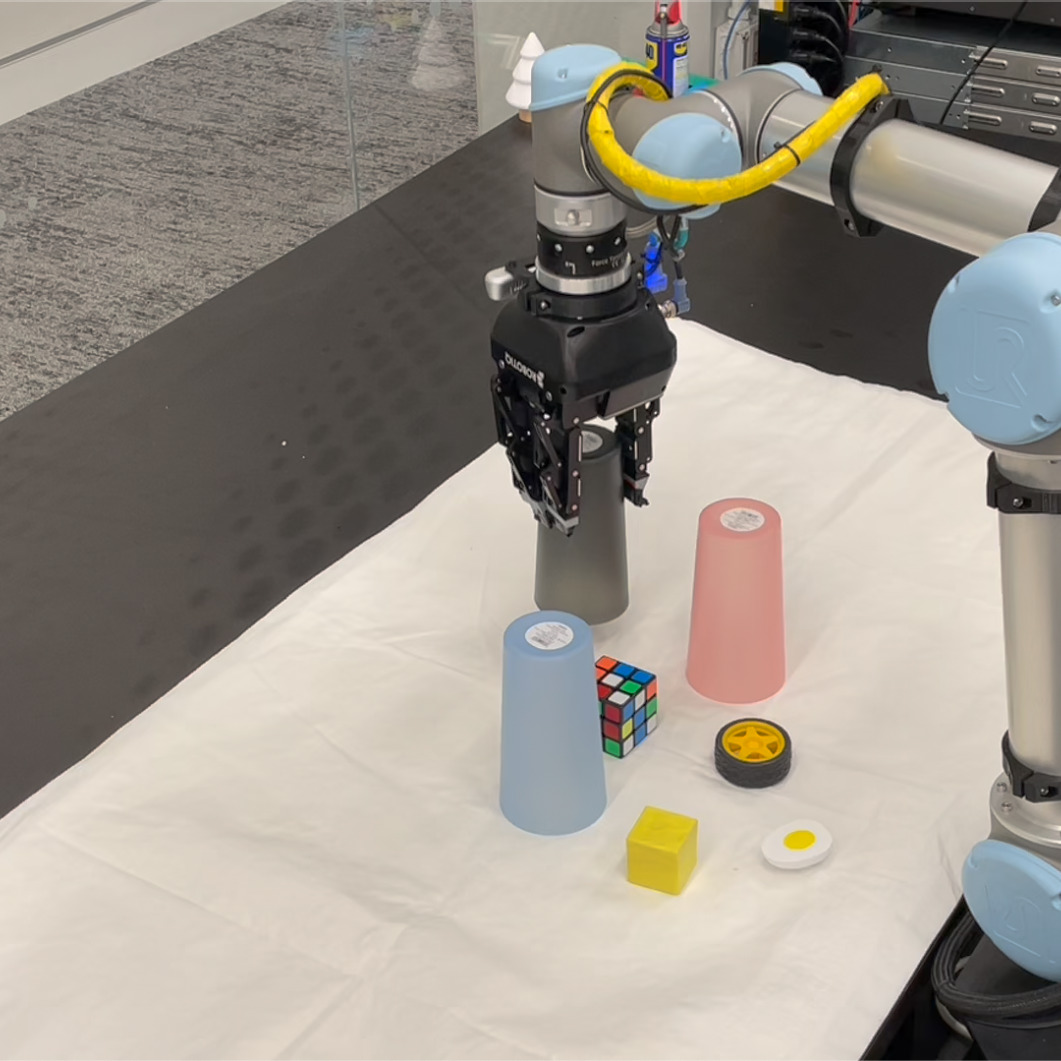}};%
        \node[empty,right = -1pt of statler2] (statler3) {\includegraphics[width=\mywidth]{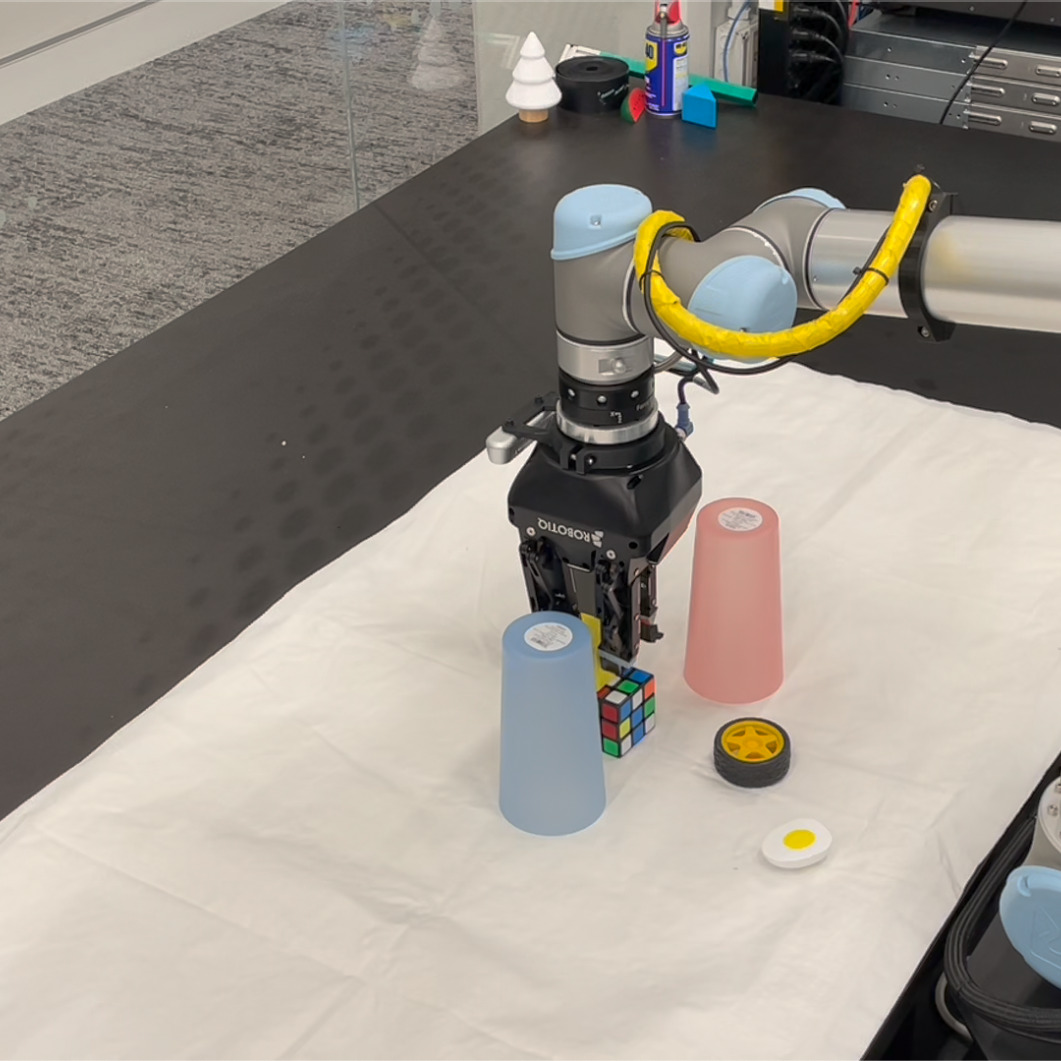}};%
        \node[empty,right = -1pt of statler3,draw=green, ultra thick] (statler4) {\includegraphics[width=\mywidth]{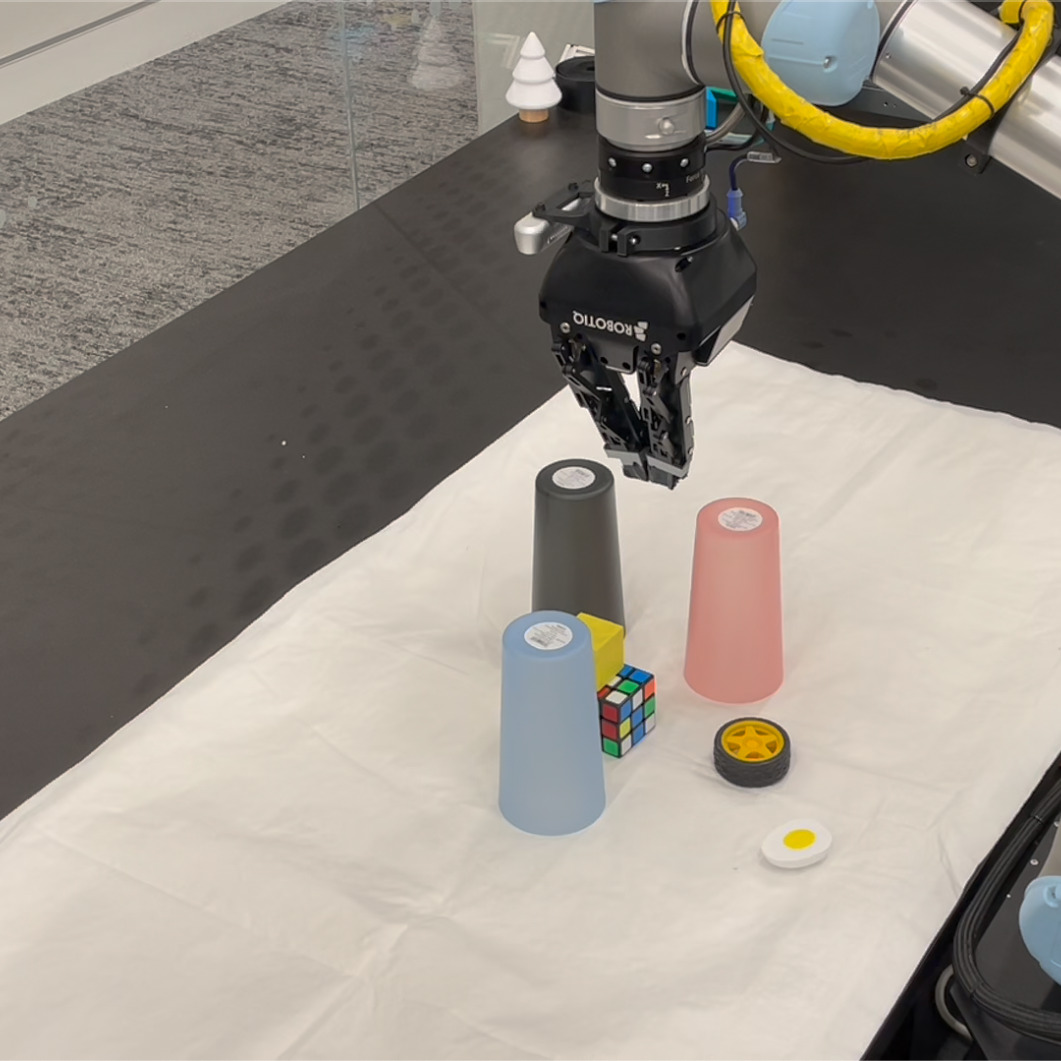}};%
        \node [circle,draw=black, fill=white, inner sep=1pt, minimum size=5pt, above right=5pt of statler4.south west, anchor=south west] (check) {\greencheck};
        \node[below = 2pt of statler0.south west] (line-start) {};
        \node[below = 2pt of statler4.south east] (line-end) {};
        \draw[thick,-stealth] (line-start) -- (line-end) node [midway,fill=white] {\footnotesize Time};
    \end{tikzpicture}
    \caption{A comparison of the resulting behavior for (top) Code-as-Policies\xspace and (bottom) our Statler\xspace model for the real robot experiments for the given multi-sentence instruction. %
    Frames %
    correspond to instances when the robot has placed an object. In order to successfully carry out the instruction, the robot must subsequently remove the black cup immediately after using it to cover the yellow block so that it can place the yellow block on the Rubik's cube. However, the the baseline Code-as-Policies\xspace (top row, third frame) fails to move the black cup aside, leaving the yellow block covered. It then places a wrong object on top of the Rubik's cube.
    }\label{fig:real-robot-exp}
\end{figure}

In order to validate Statler on a real robot, we implement it on a UR5 arm in a similar tabletop domain as the simulated experiments. We use MDETR~\cite{kamath2021mdetr}, an open-vocabulary segmentation model, to obtain segmentation masks for objects within an RGB image captured by a RealSense camera mounted on the gripper. Using the predicted object mask and the depth image, the object point-cloud can be obtained, from which we estimate its center position and bounding box. All of the primitive functions are identical to those used in simulation. In this domain, the robot is asked to stack objects and cover objects with different colored cups. At any point, an object is only permitted to be covered by at most a single object or cover. If the robot is asked to manipulate the bottom object, it must put away the top one. If a new cover or object is to be stacked on it, the existing one must be removed.

We evaluate the performance of Statler vs.\ CaP in the real robot domain using $10$ episodes. Statler achieves episode and step (in parentheses) success rate of $\nfrac{40\%}{(70\%)}$, where $67\%$ of the failure cases are due to LLM reasoning while others are caused by either perception or manipulation issues. The success rate for CaP is $\nfrac{20\%}{(46\%)}$, where LLM reasoning accounts for $88\%$ of failures. In Figure~\ref{fig:real-robot-exp}, we also provide a short example where the CaP baseline fails. The difficulty is in recognizing that yellow block is hidden under the black cup, which must be removed before picking up the yellow block as Statler\xspace correctly spots. Instead, CaP is not aware of this and tries to pick up the yellow block nonetheless, which leads MDETR to incorrectly detect the toy wheel that has yellow color in it as the yellow block.

\subsection{State-Maintenance Ablations}
To better understand Statler\xspace's state-maintenance strategy, we consider three different approaches to tracking the state.

\begin{codefloat}
\begin{lstlisting}[caption={Portion of Statler\xspace-Auto prompt.},label={lst:statler-auto},firstnumber=auto]
Your task is to maintain the status of these items using a JSON dictionary and update the status of the corresponding items after a new query.
This JSON dictionary will be commented, meaning that the starting character of each line is #.@
\end{lstlisting}
\end{codefloat}

The first (Statler\xspace-Unified) employs a single LLM as both the world-state reader and writer using a prompt that interleaves Statler\xspace's reader and writer prompts. At each step, the LLM first generates the action and then predicts the state that results from executing that action. The LLM then uses the resulting state when reasoning over the next query. Using a single LLM is conceptually simple, but it incurs an added burden for reasoning and generalization. Inspired by InstructGPT~\cite{ouyang2022training}, the second (Statler\xspace-Auto) does not receive any in-context state-update examples for the world-state writer. Instead, we provide a natural language description of how the state should be maintained. Prompt~\ref{lst:statler-auto} shows the relevant portion of the prompt. With an instruction and no in-context state-update examples, we ran our model on the same set of tasks. The third (Statler\xspace w/o State) ablates the world-state maintenance components of Statler\xspace entirely, resulting in a model that reduces to Code-as-Policies.

\begin{table}[!ht]
    \centering
    \caption{Ablation episode (individual step) success rates.}\label{tb:ablations}
    \setlength{\tabcolsep}{2.5pt}
    \begin{tabularx}{1.0\linewidth}{lYYY}%
        \toprule
        & {Pick \& Place} & {Disinfection} & {Weight}\\
        \midrule
       Statler\xspace w/o State & 
        \nfrac{0.00}{(0.54)} & %
        \nfrac{0.00}{(0.68)} & %
        \nfrac{0.00}{(0.84)} \\
       Statler\xspace-Unified &
        \nfrac{0.40}{(0.85)} & %
        \nfrac{0.35}{(0.79)} & %
        \nfrac{0.50}{(0.92)} \\
       Statler\xspace-Auto &
        \nfracb{0.75}{(0.88)} & %
        \nfracb{0.45}{(0.82)} & %
        \nfrac{0.40}{(0.90)} \\
        \midrule
       Statler\xspace (ours) &
        \nfrac{0.50}{(0.88)} & %
        \nfrac{0.40}{(0.82)} & %
        \nfracb{0.55}{(0.93)} \\
       \bottomrule
    \end{tabularx}
\end{table}
Table~\ref{tb:ablations} compares the performance of Statler\xspace to the three variations in terms of both their full-episode completion rates (using $20$ episodes for each domain) as well their individual step success rates. Without maintaining the world-state, Statler\xspace w/o State fails to complete any episodes (recall that an episode is considered to be a failure if the model fails to respond to one of the user queries during the episode) and results in individual step success rates that are significantly lower than Statler\xspace. Meanwhile, we see that Statler\xspace's use of separate LLMs for the world-state reader and world-state writer results in consistently higher episode success rates compared with the use of a unified reader and writer (Statler\xspace-Unified). The individual step success rates are higher for Pick-and-Place and Block Disinfection, and comparable for Relative Weight Reasoning. With regards to Statler\xspace's use of separate LLMs for the world-state writer and reader, we note that in-context learning has been shown to be sensitive to variations in prompt templates, the order of examples, and the examples used
\cite{dong2022survey,zhao2021calibrate}. In light of this, it is plausible that the performance gains that we achieve by dividing our reader and writer may be attributed in part to this sensitivity, allowing the models to, in effect, become specialized at their respective tasks.  Interestingly, Statler\xspace-Auto performs noticeably better than Statler\xspace and Statler\xspace-Unified with regards to the episode success rate on the Pick-and-Place and Block Disinfection domains, but comparable to Statler\xspace in terms of the individual success rates, and worse for Relative Weight Reasoning.

%% file: papers/statler/sections/conclusion.tex
\label{sec:conclusion}
We presented Statler\xspace, a language model that maintains an explicit representation of state to support longer-horizon robot reasoning tasks. Integral to Statler\xspace are a world-state reader that responds to a user query taking into account the current internal state, and a world-state writer that maintains the world state. 
Evaluations on various simulated and real robot manipulation tasks reveal that Statler\xspace significantly outperforms contemporary models on  non-trivial tasks that require reasoning over the past. Ablations demonstrate the contributions of our world-state reader and writer, and suggest Statler\xspace's flexibility to the state representation.

%% file: contents/conclusion.tex
\chapter{Conclusion and Discussion}

This thesis investigated two major components required for \emph{Embodied Spatial Intelligence}:
\begin{enumerate}
    \item \textbf{Robotic scene representation through implicit modeling}, and
    \item \textbf{Embodied spatial reasoning with hybrid systems}.
\end{enumerate}

\section*{Summary of Contributions}

On scene representation, this thesis first showed how self-supervised learning can enable camera self-calibration from image sequences without a calibration target (Section~\ref{sec:self-calib})~\cite{fang2022selfcalib}. It then introduced \emph{DeFiNe}, a continuous depth field that achieves state-of-the-art generalization by reducing architectural bias and introducing a novel 3D data augmentation strategy (Section~\ref{sec:define})~\cite{guizilini2022depth}. Finally, it presented \emph{NeRFuser}, a method to scale typically local neural radiance fields to building- and city-scale scenes via NeRF registration and blending (Section~\ref{sec:nerfuser})~\cite{fang2023nerfuser}. Collectively, these contributions advance the creation of high-fidelity, large-scale 3D scene representations from 2D observations. While implicit representations have enabled remarkable progress in novel view synthesis~\cite{Mildenhall2020NeRFRS} and 3D semantic search~\cite{kerr2023lerf}, their application to robotics tasks with strong classical alternatives, such as visual SLAM~\cite{mur2015orb}, remains competitive but not yet universally superior~\cite{zhu2022nice, Zhu2023NICERSLAMNI}.

On spatial reasoning, this thesis introduced \emph{MANGO}, a benchmark that probes the mapping and navigation competence of LLMs from textual traversals, revealing significant gaps compared to human performance (Section~\ref{sec:mango})~\cite{ding2024mango}. To bridge the gap between 3D perception and language-based reasoning, it proposed \emph{Transcrib3D}, which translates local 3D geometry into text and uses tool-augmented, iterative reasoning to achieve state-of-the-art 3D referring-expression resolution (Section~\ref{sec:transcrib3d})~\cite{fang2024transcrib3d}. Lastly, it presented \emph{Statler}, a dual-LLM architecture that explicitly maintains a symbolic world state to improve long-horizon planning and execution (Section~\ref{sec:statler})~\cite{yoneda2023statler}.

\section{A Paradigm Shift: The Rise of 2D Large-Scale Models}

During the course of this PhD, the research landscape was reshaped by the ascent of large-scale models trained predominantly on 2D data. Vision-language models (VLMs) have markedly improved language-grounded robotics, from navigation~\cite{zhang2024navid, shah2023lm, min2021film} to manipulation~\cite{huang2023voxposer, shen2023distilled, kim2024openvla}. This trend has culminated in vision-language-action (VLA) models that are trained end-to-end to output robot actions directly from 2D images and language commands~\cite{team2024octo, brohan2023rt, deng2024can, Black20240AV, Intelligence202505AV}.

This synergy extends to generative models. Language-conditioned video generators like Veo3~\cite{deepmind_veo3_2025} can synthesize scenes with plausible physics and notable 3D consistency. More impressively, action-conditioned world models like Genie3~\cite{bruce2024genie, deepmind_genie3_2025} can generate minute-long, visually consistent video sequences from actions alone, apparently without an explicit 3D rendering engine. In a sense, these models demonstrate an \emph{emergence} of 3D understanding from massive 2D datasets.

\section{A New "Cloud" on the Horizon: Are 2D Models Sufficient?}

This rapid progress prompts a critical question: are 2D large models the final answer to embodied intelligence, with progress bounded only by scale? While this is a compelling possibility, I recall the “two clouds” metaphor from Lord Kelvin 125 years ago. At a time when the physics community believed its theoretical foundations were largely complete, Kelvin highlighted two lingering anomalies. These seemingly small questions ultimately led to the revolutions of general relativity and quantum mechanics.

Similarly, a few "clouds" linger over the apparent dominance of 2D models. The primary cloud concerns their persistent weakness on geometric and arithmetic tasks~\cite{ramachandran2025well, gao2025vision}. Performance on tasks like precise counting, metric depth estimation, and 3D object detection still lags behind specialized, state-of-the-art models. Indeed, even the most capable LLMs today struggle with simple character counting or arithmetic without deferring to external tools. Built upon the same Transformer architecture, it is unsurprising that VLMs inherit these limitations. This deficit is also apparent in generation~\cite{huang2025jog3r,el2024probing,wu2025geometry,zhang2025world}; while generated videos are visually plausible, close inspection reveals that their implicit 3D structure often violates geometric principles, such as maintaining consistent vanishing points. Thus, the current emergence of 3D from 2D pretraining is, at best, approximate.

\section{The Role of 3D in the Large-Model Era}
Given the abundance of 2D data and the relative scarcity of 3D labels, 3D should not be seen as a competitor to 2D models, but as a complementary, high-leverage component. Its role is threefold:

\begin{enumerate}
    \item \textbf{3D as Ground Truth for Validation and Diagnosis.} High-quality 3D signals are indispensable for rigorously evaluating whether models pretrained on 2D data have developed genuine spatial understanding. This applies to both perception (are the model's beliefs about the world geometrically sound?) and generation (are the model's simulated worlds physically consistent?).
    
    \item \textbf{3D as a Conditioning for Consistent Generation.} While large-scale video models demonstrate impressive short-term consistency, maintaining coherence over long horizons remains a fundamental challenge. Because the physical world is governed by time-consistent 3D geometry, an explicit 3D representation can serve as a powerful memory or scaffold to condition video generation, enforcing global consistency. Early results in 3D-conditioned generation are already promising~\cite{ji2025cogen, wu2025video, liu2025revision}.
    
    \item \textbf{3D as a Target for Supervised Fine-Tuning.} After broad pretraining on 2D internet data, smaller, high-quality multi-view 3D datasets can be used to efficiently instill robust spatial priors in a model. This targeted fine-tuning can improve downstream spatial reasoning and control, providing a crucial bridge before more expensive reinforcement learning or deployment-specific adaptation.
\end{enumerate}

\section*{Closing Remarks}
The path forward in robotics will undoubtedly be dominated by scaling models on vast 2D datasets, as this approach has proven tractable and effective for learning general-purpose semantic capabilities. The emergent 3D understanding from this process will only improve with scale. However, for embodied agents that must act with precision, safety, and long-horizon consistency in the physical world, this thesis argues that 3D is indispensable.

The research presented here supports a vision where the general semantic power of 2D large models is grounded and refined by the geometric and physical rigor of 3D structure. Explicit 3D serves as a critical tool for inducing, conditioning, and verifying the spatial intelligence of these systems. The strategic fusion of scalable 2D learning with structured 3D knowledge represents a practical and robust path toward the ultimate goal of creating truly capable \emph{Embodied Spatial Intelligence}.